%% file: main.tex
\documentclass[12pt,twoside,phd]{dms}
\usepackage[utf8]{inputenc} 
\usepackage[T1]{fontenc}    %

\usepackage{lmodern}

\anglais

\def\sloppy{%
  \tolerance 500
  \emergencystretch 3em%
  \hfuzz .5pt
  \vfuzz\hfuzz}
\usepackage{graphicx,amssymb,icomma}

\usepackage[labelfont=bf, width=\linewidth]{caption}

\usepackage[pdfencoding=auto]{hyperref}  
\hypersetup{colorlinks=true,allcolors=black}
\usepackage{hypcap}   
\usepackage{bookmark} 
\numberwithin{equation}{section}
\numberwithin{table}{chapter}
\numberwithin{figure}{chapter}

\usepackage[onehalfspacing]{setspace}
\usepackage[utf8]{inputenc} 
\usepackage[T1]{fontenc}    
\usepackage{url}            
\usepackage{booktabs}       
\usepackage{amsfonts}       
\usepackage{nicefrac}       
\usepackage{microtype}      
\usepackage{xcolor}         

\usepackage{graphicx}
\usepackage{amsthm}
\usepackage{mathtools}
\usepackage{pifont}
\usepackage{nicefrac,xfrac}

\usepackage{threeparttable}

\usepackage[para]{footmisc}

\makeatletter
\newcommand{\printfnsymbol}[1]{%
  \textsuperscript{\@fnsymbol{#1}}%
}
\makeatother

\usepackage{subcaption}
\usepackage{amsmath}
\usepackage{cleveref}
\usepackage{natbib}
\usepackage{cancel}
\usepackage{multirow}
\usepackage{wrapfig}
\usepackage{multicol}

\usepackage{epsfig}
\usepackage{wrapfig}
\usepackage{enumitem}

\usepackage{float}
\usepackage{cancel}
\usepackage{listings}
\usepackage{tabularx}
\usepackage[acronym]{glossaries}

\usepackage{amsmath,amsfonts,bm}
\usepackage{mathtools, stmaryrd}

\usepackage{longtable}
\usepackage{afterpage}

\usepackage{placeins}

\allowdisplaybreaks

\usepackage{empheq}
\newcommand*\widefbox[1]{\fbox{\hspace{2em}#1\hspace{2em}}}

\input{include/math_commands}
\input{include/macros}

\usepackage[compact]{titlesec}
\titlespacing*{\section}{0pt}{12pt plus 4pt minus 2pt}{4pt plus 2pt minus 2pt}
\titlespacing*{\subsection}{0pt}{12pt plus 4pt minus 2pt}{4pt plus 2pt minus 2pt}
\titlespacing*{\subsubsection}{0pt}{12pt plus 4pt minus 2pt}{4pt plus 2pt minus 2pt}
\makeatletter 
\renewcommand{\@seccntformat}[1]{\csname the#1\endcsname\ }
\makeatother

\usepackage{titletoc}
\dottedcontents{section}[3em]{}{2.1em}{5pt}
\dottedcontents{subsection}[6.1em]{}{3em}{5pt}
\dottedcontents{subsubsection}[10.1em]{}{4em}{5pt}

\usepackage{etoolbox}
\makeatletter
\pretocmd{\chapter}{\addtocontents{toc}{\protect\addvspace{10\p@}}}{}{}
\pretocmd{\section}{\addtocontents{toc}{\protect\addvspace{10\p@}}}{}{}
\makeatother

\usepackage{chngcntr}
\counterwithin{figure}{chapter}
\counterwithin{table}{chapter}

\newacronym{cnf}{CNF}{Continuous Normalizing Flows}
\newacronym{gan}{GAN}{Generative Adversarial Networks}
\newacronym{msflow}{MRCNF}{Multi-Resolution Continuous Normalizing Flow}
\newacronym{msflow-image}{MRCNF}{Multi-Resolution Continuous Normalizing Flow}
\newacronym{msflow-im}{MRCNF}{Multi-Resolution Continuous Normalizing Flow}
\newacronym{msflow-wv}{MRCNF-Wavelet}{Multi-Resolution Continuous Normalizing Flow - Wavelet}
\newacronym{bpd}{BPD}{Bits-per-dimension}
\newacronym{mrf}{MRF}{Markov Random Field}
\newacronym{fid}{FID}{Fr\'echet Inception Distance}
\newacronym{ood}{OoD}{out-of-distribution}
\newacronym{mrcnf}{MRCNF}{Multi-Resolution Continuous Normalizing Flows}
\newacronym{ik}{IK}{Inverse Kinematics}
\newacronym{smpl}{SMPL}{SMPL}

\makeglossaries

\begin{document}

\version{1}



\title{Conditional Generative Modeling for Images, 3D Animations, and Video}

\author{Vikram Voleti}

\copyrightyear{2023}

\department{Département d'informatique et de recherche opérationnelle}

\date{July 24, 2023} 

\sujet{Informatique}

\president{Aaron Courville}

\directeur{Christopher Pal}


\membrejury{Ioannis Mitliagkas}

\examinateur{Greg Mori}   



\repdoyen{Aaron Courville} 


\maketitle

\maketitle


\francais

\chapter*{Résumé}

La modélisation générative pour la vision par ordinateur a connu d'immenses progrès ces dernières années, révolutionnant notre façon de percevoir, comprendre et manipuler les données visuelles. Ce domaine en constante évolution a connu des avancées dans la génération d'images, l'animation 3D et la prédiction vidéo, débloquant ainsi diverses applications dans plusieurs domaines tels que le divertissement, le design, la santé et l'éducation. Alors que la demande de systèmes de vision par ordinateur sophistiqués ne cesse de croître, cette thèse s'efforce de stimuler l'innovation dans le domaine en explorant de nouvelles formulations de modèles génératifs conditionnels et des applications innovantes dans les images, les animations 3D et la vidéo.

Notre recherche se concentre sur des architectures offrant des transformations réversibles du bruit et des données visuelles, ainsi que sur l'application d'architectures encodeur-décodeur pour les tâches génératives et la manipulation de contenu 3D. Dans tous les cas, nous incorporons des informations conditionnelles pour améliorer la synthèse des données visuelles, améliorant ainsi l'efficacité du processus de génération ainsi que le contenu généré.

Les techniques génératives antérieures qui sont réversibles entre le bruit et les données et qui ont connu un certain succès comprennent les flux de normalisation et les modèles de diffusion de débruitage. La variante continue des flux de normalisation est alimentée par les équations différentielles ordinaires neuronales (Neural ODEs) et a montré une certaine réussite dans la modélisation de la distribution d'images réelles. Cependant, elles impliquent souvent un grand nombre de paramètres et un temps d'entraînement élevé. Les modèles de diffusion de débruitage ont récemment gagné énormément en popularité en raison de leurs capacités de généralisation, notamment dans les applications de texte vers image.

Dans cette thèse, nous introduisons l'utilisation des Neural ODEs pour modéliser la dynamique vidéo à l'aide d'une architecture encodeur-décodeur, démontrant leur capacité à prédire les images vidéo futures malgré le fait d'être entraînées uniquement à reconstruire les images actuelles. Dans notre prochaine contribution, nous proposons une variante conditionnelle des flux de normalisation continus qui permet une génération d'images à résolution supérieure à partir d'une entrée à résolution inférieure. Cela nous permet d'obtenir une qualité d'image comparable à celle des flux de normalisation réguliers, tout en réduisant considérablement le nombre de paramètres et le temps d'entraînement.

Notre prochaine contribution se concentre sur une architecture encodeur-décodeur flexible pour l'estimation et l'édition précises de la pose humaine en 3D. Nous présentons un pipeline complet qui prend des images de personnes en entrée, aligne automatiquement un personnage 3D humain/non humain spécifié par l'utilisateur sur la pose de la personne, et facilite l'édition de la pose en fonction d'informations partielles.

Nous utilisons ensuite des modèles de diffusion de débruitage pour la génération d'images et de vidéos. Les modèles de diffusion réguliers impliquent l'utilisation d'un processus gaussien pour ajouter du bruit aux images propres. Dans notre prochaine contribution, nous dérivons les détails mathématiques pertinents pour les modèles de diffusion de débruitage qui utilisent des processus gaussiens non isotropes, présentons du bruit non isotrope, et montrons que la qualité des images générées est comparable à la formulation d'origine. Dans notre dernière contribution, nous concevons un nouveau cadre basé sur les modèles de diffusion de débruitage, capable de résoudre les trois tâches vidéo de prédiction, de génération et d'interpolation. Nous réalisons des études d'ablation en utilisant ce cadre et montrons des résultats de pointe sur plusieurs ensembles de données.

Nos contributions sont des articles publiés dans des revues à comité de lecture. Dans l'ensemble, notre recherche vise à apporter une contribution significative à la poursuite de modèles génératifs plus efficaces et flexibles, avec le potentiel de façonner l'avenir de la vision par ordinateur.

\textbf{Mots-clés:} Apprentissage profond, vision par ordinateur, modèles génératifs, apprentissage de la représentation, modèles de diffusion


\anglais
\chapter*{Summary}

Generative modeling for computer vision has shown immense progress in the last few years, revolutionizing the way we perceive, understand, and manipulate visual data. This rapidly evolving field has witnessed advancements in image generation, 3D animation, and video prediction that unlock diverse applications across multiple fields including entertainment, design, healthcare, and education. As the demand for sophisticated computer vision systems continues to grow, this dissertation attempts to drive innovation in the field by exploring novel formulations of conditional generative models, and innovative applications in images, 3D animations, and video.

Our research focuses on architectures that offer reversible transformations of noise and visual data, and the application of encoder-decoder architectures for generative tasks and 3D content manipulation. In all instances, we incorporate conditional information to enhance the synthesis of visual data, improving the efficiency of the generation process as well as the generated content.

Prior successful generative techniques which are reversible between noise and data include normalizing flows and denoising diffusion models. The continuous variant of normalizing flows is powered by Neural Ordinary Differential Equations (Neural ODEs), and have shown some success in modeling the real image distribution. However, they often involve huge number of parameters, and high training time. Denoising diffusion models have recently gained huge popularity for their generalization capabilities especially in text-to-image applications.

In this dissertation, we introduce the use of Neural ODEs to model video dynamics using an encoder-decoder architecture, demonstrating their ability to predict future video frames despite being trained solely to reconstruct current frames. In our next contribution, we propose a conditional variant of continuous normalizing flows that enables higher-resolution image generation based on lower-resolution input. This allows us to achieve comparable image quality to regular normalizing flows, while significantly reducing the number of parameters and training time.

Our next contribution focuses on a flexible encoder-decoder architecture for accurate estimation and editing of full 3D human pose. We present a comprehensive pipeline that takes human images as input, automatically aligns a user-specified 3D human/non-human character with the pose of the human, and facilitates pose editing based on partial input information.

We then proceed to use denoising diffusion models for image and video generation. Regular diffusion models involve the use of a Gaussian process to add noise to clean images. In our next contribution, we derive the relevant mathematical details for denoising diffusion models that use non-isotropic Gaussian processes, present non-isotropic noise, and show that the quality of generated images is comparable with the original formulation. In our final contribution, devise a novel framework building on denoising diffusion models that is capable of solving all three video tasks of prediction, generation, and interpolation. We perform ablation studies using this framework, and show state-of-the-art results on multiple datasets.

Our contributions are published articles at peer-reviewed venues. Overall, our research aims to make a meaningful contribution to the pursuit of more efficient and flexible generative models, with the potential to shape the future of computer vision.

\textbf{Keywords}: Deep learning, computer vision,
generative models, representation learning, diffusion models


\cleardoublepage
\pdfbookmark[chapter]{\contentsname}{toc}  
\tableofcontents
\cleardoublepage
\phantomsection  
\listoftables
\cleardoublepage
\phantomsection
\listoffigures


\chapter*{List of acronyms and abbreviations}
\begin{longtable}{p{.15\textwidth} p{.8\textwidth}}
  BPD & Bits-per-dimension \\
  CNF & Continuous Normalizing Flow \\
  CNN & Convolutional Neural Network \\
  DDPM & Denoising Diffusion Probabilistic Model \\
  EDS & Expected Denoised Sample \\
  FDP & Forward Diffusion Process \\
  FID & Fr\'echet Inception Distance \\
  FVD & Fr\'echet Video Distance \\
  GAN & Generative Adversarial Network \\
  GE & Geodesic Error \\
  GFF & Gaussian Free Field \\
  GPU & Graphics Processing Unit \\
  ICML & International Conference on Machine Learning \\
  IK & Inverse Kinematics \\
  IS & Inception Score \\
  IVP & Initial Value Problem \\
  MCVD & Masked Conditional Video Diffusion \\
  MLE & Maximum Likelihood Estimation \\
  MPJPE & Mean Per Joint Position Error \\
  MRCNF & Multi-Resolution Continuous Normalizing Flow \\
  NeurIPS & Annual Conference on Neural Information Processing Systems \\
  NI-DDPM & Non-Isotropic Denoising Diffusion Probabilistic Model \\
  NI-SMLD & Non-Isotropic Score Matching Langevin Dynamics \\
  NIVE & Non-Isotropic Variance Exploding \\
  NIVP & Non-Isotropic Variance Preserving \\
  ODE & Ordinary Differential Equations \\
  OoD & Out-of-Distribution \\
  PA-MPJPE & Procrustes-Aligned Mean Per Joint Position Error \\
  PSNR & Peak Signal-to-Noise Ratio \\
  RBN & Restricted Boltzmann Machine \\
  RDP & Reverse Diffusion Process \\
  RNN & Recurrent Neural Network \\
  SDE & Stochastic Differential Equation \\
  SGD & Stochastic Gradient Descent \\
  SIGGRAPH & Special Interest Group on Computer Graphics and Interactive Techniques (annual conference on computer graphics organized by ACM SIGGRAPH) \\
  SMLD & Score Matching Langevin Dynamics \\
  SMPL & Skinned Multi-Person Linear model \\
  SOTA & State-of-the-Art \\
  SPADE & SPatially-Aaptive DE-normalization \\
  SPATIN & SPAce-TIme-Adaptive Normalization \\
  SSIM & Structural SIMilarity \\
  VAE & Variational Autoencoder \\
  VE & Variance Exploding \\
  VP & Variance Preserving \\
\end{longtable}

\printglossary


\chapter*{Acknowledgements}

I feel extremely privileged to have had the chance to pursue a PhD at Mila. I've learned an immense amount from amazing peers and mentors. 
I've made many wonderful friends over the course of my PhD, and I will always be grateful to everyone at this incredible lab.

Most importantly, I want to thank my advisor Chris for being the most supportive advisor one could ask for. His enthusiasm, positivity and optimism right from when I met him as a prospective student is something I hope to inculcate. He has been genuinely happy about my successes within and outside of Mila, and equally supportive during difficult times. I am grateful for his mentorship and guidance, which has been instrumental in shaping my career as a researcher. I would also like to thank Adam Oberman for his exceptional empathy and invaluable guidance. I am also grateful to Yoshua Bengio and Valerie Pisano (among so many others) for running a lab where one is surrounded by incredibly smart people, and providing us an environment where ideas are exchanged freely. Thank you Linda Peinthiere and Celine B\'egin, without you, we'd be lost in an administrative maze.

I've also been fortunate to have been mentored by several amazing researchers during my internships. I'd like to thank Bryan Seybold and Sourish Chaudhuri at Google for being great mentors and allowing me to dip my toe in industry-applied research. I am grateful for their guidance during the internship, as well as their support beyond the program, including providing mock interview practice. I'd like to thank Boris Oreshkin and the DeepPose team including Florent Bocquel\'et, Louis-Simon M\'enard, F\'elix Harvey, Jeremy Cowles and others for all their help and guidance during my internship at Unity Technologies. Boris has been a wonderful collaborator and mentor. I've immensely enjoyed all our conversations, and I feel privileged to have learned so much from him: his humility, his attitude of approaching problems with a can-do spirit, his skill of anticipating how our current research can be tied to future benefits for the organization. 
I'd like to thank Yashar Mehdad and Barlas O\u{g}uz at Meta for giving me the resources and guidance to do large-scale text-to-3D research. It was inspiring to work with a team full of smart people who continued to perform excellent research despite only recently pivoting to computer vision. Their continual support throughout the internship helped me bridge the gap between academic research and industrial goals. 

There were also several co-authors and friends at Mila who have contributed substantially to the articles in this thesis as well as to my own personal development, including but not limited to ---- David Kanaa, Samira Kahou, Vincent Michalski for their humility and experienced insights; Sai Rajeshwar, Sandeep Subramanian for all the conversations over dosa; Tegan Maharaj for the parties and late night dinners; Anirudh Goyal, Krishna Murthy Jatavallabhula, Florian Golemo for their timely guidance and collaboration; Chris Finlay for his opportune collaboration and fun conversations; Roger Girgis, Martis Weiss, Mattie Tesfaldet, Chris Beckham for being the best friends one could ask for during this tumultuous journey; Alexia Jolicoeur-Martineau for being the best collaborator and co-author I have had the good fortune to work with. Thank you for giving me the opportunity to work with you, I've learned a lot from our interactions.

I thank Graham Taylor and Alexander Wong, who supported me during my PhD and made it easier to live and work outside of Montreal. I also thank Jason Deglint, Katie Thomas, and the BlueLion team for the wonderful conversations and fun meetups. I am grateful to Microsoft Research for granting me a Diversity Award so I could take care of important financial expenses.

Finally, I am indebted to my family, my father and mother for supporting me in all my endeavours in immeasurable and inconceivable ways throughout my life. I am also indebted to my wife, Kirti, who supported me through all my career aspirations, while excelling in her own PhD. Thank you for your patience and understanding, especially during times of uncertainty and self-doubt, and for being my partner in life.

 %
 %

\NoChapterPageNumber
\cleardoublepage


\include{Introduction}

\include{Background}

\include{EncODEDec}

\include{MRCNF}

\include{ProtoResSMPL}

\include{NIDDPM}

\include{MCVD}

\include{Conclusion}



\bibliographystyle{plainnat} 
\bibliography{ref.bib} 

\appendix

\end{document}

%% file: include/math_commands.tex
\usepackage{amsmath,amsfonts,bm}

\def\eqref#1{equation~\ref{#1}}

\def\1{\bm{1}}

\def\eps{{\epsilon}}

\def\rvepsilon{{\bm{\epsilon}}}
\def\rvtheta{{\bm{\theta}}}
\def\rva{{\mathbf{a}}}
\def\rvb{{\mathbf{b}}}
\def\rvc{{\mathbf{c}}}

\def\rvf{{\mathbf{f}}}
\def\rvg{{\mathbf{g}}}

\def\rvu{{\mathbf{i}}}

\def\rvp{{\mathbf{p}}}

\def\rvs{{\mathbf{s}}}

\def\rvu{{\mathbf{u}}}
\def\rvv{{\mathbf{v}}}
\def\rvw{{\mathbf{w}}}
\def\rvx{{\mathbf{x}}}
\def\rvy{{\mathbf{y}}}
\def\rvz{{\mathbf{z}}}

\def\rmA{{\mathbf{A}}}

\def\rmC{{\mathbf{C}}}

\def\rmG{{\mathbf{G}}}

\def\rmI{{\mathbf{I}}}

\def\rmK{{\mathbf{K}}}
\def\rmL{{\mathbf{L}}}
\def\rmM{{\mathbf{M}}}

\def\rmP{{\mathbf{P}}}
\def\rmQ{{\mathbf{Q}}}

\def\rmW{{\mathbf{W}}}

\def\vzero{{\bm{0}}}

\def\vmu{{\bm{\mu}}}

\def\evtheta{{\theta}}

\def\mK{{\bm{K}}}

\def\mLambda{{\bm{\Lambda}}}
\def\mSigma{{\bm{\Sigma}}}

\DeclareMathAlphabet{\mathsfit}{\encodingdefault}{\sfdefault}{m}{sl}
\SetMathAlphabet{\mathsfit}{bold}{\encodingdefault}{\sfdefault}{bx}{n}

\def\gC{{\mathcal{C}}}

\def\gG{{\mathcal{G}}}

\def\gI{{\mathcal{I}}}

\def\gL{{\mathcal{L}}}

\def\gN{{\mathcal{N}}}

\def\gU{{\mathcal{U}}}

\newcommand{\E}{\mathbb{E}}

\newcommand{\Var}{\mathrm{Var}}

\usepackage{url}

\usepackage{color}
\usepackage{graphicx}
\usepackage{float}
\usepackage{subcaption}
\usepackage{multirow}
\usepackage{siunitx}
\newcommand{\Norm}[1]{\left\Vert#1\right\Vert}
\newcommand{\abs}[1]{\vert#1\vert}
\newcommand{\Abs}[1]{\left\vert#1\right\vert}

\newcommand{\bq}{\begin{equation}}
\newcommand{\eq}{\end{equation}}

\newcommand{\loss}{\ell}

\newcommand{\D}{{\mathrm{d}}}
\newcommand{\EE}{\mathbb{E}}

\newcommand{\barx}{\bar x}
\newcommand{\xbar}{\barx}
\newcommand{\balpha}{\bar\alpha}

\newcommand\numberthis{\addtocounter{equation}{1}\tag{\theequation}}

\newcommand\conj{H}
\newcommand\trns{T}
\newcommand{\Real}{\text{Real}}

\newcommand{\complex}{\text{c}}

\allowdisplaybreaks

\newcommand{\ud}{\mathrm{d}}

\newcommand{\VE}{\text{VE}}
\newcommand{\VP}{\text{VP}}
\newcommand{\NIVE}{\text{NIVE}}
\newcommand{\NIVP}{\text{NIVP}}
\newcommand{\SMLD}{\text{SMLD}}
\newcommand{\DDPM}{\text{DDPM}}
\newcommand{\DDIM}{\text{DDIM}}

\newcommand{\GFFDDPM}{\operatorname{NI-DDPM}}
\newcommand{\GFFDDIM}{\operatorname{NI-DDIM}}
\newcommand{\GFFSMLD}{\operatorname{NI-SMLD}}

\newcommand{\x}{\mathbf{x}}

\usepackage{graphicx}
\usepackage{subcaption}
\usepackage{caption}

\newcommand{\Eb}[2]{\E_{#1}\!\left[#2\right]}

\newcommand{\bI}{\mathbf{I}}

\newcommand{\bx}{\mathbf{x}}

\newcommand{\bepsilon}{{\boldsymbol{\epsilon}}}
\newcommand{\bmu}{{\boldsymbol{\mu}}}

\newcommand\Tstrut{\rule{0pt}{2.6ex}}         

\newcommand\Tstrutsmall{\rule{0pt}{2.2ex}}         

%% file: include/macros.tex
\usepackage{xcolor}

\DeclareMathOperator{\smpl}{\textsc{SMPL}}
\DeclareMathOperator{\geodesic}{\textsc{GE}}
\DeclareMathOperator{\mpjpe}{\textsc{MPJPE}}
\DeclareMathOperator{\tr}{\textrm{tr}}
\DeclareMathOperator{\smplis}{\textsc{SMPL-SI}}
\DeclareMathOperator{\kernel}{\textsc{k}}

\usepackage{xparse} \DeclarePairedDelimiterX{\Iintv}[1]{\llbracket}{\rrbracket}{\iintvargs{#1}}
\NewDocumentCommand{\iintvargs}{>{\SplitArgument{1}{,}}m}
{\iintvargsaux#1} %
\NewDocumentCommand{\iintvargsaux}{mm} {#1\mkern1.5mu,\,\mkern1.5mu#2}

\newcommand{\indep}{\perp \!\!\! \perp}

\makeatletter
\newenvironment{sqcases}{%
  \matrix@check\sqcases\env@sqcases
}{%
  \endarray\right\rbrack%
}
\def\env@sqcases{%
  \let\@ifnextchar\new@ifnextchar
  \left\lbrack
  \def\arraystretch{1.2}%
  \array{c@{}}%
}
\makeatother

%% file: Introduction.tex
\anglais

\counterwithin{figure}{chapter}
\counterwithin{table}{chapter}

\chapter{Introduction}

\section{Computer vision and machine learning}

Computer vision is a field of study that aims to enable machines to understand and interpret visual information, including images and videos. The goal of computer vision systems is to replicate the capabilities of human vision, such as object detection, motion tracking, and understanding spatial relationships. This field has seen significant progress over the past several decades, especially due to machine learning in the recent past. In particular, generative modeling for computer vision has gained significant prominence in the past few years, and promotes a wide variety of applications.

Early work in computer vision focused on simple image processing techniques, such as edge detection and thresholding. In the 1960s and 1970s, researchers began to develop more sophisticated algorithms for image analysis, including pattern recognition and feature extraction. One of the most influential works from this era was the book ``Digital Picture Processing'' by A. Rosenfeld and A. C. Kak (1976) \citep{rosenfeld1969picture}, which provided a comprehensive overview of image processing techniques.

In the 1980s and 1990s, computer vision research shifted towards more advanced techniques, such as machine learning and neural networks. The seminal book ``Parallel Distributed Processing'' by D. E. Rumelhart and J. L. McClelland (1986) \citep{mcclelland1986parallel} introduced the idea of using neural networks for image analysis, which paved the way for the development of deep learning techniques in the 21st century.

Today, computer vision systems are used in a wide range of applications, from autonomous vehicles to medical imaging to facial recognition systems. Recent advances in deep learning have enabled machines to achieve human-level performance on many computer vision tasks, including object detection and recognition, image segmentation, and pose estimation.

The following sections introduce generative modeling in computer vision and applications. Relevant generative modeling techniques are discussed, with a special focus on two methods relevant to this thesis : continuous normalizing flows, and denoising diffusion models.

\section{Generative modeling}
\label{1_gen}

Generative modeling is a sub-field of machine learning that aims to learn the underlying probability distribution of a given dataset, and use this knowledge to generate new, realistic samples that are similar to the original data. This has many applications, such as image and video synthesis, data augmentation, style transfer, etc.

One of the earliest and most popular generative models is the Restricted Boltzmann Machine (RBM), which was introduced by Smolensky in 1986~\citep{smolensky1986information}. RBMs are a type of neural network that learn to model the joint probability distribution of the input data. RBMs have been used in a variety of applications, including image and speech recognition, recommendation systems, and collaborative filtering.

Another important early generative model is the Autoencoder, which was first introduced by Rumelhart et al. in 1986~\citep{rumelhart1986learning}. Autoencoders are neural networks that learn to encode input data into a lower-dimensional representation, and then decode this representation back into the original data. Autoencoders can be used for a variety of tasks, such as data compression, denoising, and anomaly detection.

In recent years, generative modeling has emerged as an exciting area of research. In the context of computer vision, generative models can be used to create realistic images, videos, and other visual media. Generative modeling for computer vision is typically based on deep learning architectures such as Generative Adversarial Networks (GANs)~\citep{goodfellow2014gan}, Variational Auto-Encoders (VAEs)~\citep{kingma2013auto}, Normalizing flows~\citep{dinh2015nice, dinh2016density}, Continuous normalizing flows~\citep{chen2018neural}, and Denoising diffusion models~\citep{song2019generative, ho2020ddpm}.

GANs were introduced in 2014 by Goodfellow et al. and have since become one of the most popular and widely used generative models for computer vision. GANs consist of two neural networks, a generator network and a discriminator network, that are trained together in a game-like setting. The generator network learns to create realistic images that can fool the discriminator network into thinking they are real, while the discriminator network learns to distinguish between real and fake images.

VAEs, on the other hand, are based on the idea of learning a low-dimensional representation of the data that can be used to generate new samples. They were introduced in 2014 by Kingma and Welling, and have since become a popular choice for generative modeling in computer vision. VAEs consist of an encoder network that learns to map input images to a latent space, and a decoder network that learns to generate new images from the latent space.

Normalizing flows are a family of generative models that learn to transform a simple distribution, such as a standard normal distribution, into a complex distribution that resembles the target distribution of the data. Normalizing flows have shown impressive results in image generation, and have been used to create high-resolution images.

The upcoming sections introduce two specific generative modeling techniques: Continuous normalizing flows, and Denoising diffusion Models. These techniques have been chosen due to their direct relevance to the contributions presented in the following chapters. Furthermore, conditional variants, and the diverse applications  of generative models will be discussed.

\subsection{Continuous normalizing flows}

Continuous Normalizing Flows (CNFs) are another class of generative models that have shown great promise in recent years that leverage Neural Ordinary Differential Equations (Neural ODEs). Instead of specifying a fixed sequence of transformations like Normalizing Flows, Neural ODEs learn a continuous-time dynamics model that evolves over time by using a neural network to approximate the solution to an ordinary differential equation. This approach provides a flexible and adaptive way of modeling complex systems.
Neural ODEs have been successfully applied to a range of applications, including image and speech generation, data imputation, and scientific simulations.

Continuous Normalizing Flows go a step further, and assume that on one end of these Neural ODE-based transformations is a simple distribution, such as a standard normal distribution, resulting in a flexible and expressive generative model.
CNFs have been shown to be effective in a range of applications, including image and speech generation, and have the potential to be used in a wide range of scientific and engineering applications. However, they can be challenging to train, and there is ongoing research to develop more efficient and effective training algorithms for these models.

\subsection{Denoising diffusion models}

Denoising diffusion models are a relatively new class of generative models that learn a stochastic process that gradually transforms a known distribution, such as a standard normal distribution, into the target distribution of the data. Unlike normalizing flows which learn a series of invertible transformations, diffusion models leverage a series of noise processes that act as a stochastic diffusion process. This makes diffusion models well-suited for modeling complex high-dimensional distributions, and they have shown great promise in a range of applications, including image synthesis and denoising tasks.

The ability of diffusion models to generate high-quality data with realistic textures and fine details has been demonstrated in several recent studies~\citep{ho2020ddpm, chen2021wavegrad, voleti2022MCVD, poole2023dreamfusion}, where they have been used to generate images, audio, videos, and 3D objects that are virtually indistinguishable from real-world data. Furthermore, diffusion models have been shown to have a number of advantages over other generative models, including improved training stability, better likelihood estimation, and the ability to handle missing data. Despite these successes, there are still many challenges to be addressed in the development and application of diffusion models, and ongoing research is focused on improving their scalability, robustness, and efficiency.

\section{Conditional generative modeling}

Conditional generative modeling involves generating new data samples based on a given set of conditions. The conditioning aspect of the modeling refers to providing additional information to the generative model in the form of inputs or labels, which are used to guide the generation process. In other words, the model is trained to generate data samples that are conditioned on specific inputs or labels. This approach is particularly useful when the generated data needs to meet certain criteria, or follow specific patterns. For example, in image generation, the conditional inputs may include the desired object category, color, or size. The conditioning information can be incorporated into the model architecture in various ways, such as by adding an additional input layer, or by modifying the loss function to account for the conditioning information.

Text-to-image diffusion models are a type of conditional generative model that involves generating high-quality images based on textual descriptions~\citep{rombach2022high}.
The conditioning information in this case includes the textual description of the image to be generated, such as the color, shape, and position of various objects. The generation process in text-to-image diffusion models involves iteratively refining the image based on the given textual input.
One of the key advantages of text-to-image diffusion models is their ability to generate highly customized images based on specific textual inputs. This makes them particularly useful in applications such as e-commerce, where personalized images can be generated based on the customer's description of the product they want to purchase. This can be expanded to other modalities such as audio, video, etc.

\section{Applications}

Generative modeling for computer vision has many exciting applications, such as image and video synthesis, data augmentation, and style transfer. These models can be used to create new, realistic images and videos from scratch, and to manipulate existing visual media in creative ways. For example, generative models can be used to translate images into different styles~\citep{gatys2016neural} such as converting a photo into a painting or a sketch, or to create realistic animations of objects and scenes~\citep{unterthiner2018towards,voleti2022MCVD}.

Another important application of generative modeling in computer vision is in data augmentation. Data augmentation refers to the process of creating new training samples by applying random transformations to existing samples. This technique enhances deep learning models by expanding the training set and boosting performance. Generative models can generate new, realistic training samples to augment the training set~\citep{zhang2021datasetgan}, potentially improving the accuracy and robustness of the model.

One of the most prominent applications of image, 3D, and video generation is in the entertainment industry. The ability to generate realistic images and animations has enabled filmmakers and game developers to create stunning visual effects and immersive worlds that were previously impossible to achieve using traditional methods~\citep{mildenhall2020nerf}.

In advertising, image and video generation enables realistic product simulations and virtual try-on experiences for customers~\citep{jiang2022clothformer}. This can enhance sales and reduce returns by allowing customers to visualize the product's fit before making a purchase.

In healthcare, image and video generation can be used for a variety of applications, such as training medical professionals and developing new treatments. For example, medical simulations can be generated to train surgeons and other healthcare professionals in complex procedures, while virtual environments can be created to simulate and test new medical treatments before they are used on patients~\citep{Mirchi2020}.

Overall, conditional generative modeling is a powerful technique for generating highly customized and realistic data samples. It has broad applications across various industries and fields. The ability to create realistic images, 3D objects, and videos based on specific conditions brings numerous benefits. As generative modeling techniques advance, we anticipate even more exciting applications in image, 3D, and video generation in the future.


\section{Thesis overview}

This thesis aims to explore various aspects of conditional generative modeling. The following is a list of peer-reviewed publications based on these contributions:
\begin{itemize}
\item \textit{NeurIPS 2022} - ``MCVD: Masked Conditional Video Diffusion for Prediction, Generation, and Interpolation'', V. Voleti, A. Jolicoeur-Martineau, C. Pal \href{https://arxiv.org/abs/2205.09853}{arXiv}
\item
\textit{NeurIPS 2022 Workshop} - ``Score-based Denoising Diffusion with Non-Isotropic Gaussian Noise Models'', V. Voleti, C. Pal, A. Oberman \href{https://arxiv.org/abs/2210.12254}{arXiv}
\item
\textit{SIGGRAPH Asia 2022} - ``SMPL-IK: Learned Morphology-Aware Inverse Kinematics for AI-Driven Artistic Workflows'', V. Voleti, B. N. Oreshkin, F. Bocquelet, F. G. Harvey, L. Ménard, C. Pal
\href{https://arxiv.org/abs/2208.08274}{arXiv}
\item 
\textit{ICML 2021 Workshop} - ``Improving Continuous Normalizing Flows using a Multi-Resolution Framework'', V. Voleti, C. Finlay, A. Oberman, C. Pal
\item
\textit{NeurIPS 2019 Workshop} - ``Simple Video Generation using Neural ODEs'', V. Voleti*, D. Kanaa*, S. E. Kahou, C. Pal
\href{https://arxiv.org/abs/2109.03292}{arXiv}

\end{itemize}

Through a comprehensive literature review, original research, and analysis, this thesis provides an understanding of the application of (conditional) generative modeling in the domains of:
\begin{itemize}
    \item images: \Cref{chap:MRCNF,chap:NIDDPM},
    \item 3D animations: \Cref{chap:ProtoResSMPL}, and
    \item video: \Cref{chap:EncODEDec,chap:MCVD}.
\end{itemize}

The chapters are arranged in the chronological order of publication of the respective articles.
\Cref{chap:Background} provides essential background information that lays the foundation for subsequent chapters.
\Cref{chap:EncODEDec} employs Neural Ordinary Differential Equations (Neural ODEs) in an encoder-decoder framework. \Cref{chap:MRCNF} builds on Continuous Normalizing Flows (CNFs), a development on Neural ODEs. \Cref{chap:ProtoResSMPL} utilizes an encoder-decoder framework for 3D pose estimation. \Cref{chap:NIDDPM,chap:MCVD} further develop and employ denoising diffusion models.

\Cref{chap:Background} consists of relevant background information that is useful for the later chapters. \Cref{section:neuralode} introduces Neural ODEs, and \Cref{sec:bg_CNF} presents Continuous Normalizing Flows (CNFs). These concepts shall be helpful to understand \Cref{chap:EncODEDec,chap:MRCNF}. \Cref{sec:ddpm} introduces denoising diffusion models, including relevant mathematical details. It involves the derivations of two broad streams of diffusion models called Denoising Diffusion Probabilistic Models (DDPM) (\Cref{sec:ddpm}) and Score-Matching Langevin Dynamics (SMLD) (\Cref{sec:smld}). These concepts shall be helpful to understand \Cref{chap:NIDDPM,chap:MCVD}.

\Cref{chap:EncODEDec} describes our novel application of Neural ODEs to video prediction. \Cref{EncODEDec:related} details relevant prior work. \Cref{EncODEDec:neuralode} provides an overview of Neural ODEs. \Cref{EncODEDec:method} explains our method of using Neural ODEs to model the dynamics of a video. \Cref{EncODEDec:exp} demonstrates the application of our method on the MovingMNIST dataset. \Cref{EncODEDec:future} discusses the limitations of our method and future work. This work was published at a workshop at NeurIPS 2019.

\Cref{chap:MRCNF} derives a conditional variant of continuous normalizing flows called Multi-Resolution Continuous Normalizing Flows (MRCNF), and applies it to image generation. \Cref{background} details background information, and \Cref{method} builds on them to derive MRCNF. \Cref{section:related} reviews prior related works, with a focus on WaveletFlow in \Cref{subsec:WaveletFlow}. \Cref{mrcnf:results} presents experimental results on image generation, showcasing MRCNF's efficiency in terms of number of parameters and training time. It also includes various conclusions from ablation studies. \Cref{ood} analyzes the out-of-distribution (OoD) properties of MRCNFs, and shows that they are similar to those of other normalizing flows. This work was published as a workshop paper at ICML 2021, and a version is currently under review for a journal.

\Cref{chap:ProtoResSMPL} focuses on incorporating a 3D human pose prior in inverse kinematics i.e. estimation of full 3D pose from partial inputs. \Cref{protores:bg} expands on relevant background information. \Cref{{sec:smpl-ik}} presents our learned morphology-aware inverse kinematics module involving an encoder-decoder architecture that is flexible in input space. \Cref{sec:smpl_is} details our proposed method to apply the estimated pose on non-human 3D characters. \Cref{sec:proposed_ai_driven_flow} presents our proposed artistic workflow for 3D scene authoring from an image. \Cref{app:smpl_ik_details} provides details on the neural network architecture, training, and evaluation. \Cref{protores:res,protores:lim} report experimental results and limitations of our method, and \Cref{protores:demos} provides demo videos. This work was published at SIGGRAPH Asia 2022.

\Cref{chap:NIDDPM} presents a novel mathematical derivation of Non-Isotropic Denoising Diffusion Models, a variant of denoising diffusion models that uses an underlying non-isotropic Gaussian noise model.
\Cref{sec:niddpm} derives Non-Isotropic DDPM (NI-DDPM), and \Cref{sec:nismld} derives Non-Isotropic SMLD (NI-SMLD). \Cref{figs:IvsNIDDPM} is dedicated to a direct comparison between DDPM and NI-DDPM. \Cref{GFF} presents an instantiation of a non-isotropic noise process called Gaussian Free Fields (GFF), and \Cref{NIDDPM_exps} reports results from experiments on image generation from GFFs using NIDDPM. This work was published as a workshop paper at NeurIPS 2022.

\Cref{chap:MCVD} presents a novel application of denoising diffusion models to modeling video in a conditional fashion. \Cref{MCVD:method} derives the use of diffusion models to solve three video-related tasks : video prediction, video generation, video interpolation, all using a single model. \Cref{MCVD:arch} explains the model architecture used. \Cref{MCVD:related} provides a direct comparison with prior relevant methods, and discusses other prior methods. \Cref{MCVD:exp} details state-of-the-art results on all tasks, and \Cref{MCVD:ablation} mentions various ablation studies conducted. \Cref{MCVD:videos} provides qualitative results of generated videos. This work was published as a conference paper at NeurIPS 2022.

Overall, I hope this thesis provides a comprehensive presentation of these key modern methods for constructing conditional generative models for images, 3D animations, and video, and their potential impact on the field of computer vision and deep learning.

%% file: Background.tex
\anglais
\counterwithin{figure}{chapter}
\counterwithin{table}{chapter}

\chapter{Background}
\label{chap:Background}

This chapter introduces the relevant details of ODEs (\Cref{sec:bg_ODE}), Neural ODEs (\Cref{section:neuralode}), and Continuous Normalizing Flows (\Cref{sec:bg_CNF}). These concepts will be helpful for \Cref{chap:MRCNF,chap:EncODEDec}. It then introduces the main mathematical details of DDPM~\citep{ho2020ddpm} (\Cref{sec:ddpm}) and SMLD~\citep{song2019generative} (\Cref{sec:smld}). These concepts will be helpful for \Cref{chap:NIDDPM,chap:MCVD}.

\input{NeuralODE}

\input{DDPM_SMPLD}

%% file: NeuralODE.tex
\section{Ordinary Differential Equations (ODEs)}
\label{sec:bg_ODE}

Ordinary Differential Equations (ODEs) are mathematical tools used to model dynamic systems that evolve over time \citep{courant1965introduction, apostol1991calculus, strang1991calculus}. They are used to describe phenomena such as population growth, chemical reactions, motion of particles, heat transfer, and electrical circuits, to name a few, and have a wide range of applications in various fields of science and engineering, including physics, chemistry, biology, economics, and engineering. ODEs come in various forms, from simple linear equations to complex non-linear systems, and can be solved analytically or numerically. Many numerical methods have been developed to solve ODEs, making them an essential tool for modeling and simulating various dynamical systems in a variety of fields.

\subsection{Initial Value Problem (IVP)}

Typically, the context of ODEs begins with the Initial Value Problem. For a state $x(t)$ that changes dynamically with time $t$, if the rate of change is described by a function $f$, and if the initial value of the state $x$ at time $t_0$ is given, then what is the value of $x(t)$ at some other time step $t_1$?
\begin{align}
\frac{dx(t)}{dt} = f(x(t), t);\ \ x(t_0) \ \small{\text{is given}};\ \ \normalsize{x(t_1) = \ ?}
\end{align}

Here, $f$ is called the differential of $x$ with respect to time $t$. Many physical processes follow this template of the Initial Value Problem (IVP).

The solution to the Initial Value Problem is:
\begin{align}
x(t_1) = x(t_0) + \int_{t_0}^{t_1} f(x(t), t)\ dt .
\end{align}

As an example,
\begin{align*}
\frac{dx}{dt} = 2t;\ x(0) &= 2;\ x(1) =\ ? \numberthis\\
\implies x(1) &= x(0) + \int_0^1 2t\ dt , \\
&= x(0) + (t^2|_{t=1} - t^2|_{t=0}) , \\
&= 2 + 1^2 - 0^2 , \\
&= 3. \numberthis
\end{align*}

\subsection{Numerical integration}

The above example involved the use of analytical integration for $\int_{t_0}^{t_1} f(x(t), t)\ dt$. However, in some cases, it may not be possible to use simple integration to estimate the solution. For example, the solution to the following IVP requires some non-trivial simplification:
\begin{align*}
\frac{dx}{dt} = 2xt\ ;\ x(0) &= 3;\ x(1) =\ ? \numberthis \\
\implies \int\frac{1}{2x}\ dx &= \int t\ dt , \\
\implies \frac{1}{2}\log{x} &= \frac{1}{2}t^2 + c_0 , \\
\implies x(t) &= ce^{t^2} . \numberthis \\
\text{\small{We know that}}\ x(0) &= 3 \implies  c = 2 ,\\
\implies x(t) &= 2e^{t^2} , \\
\implies x(1) &= 5.436 . \numberthis
\end{align*}

Hence, the analytic solution is:
\begin{align*}
\text{\underline{Analytic solution}:}\ \ x(t) &= 2e^{t^2} \implies x(1) = 5.436 . \numberthis \\
\end{align*}

\begin{figure}[!hbt]
\begin{center}
\includegraphics[width=0.45\linewidth]{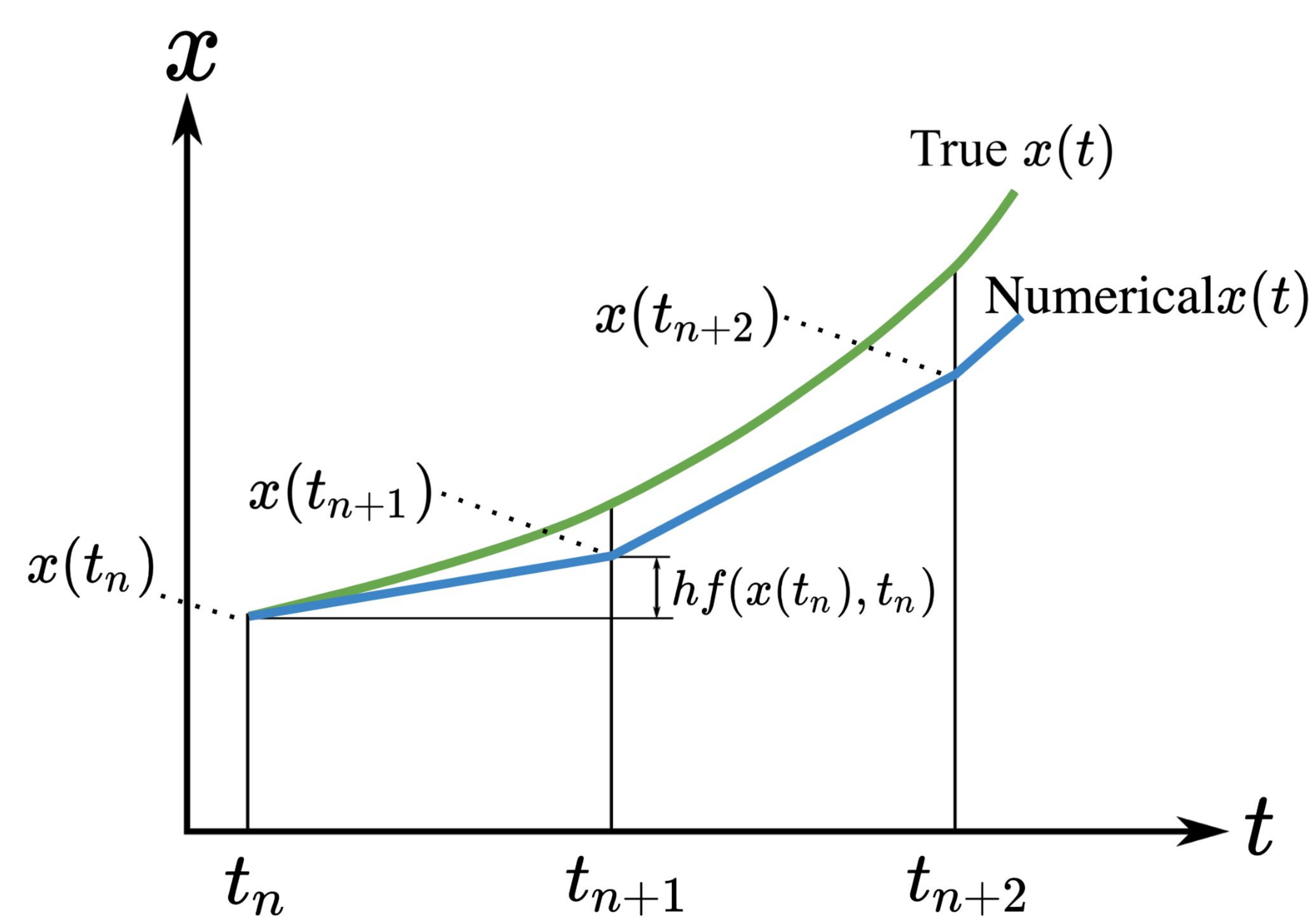}
\end{center}
\caption{Illustration of 1st-order Runge-Kutta / Euler’s method of
numerical integration.
}
\label{fig:ode}
\end{figure}

In such cases, in order to automate the integration process in a computer, the integration is approximated using Numerical Integration methods, or ODE Solvers. There are several ODE Solvers, the simplest for them being
the Euler method. This is illustrated in \Cref{fig:ode}, and is described below:
\begin{align}
\begin{cases}
t_{n+1} &= t_n + h , \\
x(t_{n+1}) &= x(t_n) + h\ f(x(t_n), t_n) .
\end{cases}
\label{eq:euler1}
\end{align}

Hence, using Euler's method in \Cref{eq:euler1}, the numerical solution to the above problem is:
\begin{align*}
\text{\underline{Numerical solution}:} \\
\text{Let}\ h &= 0.25. \numberthis\\
x(0.25) &= x(0) + 0.25*f(x(0), 0) ,\\
&= 3 + 0.25 * (2 * 3 * 0) ,\\
&= 3 . \numberthis\\
x(0.5) &= x(0.25) + 0.25*f(x(0.25), 0.25) ,\\
&= 3 + 0.25 * (2 * 3 * 0.25) ,\\
&= 3.375 .\numberthis\\
x(0.75) &= x(0.5) + 0.25 * f(x(0.5), 0.5),\\
&= 3.375 + 0.25 * (2 * 3.375 * 0.5),\\
&= 4.21875 .\numberthis \\
x(1) &= x(0.75) + 0.25 * f(x(0.75), 0.75),\\
&= 4.21875 + 0.25 * (2*4.21875*0.75),\\
&= 5.8008. \numberthis
\end{align*}

We can see that the analytic solution 5.436 and the numerical solution 5.8008 are not equal. Indeed, the numerical solution is only an approximation of the analytical solution. Better approximations are obtained using higher order methods such as Runge-Kutta methods~\citep{pontyagin1962mathematical, kutta1901beitrag,hairer2000solving}, multi-step algorithms such as the Adams-Moulton-Bashforth methods~\citep{butcher2016numerical, hairer2000solving, quarteroni2000matematica}, etc. More advanced algorithms provide better control over the approximation error and the accuracy~\citep{press2007numerical}. For example, the 4th order Runge-Kutta method is used as the default ODE solver in many applications:
\begin{align}
\begin{cases}
t_{n+1} &= t_n + h,\\
s_1 &= f(x(t_n),\ t_n),\\
s_2 &= f(x(t_n) + \frac{h}{2}s_1,\ t_n + \frac{h}{2}) ,\\
s_3 &= f(x(t_n) + \frac{h}{2}s_2,\ t_n + \frac{h}{2}) ,\\
s_4 &= f(x(t_n) + hs_3,\ t_n + h) ,\\
x(t_{n+1}) &= x(t_n) + \frac{h}{6}(s_1 + 2s_2 + 2s_3 + s_4).
\end{cases}
\end{align}

Hence, we shall now use ``ODESolve'' as a placeholder for the user's choice of ODE Solver. The solution to the Initial Value Problem is thus:
\begin{align}
x(t_1) = \text{ODESolve}(\ f(x(t), t),\ x(t_0),\ t_0,\ t_1\ ),
\label{eq:odesolve}
\end{align}
where $x(t_1)$ is the value of the state $x$ to be estimated at time step $t_1$, $x(t_0)$ is the initial value of $x$ at initial time step $t_0$, and $f$ is the differential function of $x$.


Suppose $f$ is continuously differentiable. Then, considering the solution curves as plotted on a plane with time $t$ on one axis and state $x$ on the other axis/axes,
\begin{enumerate}
    \item the solution curves for this differential equation completely fill the plane, and
    \item the solution curves of different solutions do not intersect.
\end{enumerate}

\begin{figure}[!tbh]
\begin{center}
\includegraphics[width=0.45\linewidth]{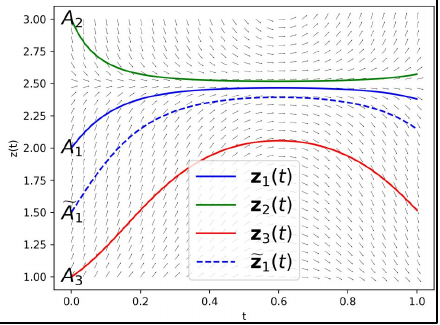}
\end{center}
\caption{Illustration of the solution to ODEs being flows (taken from \citet{yan2020robustness}).
}
\label{fig:flows}
\end{figure}

This means the solution of an ODE is a \textit{flow}, it involves non-intersecting solution curves. This is illustrated in \Cref{fig:flows}, as provided in \citet{yan2020robustness}.

\clearpage

\section{Neural Ordinary Differential Equations (Neural ODEs)}
\label{section:neuralode}

Neural ODEs formulate ODEs such that the differential function is a neural network~\citep{chen2018neural}. This represents a paradigm shift in the way ODEs were typically solved. Whereas earlier ODEs governing natural phenomena were hand-designed, now using the framework of Neural ODEs, the ODE could be learnt through backpropagation into a parameterized neural network.

Suppose the problem of classification is taken up. Given an input data point $\rvx(t_0)$, it is transformed into a feature $\rvx(t_1)$ using a Neural ODE $f_\theta$ parameterized by $\theta$. Then, a classification loss function is applied on the feature. In general, any objective function $L$ can be applied on $\rvx(t_1)$. Then, the neural network is trained through backpropagation to update its parameters $\theta$ to minimize the loss $L$.

\begin{figure}[!hbt]
\begin{center}
\includegraphics[width=\linewidth]{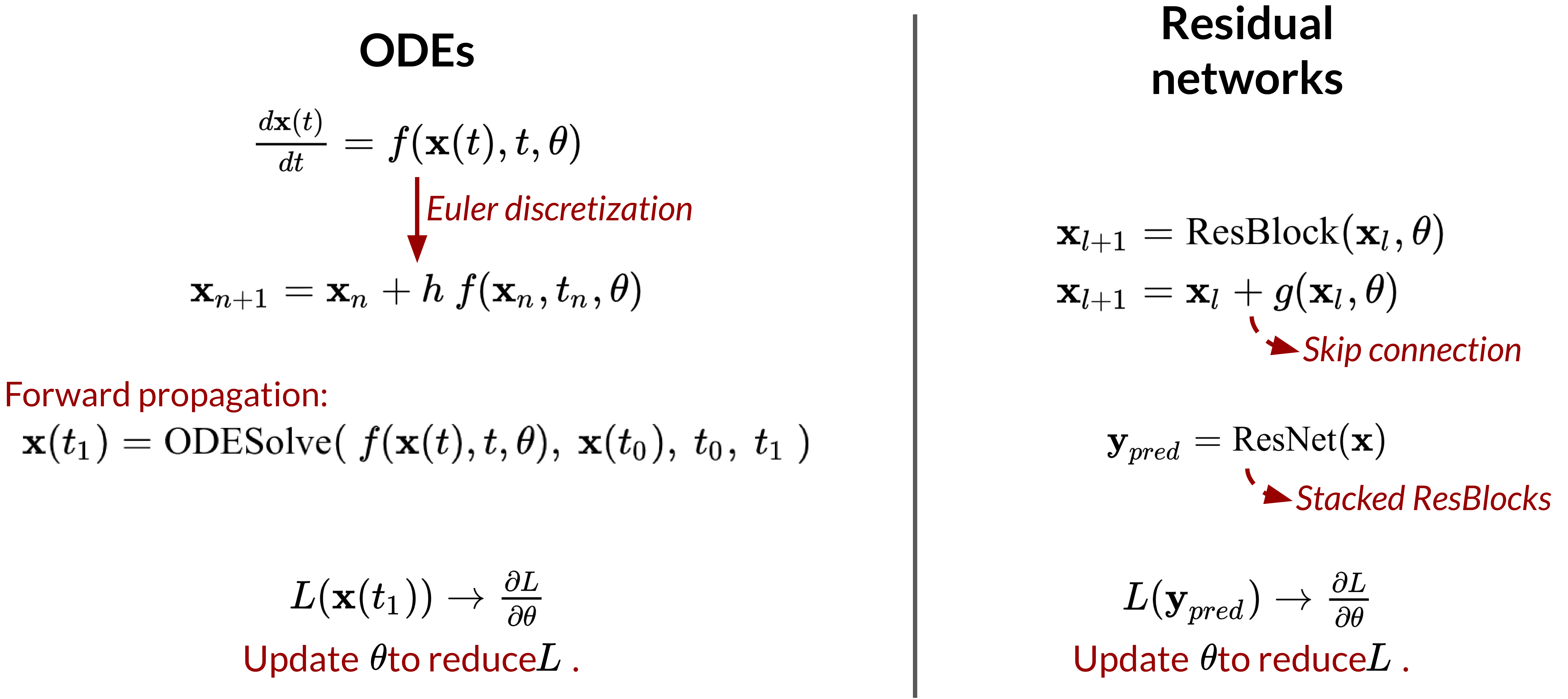}
\end{center}
\caption{Illustration of the solution to ODEs being flows.
}
\label{fig:resnet}
\end{figure}

There are parallels that could be drawn with a residual network, if the residual network shares its parameters across all layers. This is illustrated in \Cref{fig:resnet}.

\subsection{Adjoint method}
\label{subsec:adj}

However, computing the gradient of the loss $L$ with respect to the parameters $\theta$ i.e. $\partial L/\partial \theta$, incurs a high memory cost, since all activations of all iterations of ODESolve need to be stored in memory to complete backpropagation. This is sub-optimal, can we do better?

It turns out that there is a way to compute $\partial L/\partial \theta$ without having to save all activations from $t_0$ to $t_1$. This is made possible by what is called the ``adjoint method''~\citep{pontyagin1962mathematical}. The adjoint $\rva(t)$ of a state $\rvx(t)$ is defined as:

\begin{align}
\text{adjoint}\ \mathbf{a}(t) &= \frac{\partial L}{\partial \mathbf{x}} ,\\
 \frac{d\mathbf{a}}{dt} &= -\mathbf{\mathbf{a}}(t)^\top \frac{\partial f(\mathbf{x}(t),\ t,\ \theta)}{\partial \mathbf{x}}.
\end{align}
The adjoint $a(t_1)$ can be computed from the loss $L$ and the final state $\rvx(t_1)$ after one forward propagation pass through the ODESolve function:
\begin{align}
\mathbf{x}(t_1) &= \text{ODESolve}(\ f(\mathbf{x}(t), t, \theta),\ \mathbf{x}(t_0),\ t_0,\ t_1\ ) , \label{eq:x_forward} \\
\implies \mathbf{a}(t_1) &= \frac{\partial L}{\partial \mathbf{x}(t_1)} .\label{eq:adj_forward}
\end{align}
Then, \citet{pontyagin1962mathematical} showed that $\partial L/\partial \theta$ can be computed by solving another ODE involving the adjoint, in the reverse direction from $t_1$ to $t_0$:
\begin{align}
\frac{\partial L}{\partial \theta} = \int_{t_1}^{t_0} -\mathbf{a}(t)^\top \frac{\partial f(\mathbf{x}(t),\ t,\ \theta)}{\partial \theta}\ dt .
\label{eq:adj_loss}
\end{align}
This can be computed using our ODESolve function from $t_1$ to $t_0$, with the initial value as $\vzero$, and the differential as defined in \cref{eq:adj_loss}:
\begin{align}
\frac{\partial L}{\partial \theta} = \text{ODESolve}(-\mathbf{\mathbf{a}}(t)^\top \frac{\partial f(\mathbf{x}(t),\ t,\ \theta)}{\partial \theta},\ \ \ \mathbf{0}_{|\theta|}\ \ ,\ t_1,\ t_0) .
\end{align}
However, for this ODESolve to work, the values of $\rva(t)$ and $\rvx(t)$ at all intermediate steps of numerical integration are required. Hence, two other ODESolves are performed from $t_1$ to $t_0$: one to compute $\rva(t)$, and the other for $\rvx(t)$. The initial values for these ODESolves are $\rva(t_1)$ and $\rvx(t_1)$, which were computed in the forward pass in \Cref{eq:x_forward,eq:adj_forward}:
\begin{align}
\rvx(t_0) &= \text{ODESolve}(\quad\quad\ f(\mathbf{x}(t), t, \theta) \quad\quad,\ \mathbf{x}(t_1),\ t_1,\ t_0) . \\
\rva(t_0) &= \text{ODESolve}(-\mathbf{\mathbf{a}}(t)^\top \frac{\partial f(\mathbf{x}(t),\ t,\ \theta)}{\partial \mathbf{x}},\ \mathbf{a}(t_1),\ t_1,\ t_0) .
\end{align}
These three ODESolves can be combined into a single ODESolve:
\begin{align}
\begin{sqcases}
\rvx(t_0) \\
\rva(t_0) \\
\frac{\partial L}{\partial \theta}
\end{sqcases}
= \text{ODESolve}
\left(
\begin{sqcases}
f(\mathbf{x}(t), t, \theta) \\
-\mathbf{\mathbf{a}}(t)^\top \frac{\partial f(\mathbf{x}(t),\ t,\ \theta)}{\partial \mathbf{x}}\\
-\mathbf{\mathbf{a}}(t)^\top \frac{\partial f(\mathbf{x}(t),\ t,\ \theta)}{\partial \theta}
\end{sqcases}
,
\begin{sqcases}
\rvx(t_1) \\
\rva(t_1) \\
\mathbf{0}_{|\theta|}
\end{sqcases}
,
t_1,
t_0
\right).
\end{align}
Thus, at the end of ODESolve, $\partial L/\partial \theta$ is directly obtained in the third part of the augmented state. $\partial L/\partial \theta$ is then used to update $\theta$ using gradient descent, thus training the Neural ODE.

Because a Neural ODE ultimately describes an ODE, the fundamental theorem of ODEs applies to the solution $f$: Neural ODEs describe a homeomorphism/\textit{flow} i.e. they preserve dimensionality, and they form non-intersecting solution trajectories. Moreover, Neural ODEs are reversible architectures : the same ODE can be solved forwards or backwards.

\newpage
\section{Continuous Normalizing Flows (CNFs)}
\label{sec:bg_CNF}

Neural ODEs can then be used in a generative modeling framework by setting the final state to be a sample from a known distribution, typically a noise distribution such as the standard normal. Thus, a Neural ODE can then be trained to map between a data distribution and the normal distribution. Since a Neural ODE describes a (geometric) flow, and in this framework is used to map to the normal distribution, this framework is a \textit{normalizing flow}. To distinguish it from the usual normalizing flow in literature~\citep{dinh2016density}, considering the fact that it operates in continuous space while normalizing flows operate in discrete steps, this framework is called ``Continuous Normalizing Flow'' (CNF).

Suppose a CNF $g$ transforms its state $\rvv(t)$ using a Neural ODE, with the differential defined by the neural network $f$ parameterized by $\theta$. $\rvv(t_0)=\rvx$ is, say, an image, and at the final time step $\rvv(t_1)=\rvz$ is a sample from a known noise distribution.
\begin{align}
\frac{\D \rvv(t)}{\D t} &= f(\rvv(t), t, \theta), \\
\implies \rvv(t_1) &= g(\rvv(t_0)) = \rvv(t_0) + \int_{t_0}^{t_1} f(\rvv(t), t, \theta)\ \D t , \\
\implies \rvz &= g(\rvx) = \rvx + \int_{t_0}^{t_1} f(\rvv(t), t, \theta)\ \D t .
\label{eq:neural_ode1}
\end{align}
This integration is typically performed by an ODE solver, as shown in ~\Cref{eq:odesolve}. Since this integration can be run backwards as well to obtain the same $\rvv(t_0)$ from $\rvv(t_1)$, a \gls{cnf} is a reversible model.

CNFs can now be trained by maximizing the likelihood of real data points under the model. This is equivalent to transforming the real data point $\rvv(t_0)$ to the the final state $\rvv(t_1)$, and maximizing the likelihood of $\rvv(t_1)$ under the standard normal distribution. Given this maximum likelihood objective function, the neural network $f$ in the CNF can be optimized as described in \Cref{subsec:adj}.

The change-of-variables formula describes how the likelihood of the real data point $\rvx$ can be computed as a combination of the likelihood of the final point $\rvz$ and the change in the likelihood under the CNF transformation $g$:
\begin{align}
\label{eq:change_of_variables1}
    \log p(\rvx) = \log \Abs{\det\frac{\partial g}{\partial \rvv}} + \log p(\rvz) = \Delta\log p_{\rvv(t_0) \rightarrow \rvv(t_1)} + \log p(\rvz) .
\end{align}
The second term, $\log p(\rvz)$, is computed as the log probability of $\rvz$ under a known noise distribution, typically the standard normal $\mathcal{N}(\mathbf{0}, \mathbf{I})$. However, the log determinant of the Jacobian (first term on the right) is often intractable. Previous works on normalizing flows have found some ways to estimate this efficiently.

\citet{chen2018neural} and \citet{grathwohl2019ffjord} instead proposed a more efficient variant in the \gls{cnf} context, the instantaneous change-of-variables formula:
\begin{align}
&\frac{\partial \log p(\rvv(t))}{\partial t} = -\text{Tr}\left(\frac{\partial f_\theta}{\partial \rvv(t)}\right) ,
\\
\implies
&\Delta\log p_{\rvv(t_0) \rightarrow \rvv(t_1)} = \int_{t_0}^{t_1} -\text{Tr}\left(\frac{\partial f_\theta}{\partial \rvv(t)} \right) \D t .
\label{eq:inst1}
\end{align}
Here, ``Tr'' implies the trace operation i.e. the sum of the diagonal elements in the matrix.

Hence, the change in log-probability of the state of the Neural ODE i.e. $\Delta\log p_{\rvv(t_0) \rightarrow \rvv(t_1)}$ is expressed as another differential equation. The ODE solver now solves both differential equations \Cref{eq:neural_ode1,eq:inst1} by augmenting the original state. Thus, a \gls{cnf} forward pass provides both the final state $\rvv(t_1)$ as well as the change in log probability $\Delta\log p_{\rvv(t_0) \rightarrow \rvv(t_1)}$ together:
\begin{align}
\begin{sqcases}
\rvz \\
\Delta\log p_{\rvv(t_0) \rightarrow \rvv(t_1)}
\end{sqcases}
= \text{ODESolve}
\left(
\begin{sqcases}
f(\mathbf{v}(t), t, \theta) \\
-\text{Tr}\left(\frac{\partial f_\theta}{\partial \rvv(t)} \right)
\end{sqcases}
,
\begin{sqcases}
\rvx\ \  \\
\mathbf{0}\ \ 
\end{sqcases}
,
t_0,
t_1
\right) .
\end{align}
Thus, having estimated $\rvz$ and $\Delta\log p_{\rvv(t_0) \rightarrow \rvv(t_1)}$, $\log p(\rvx)$ can be computed using \Cref{eq:change_of_variables1}. Taking the objective function to be maximizing $\log p(\rvx)$ i.e. maximizing the likelihood of real data under the model, the CNF can be trained as described in \Cref{subsec:adj}.

%% file: DDPM_SMPLD.tex
\newpage
\section{Denoising Diffusion Probabilistic Models (DDPM)}
\label{sec:ddpm}

The formulations related to DDPM~\citep{ho2020ddpm} are introduced here. \Cref{chap:NIDDPM} builds on this and derives a formulation with non-isotropic DDPM. \Cref{chap:MCVD} applies DDPM to video prediction, generation and interpolation.

DDPM focuses on modeling the diffusion process of noisy samples to approximate the clean data distribution. In its practical implementation, DDPM uses a neural network to estimate the noise to be subtracted from the noisy data to make it cleaner. At training time, a noise sample is added to a clean data sample, and the neural network is trained to predict the noise sample from the noisy data. At sample time, a noise sample is iteratively cleaned little by little, by estimating the noise to be subtracted at each step.

\subsection{Forward (data to noise) for DDPM}

In DDPM, for a fixed sequence of positive scales $0 < \beta_1 < \cdots < \beta_L < 1$, $\rvepsilon_t \sim \gN(\vzero, \rmI)$ is a noise sample from a standard normal distribution $\gN$ with zero mean $\vzero$ and identity covariance matrix $\rmI$, and $\rvx_0$ is a clean data point, the \textbf{transition ``forward''} noising process is:
\begin{align}
&p_{\beta_t}^{\DDPM} (\rvx_t \mid \rvx_{t-1}) = \gN(\rvx_t \mid \sqrt{1 - \beta_t} \rvx_{t-1}, \beta_t \rmI), \\
\implies
&\rvx_t = \sqrt{1 - \beta_t}\rvx_{t-1} + \sqrt{\beta_t} \rvepsilon_t
\label{eq:DDPM_forward_eq}
\end{align}
Then, the \textbf{cumulative ``forward''} noising process can be derived as:
\begin{align}
&p_{t}^{\DDPM} (\rvx_t \mid \rvx_0) = p_{\beta_t}^{\DDPM} (\rvx_t \mid \rvx_{t-1})\ p_{\beta_t}^{\DDPM} (\rvx_{t-1} \mid \rvx_{t-2})\ \cdots\ p_{\beta_t}^{\DDPM} (\rvx_1 \mid \rvx_0).
\end{align}
Using $\balpha_t = \prod_{s=1}^t (1 - \beta_s)$, the cumulative ``forward'' noising process can be simplified to:
\begin{align}
p_t^{\DDPM} (\rvx_t \mid \rvx_0) &= \gN(\rvx_t \mid \sqrt{\balpha_t} \rvx_0, (1 - \balpha_t) \rmI), \label{eq:ddpm_forw} \\
\implies \rvx_t &= \sqrt{\balpha_t}\rvx_0 + \sqrt{1 - \balpha_t} \rvepsilon
, \text{ where } \rvepsilon \sim \gN(\vzero, \rmI), \label{eq:DDPM_noise}
\\
\implies \rvepsilon &= \frac{\rvx_t - \sqrt{\balpha_t}\rvx_0}{\sqrt{1 - \balpha_t}} .
\end{align}

\subsection{Score for DDPM}

Then, the \textbf{score} i.e. $\nabla_{\rvx_t} \log p_{t} (\rvx_t \mid \rvx_0)$ can be calculated as:
\begin{align}
&\log p_t^{\DDPM} (\rvx_t \mid \rvx_0) = \log \text{(const)} - \frac{1}{2(1 - \balpha_t)}(\rvx_t - \sqrt{\balpha_t}\rvx_0)^\trns(\rvx_t - \sqrt{\balpha_t}\rvx_0) ,
\\
\implies &\text{Score }\ \rvs = \nabla_{\rvx_t} \log p_{t}^{\DDPM} (\rvx_t \mid \rvx_0) = -\frac{1}{(1 - \balpha_t)}(\rvx_t - \sqrt{\balpha_t} \rvx_0) 
\label{eq:score_DDPM_x} , \\
&\qquad\ \quad = -\frac{1}{\sqrt{1 - \balpha_t}}\left[\frac{\rvx_t - \sqrt{\balpha_t}\rvx_0}{\sqrt{1 - \balpha_t}}\right] = -\frac{1}{\sqrt{1 - \balpha_t}}\rvepsilon .
\label{eq:score_DDPM}
\end{align}

\subsection{Score-matching objective function for DDPM}

The score-matching objective for DDPM at noise level $t$ is the expected Mean Square Error (MSE) between the true score in \cref{eq:score_DDPM_x}, and the predicted score from a neural network $\rvs_\rvtheta$:
\begin{align}
&\loss^\DDPM(\rvtheta; \balpha_t) \triangleq\ \frac{1}{2} \EE_{p_{t}(\rvx_t \mid \rvx_0) p(\rvx_0)} \bigg[ \Norm{\rvs_\rvtheta (\rvx_t, \balpha_t) + \frac{1}{(1 - \balpha_t)}(\rvx_t - \sqrt{\balpha_t} \rvx_0) }_2^2 \bigg] .
\end{align}
$\rvs_\rvtheta$ predicts the score from the noisy image $\rvx_t$, and the noise level $\balpha_t$ (or just the time step $t$).

The overall loss is the weighted sum of the losses at each step:
\begin{align}
& \gL^\DDPM(\rvtheta; \{\balpha_t\}_{t=1}^{L}) \triangleq \EE_t\ \lambda(\balpha_t) \ \loss(\rvtheta; \balpha_t) .
\label{eq:DDPM_loss}
\end{align}
To let the loss have equal weight across all noise levels, the weight $\lambda(\balpha_t)$ is the inverse of the variance of the true score at that noise level.

\subsection{Variance of score for DDPM}
\begin{align*}
&\EE\left[ \Norm{ \nabla_{\rvx_t} \log p_{t}^\DDPM (\rvx_t \mid \rvx_0) }_2^2 \right] 
= \EE\left[ \Norm{ -\frac{1}{(1 - \balpha_t)}(\rvx_t - \sqrt{\balpha_t} \rvx_0) }_2^2 \right] , \\
&= \EE\left[ \Norm{ \frac{\sqrt{1 - \balpha_t}\rvepsilon}{(1 - \balpha_t)} }_2^2 \right]
= \frac{1}{1 - \balpha_t}\EE\left[ \Norm{\rvepsilon}_2^2 \right] = \frac{1}{1 - \balpha_t} .
\numberthis
\label{eq:DDPM_variance}
\end{align*}

\subsection{Overall objective function for DDPM}

The overall objective function in \cite{ho2020ddpm} used the inverse of the variance of the score at each time step (from \cref{eq:DDPM_variance}) as the weight $\lambda(\balpha_t)$:
\begin{align}
&\lambda^\DDPM(\balpha_t) \propto 1/\EE\big[ \Norm{ \nabla_{\rvx_t} \log p_{t}^\DDPM (\rvx_t \mid \rvx_0) }_2^2 \big] ,
\\
\implies &\lambda^\DDPM(\balpha_t) = 1 - \balpha_t .
\end{align}
Then, the overall objective in \cref{eq:DDPM_loss} changes to:
\begin{align}
\gL^\DDPM(\rvtheta; \{\balpha_t\}_{t=1}^{L}) &\triangleq  \EE_{t, p_{t}(\rvx_t \mid \rvx_0) p(\rvx_0)} \bigg[ \Norm{\sqrt{1 - \balpha_t} \rvs_\rvtheta (\rvx_t, \balpha_t) + \frac{(\rvx_t - \sqrt{\balpha_t} \rvx_0)}{\sqrt{1 - \balpha_t}} }_2^2 \bigg] , \\
&= \EE_{t, p_{t}(\rvx_t \mid \rvx_0) p(\rvx_0)} \bigg[ \Norm{\sqrt{1 - \balpha_t} \rvs_\rvtheta (\rvx_t, \balpha_t) + \rvepsilon }_2^2 \bigg] .
\numberthis
\end{align}

\subsection{Noise-matching objective for DDPM}

Upon inspection of \cref{eq:score_DDPM}, one can recognize that the score is a $-1/\sqrt{1 - \balpha_t}$ factor of $\rvepsilon$, hence only $\rvepsilon$ needs to be estimated:
\begin{align}
\rvs_\rvtheta(\rvx_t, \balpha_t) = -\frac{1}{\sqrt{1 - \balpha_t}} \rvepsilon_\rvtheta(\rvx_t, \balpha_t) .
\end{align}

In this case, the overall objective function changes to the noise-matching objective:
\begin{align*}
\gL^\DDPM(\rvtheta; \{\balpha_t\}_{t=1}^{L}) 
&\triangleq \EE_{t, \rvepsilon, \rvx_0} \left[ \Norm{-\rvepsilon_\rvtheta (\rvx_t, \balpha_t) + \rvepsilon }_2^2 \right] , \\
&= \EE_{t, \rvepsilon, \rvx_0} \left[ \Norm{ \rvepsilon - \rvepsilon_\rvtheta (\sqrt{\balpha_t}\rvx_0 + \sqrt{1 - \balpha_t}\rvepsilon, \balpha_t)}_2^2 \right] .
\numberthis
\label{eq:DDPM_final_loss}
\end{align*}

This \cref{eq:DDPM_final_loss} is Equation 14 in the DDPM paper~\citep{ho2020ddpm}. The DDPM paper~\citep{ho2020ddpm} retains conditioning of $\rvepsilon_\rvtheta$ on $\balpha_t$ (or just $t$), but the SMLD paper~\citep{song2019generative} omits it.

\subsection{Reverse (noise to data) in DDPM}
The goal is to estimate the \textbf{reverse} transition probability $q_t^\DDPM(\rvx_{t-1} \mid \rvx_t)$. However, this is intractable to compute, but it is possible to estimate it conditioned on $\rvx_0$, using Bayes' theorem:
\begin{align}
q_t^\DDPM(\rvx_{t-1} \mid \rvx_t, \rvx_0) &= \frac{q_t^\DDPM(\rvx_{t} \mid \rvx_{t-1}) \quad q_t^\DDPM(\rvx_{t-1} \mid \rvx_0)}{q_t^\DDPM(\rvx_{t} \mid \rvx_0)} ,
\\
&= \frac{\gN(\rvx_t \mid \sqrt{1 - \beta_t} \rvx_{t-1}, \beta_t \rmI) \quad \gN(\rvx_{t-1} \mid \sqrt{\balpha_{t-1}} \rvx_0, (1 - \balpha_{t-1}) \rmI)}{\gN(\rvx_t \mid \sqrt{\balpha_t} \rvx_0, (1 - \balpha_t) \rmI)} .
\end{align}
This can be simplified to:
\begin{align*}
\implies &q_t^\DDPM(\rvx_{t-1} \mid \rvx_t, \rvx_0) = \gN(\rvx_{t-1} \mid \tilde\vmu_{t-1}(\rvx_t, \rvx_0), \tilde\beta_{t-1}\rmI) \text{, where} \\
&\tilde\vmu_{t-1}(\rvx_t, \rvx_0) = \frac{\sqrt{\balpha_{t-1}}\beta_t}{1 - \balpha_t}\rvx_0 + \frac{\sqrt{1 - \beta_t}(1 - \balpha_{t-1})}{1 - \balpha_t}\rvx_t ;\quad \tilde\beta_{t-1} = \frac{1 - \balpha_{t-1}}{1 - \balpha_t}\beta_t .
\numberthis
\label{eq:q_tminus1}
\\
\implies &\rvx_{t-1} = \tilde\vmu_{t-1}(\rvx_t, \rvx_0) + \tilde\beta_{t-1} \rvz \quad \text{where } \rvz \sim \gN(\vzero, \rmI) \text{ is a noise sample.}
\numberthis
\label{eq:NIDDPM_sample}
\end{align*}

Given $\rvx_t$, \cref{eq:DDPM_noise} is used to estimate $\rvx_0$ by first estimating noise $\rvepsilon$ using a \textbf{neural network} $\rvepsilon_\rvtheta(\rvx_t, t)$:
\begin{gather*}
\hat\rvx_0 = \frac{1}{\sqrt{\balpha_t}}(\rvx_t - \sqrt{1 - \balpha_t}\rvepsilon_{\rvtheta}(\rvx_t, t)) .
\numberthis
\label{eq:hatx0}
\\ 
[\because \rvx_t = \sqrt{\balpha_t}\rvx_0 + \sqrt{1 - \balpha_t} \rvepsilon \text{ from \cref{eq:DDPM_noise}, and loss is minimized when } \rvepsilon_{\rvtheta^*}(\rvx_t) = \rvepsilon.]
\end{gather*}
Hence, using $\hat\rvx_0$ estimated from $\rvx_t$ using \cref{eq:hatx0}, $\rvx_{t-1}$ is computed from \cref{eq:NIDDPM_sample} as:
\begin{gather*}
\rvx_{t-1} = \left[\frac{\sqrt{\balpha_{t-1}}\beta_t}{1 - \balpha_t}\hat\rvx_0 + \frac{\sqrt{1 - \beta_t}(1 - \balpha_{t-1})}{1 - \balpha_t}\rvx_t \right] + \sqrt{\tilde\beta_{t-1}} \rvz .
\numberthis
\label{eq:DDPM_x_tminus1}
\end{gather*}
where $\rvz \sim \gN(\vzero, \rmI)$ is a noise sample.

\subsection{Sampling in DDPM}


\cite{ho2020ddpm} splits sampling at each time step into 2 steps:
\begin{align*}
&\text{\textit{Step 1 (from \cref{eq:hatx0}):  }} \hat\rvx_0 = \frac{1}{\sqrt{\balpha_t}}(\rvx_t - \sqrt{1 - \balpha_t}\rvepsilon_{\rvtheta^*}(\rvx_t, \balpha_t)) .
\numberthis
\\
&\text{\textit{Step 2 (from \cref{eq:DDPM_x_tminus1}):  }} \rvx_{t-1} = \frac{\sqrt{\balpha_{t-1}}\beta_t}{1 - \balpha_t}\hat\rvx_0 + \frac{\sqrt{1 - \beta_t}(1 - \balpha_{t-1})}{1 - \balpha_t}\rvx_t + \sqrt{\tilde\beta_{t-1}} \rvz_{t-1} .
\numberthis \label{eq:ddpm_step2}
\end{align*}
This can be simplified as:
\begin{align*}
\implies \tilde\vmu_{t-1}(\rvx_t, \hat\rvx_0) &= \frac{\sqrt{\balpha_{t-1}}\beta_t}{1 - \balpha_t}\left(\frac{1}{\sqrt{\balpha_t}}\big(\rvx_t - \sqrt{1 - \balpha_t}\rvepsilon_{\rvtheta^*}(\rvx_t)\big)\right) + \frac{\sqrt{1 - \beta_t}(1 - \balpha_{t-1})}{1 - \balpha_t}\rvx_t , \\
&\quad = \frac{\sqrt{\balpha_{t-1}}}{\sqrt{\balpha_t}} \frac{\beta_t}{1 - \balpha_t}\rvx_t - \frac{\sqrt{\balpha_{t-1}}}{\sqrt{\balpha_t}} \frac{\beta_t}{\sqrt{1 - \balpha_t}} \rvepsilon_{\rvtheta^*}(\rvx_t) + \frac{\sqrt{1 - \beta_t}}{1 - \balpha_t}\big(1 - \balpha_{t-1}\big)\rvx_t , \\
&\quad = \frac{1}{\sqrt{1 - \beta_t}} \frac{\beta_t}{1 - \balpha_t}\rvx_t + \frac{\sqrt{1 - \beta_t}}{1 - \balpha_t}\left(1 - \frac{\balpha_t}{1 - \beta_t}\right)\rvx_t - \frac{1}{\sqrt{1 - \beta_t}} \frac{\beta_t}{\sqrt{1 - \balpha_t}} \rvepsilon_{\rvtheta^*}(\rvx_t) , \\
&\quad = \frac{1}{\sqrt{1 - \beta_t}}\left[ \frac{\beta_t}{1 - \balpha_t}\rvx_t + \frac{(1 - \beta_t)}{1 - \balpha_t}\left(1 - \frac{\balpha_t}{1 - \beta_t}\right)\rvx_t - \frac{\beta_t}{\sqrt{1 - \balpha_t}} \rvepsilon_{\rvtheta^*}(\rvx_t) \right] , \\
&\quad = \frac{1}{\sqrt{1 - \beta_t}}\left[ \frac{\beta_t + 1 - \beta_t - \balpha_t}{1 - \balpha_t}\rvx_t - \frac{\beta_t}{\sqrt{1 - \balpha_t}} \rvepsilon_{\rvtheta^*}(\rvx_t) \right] , \\
&\quad = \frac{1}{\sqrt{1 - \beta_t}}\left( \rvx_t - \frac{\beta_t}{\sqrt{1 - \balpha_t}} \rvepsilon_{\rvtheta^*}(\rvx_t) \right) . \\
\implies \rvx_{t-1}
&= \frac{1}{\sqrt{1 - \beta_t}}\left(\rvx_t - \frac{\beta_t}{\sqrt{1 - \balpha_t}} \rvepsilon_{\rvtheta^*}(\rvx_t)\right) + \sqrt{\tilde\beta_{t-1}}\  \rvz_{t-1} , 
\numberthis
\\
&= \frac{1}{\sqrt{1 - \beta_t}}\left(\rvx_t + \beta_t \rvs_{\rvtheta^*}(\rvx_t, \balpha_t)\right) + \sqrt{\tilde\beta_{t-1}}\ \rvz_{t-1} .
\numberthis
\end{align*}

However, an alternative sampler mentioned in \cite{song2020sde} contains $\beta_{t-1}$ instead of $\tilde\beta_{t-1}$:
\begin{align}
\rvx_{t-1}
&= \frac{1}{\sqrt{1 - \beta_t}}\left(\rvx_t - \frac{\beta_t}{\sqrt{1 - \balpha_t}} \rvepsilon_{\rvtheta^*}(\rvx_t)\right) + \sqrt{\beta_{t-1}} \rvz_{t-1} ,
\\
&= \frac{1}{\sqrt{1 - \beta_t}}\left(\rvx_t + \beta_t \rvs_{\rvtheta^*}(\rvx_t, \balpha_t)\right) + \sqrt{\beta_{t-1}} \rvz_{t-1} .
\end{align}

Another alternative sampling technique called Denoising Diffusion Implicit Models (DDIM) was introduced by \citet{song2020ddim}, as discussed below.

\subsection{Sampling using DDIM}

DDIM~\citep{song2020ddim} replaces \textit{Step 2} in \cref{eq:ddpm_step2} with the DDPM forward process \cref{eq:DDPM_noise}:
\begin{align*}
\text{\textit{Step 1: }} \hat\rvx_0 &= \frac{1}{\sqrt{\balpha_t}}(\rvx_t - \sqrt{1 - \balpha_t}\rvepsilon_{\rvtheta^*}(\rvx_t)) .
\numberthis\\
\text{\textit{Step 2: }} \rvx_{t-1} &= \sqrt{\balpha_{t-1}} \hat\rvx_0 + \sqrt{1 - \balpha_{t-1}} \rvepsilon_{\rvtheta^*}(\rvx_t) .
\numberthis
\label{eq:DDIM_step2}
\end{align*}

This is derived from the following distributions from \cite{song2020ddim}:
\begin{align*}
p_{L}^\DDIM(\rvx_L \mid \rvx_0) &= \gN(\rvx_L \mid \sqrt{\balpha_L}\rvx_0, (1 - \balpha_L)\rmI) ,
\numberthis
\\
q_{t-1}^\DDIM(\rvx_{t-1} \mid \rvx_t, \rvx_0) &= \gN\left(\rvx_{t-1} \mid \sqrt{\balpha_{t-1}}\rvx_0 + \sqrt{1 - \balpha_{t-1}}\frac{\rvx_t - \sqrt{\balpha_t}\rvx_0}{\sqrt{1 - \balpha_t}}, \vzero \right) ,
\numberthis
\\
\implies p_{t}^\DDIM(\rvx_{t} \mid \rvx_0) &= \gN\left(\rvx_{t} \mid \sqrt{\balpha_{t}}\rvx_0, (1 - \balpha_{t})\rmI \right) .
\numberthis
\end{align*}
\textbf{Proof by induction:}
From 2.115 in \cite{bishop2006pattern}:

For a random variable $\rvu$ distributed as a normal with mean $\vmu$ and covariance matrix $\mLambda^{-1}$, and a dependent variable $\rvv$ conditionally distributed as a normal with mean $\rmA\rvu + \rvb$ and covariance matrix $\rmL^{-1}$:
\begin{align}
p(\rvu) &= \gN(\rvu \mid \vmu, \mLambda^{-1}),
\\
p(\rvv \mid \rvu) &= \gN(\rvv \mid \rmA\rvu + \rvb, \rmL^{-1}) ,
\end{align}
the marginal probability of $\rvv$ is distributed as:
\begin{align}
\implies p(\rvv) &= \gN(\rvv \mid \rmA\vmu + \rvb, \rmL^{-1} + \rmA \mLambda^{-1} \rmA^\trns) . \label{eq:marginal}
\end{align}
In the case of DDPM, considering $p(\rvu) = p_t^{\DDPM}(\rvx_t \mid \rvx_0)$, and $p(\rvv \mid \rvu) = q_{t-1}^{\DDPM}(\rvx_{t-1} \mid \rvx_0)$:
\begin{align*}
p_{t}^{\DDPM}(\rvx_t \mid \rvx_0) &= \gN(\rvx_t \mid \sqrt{\balpha_t}\rvx_0, (1 - \balpha_t)\rmI) \text{ from \cref{eq:ddpm_forw}} \text{, and} \\
q_{t-1}^{\DDPM}(\rvx_{t-1} \mid \rvx_t, \rvx_0) &= \gN\left( \rvx_{t-1} \mid \sqrt{\balpha_{t-1}}\rvx_0 + \sqrt{1 - \balpha_{t-1}}\frac{\rvx_t - \sqrt{\balpha_t}\rvx_0}{\sqrt{1 - \balpha_t}}, \vzero \right) \text{ from \cref{eq:q_tminus1}},
\end{align*}
then the marginal $p(\rvv) = q_{t-1}^{\DDPM}(\rvx_{t-1} \mid \rvx_0)$ is computed using \cref{eq:marginal} as:
\begin{align*}
\implies q_{t-1}^{\DDPM}(\rvx_{t-1} \mid \rvx_0) &= \gN\bigg( \rvx_{t-1} \mid \sqrt{\balpha_{t-1}}\rvx_0 + \sqrt{1 - \balpha_{t-1}}\frac{\sqrt{\balpha_t}\rvx_0 - \sqrt{\balpha_t}\rvx_0}{\sqrt{1 - \balpha_t}},\\
&\qquad\qquad\qquad \vzero + \frac{1 - \balpha_{t-1}}{1 - \balpha_{t}}(1 - \balpha_t)\rmI \bigg) , \numberthis\\
&= \gN\left( \rvx_{t-1} \mid \sqrt{\balpha_{t-1}}\rvx_0, (1 - \balpha_{t-1})\rmI \right) . \numberthis
\end{align*}
Hence $\rvx_{t-1} = \sqrt{\balpha_{t-1}} \hat\rvx_0 + \sqrt{1 - \balpha_{t-1}} \rvepsilon_{\rvtheta^*}(\rvx_t)$ in \cref{eq:DDIM_step2}.

\subsection{Expected Denoised Sample (EDS) for DDPM}

From \cite{saremi2019neb}, for isotropic Gaussian noise, we know that the expected denoised sample $\rvx_0^*(\rvx_t, \balpha_t) \triangleq \EE_{\rvx_0 \sim q_{t}(\rvx_0 \mid \rvx_t)}[\rvx_0]$ and the optimal score $\rvs_{\rvtheta^*}(\rvx_t, \balpha_t)$ are related as:
\begin{align}
&\rvs_{\rvtheta^*}(\rvx_t, \balpha_t) = \EE\left[ \Norm{ \nabla_{\rvx_t} \log p_{t} (\rvx_t \mid \rvx_0) }_2^2 \right] (\rvx_0^* (\rvx_t, \balpha_t) - \rvx_t) .
\label{eq:EDS_defn}
\end{align}
For DDPM, using \cref{eq:DDPM_variance}, this simplifies to:
\begin{align}
&\rvs_{\rvtheta^*}(\rvx_t, \balpha_t) = \frac{1}{1 - \balpha_t}\left(\rvx_0^*(\rvx_t, \balpha_t) - \rvx_t\right) , \\
\implies &\rvx_0^*(\rvx_t, \balpha_t) = \rvx_t + (1 - \balpha_t)\ \rvs_{\rvtheta^*}(\rvx_t, \balpha_t) = \rvx_t - \sqrt{1 - \balpha_t} \ \rvepsilon_{\rvtheta^*}(\rvx_t, \balpha_t) .
\end{align}

While so far the noising/denoising processes have been considered discrete, the following section takes up the continuous formulation of DDPM using Stochastic Differential Equations (SDEs).

\subsection{SDE formulation : Variance Preserving (VP) SDE}

In the continuous formulation, the discrete $\beta_t$ is now a predefined continuous $\beta(t)$. For DDPM i.e. Variance Preserving (VP) SDE, given $\rvw$ is a Weiner process i.e. standard Brownian motion, the \textbf{forward equation} and \textbf{transition probability} are (derived below):
\begin{align}
\D \rvx &= -\frac{1}{2}\beta(t)\rvx\ \D t + \sqrt{\beta(t)}\ \D \rvw ,
\\
p_{0t}^\VP(\rvx(t) \mid \rvx(0)) &= \gN \left(\rvx(t) \mid \rvx(0)\ e^{-\frac{1}{2}\int_0^t \beta(s) \D s}, \rmI - \rmI e^{-\int_0^t \beta(s) \D s} \right) .
\label{eq:VPSDE_forward}
\end{align}



\subsubsection{Derivations}:
\label{subsubsec:ddpm_sde_deriv}

\textbf{Forward process}: We know from \cref{eq:DDPM_forward_eq} that:
\begin{align}
\rvx_t &= \sqrt{1 - \beta_t}\rvx_{t-1} + \sqrt{\beta_t} \rvepsilon_{t-1} . \\
\implies \rvx(t + \Delta t) &= \sqrt{1 - \beta(t + \Delta t) \Delta t}\ \rvx(t) + \sqrt{\beta(t + \Delta t) \Delta t}\ \rvepsilon(t) , \\
&\approx \left(1 - \frac{1}{2}\beta(t + \Delta t)\Delta t\right) \rvx(t) + \sqrt{\beta(t + \Delta t) \Delta t}\ \rvepsilon(t) ,\\
&\approx \rvx(t) - \frac{1}{2}\beta(t)\Delta t\ \rvx(t) + \sqrt{\beta(t) \Delta t}\ \rvepsilon(t) ,\\
\implies \D \rvx &= -\frac{1}{2}\beta(t)\rvx\ \D t + \sqrt{\beta(t)}\ \D \rvw .
\numberthis
\end{align}

We know from eq. 5.50 and 5.51 in \cite{sarkka2019applied} that, given that a random variable $\rvx$ follows a stochastic process with drift coefficient $\rvf(\rvx, t)$ and diffusion coefficient $\rmG(\rvx, t)$:
\begin{align}
&\D \rvx = \rvf(\rvx, t) \D t + \rmG(\rvx, t) \D \rvw ,
\end{align}
the mean $\vmu$ and covariance $\mSigma_{\text{cov}}$ follow the following differential processes:
\begin{align}
&\frac{\D \vmu}{\D t} = \EE_\rvx[\rvf(\rvx, t)], \label{eq:VPSDE_mean} \\
&\frac{\D \mSigma_{\text{cov}}}{\D t} = \EE_\rvx[\rvf(\rvx, t) (\rvx - \vmu)^\trns] + \EE_\rvx[(\rvx - \vmu) \rvf(\rvx, t)^\trns] + \EE_\rvx[\rmG(\rvx, t) \rmQ \rmG^\trns(\rvx, t)], \label{eq:VPSDE_cov}
\end{align}
where $\rvw$ is Brownian motion, $\rmQ$ is the PSD of $\rvw$. For Gaussian noise, $\rmQ = \rmI$.

\vspace{1em}

Hence, \textbf{Mean} (from eq. 5.50 in \cite{sarkka2019applied} i.e. \cref{eq:VPSDE_mean} above):
\begin{align*}
&\D \rvx = \rvf\ \D t + \rmG\ \D \rvw
\implies \frac{\D \vmu}{\D t} = \EE_\rvx[\rvf] .
\numberthis
\end{align*}
For DDPM, $\rvf = -\frac{1}{2}\beta(t)\rvx$.
\begin{align}
&\therefore \frac{\D \vmu_{\DDPM}(t)}{\D t} = \EE_\rvx[-\frac{1}{2}\beta(t)\rvx] = -\frac{1}{2}\beta(t)\EE_\rvx(\rvx) = -\frac{1}{2}\beta(t)\vmu_{\DDPM}(t) , \\
&\implies \frac{\D \vmu_{\DDPM}(t)}{\vmu_{\DDPM}(t)} = -\frac{1}{2}\beta(t) \D t \implies \log \vmu_{\DDPM}(t) \vert_{0}^{t} = -\frac{1}{2}\int_0^t \beta(s) \D s , \\
&\implies \log \frac{\vmu_{\DDPM}(t)}{\vmu(0)} = -\frac{1}{2}\int_0^t \beta(s) \D s ,\\
&\implies \vmu_{\DDPM}(t) = \vmu(0)\ e^{-\frac{1}{2}\int_0^t \beta(s) \D s} .
\end{align}

\vspace{1.5em}

\textbf{Covariance} (from eq. 5.51 in \cite{sarkka2019applied} i.e. \cref{eq:VPSDE_cov} above):
\begin{align*}
&\D \rvx = \rvf\ \D t + \rmG\ \D \rvw , \\
\implies 
&\frac{\D \mSigma_{\text{cov}}}{\D t} = \EE_\rvx[\rvf (\rvx - \vmu)^\trns] + \EE_\rvx[(\rvx - \vmu) \rvf^\trns] + \EE_\rvx[\rmG\rmG^\trns] .
\numberthis
\end{align*}
For DDPM, $\rvf = -\frac{1}{2}\beta(t)\rvx, \vmu = \vzero, \rmG = \sqrt{\beta(t)}\rmI$.
\begin{align}
&\therefore \frac{\D \mSigma_{\DDPM}(t)}{\D t} = \EE_\rvx[-\frac{1}{2}\beta(t)\rvx\rvx^\trns] + \EE_\rvx[\rvx(-\frac{1}{2}\beta(t)\rvx)^\trns] + \EE_\rvx[\sqrt{\beta(t)}\rmI\sqrt{\beta(t)}\rmI] , \\
&\qquad\qquad\qquad = -\beta(t)\mSigma_{\DDPM}(t) + \beta(t)\rmI = \beta(t)(\rmI - \mSigma_{\DDPM}(t)) , \\
&\implies \frac{\D \mSigma_{\DDPM}(t)}{\rmI - \mSigma_{\DDPM}(t)} = \beta(t) \D t \implies -\log(\rmI - \mSigma_{\DDPM}(t)) \vert_{0}^{t} = \int_0^t \beta(s) \D s , \\
&\implies -\log(\rmI - \mSigma_{\DDPM}(t)) + \log(\rmI - \mSigma_{\rvx}(0)) = \int_0^t \beta(s) \D s , \\
&\implies \frac{\rmI - \mSigma_{\DDPM}(t)}{\rmI - \mSigma_{\rvx}(0)} = e^{-\int_0^t \beta(s) \D s} \implies \mSigma_{\DDPM}(t) = \rmI - e^{-\int_0^t \beta(s) \D s} (\rmI - \mSigma_{\rvx}(0)) , \\
&\implies \mSigma_{\DDPM}(t) = \rmI + e^{-\int_0^t \beta(s) \D s} (\mSigma_{\rvx}(0) - \rmI) .
\end{align}

\vspace{1.5em}

For each data point $\rvx(0)$, $\vmu(0) = \rvx(0)$, $\mSigma_{\rvx}(0) = \vzero$:
\begin{align}
&\implies \vmu_{\DDPM}(t) = \rvx(0)\ e^{-\frac{1}{2}\int_0^t \beta(s) \D s}, \\
&\qquad \quad \mSigma_{\DDPM}(t) = \rmI + e^{-\int_0^t \beta(s) \D s} (\vzero - \rmI) = \rmI - \rmI e^{-\int_0^t \beta(s) \D s} .
\end{align}
$\therefore \DDPM \text{ i.e. } p_{0t}^\VP(\rvx(t) \mid \rvx(0)) = \gN \left(\rvx(t) \mid \rvx(0)\ e^{-\frac{1}{2}\int_0^t \beta(s) \D s}, \rmI - \rmI e^{-\int_0^t \beta(s) \D s} \right)$ in \cref{eq:VPSDE_forward}.

\vspace{1.5em}

\textbf{Calculating} $\int_0^t \beta(s) \D s$ using a linear beta schedule:
\begin{align}
&\beta(t) = \beta_{\min} + t (\beta_{\max} - \beta_{\min})
\implies \int_0^t \beta(s) \D s = t \beta_{\min} + \frac{t^2}{2}(\beta_{\max} - \beta_{\min}) .
\end{align}

\newpage
\section{Score Matching Langevin Dynamics (SMLD) }
\label{sec:smld}

The formulations related to SMLD~\citep{song2019generative, song2020improved} are introduced here. \Cref{chap:NIDDPM} builds on this and derives a formulation for non-isotropic SMLD.

While DDPM and SMLD both provide excellent generative sample quality, there are key differences in their approaches. In the following sections, the same mathematical derivations as for DDPM are repeated for SMLD so that the differences are made clear. DDPM focuses on modeling the diffusion process of noisy samples to approximate the clean data distribution. In contrast, SMLD directly estimates the gradient from noisy to clean samples i.e. the score function, and employs Langevin dynamics to traverse from noise to data. While DDPM defines a variance preserving SDE, SMLD defines a variance exploding SDE, as will be detailed below.

In their practical implementations, DDPM and SMLD employ different approaches to achieve their objectives. DDPM uses a neural network to estimate the noise to be subtracted from the noisy data to make it cleaner. SMLD uses a neural network to estimate the score i.e. the gradient from noisy to cleaner data. As shall be seen, the score and noise are inter-related, so training to predict one is equivalent to predicting the other.

\subsection{Forward (data to noise) for SMLD}

In SMLD, for a fixed sequence of positive scales $0 < \sigma_1 < \cdots < \sigma_L < 1$, and a noise sample $\rvepsilon \sim \gN(\vzero, \rmI)$, and a clean data point $\rvx_0$, the \textbf{cumulative} ``\textbf{forward}'' process is:
\begin{align}
&q_{\sigma_t}^\SMLD (\rvx_i \mid \rvx_0) = \gN(\rvx_i \mid \rvx, \sigma_i^2 \rmI) \implies \rvx_i = \rvx_0 + \sigma_i \rvepsilon .
\end{align}
The \textbf{transition} ``\textbf{forward}'' process can be derived as:
\begin{align}
&q_{\sigma_i}^\SMLD (\rvx_{i+1} \mid \rvx_i) = \gN(\rvx_{i+1} \mid \rvx_i, (\sigma_{i+1}^2 - \sigma_i^2)\rmI) \implies \rvx_i = \rvx_{i-1} + \sqrt{\sigma_i^2 - \sigma_{i-1}^2}\rvepsilon_{i-1} . \label{eq:SMLD_forward}
\end{align}

\subsection{Score for SMLD}

For isotropic Gaussian noise as in SMLD,
\begin{align}
&q_{\sigma_t}^\SMLD (\rvx_i \mid \rvx_0) = \gN(\rvx_i \mid \rvx_0, \sigma_i^2 \rmI) , \\
\implies &\nabla_{\rvx_i} \log q_{\sigma_i}^\SMLD (\rvx_i \mid \rvx_0) = -\frac{1}{\sigma_i^2}(\rvx_i - \rvx_0) = -\frac{1}{\sigma_i}\rvepsilon .
 \end{align}

\subsection{Score-matching objective function for SMLD}

The objective function for SMLD at noise level $\sigma_i$ is:
\begin{align}
&\loss^\SMLD(\rvtheta; \sigma_i) \triangleq\ \frac{1}{2} \EE_{q_{\sigma_i}^\SMLD(\rvx_i \mid \rvx_0) p(\rvx_0)} \bigg[ \Norm{\rvs_\rvtheta (\rvx_i, \sigma_i) + \frac{1}{\sigma_i^2} (\rvx_i - \rvx_0) }_2^2 \bigg] .
\end{align}

\subsection{Variance of score for SMLD}
\begin{align}
\EE\left[ \Norm{ \nabla_{\rvx_i} \log q_{\sigma_i}^\SMLD (\rvx_i \mid \rvx_0) }_2^2 \right] &= \EE\left[ \Norm{ -\frac{(\rvx_i - \rvx_0)}{\sigma_i^2} }_2^2 \right] ,
= \frac{1}{\sigma_i^2}\EE\left[ \Norm{\rvepsilon}_2^2 \right] = \frac{1}{\sigma_i^2} .
\label{eq:SMLD_var}
\end{align}

\subsection{Overall objective function for SMLD}

\cite{song2019generative, song2020improved} chose a geometric series of $\sigma_i$'s, i.e. $\sigma_{i-1}/\sigma_i = \gamma$. The overall objective function was a weighted combination of the objectives at different noise levels, the weight $\lambda(\sigma_i)$ being the inverse of the variance of the score from \cref{eq:SMLD_var} i.e. $\lambda(\sigma_i) = \sigma_i^2$:
\begin{align*}
\gL^\SMLD(\rvtheta; \{\sigma_i\}_{i=1}^{L}) &\triangleq \frac{1}{2L} \sum_{i=1}^{L} \EE_{q_{\sigma_i}^{\SMLD}(\rvx_i \mid \rvx_0) p(\rvx_0)} \bigg[ \Norm{\sigma_i \rvs_\rvtheta (\rvx_i, \sigma_i) + \frac{(\rvx_i - \rvx_0)}{\sigma_i} }_2^2 \bigg] , \\
&= \frac{1}{2L} \sum_{i=1}^{L} \EE_{q_{\sigma_i}^{\SMLD}(\rvx_i \mid \rvx_0) p(\rvx_0)} \bigg[ \Norm{\sigma_i \rvs_\rvtheta (\rvx_i, \sigma_i) + \rvepsilon }_2^2 \bigg] .
\numberthis
\end{align*}

\subsection{Unconditional SMLD score estimation}

\citet{song2020improved} discovered that empirically the estimated score was proportional to $\frac{1}{\sigma}$. So an unconditional score model is:
\begin{align}
\rvs_\rvtheta(\rvx_i, \sigma_i) = -\frac{1}{\sigma_i} \rvepsilon_\rvtheta(\rvx_i) .
\end{align}

In this case, the overall objective function changes to:
\begin{align*}
\gL^\SMLD(\rvtheta; \{\sigma_i\}_{i=1}^{L}) &\triangleq \frac{1}{2L} \sum_{i=1}^{L} \EE_{q_{\sigma_i}^{\SMLD}(\rvx_i \mid \rvx_0) p(\rvx_0)} \bigg[ \Norm{ \rvepsilon - \rvepsilon_\rvtheta (\rvx_i) }_2^2 \bigg] ,
\numberthis
\\
&= \frac{1}{2L} \sum_{i=1}^{L} \EE_{q_{\sigma_i}^{\SMLD}(\rvx_i \mid \rvx_0) p(\rvx_0)} \bigg[ \Norm{ \rvepsilon - \rvepsilon_\rvtheta (\rvx_0 + \sigma_i \rvepsilon) }_2^2 \bigg] .
\end{align*}

\subsection{Sampling in SMLD}

Unlike DDPM, SMLD does not explicitly define a reverse process. Instead, \citet{song2019generative, song2020improved} use an iterative variant of Langevin Sampling called Annealed Langevin Sampling to transform from noise to data. $i=0$ corresponds to data, and $i=L$ corresponds to noise, hence time order for noise to data is $L$ to $0$.

Forward : $\rvx_i = \rvx_{i-1} + \sqrt{\sigma_i^2 - \sigma_{i-1}^2}\rvepsilon_{i-1}$.

\underline{Reverse}: Using Annealed Langevin Sampling from ~\citet{song2019generative, song2020improved}:
\begin{align*}
&\rvx_L^M \sim \gN(\vzero, \sigma_{\max}\rmI).\\
&\begin{rcases}
&\rvx_i^M = \rvx_{i+1}^0 . \\
&\alpha_i = \eps \sigma_i^2/\sigma_{\min}^2. \\
&\rvx_i^{m-1} \leftarrow \rvx_i^m + \alpha_i \rvs_{\rvtheta^*}(\rvx_i^m, \sigma_i) + \sqrt{2\alpha_i} \rvepsilon_i^{m-1} , m=M,\cdots,0 .\\
\implies &\rvx_i^{m-1} \leftarrow \rvx_i^m - \frac{\alpha_i}{\sigma_i} \rvepsilon_{\rvtheta^*}(\rvx_i^m) + \sqrt{2\alpha_i} \rvepsilon_i^{m-1} , m=M,\cdots,0 .
\end{rcases}
i = L, \cdots, 1
\numberthis
\end{align*}

Using Consistent Annealed Sampling from \citet{jolicoeur2020adversarial}:
\begin{align*}
&\alpha_i = \eps \sigma_i^2/\sigma_{\min}^2 = \eta \sigma_i^2;\ \beta = \sqrt{1 - \gamma^2(1 - \eps/\sigma_{\min}^2)^2};\ \gamma = \sigma_i/\sigma_{i-1} ; \sigma_i > \sigma_{i-1} . \\
&\rvx_{i-1} \leftarrow \rvx_i + \alpha_i \rvs_{\rvtheta^*}(\rvx_i, \sigma_i) + \beta \sigma_{i-1} \rvepsilon_{i-1}, i = L, \cdots, 1 . \\
\implies &\rvx_{i-1} \leftarrow \rvx_i - \eta \sigma_i \rvepsilon_{\rvtheta^*}(\rvx_i) + \beta \sigma_{i-1} \rvepsilon_{i-1}, i = L, \cdots, 1 .
\numberthis
\end{align*}

\subsection{Expected Denoised Sample (EDS) for SMLD}

From \cite{saremi2019neb}, for isotropic Gaussian noise, we know that the expected denoised sample $\rvx_0^*(\rvx_i, \sigma_i) \triangleq \EE_{\rvx_0 \sim q_{\sigma_i}(\rvx_0 \mid \rvx_i)}[\rvx_0]$ and the optimal score $\rvs_{\rvtheta^*}(\rvx_i, \sigma_i)$ are related as:
\begin{align*}
&\rvs_{\rvtheta^*}(\rvx_i, \sigma_i) = \frac{1}{\sigma_i^2}(\rvx_0^*(\rvx_i, \sigma_i) - \rvx_i) , \\
\implies &\rvx_0^*(\rvx_i, \sigma_i) = \rvx_i + \sigma_i^2 \rvs_{\rvtheta^*}(\rvx_i, \sigma_i) = \rvx_i - \sigma_i \rvepsilon_{\rvtheta^*}(\rvx_i) .
\numberthis
\end{align*}

\subsection{SDE formulation : Variance Exploding (VE) SDE}

For SMLD i.e. Variance Exploding (VE) SDE, the forward equation and transition probability are derived (below) as:
\begin{align}
    \D \rvx &= \sqrt{\frac{\D [\sigma^2(t)]}{\D t}}\ \D \rvw \label{eq:VESDE_d} , \\
    p_{0t}^\VE(\rvx(t) \mid \rvx(0)) &= \gN \left(\rvx(t) \mid \rvx(0), \sigma^2(t) \rmI \right) . \label{eq:VESDE_p}
\end{align}

\subsubsection{Derivations}:

\textbf{Forward process}: We know from \cref{eq:SMLD_forward} that:
\begin{align*}
\rvx_i &= \rvx_{i-1} + \sqrt{\sigma_i^2 - \sigma_{i-1}^2}\rvepsilon_{i-1} . \\
\implies \rvx(t + \Delta t) &= \rvx(t) + \sqrt{(\sigma^2(t + \Delta t) - \sigma^2(t)) \Delta t}\ \rvepsilon(t) ,\\
&\approx \rvx(t) + \sqrt{\frac{\D [\sigma^2(t)]}{\D t} \Delta t}\ \rvw(t) . \\
\implies \D \rvx &= \sqrt{\frac{\D [\sigma^2(t)]}{\D t}}\ \D \rvw .
\numberthis
\label{eq:SDE:SMLD}
\end{align*}
Thus, \cref{eq:VESDE_d} is derived.

\vspace{1.5em}

\textbf{Mean} $\vmu$ and \textbf{Covariance} $\mSigma_{\text{cov}}$ (from eq. 5.50 and eq.5.51 in \cite{sarkka2019applied}) for a random variable $\rvx$ that changes according to a stochastic process with drift and diffusion coefficients $\rvf$ and $\rmG$, change as:
\begin{align}
\D \rvx = \rvf\ \D t + \rmG\ \D \rvw 
\implies 
&\frac{\D \vmu}{\D t} = \EE_\rvx[\rvf] , \\
&\frac{\D \mSigma_{\text{cov}}}{\D t} = \EE_\rvx[\rvf (\rvx - \vmu)^\trns] + \EE_\rvx[(\rvx - \vmu) \rvf^\trns] + \EE_\rvx[\rmG\rmG^\trns] .
\numberthis
\end{align}

For SMLD i.e. VE SDE, $\rvf = \vzero, \vmu = \vzero, \rmG = \sqrt{\frac{\D [\sigma^2(t)]}{\D t}} \rmI$.
\begin{align*}
&\frac{\D \vmu_{\SMLD}(t)}{\D t} = \EE_\rvx[\vzero] = \vzero , \\
\implies &\vmu_{\SMLD}(t) = \vmu(0) = \rvx(0). \\
&\frac{\D \mSigma_{\SMLD}(t)}{\D t} = \EE_\rvx\left[\vzero + \vzero + \sqrt{\frac{\D [\sigma^2(t)]}{\D t}} \sqrt{\frac{\D [\sigma^2(t)] \rmI}{\D t}}\right] = \frac{\D [\sigma^2(t)]}{\D t}\rmI, \\
\implies &\mSigma_{\SMLD}(t) = \sigma^2(t) \rmI .
\end{align*}

$\therefore \text{SMLD i.e. } p_{0t}^\VE(\rvx(t) \mid \rvx(0)) = \gN \left(\rvx(t) \mid \rvx(0), \sigma^2(t) \rmI \right)$.
Thus, \cref{eq:VESDE_p} is derived.

%% file: EncODEDec.tex
\anglais
\counterwithin{figure}{chapter}
\counterwithin{table}{chapter}

\chapter{Simple video generation using Neural ODEs~\citep{voleti2019simple}}
\label{chap:EncODEDec}

\setcounter{section}{-1}
\section{Prologue to article}
\label{chap:pro_EncODEDec}

\subsection{Article details}

\textbf{Simple video generation using Neural ODEs}. Vikram Voleti*, David Kanaa*, Samira Ebrahimi Kahou, Christopher Pal (*denotes equal contribution). \textit{Advances in Neural Information Processing Systems (NeurIPS) 2019 Workshop}

\textit{Personal contribution}:
The project began with discussions between the authors at Mila during Christopher Pal's research group meetings. The idea was to try to see if Neural ODEs were capable of modeling the dynamics of a video, to such an extent as to predict future frames. Vikram Voleti proposed the preliminary experiments to test the hypothesis, and wrote the code. Vikram Voleti and David Kanaa wrote further code, performed several experiments with various settings, discussed the mathematical foundations of the method, proposed further ideas about the architecture of the model, coded two variants in the architecture, performed experiments using the Moving MNIST dataset, and wrote parts of the paper. Samira Ebrahimi Kahou and Christopher Pal provided advice and guidance throughout the project and wrote parts of the paper.

\subsection{Context}

Despite having been studied to a great extent, the task of conditional generation of sequences of frames---or videos---remains extremely challenging. It is a common belief that a key step towards solving this task resides in modelling accurately both spatial and temporal information in video signals. A promising direction to do so has been to learn latent variable models that predict the future in latent space and project back to pixels, as suggested in literature. Recently, Neural ODEs were proposed as a way to model continuous dynamics using neural networks. However, whether they were capable of modeling the dynamics in videos had not been explored yet.

\subsection{Contributions}

Building on top of a family of models introduced in prior works, Neural ODE, this work investigates an approach that models time-continuous dynamics over a continuous latent space with a differential equation with respect to time. The intuition behind this approach is that these trajectories in latent space could then be extrapolated to generate video frames beyond the time steps for which the model is trained.  We show that our approach yields promising results in the task of future frame prediction on the Moving MNIST dataset with 1 and 2 digits. To the best of our knowledge, this is the first work that explores the use of Neural ODEs in the space of video generation.

\subsection{Research impact}

Our main hypothesis that Neural ODEs are capable of capturing the dynamics of a video has been re-validated and improved upon by future works. This work influenced several future publications that used Neural ODEs as part of generative models for images~\citep{finlay2020rnode,ghosh2020steer,wang2021image,voleti2021improving}, as well as improved video generation~\citep{Yildiz2019ODE2VAEDG,park2020vid,Xu2023ControllableVG,auzina2023invariant}. Some of these works expanded upon topics we introduced in this paper, such as interpolation of video by oversampling the ODE trajectory.

\newpage
\section{Introduction}

Conditional frame generation in videos (interpolation and/or extrapolation) remains a challenging task despite having been well studied in the literature. It involves encoding the first few frames of a video into a good representation that could be used for subsequent tasks, i.e. prediction of the next few frames.
Solving the task of conditional frame generation in videos requires one to identify, extract and understand latent concepts in images, as well as adequately model both spatial and temporal factors of variation. Typically, an encoder-decoder architecture is used to first encode conditioning frames into latent space, and then recurrently predict future latent points and decode them into pixel space to render future frames.

In this paper, we investigate the use of Neural Ordinary Differential Equations~\citep{chen2018neural} (Neural ODEs) for video generation. The intuition behind this is that we would like to enforce the latent representations of video frames to follow continuous dynamics. Following a dynamic means that frames close to each other in the space-time domain (for example, any video of a natural scene) are close in the latent space. This implies that if we connect the latent embeddings of contiguous video frames, we should be able to obtain trajectories that can be solved for with the help of ordinary differential equations.

Since these trajectories follow certain dynamics in latent space, it should also be possible to extrapolate these trajectories in latent space to future time steps, and decode those latent points to predict future frames. In this paper, we explore this possibility by experimenting on a simple video dataset --- Moving MNIST~\citep{srivastava2015unsupervised} --- and show that Neural ODEs do offer the advantages described above in predicting future video frames.

Our main contributions are:
\begin{itemize}
    \item we repurpose the encoder-decoder architecture for video generation with Neural ODEs,
    \item we show promising results on 1-digit and 2-digit Moving MNIST,
    \item we discuss the future directions of work and relevant challenges.
\end{itemize}

To the best of our knowledge, this is the first work that explores the use of Neural ODEs in the space of video generation.

\section{Related work}
\label{EncODEDec:related}
%
%
Early work on using deep learning algorithms to perform video generation has tackled the problematic in various ways, from using stacked regular LSTM layers~\citep{srivastava2015unsupervised} to combining convolution with LSTM modules in order to extract local spatial information which correlates with long-term temporal dependencies~\citep{xingjian2015convolutional}. \citet{prabhat2017deep} show 3D convolution can be effectively used to extract spatio-temporal information from sequences of images for extreme weather detection. \citet{wang2018video} use a generative model guided with segmentation maps to generate single step future frame. While their results may be interesting, the model might rely too much on segmentation due to the conditioning.

Other work in the recent literature \citep{babaeizadeh2018stochastic, denton2018SVGLP, Lee2018StochasticAV} incorporate stochastic components to their model that encodes the conditioning frames into latent space similar to a prior distribution, which is then sampled from to predict the next frames. This ensures the uncertainty over possible futures is taken into account.

More recently, \citep{Castrejn2019ImprovedCV} show that using a hierarchy of latent variables to improve the expressiveness of their generative model can lead to noticeably better performance on the task of video generation. \citep{clark2019adversarial} use a Generative Adversarial Network combined with separable spatial and temporal attention models applied on latent feature maps in order to  handle spatial and temporal consistency separately.

While some of the above methods have yielded state-of-the-art results, some still struggle to produce smooth motions and for those who do produce continuously smooth ones, they enforce it through temporal regularisation in the optimisation objective, or through a specific training procedure. Drawing from recent work on using parameterised ODE estimators~\citep{chen2018neural} to model continuous-time dynamics, we choose to approach this problem with the intuition that we would like the video frames to be smoothly connected in latent space according to some continuous dynamics which we would learn. Unlike recurrent neural networks and other purely auto-regressive approaches which require observations to occur at uniform intervals, and may require three different models to extrapolate forward and backwards, or interpolate, continuously-defined dynamics should naturally allow to process observations occurring at non-uniform intervals and generate at any time, thus reducing the number of models required to perform extrapolation and interpolation to one.

\section{Neural Ordinary Differential Equations (Neural ODEs)}
\label{EncODEDec:neuralode}

Neural Ordinary Differential Equations~\citep{chen2018neural} (Neural ODEs) represent a family of parameterised algorithms designed to model the evolution across time of any system, of state $\boldsymbol{\xi}(t)$ at an arbitrary time $t$, governed by continuous-time dynamics satisfying a Cauchy (or initial value) problem
\begin{equation*}
     \begin{cases}
        \,\boldsymbol{\xi}(t_0) &= \boldsymbol{\xi}_0 , \\ \\
        \,\dfrac{\partial\boldsymbol{\xi}}{\partial t}(t) &= f(\boldsymbol{\xi}(t), t) .
     \end{cases}
\end{equation*}
By approximating the differential with an estimator $f_{\theta} \simeq f$ parameterized by $\theta$, such as a neural network, these methods allow to learn such dynamics (or, trajectories) from relevant data.
Thus formalised, the state $\xi(t)$ of such a system is defined at all times, and can be computed at any desired time using a numerical ODE solver, which will evaluate the dynamics $f_{\theta}$ to determine the solution.

\[ (\boldsymbol{\xi}_0, \boldsymbol{\xi}_1, \dots, \boldsymbol{\xi}_n) = \text{\ttfamily ODEsolver}(f_{\theta} , \boldsymbol{\xi}_0, (t_0, t_1, \dots, t_n)). \]

For any single arbitrary time value $t_i$, a call to the {\ttfamily ODEsolver} computes a numerical approximation of the integral of the dynamics from the initial time value $t_0$ to $t_i$.
\[ \boldsymbol{\xi}_i = \text{\ttfamily ODEsolver}(f_{\theta} , \boldsymbol{\xi}_0, (t_0, t_i)) \, \simeq \, \boldsymbol{\xi}_0 + \int_{t_0}^{t_i} f_{\theta}\left(\boldsymbol{\xi}(s), s\right)\,ds \, = \, \boldsymbol{\xi}(t_i).\]

There exist in the literature a plethora of algorithms to perform numerical integration of differential equations. Amongst the most common are : the simplest, Euler's method; higher order methods such as Runge-Kutta methods~\citep{pontyagin1962mathematical, kutta1901beitrag,hairer2000solving}; as well as multistep algorithms like the Adams-Moulton-Bashforth methods~\citep{butcher2016numerical, hairer2000solving, quarteroni2000matematica}. More advanced algorithms have been proposed to provide better control over the approximation error and the accuracy~\citep{press2007numerical}.
In their implementation\footnote{https://github.com/rtqichen/torchdiffeq}~\citep{chen2018neural} use a variant of the fifth-order Runge-Kutta with adaptive stepsize and local truncation error monitoring to ensure accuracy.

The optimisation of the Neural ODE is performed through the framework of adjoint sensitivity~\citep{pontyagin1962mathematical} which can be formalised as follows. Provided a scalar-valued objective function:
\[L(\theta) = L\left(\boldsymbol{\xi}_0 + \int_{t_0}^{t_i} f_{\theta}\left(\boldsymbol{\xi}(s), s\right)\,ds\right),\]
the gradient of the objective with respect to the model's parameters follows the differential system:
\begin{align*}
        \dfrac{d\boldsymbol{a}(t)}{dt}   &= - \,\, \boldsymbol{a}(t)^\top \, \dfrac{\partial f(\boldsymbol{\xi}(t), t, \theta)}{\partial \boldsymbol{\xi}} ,\\
        \dfrac{dL}{d\theta} \,\,&= - \int_{t_i}^{t_0} \boldsymbol{a}(s)^\top \, \dfrac{\partial f(\boldsymbol{\xi}(s), s, \theta)}{\partial \theta} ds,
\end{align*}
where the $\boldsymbol{a}(t) = \partial L/\partial \boldsymbol{\xi}$ is the adjoint.

\section{Our approach}
\label{EncODEDec:method}

Our approach combines the familiar encoder-decoder architecture of neural network models with a Neural ODE that works in the latent space.
\begin{enumerate}
    \item We encode the conditioning frames into a point in latent space
    \item We feed this latent embedding to a Neural ODE as the ``initial value'' at time $t=0$, and use it to predict latent points corresponding to future time steps.
    \item We decode each of these latent points into frames in pixel space at different time steps
\end{enumerate}

More formally, in accordance with established formulations of the task of video prediction, let us assume a setting in which we have a set of $m$ contextual frames $\mathcal{C} = \{ (x_{i}, t_{i}) \}_{i \in \Iintv{0, m}}$. We seek to learn a predictive model such that, provided $\mathcal{C}$, we can make predictions $\mathcal{P}(\mathcal{C})= \{ (x_{j}, t_{j}) \}_{j \in \Iintv{m, m+n}}$ about the evolution of the video across time, arbitrarily in the future or past (extrapolation) or even in between observed frames (interpolation).

Let $x(t)$ denote the continuous signal representing the video stream from which $\mathcal{C}$ is sampled, that is : \[\forall (x_i, t_i) \in \mathcal{C}, \, x(t_i) = x_i.\]
The temporal changes in the raw signal $x(t)$ can be interpreted as effects of temporal variations in the latent concepts embedded within it. For example, suppose we have a video of a ball moving, any temporal change in the video will be observed only on pixels related to the latent notion of "moving ball". Because the concept "ball" follows some motion, the related pixels will change accordingly.
From this statement it follows the intuition to model dynamics in latent space and capture spatial characteristics separately. Thus we learn a predictor $\mathcal{P}$ which 
\begin{itemize}
      \item learns a latent representation of the observed discrete sequence of frames that captures spatial factors of variation, as well as
      \item infers plausible latent continuous dynamics from which the aforementioned discrete sequence may be sampled i.e. which better explains the temporal variations within the sequence.
\end{itemize}

The proposed model follows the formalism of latent variable model proposed by~\citep{chen2018neural} in which the latent at the current time value $z(t_{m})$ is sampled from a distribution $\mathbb{P}_{Z}$, the latent generative process is defined by an ODE that determines the trajectory followed in latent space from the initial condition $z(t_{m})$, and a conditional $\mathbb{P}_{X|Z}$ with respect the latent vectors predicted along the trajectory at provided times is used to independently sample predicted images:

\vspace{-0.5em}

\begin{align*}
        z_{m} &\sim \mathbb{P}_{Z} \left( \cdot \right) , \\
        z(t_{i}) &= \mathcal{I} (f_{\theta}, z_{m}, t_{m}, t_{i}) = z_{m} + \int_{t_{m}}^{t_{i}} f\left(z(s), s; \theta \right)\,ds &\forall t_{i} &\in \Iintv{t_{m}, t_{m+n}}, \\
        x(t_{i}) &\sim \mathbb{P}_{X|Z} \left( \cdot\,|\, z(t_{i}) \right),\quad x(t_{i}) \indep x(t_{j}) &\forall t_{i}, t_{j} &\in \Iintv{t_{m}, t_{m+n}} .
\end{align*}

In practice, we use an approximate posterior $q_{\phi}(\cdot\,|\,\mathcal{C})$ instead of $\mathbb{P}_{Z}$, and similarly, instead of $\mathbb{P}_{X|Z}$, we use an estimator $p_{\psi}(\cdot\,|\,z(t_{m}))$. Together, these estimators function as an \emph{encoder-decoder} pair between the space of image pixels and that of latent representations.

We investigate a deterministic setting where a unique and non-recurrent pair encoder-decoder is used to process every frame. The encoder projects a frame $(x_{i}, t_{i})$ onto an embedding $z_{t_{i}} = z_{i} = q_{\phi}(x_{t_{i}})$, then the ODE defining the latent dynamics is integrated to produce the value of the latent embedding $z_{t_{i}} = \mathcal{I} (f_{\theta}, z_{0}, t_{0}, t_{i})$. Finally, the decoder is used to project $z(t_{j})$ back into an image $\hat{x}_{t_{j}} = p_{\psi}(z_{t_{j}})$. In terms of objective function used to optimise the parameters of the model, we use a combination of an $L_2$ reconstruction in pixel space, and an $L_2$ distance between the latent points predicted by the NeuralODE and the embeddings of each frame:

\vspace{-0.5em}

\begin{align*}
    \mathcal{L}(\phi, \theta, \psi) &= \,\sum_{\Iintv{t_{m}, t_{m+n}}} \| x_{t_{i}} - p_{\psi} \circ \mathcal{I} (f_{\theta}, q_{\phi}(x_{t_{0}}), t_{0}, t) \|_{2}^{2} \\
    &\qquad \qquad \qquad + \| q_{\phi}(x_{t_{i}}) - \mathcal{I} (f_{\theta}, q_{\phi}(x_{t_{0}}), t_{0}, t_{i}) \|_{2}^{2} .
    \numberthis
    \label{eq:loss2}
\end{align*}

The latter component of the objective function is meant to ensure that we learn a compact latent subspace to which both the learnt dynamics and the encoder project. More precisely, it enforces the latent representation predicted by the Neural ODE to match that estimated for each time step by the encoder.

We also inquire into the sequence-to-sequence architecture~\citep{chen2018neural}, where

\begin{align*}
    \mathbb{P}_{Z} = \mathcal{N}(\mu(\mathcal{C}), \sigma^{2}(\mathcal{C})), &\quad \mathbb{P}_{X|Z} = \mathcal{N}(\mu(z(t_{m})), \sigma^{2}(z(t_{m}))), \\
    &\text{thus,} \\
    q_{\phi}(\cdot\,|\,\mathcal{C}) = (\mu_{\phi}(\mathcal{C}),\sigma^{2}_{\phi}(\mathcal{C})), &\quad p_{\psi}(\cdot\,|\,z(t_{m})) = (\mu_{\psi}(z(t_{m})), \sigma^{2}_{\psi}(z(t_{m}))).
\end{align*}

In practice, $\sigma_{\psi}(z(t_{m}))$ is set to a constant value $\sigma = 1$ and $\mu_{\psi}(z(t_{m})) = x_{m}$, the true frame observed at time $t_{m}$. In this setting, the variational encoder $q_{\phi}$ used is based on an RNN model over the context $\mathcal{C} = \{ (x_{i}, t_{i}) \}_{i \in \Iintv{0, m}}$, whereas the decoder $p_{\psi}$ is non-recurrent---hypothesis of independence between generated frames; the temporal dependencies are modelled by the ODE. At training time, the entire estimator is optimised as a variational auto-encoders~\citep{kingma2013auto, rezende2014stochastic} through the maximisation of the Evidence Lower Bound (ELBO):

\begin{equation}\label{eq:elbo}
    \mathcal{E}(\phi, \theta, \psi) = \,\underbrace{\sum_{\Iintv{t_{m}, t_{m+n}}} -\,\mathbb{E}_{z_{m} \sim q_{\phi}(\cdot\,|\,\mathcal{C})} [\log p_{\psi}(\hat{x}_{t}\,|\,\mathcal{I} (f_{\theta}, z_{m}, t_{m}, t)]}_{\text{reconstruction term}}
    + \mathrm{D}_{KL}(q_{\phi}(\cdot\,|\,\mathcal{C})\,\|\,\mathcal{N}(0, \boldsymbol{I})) .
\end{equation}

\section{Experiments on Moving MNIST}
\label{EncODEDec:exp}

We explore two different methods of combining an encoder-decoder framework with ODEs for 1-digit and 2-digit Moving MNIST~\citep{srivastava2015unsupervised}. In each case, we use the first 10 frames as both input to the model and as ground truth for reconstruction, which is the output of the model. We then check how the model performs on the subsequent 10 frames.

\subsection{1-digit Moving MNIST with non-RNN Encoder}

\begin{figure*}[t]
\centering
\begin{tabular}{cc}
\includegraphics[width=.45\textwidth]{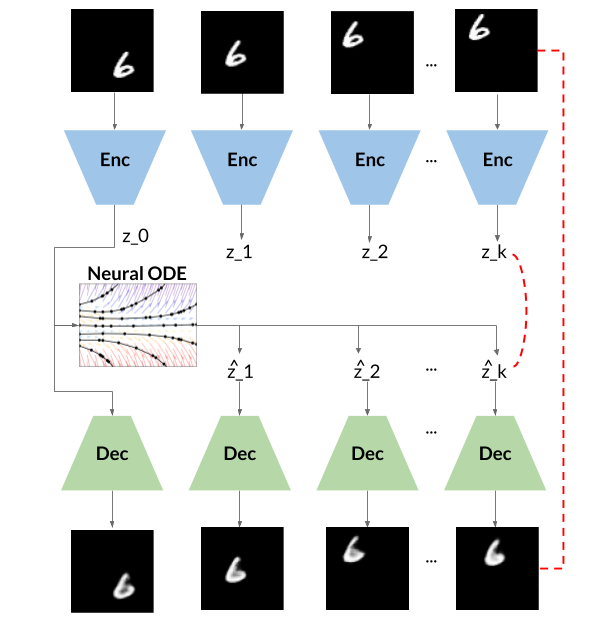} &
\includegraphics[width=.45\textwidth]{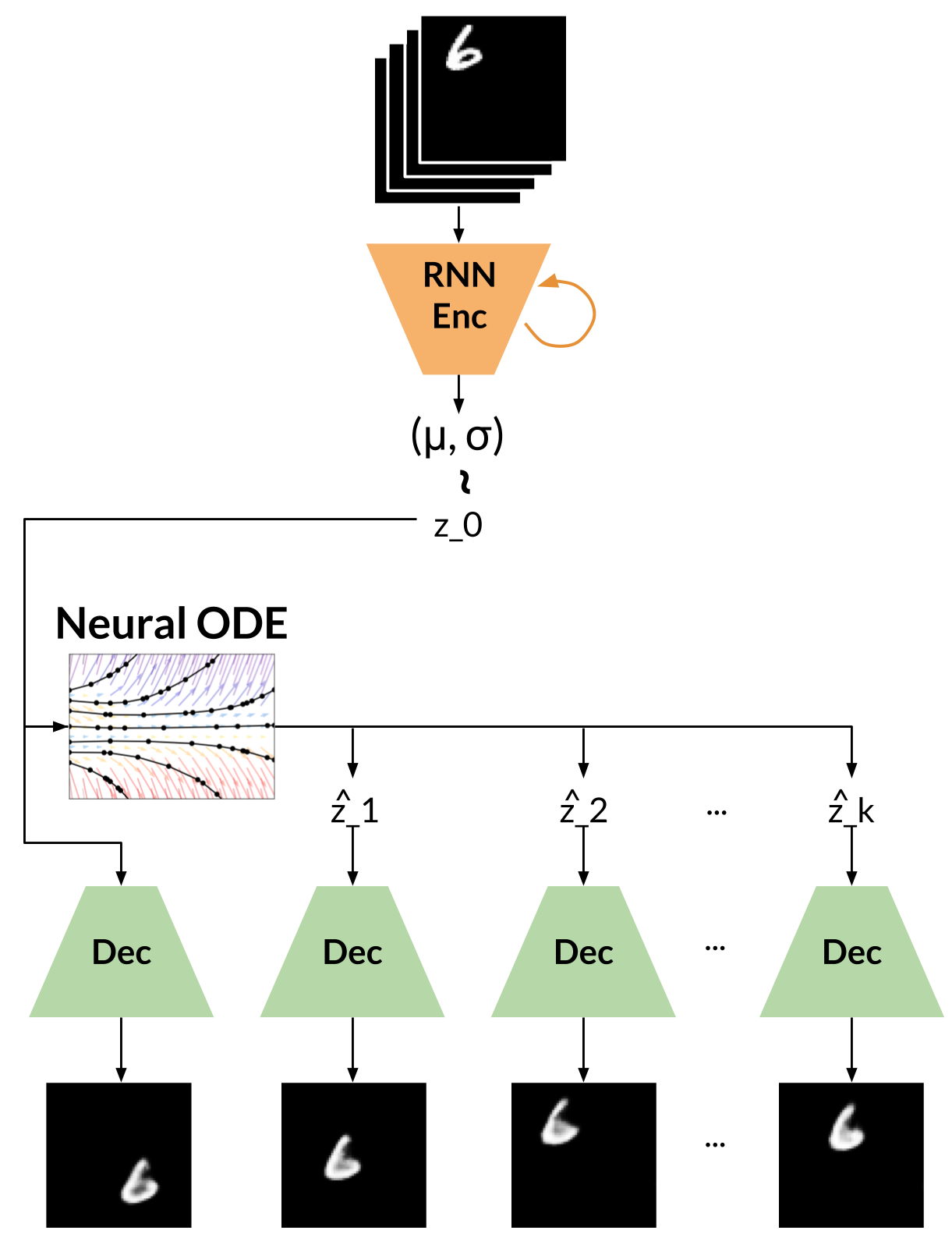}\\
(a) & (b)
\end{tabular}
\vspace{-.5em}
\caption{Architectures for Encoder-ODE-Decoder}
\label{fig:arch}
\end{figure*}

This method, corresponding to \autoref{eq:loss2}, involves an encoder and a decoder that each act on a single frame to embed and decode, respectively, a latent representation. Figure~\ref{fig:arch} (a) shows this architecture. Here, we try to enforce this representation to follow a continuous dynamics in latent space such that there is a one-to-one mapping between the raw pixel space and the latent space from both the encoder side as well as the decoder side.

This model takes one frame as the conditioning input, encodes it, feeds it to the ODE which then predicts the latent representations of the first 10 time steps (including the one which was fed to it), each of which is then decoded to pixel space. We then compute a loss between the reconstructed output and the original input. In addition, each frame of the original video is also encoded separately, and we compute another loss on the encoded latent representations and those predicted by the ODE. This is to enforce the latent representations provided by the encoder to follow the dynamics implicit in the Neural ODE.

\begin{figure*}[t]
\centering
\begin{tabular}{cc}
Reconstruction (1-10) \hfill Prediction (11-20) \hfill\\
\hfill 1 \hfill 2 \hfill 3 \hfill 4 \hfill 5 \hfill 6 \hfill 7 \hfill 8 \hfill 9 \hfill 10 \hfill 11 \hfill 12 \hfill 13 \hfill 14 \hfill 15 \hfill 16 \hfill 17 \hfill 18 \hfill 19 \hfill 20 \hfill\\
\includegraphics[width=.95\textwidth]{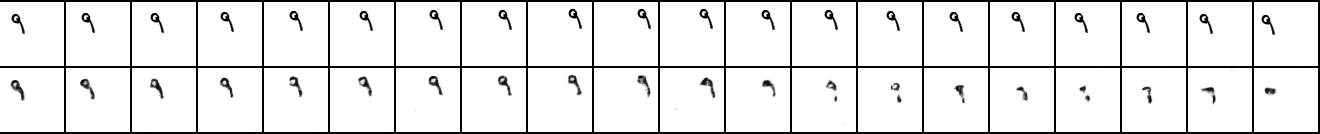}\\
(a) Train \\
\includegraphics[width=.95\textwidth]{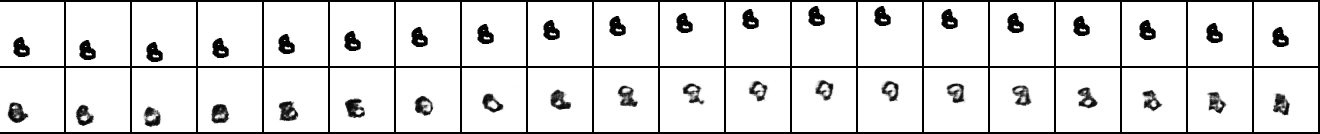}\\
(b) Validation
\end{tabular}
\vspace{-.5em}
\caption{Samples predicted at 20 time steps, conditioned on the first 10 time steps with frames from (a) train set and (b) validation set using Non-RNN Encoder --- ODE --- Decoder (Figure~\ref{fig:arch}a). In each, top row are original samples, and bottom row are predicted samples. For this figure, we use the trained model to reconstruct the first 10 frames and then predict the next 10 frames}
\label{fig:arch1_samples}
\end{figure*}

We used 1000 video sequences of length 10 as conditioning input (as well as reconstruction output), and a batch size of 100. The encoder and decoder have inverted architectures with the same number of channels in their respective orders. Figure~\ref{fig:arch1_samples} shows samples from using this architecture.

\subsection{1-digit Moving MNIST with RNN Encoder}

While the previous architecture works pretty well on the training samples, We see that it does not work very well on validation data. We believe that there are two things that must be corrected:

\begin{itemize}
    \item The encoder must be conditioned on multiple frames.
    \item The latent representation provided by the encoder must be stochastic in nature
\end{itemize}

Since the Neural ODE is only seeing the first frame of the video to base its latent dynamic trajectories on, it is a highly constrained problem.
However, we would like to relax this constraint by conditioning the Neural ODE on multiple frames, which is commonly practiced in video prediction/generation.

We would also like to make the model stochastic. The previous model is deterministic, so there is a high chance it simply memorizes the training data. So, given an input frame, there is exactly 1 trajectory the Neural ODE is able to generate for it, so there is no scope of any variation in the generated videos. We would like to generate different videos given the same conditioning input, since it matches with real world data.

Figure~\ref{fig:arch} (b) shows such an architecture that solves both the above issues, corresponding to \autoref{eq:elbo}. It is similar to a Variational Recurrent Neural Network~\citep{Chung2015ARL}, except here a Neural ODE handles the latent space.

We feed the first 10 frames as conditioning input to a Recurrent Neural Network (RNN). This network outputs the mean and variance of a multivariate Gaussian distribution. We sample a latent point from this distribution, and feed this to a Neural ODE as the initial value of the latent variable at $t=0$. The Neural ODE then predicts latent representations at the first 10 time steps, which we then decode independently to raw pixels. We compute a reconstruction loss between the predicted frames and the original frames in the first 10 time steps. We also add a KL-divergence loss between the predicted Gaussian distribution and the standard normal distribution, to constrain the latent representation to follow a standard normal prior.

\begin{figure*}[t]
\centering
\begin{tabular}{cc}
Reconstruction (1-10) \hfill Prediction (11-20) \hfill\\
\hfill 1 \hfill 2 \hfill 3 \hfill 4 \hfill 5 \hfill 6 \hfill 7 \hfill 8 \hfill 9 \hfill 10 \hfill 11 \hfill 12 \hfill 13 \hfill 14 \hfill 15 \hfill 16 \hfill 17 \hfill 18 \hfill 19 \hfill 20 \hfill\\
\includegraphics[width=.95\textwidth]{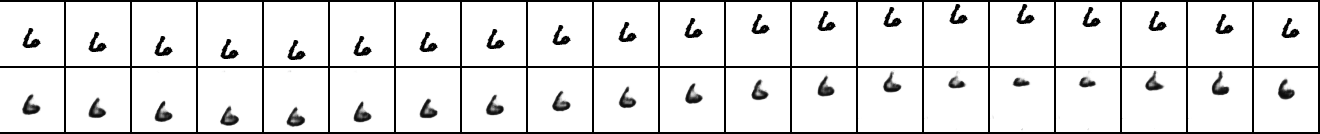}\\
\includegraphics[width=.95\textwidth]{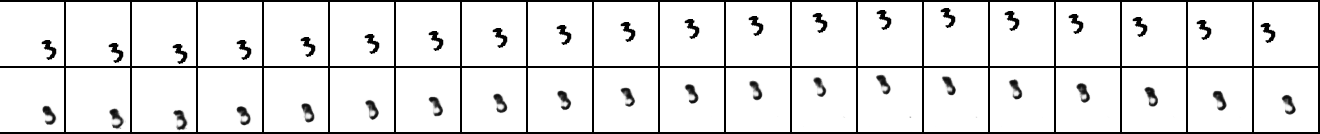}\\
(a) Train \\
\includegraphics[width=.95\textwidth]{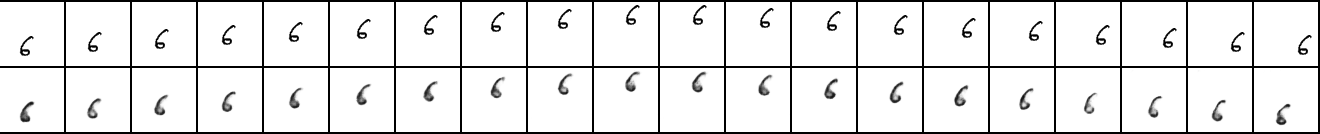}\\
\includegraphics[width=.95\textwidth]{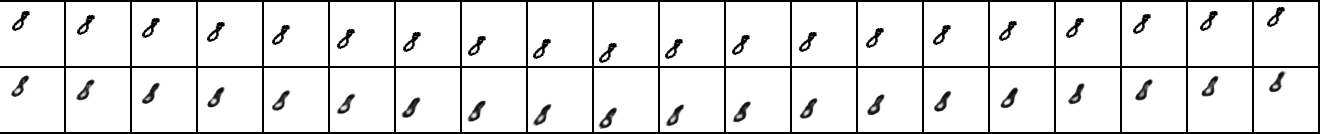}\\
\includegraphics[width=.95\textwidth]{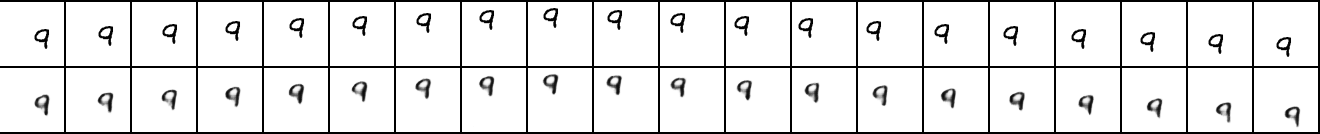}\\
\includegraphics[width=.95\textwidth]{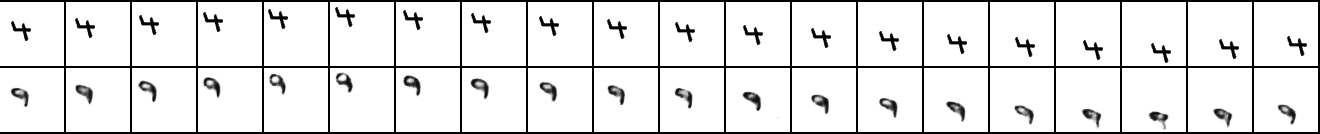}\\
\includegraphics[width=.95\textwidth]{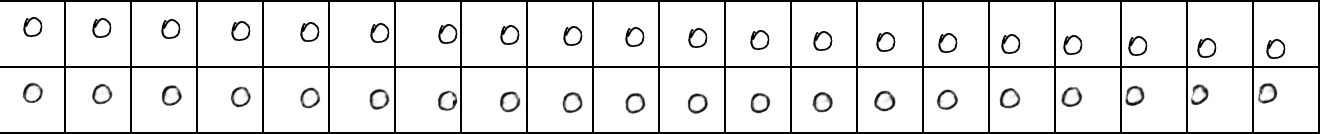}\\
\includegraphics[width=.95\textwidth]{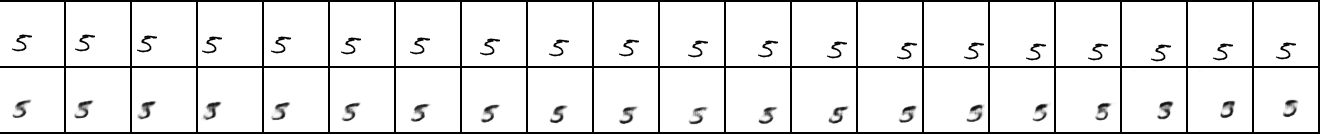}\\
\includegraphics[width=.95\textwidth]{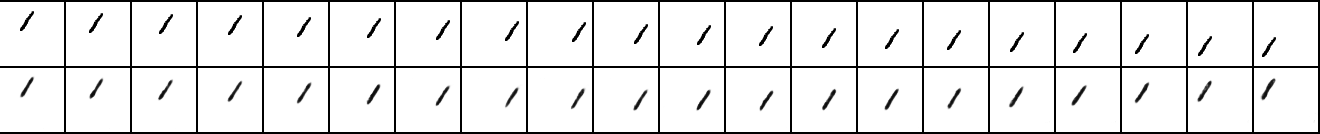}\\
\includegraphics[width=.95\textwidth]{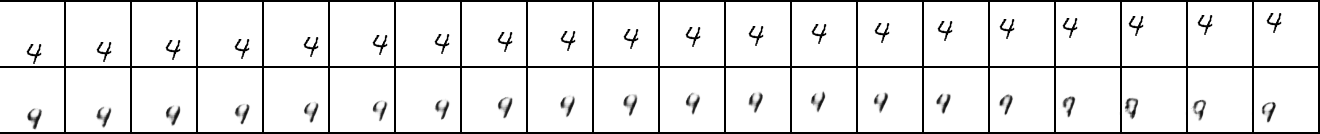}\\
(b) Validation
\end{tabular}
\caption{Predicted samples at 20 time steps conditioned on the first 10 time steps using RNN Encoder-ODE-Decoder (Figure~\ref{fig:arch}b). Original (top row) and predicted (bottom row) samples are shown for both (a) train and (b) validation sets. The model is trained to reconstruct the first 10 frames and predict the next 10 frames.}
\label{fig:arch2_samples}
\end{figure*}

The model architectures of the encoder (except the recurrent part) and the decoder are the same as in the previous model. We provide 10000 videos as training input, and use a batch size of 128. Figure~\ref{fig:arch2_samples} shows the results using this architecture. We can see that the model has been able to capture both structural information and temporal information.


\subsection{2-digit Moving MNIST with RNN Encoder}

We use the same architecture as for 1-digit Moving MNIST (Figure~\ref{fig:arch} (b)) to try to reconstruct 2-digit Moving MNIST. We used the same model settings (number of layers, number of channels, etc.) and the same optimization settings. At the time of writing this paper, we stopped the training at 2000 epochs, same as that for 1-digit Moving MNIST.

\begin{figure*}[t]
\centering
\begin{tabular}{cc}
Reconstruction (1-10) \hfill Prediction (11-20) \hfill\\
\hfill 1 \hfill 2 \hfill 3 \hfill 4 \hfill 5 \hfill 6 \hfill 7 \hfill 8 \hfill 9 \hfill 10 \hfill 11 \hfill 12 \hfill 13 \hfill 14 \hfill 15 \hfill 16 \hfill 17 \hfill 18 \hfill 19 \hfill 20 \hfill\\
\includegraphics[width=.95\textwidth]{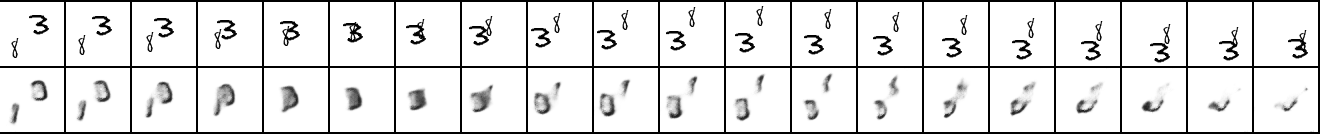}\\
(a) Train \\
\includegraphics[width=.95\textwidth]{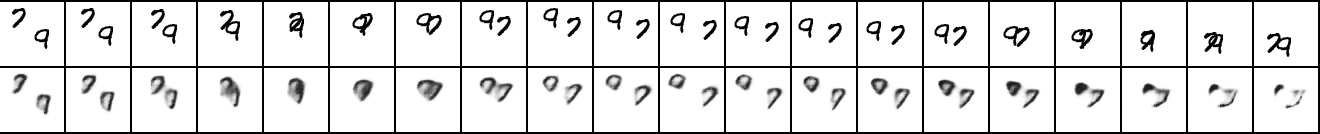}\\
\includegraphics[width=.95\textwidth]{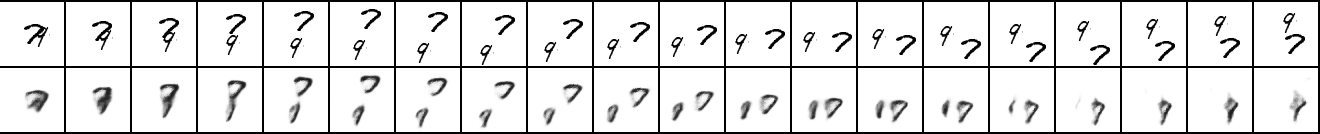}\\
(b) Validation
\end{tabular}
\caption{(top) Original and (bottom) predicted samples of RNN Encoder --- ODE --- Decoder on 2-digit Moving MNIST}
\label{fig:mmnist2_samples}
\end{figure*}

Figure~\ref{fig:mmnist2_samples} shows some samples from a model trained on 2-digit Moving MNIST. We believe the spatial trajectories of each individual digit are being recorded very well by the Neural ODE. However it would take many more epochs for the encoder and decoder to reconstruct the images better. This phenomenon of the Neural ODE training earlier than the encoder-decoder was observed while training 1-digit Moving MNIST as well.

\subsection{A note on the problem formulation}

Note the difference in our training formulation compared to the typical approach for generating videos. The typical training procedure involves conditioning the model on initial frames, and training it to predict future steps. Then in evaluation, conditioned on initial frames of unseen videos, the model is used to predict for time steps including and beyond the trained steps.

However, we formulate our training problem as \textit{reconstruction} instead of prediction i.e. conditioned on initial frames, our model is only trained to reconstruct the same frames during training. Despite this, we can still predict future frames effectively by leveraging the preserved dynamics of the training set, by following the trajectory in latent space and decoding.  

\section{Future work}
\label{EncODEDec:future}

There are several future directions we are looking at:
\begin{itemize}
    \item We would first like to improve the results for 2-digit Moving MNIST. As of the time of writing this paper, we are already making progress in this direction.
    
    \item \textbf{Scaling up}: We would like to scale this up to bigger datasets such as KTH~\citep{schuldt2004recognizing}, the Kinetics dataset~\citep{kay2017kinetics}, etc. As of the time of writing this paper, we are 
    already making progress in this direction.
    
    \item \textbf{Fair comparison}: We would like to explore how well it performs when conditioning on some frames and training to predict the subsequent frames, as it is typically tackled by many of the recent papers on video generation~\citep{denton2018SVGLP, babaeizadeh2018stochastic, Lee2018StochasticAV}, so we can make a fair comparison of our approach with these methods.
    
    \item \textbf{Disentanglement}: We would like to examine the latent representation created by the Neural ODE in this domain in more detail. We would like to explore whether it implicitly disentangles spatial and temporal information, which it seems to be doing so from the evidence so far.
    
    \item \textbf{Visualization}: We would like to visualize the latent representation in lower dimensional space, to check the evidence of trajectories as being enforced by the Neural ODE. The best reason to use Neural ODE is so that the latent representation is now more interpretable i.e. consecutive time steps lie on a lower-dimensional trajectory. We would like to prove this, and show how exploring in latent space maps to intuitive changes in the decoded frames. We plan on using tools such as t-SNE~\citep{vanDerMaaten2008} and UMAP~\citep{McInnes2018UMAPUM} for visualization.
    
    \item \textbf{Temporal interpolation of videos}: Since videos follow continuous dynamics in latent space, it is possible to sample latent points for fractional time steps, i.e. time steps that are in the continuous range \textit{between} the time steps of the original video, and decode them using the same decoder. Hence, it should be possible to increase the frame rate of any given video without additional effort. This also opens the door to the use of learned representation for other downstream tasks in video.
    
    \item \textbf{Better metric for evaluation}: We would also like to have a better metric to estimate the quality of generated videos. The shortcomings of the current metrics of PSNR and SSIM have been mentioned~\citep{Lee2018StochasticAV}. More recently, \citep{clark2019adversarial} use the popular image quality metrics of Inception Score~\citep{salimans2016improved} and FID~\citep{heusel2017gans}, however, these metrics do not necessarily account for consistency in temporal information. Since many recent models have a stochastic component, it is all the more important to be able to effectively measure the difference in the real and predict distributions.
\end{itemize}

\section{Conclusion}

In this paper, we explored the use of Neural ODEs for video prediction. We showed very promising results on the 1-digit and 2-digit Moving MNIST dataset. We investigated two different architectures, with and without a recurrent component, respectively stochastic and deterministic. Even though we formulated the training problem as reconstruction, we were able to use our model for prediction of future frames because the Neural ODE learns the temporal evolution of continuous-time dynamics governing the latent features. We discussed in detail many future directions that would be useful to support our current approach, as well as help the space of video generation. We also discussed how our approach could be directly applied to other tasks such as temporal interpolation of videos. We hope that the research community uses our approach and takes advantage of the implicit feature of Neural ODEs to model continuous dynamics.

%% file: MRCNF.tex
\anglais
\counterwithin{figure}{chapter}
\counterwithin{table}{chapter}

\chapter{Multi-Resolution Continuous Normalizing Flows
~\citep{voleti2021improving}
}
\label{chap:MRCNF}

\setcounter{section}{-1}
\section{Prologue to article}
\label{chap:pro_MRCNF}

\subsection{Article details}

\textbf{Improving Continuous Normalizing Flows using a Multi-Resolution Framework}. Vikram Voleti, Chris Finlay, Adam Oberman, Christopher Pal. \textit{International Conference on Machine Learning 2021 Workshop (also under review at a journal)}.

\textit{Personal contribution}:
This project began when Vikram Voleti contacted Chris Finlay, then a PhD candidate at McGill University with Prof. Adam Oberman. Chris Finlay had earlier worked on a publication that improved the dynamics of Neural ODEs for image generation, and Vikram had worked on a project that used Neural ODEs for video generation. Vikram Voleti and Chris Finlay brainstormed over ideas for improving image generation using the continuous normalizing flows framework of Neural ODEs. Adam Oberman and Christopher Pal provided advice and guidance throughout the project and wrote parts of the paper. With help from Adam Oberman and Christopher Pal, Vikram derived the mathematical framework. With help from Chris Finlay, Vikram designed the experiments, wrote the code, ran experiments, proposed and executed on out-of-distribution analysis, and wrote the paper.

\subsection{Context}

Neural Ordinary Differential Equations (Neural ODEs) can serve as generative models of images using the perspective of Continuous Normalizing Flows (CNFs). Such models offer exact likelihood calculation, and invertible generation/density estimation. However, they had not been used in a multi-resolution framework yet, and most implementations using normalizing flows had very high number of parameters as well as high training time. A concurrent work called WaveletFlow also used a multi-resolution framework, however it incurred high parameter cost and training time.

\subsection{Contributions}

In this work we introduce a Multi-Resolution variant of CNFs called Multi-Resolution CNF (MRCNF), by characterizing the conditional distribution over the additional information required to generate a fine image that is consistent with the coarse image. We introduce a transformation between resolutions that allows for no change in the log likelihood. We show that this approach yields comparable likelihood values for various image datasets, with improved performance at higher resolutions, with fewer parameters, using only one GPU. Further, we examine the out-of-distribution properties of MRCNFs, and find that they are similar to those of other likelihood-based generative models.

\subsection{Research impact}

This work derived the use of continuous normalizing flows in the context of image generation in a multi-resolution framework. Although there are concurrent and follow-up works that also show the validity of our approach~\citep{yu2020waveletflow}, the adoption of our proposed continuous normalizing flows framework however hasn’t been as widespread as we had hoped.

\newpage

\begin{figure*}[!thb]
\begin{center}
\includegraphics[width=0.9\linewidth]{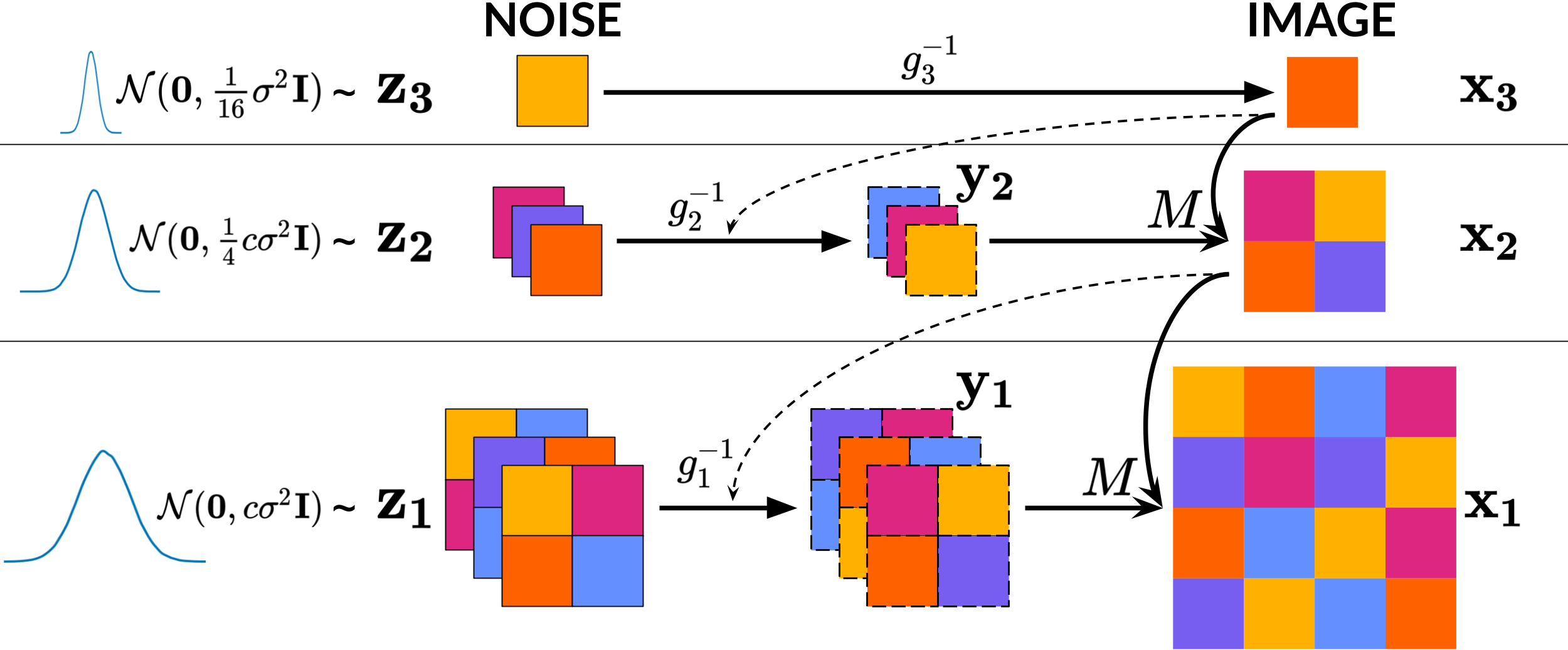}
\end{center}
\caption{The architecture of our \gls{msflow-image} method (best viewed in color). Continuous normalizing flows (CNFs) $g_s$ are used to generate images $\rvx_s$ from noise $\rvz_s$ at each resolution, with those at finer resolutions conditioned (dashed lines) on the coarser image one level above $\rvx_{s+1}$, except at the coarsest level where it is unconditional. Every finer CNF produces an intermediate image $\rvy_s$, which is then combined with the immediate coarser image $\rvx_{s+1}$ using a linear map $M$ from \autoref{eq:M} to form $\rvx_s$. The multiscale maps are defined by \autoref{eq:gen_cnf}.
}
\label{fig:mrcnf_2d_gen}
\end{figure*}

\section{Introduction}
\label{section:introduction}
Reversible generative models derived through the use of the change of variables technique~\citep{dinh2016density, kingma2018glow, ho2019flow++, yu2020waveletflow} are growing in interest as alternatives to generative models based on Generative Adversarial Networks (GANs)~\citep{goodfellow2016deep} and Variational Autoencoders (VAEs)~\citep{kingma2013auto}. While GANs and VAEs have been able to produce visually impressive samples of images, they have a number of limitations. A change of variables approach facilitates the transformation of a simple base probability distribution into a more complex model distribution. Reversible generative models using this technique are attractive because they enable efficient density estimation, efficient sampling, and computation of exact likelihoods.

A promising variation of the change-of-variable approach is based on the use of a continuous time variant of normalizing flows \citep{chen2018neural,grathwohl2019ffjord, finlay2020rnode}, which uses an integral over continuous time dynamics to transform a base distribution into the model distribution, called \gls{cnf}. This approach uses ordinary differential equations (ODEs) specified by a neural network, or Neural ODEs. \gls{cnf}s have been shown to be capable of modelling complex distributions such as those associated with images.

While this new paradigm for the generative modelling of images is not as mature as GANs or VAEs in terms of the generated image quality, it is a promising direction of research as it does not have some key shortcomings associated with GANs and VAEs. Specifically, GANs are known to suffer from mode-collapse~\citep{lin2018pacgan}, and are notoriously difficult to train~\citep{arjovsky2017towards} compared to feed forward networks because their adversarial loss seeks a saddle point instead of a local minimum~\citep{berard2019closer}. \gls{cnf}s are trained by mapping images to noise, and their reversible architecture allows images to be generated by going in reverse, from noise to images. This leads to fewer issues related to mode collapse, since any input example in the dataset can be recovered from the flow
using the reverse of the transformation learned during training. VAEs only provide a lower bound on the marginal likelihood whereas \gls{cnf}s provide exact likelihoods. Despite the many advantages of reversible generative models built with \gls{cnf}s, quantitatively such methods still do not match the widely used Fréchet Inception Distance (FID) scores of GANs or VAEs. However their other advantages motivate us to explore them further.

Furthermore, state-of-the art GANs and VAEs exploit the multi-resolution properties of images, and recent top-performing methods also inject noise at each resolution~\citep{brock2019biggan,shaham2019singan,karras2020analyzing,vahdat2020nvae}. While shaping noise is fundamental to normalizing flows, only recently have normalizing flows exploited the multi-resolution properties of images. For example, WaveletFlow~\citep{yu2020waveletflow} splits an image into multiple resolutions using the Discrete Wavelet Transform, and models the average image at each resolution using a normalizing flow. While this method has advantages, it suffers from many issues such as high parameter count and long training time.

In this work, we consider a non-trivial multi-resolution approach to continuous normalizing flows, which fixes many of these issues.
A high-level view of our approach is shown in \autoref{fig:mrcnf_2d_gen}.
%
%
Our main contributions are:
\begin{enumerate}
    \item We propose a multi-resolution transformation that does not add cost in terms of likelihood.
    \item We introduce \textbf{\gls{mrcnf}}.
    \item We achieve comparable \gls{bpd} (negative log likelihood per pixel) on image datasets using fewer model parameters and significantly less training time with only one GPU.
    \item We explore the out-of-distribution properties of (MR)CNF, and find that they are similar to non-continuous normalizing flows.
\end{enumerate}

\section{Background}
\label{background}

\subsection{Normalizing flows}

Normalizing flows~\citep{tabak2013family, jimenez2015variational, dinh2016density, papamakarios2019normalizing, kobyzev2020normalizing} are generative models that map a complex data distribution, such as real images, to a known noise distribution. They are trained by maximizing the log likelihood of their input images. Suppose a normalizing flow $g$ produces output $\rvz$ from an input $\rvx$ i.e. $\rvz = g(\rvx)$. The change-of-variables formula provides the likelihood of the image under this transformation as:
\begin{align}
\label{eq:change_of_variables}
    \log p(\rvx) = \log \Abs{\det\frac{\D g}{\D \rvx}} + \log p(\rvz) .
\end{align}
The first term on the right (log determinant of the Jacobian) is often intractable, however, previous works on normalizing flows have found ways to estimate this efficiently. The second term, $\log p(\rvz)$, is computed as the log probability of $\rvz$ under a known noise distribution, typically the standard Gaussian $\mathcal{N}(\mathbf{0}, \mathbf{I})$.
The normalizing flow is trained by maximizing the log-likelilhood of the data $\rvx$ in the real distribution i.e. $\log p(\rvx)$, using \autoref{eq:change_of_variables}.

\subsection{Continuous Normalizing Flows (CNF)}
\label{subsec:cnf}

Continuous Normalizing Flows (CNF)~\citep{chen2018neural, grathwohl2019ffjord, finlay2020rnode} are a variant of normalizing flows that operate in the continuous domain.
A \gls{cnf} creates a geometric flow between the input and target (noise) distributions, by assuming that the state transition is governed by an Ordinary Differential Equation (ODE). It further assumes that the differential function is parameterized by a neural network, this model is called a Neural ODE~\citep{chen2018neural}. Suppose \gls{cnf} $g$ transforms its state $\rvv(t)$ using a Neural ODE, with the differential defined by the neural network $f$ parameterized by $\theta$. $\rvv(t_0)=\rvx$ is, say, an image, and at the final time step $\rvv(t_1)=\rvz$ is a sample from a known noise distribution.
\begin{align}
\frac{\D \rvv(t)}{\D t} &= f(\rvv(t), t, \theta) \implies \rvv(t_1) = g(\rvv(t_0)) = \rvv(t_0) + \int_{t_0}^{t_1} f(\rvv(t), t, \theta)\ \D t .
\label{eq:neural_ode}
\end{align}
This integration is typically performed by an ODE solver. Since this integration can be run backwards as well to obtain the same $\rvv(t_0)$ from $\rvv(t_1)$, a \gls{cnf} is a reversible model.
\autoref{eq:change_of_variables} can be used to compute the change in log-probability induced by the \gls{cnf}. However, \citet{chen2018neural} and \citet{grathwohl2019ffjord} proposed a more efficient variant in the \gls{cnf} context, the instantaneous change-of-variables formula:
\begin{align}
&\frac{\partial \log p(\rvv(t))}{\partial t} = -\text{Tr}\left(\frac{\partial f_\theta}{\partial \rvv(t)}\right) ,
\\
\implies
&\Delta\log p_{\rvv(t_0) \rightarrow \rvv(t_1)} = -\int_{t_0}^{t_1}\text{Tr}\left(\frac{\partial f_\theta}{\partial \rvv(t)} \right) \D t .
\label{eq:inst}
\end{align}
Hence, the change in log-probability of the state of the Neural ODE i.e. $\Delta\log p_\rvv$ is expressed as another differential equation. The ODE solver now solves both differential equations \autoref{eq:neural_ode} and \autoref{eq:inst} by augmenting the original state with the above. Thus, a \gls{cnf} provides both the final state $\rvv(t_1)$ as well as the change in log probability $\Delta\log p_{\rvv(t_0) \rightarrow \rvv(t_1)}$ together.

Prior works~\citep{grathwohl2019ffjord, finlay2020rnode, ghosh2020steer, onken2021otflow, huang2021acc} have trained CNFs as reversible generative models of images by maximizing the image likelihood:
\begin{align}
\rvz = g(\rvx) \qquad ; \qquad
\log p(\rvx) = \Delta\log p_{\rvx \rightarrow \rvz} + \log p(\rvz) .
\label{eq:cnf}
\end{align}
where $\rvx$ is an image, $\rvz$ and $\Delta\log p_{\rvx \rightarrow \rvz}$ are computed by the CNF using \autoref{eq:neural_ode} and \autoref{eq:inst}, and $\log p(\rvz)$ is the likelihood of $\rvz$ under a known noise distribution, typically the standard Gaussian $\mathcal{N}(\mathbf{0}, \mathbf{I})$. Novel images are generated by sampling $\rvz$ from the noise distribution, and running the CNF in reverse.

\section{Our method}
\label{method}
Our method is a reversible generative model of images that builds on top of \gls{cnf}s. We introduce the notion of multiple resolutions in images, and connect the different resolutions in an autoregressive fashion. This helps generate images faster, with better likelihood values at higher resolutions, using only one GPU in all our experiments. We call this model Multi-Resolution Continuous Normalizing Flow (MRCNF).

\subsection{Multi-resolution image representation}
\label{sec:MRimage}
Multi-resolution representations of images have been explored in computer vision for decades~\citep{burt1981fast, marr2010vision, witkin1987scale, burt1983laplacian, mallat1989theory, lindeberg1990scale}. Much of the content of an image at a resolution is a composition of low-level information captured at coarser resolutions, and high-level information not present in the coarser images. We take advantage of this by first decomposing an image in \emph{resolution space} i.e. by expressing it as a series of $S$ images at decreasing resolutions: $\rvx \rightarrow (\rvx_1, \rvx_2, \dots, \rvx_S)$, where $\rvx_1=\rvx$ is the finest image, $\rvx_S$ is the coarsest, and every $\rvx_{s+1}$ is the average image of $\rvx_s$.
Thus, each $\rvx_j, j>i$ is deterministic if $\rvx_i$ is given.
This called an image pyramid, or a Gaussian Pyramid if the upsampling-downsampling operations include a Gaussian filter~
\citep{burt1981fast, burt1983laplacian, adelson1984pyramid, witkin1987scale, lindeberg1990scale}. In this work, we obtain a coarser image simply by averaging pixels in every 2x2 patch, thereby halving the width and height.
, i.e. we express a $32\times32$ image into, say, 3 images of resolutions: ($8\times8$, $16\times16$, $32\times32$), each image being an average of its immediate finer image.
However, this representation is redundant since much of the information in $\rvx_1$ is contained in $\rvx_{s>1}$. Instead,
We then express $\rvx$ as a series of high-level information $\rvy_s$ not present in the immediate coarser images $\rvx_{s+1}$, and a final coarse image $\rvx_S$, and our overall method is to map these $S$ terms to $S$ noise samples using $S$ CNFs.:
\begin{align}
\rvx \rightarrow (\rvy_1, \rvx_2) \rightarrow (\rvy_1, \rvy_2, \rvx_3) \rightarrow \dots \rightarrow (\rvy_1, \rvy_2, \dots, \rvy_{S-1}, \rvx_S) .
\label{eq:multi-resolution}
\end{align}


\subsection{Defining the high-level information \texorpdfstring{$\rvy_s$}{}}
\label{subsec:y}

\begin{figure*}[!htb]
\centering
\includegraphics[width=.5\linewidth]{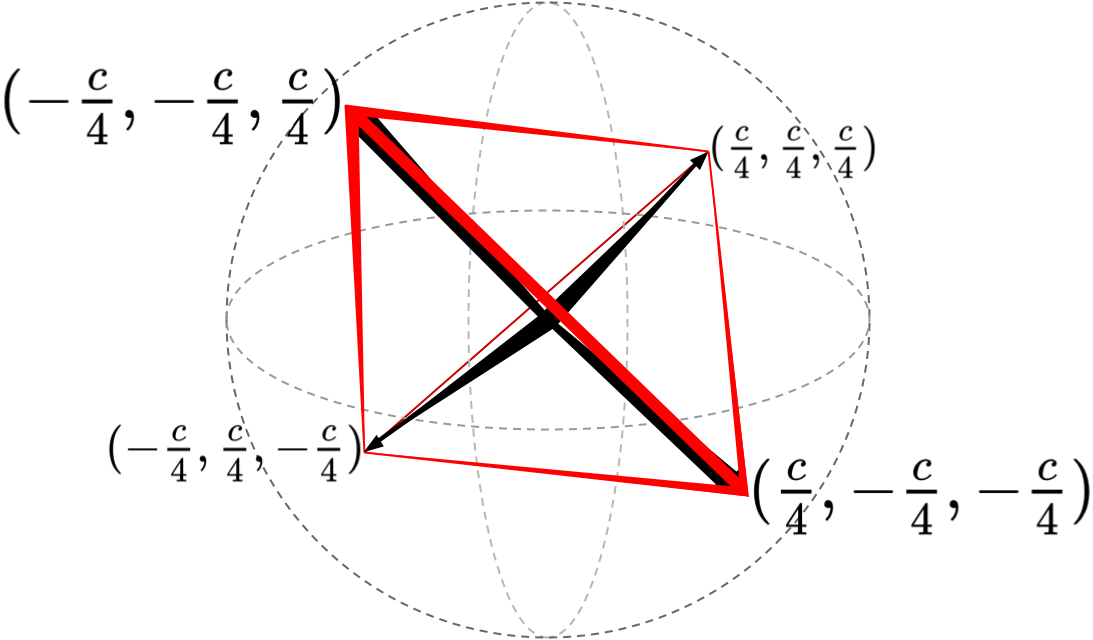}
\caption{Tetrahedron in 3D space with 4 corners. \texorpdfstring{$c = 2^{2/3}$}{} }
\label{fig:Tetrahedron}
\end{figure*}

We choose to design a linear transformation with the following properties: 1) invertible i.e. it should be possible to deterministically obtain $\rvx_s$ from $\rvy_s$ and $\rvx_{s+1}$, and vice versa ;
2) volume preserving i.e. determinant is 1, change in log-likelihood is 0 ; 3) angle preserving ; and 4) range preserving.





Consider the simplest case of 2 resolutions where $\rvx_1$ is a 2x2 image with pixel values $x_1, x_2, x_3, x_4$, and $\rvx_2$ is a 1x1 image with pixel value $\xbar = \frac{1}{4}(x_1 + x_2 + x_3 + x_4)$. We require three values $(y_1, y_2, y_3) = \rvy_1$ that contain information not present in $\rvx_2$, such that $\rvx_1$ is obtained when $\rvy_1$ and $\rvx_2$ are combined.

This could be viewed as a problem of finding a matrix $\rmM$ such that: $[x_1, x_2, x_3, x_4]^\top = \rmM\ [y_1, y_2, y_3, \xbar]^\top$. We fix the last column of $\rmM$ as $[1, 1, 1, 1]^\top$, since every pixel value in $\rvx_1$ depends on $\barx$. Finding the rest of the parameters can be viewed as requiring four 3D vectors that are spaced such that they do not degenerate the number of dimensions of their span. These can be considered as the four corners of a tetrahedron in 3D space,
under any configuration (rotated in 3D space), and any scaling of the vectors (see \autoref{fig:Tetrahedron}).

Out of the many possibilities for this tetrahedron is
the matrix that performs the Discrete Haar Wavelet Transform~\citep{mallat1989theory, mallat2009wavelet}:
\begin{align}
\label{haar}
\begin{bmatrix}
x_1 \\ x_2 \\ x_3 \\ x_4
\end{bmatrix}
=
\begin{bmatrix}
\ \ \ \frac{1}{2} & \ \ \ \frac{1}{2} & \ \ \ \frac{1}{2} & 1 \\
\ \ \ \frac{1}{2} & -\frac{1}{2} & -\frac{1}{2} & 1 \\
-\frac{1}{2} & \ \ \ \frac{1}{2} & -\frac{1}{2} & 1 \\
-\frac{1}{2} & -\frac{1}{2} & \ \ \ \frac{1}{2} & 1
\end{bmatrix}
\begin{bmatrix}
y_1 \\ y_2 \\ y_3 \\ \xbar
\end{bmatrix}
\iff
\begin{bmatrix}
y_1 \\ y_2 \\ y_3 \\ \xbar
\end{bmatrix}
=
\begin{bmatrix}
\frac{1}{2} & \ \ \ \frac{1}{2} & -\frac{1}{2} & -\frac{1}{2} \\
\frac{1}{2} & -\frac{1}{2} & \ \ \ \frac{1}{2} & -\frac{1}{2} \\
\frac{1}{2} & -\frac{1}{2} & -\frac{1}{2} & \ \ \ \frac{1}{2} \\
\frac{1}{4} & \ \ \ \frac{1}{4} & \ \ \ \frac{1}{4} & \ \ \ \frac{1}{4}
\end{bmatrix}
\begin{bmatrix}
x_1 \\ x_2 \\ x_3 \\ x_4
\end{bmatrix}
.
\end{align}
However, this transformation incurs a cost in terms of log-likelihood:
\begin{align}
\Delta\log p_{(x_1, x_2, x_3, x_4) \rightarrow (y_1, y_2, y_3, \barx)} = \log \Abs{\det (\rmM^{-1})} = \log(1/2) .
\label{eq:haar_logabsdet}
\end{align}
and is therefore not volume preserving. Other simple scaling of \autoref{haar} has been used in the past, for example multiplying the last row by 2, yielding an orthogonal transformation, such as in WaveletFlow~\citep{yu2020waveletflow}. However, this transformation neither preserves the volume i.e. the log determinant is not 0, nor the maximum i.e. the range of $\rvx_s$ changes.  

We wish to find a transformation $\rmM$ where: one of the results is the average of the inputs, $\bar x$; it is unit determinant; the columns are orthogonal; and it preserves the range of $\bar x$.
Fortunately such a matrix exists -- although we have not seen it discussed in prior literature. It can be seen as a variant of the Discrete Haar Wavelet Transformation matrix that is unimodular, i.e. has a determinant of $1$ (and is therefore volume preserving). It is also preserving the range of the images for the input and its average, , i.e. it is range preserving such that $\min(x) = \min(\barx)$ and $\max(x) = \max(\barx)$.
It is obtained by multiplying the 3 rows corresponding to the non-average terms with $2^{2/3}$. This is shown in \autoref{fig:Tetrahedron}):
\begin{align}
\label{Mdefn}
\begin{bmatrix}
x_1 \\ x_2 \\ x_3 \\ x_4
\end{bmatrix}
=
\frac{1}{a}
\begin{bmatrix}
\ \ \ c & \ \ \ c & \ \ \ c & a \\
\ \ \ c & -c & -c & a \\
-c & \ \ \ c & -c & a \\
-c & -c & \ \ \ c & a
\end{bmatrix}
\begin{bmatrix}
y_1 \\ y_2 \\ y_3 \\ \xbar
\end{bmatrix}
\hspace{-.5em}
\iff
\hspace{-.5em}
\begin{bmatrix}
y_1 \\ y_2 \\ y_3 \\ \xbar
\end{bmatrix}
=
\begin{bmatrix}
c^{-1} & \ \ \ c^{-1} & -c^{-1} & -c^{-1}\\
c^{-1} & -c^{-1} & \ \ \ c^{-1} & -c^{-1}\\
c^{-1} & -c^{-1} & -c^{-1} & \ \ \ c^{-1}\\
a^{-1} & \ \ \ a^{-1} & \ \ \ a^{-1} & \ \ \ a^{-1}
\end{bmatrix}
\begin{bmatrix}
x_1 \\ x_2 \\ x_3 \\ x_4
\end{bmatrix}
,
\end{align}
where $c = 2^{2/3}$, $a=4$. Hence, there is no cost to the log-likelihood due to the transformation:
\begin{align}
\Delta\log p_{(x_1, x_2, x_3, x_4) \rightarrow (y_1, y_2, y_3, \barx)} = \log \Abs{\det (\rmM^{-1})} = \log(1) = 0 .
\label{eq:Mdefn_logabsdet}
\end{align}

This can be scaled up to larger spatial regions by performing the same calculation for each 2x2 patch. Let $M$ be the function that uses matrix $\rmM$ from above and combines every pixel in $\rvx_{s+1}$ with the three corresponding pixels in $\rvy_s$ to make the 2x2 patch at that location in $\rvx_s$ using \autoref{Mdefn}:
\begin{align}
\rvx_s = M(\rvy_s, \rvx_{s+1}) \iff \rvy_s, \rvx_{s+1} = M^{-1}(\rvx_s) .
\label{eq:M}
\end{align}
\autoref{eq:change_of_variables} can be used to compute the change in log likelihood from this transformation $\rvx_s \rightarrow (\rvy_s, \rvx_{s+1})$:
\begin{align*}
\log p(\rvx_s)
&= \Delta \log p_{\rvx_s \rightarrow (\rvy_s, \rvx_{s+1})} + \log p(\rvy_s, \rvx_{s+1}) ,
\label{eq:multires_logp} \\
&= \log \Abs{\det (M^{-1})} + \log p(\rvy_s, \rvx_{s+1}) ,
\numberthis
\end{align*}
where $\log \Abs{\det (M^{-1})}$ can be determined for the Wavelet transform in \autoref{haar} using:
\begin{align}
\log \Abs{\det (M^{-1})} = \text{dims}(\rvx_{s+1})\log (1/2) ,
\label{eq:logabsdet_half}
\end{align}
where ``dims'' is the number of pixels times the number of channels (typically 3) in the image, and for our unimodular transform in \autoref{Mdefn} using \autoref{eq:Mdefn_logabsdet} as
:
\begin{align}
\log \Abs{\det (M^{-1})} = 0 .
\label{eq:logabsdet0}
\end{align}

\subsection{Multi-Resolution Continuous Normalizing Flows (MRCNF)}

Using the multi-resolution image representation in \autoref{eq:multi-resolution}, we characterize the conditional distribution over the additional degrees of freedom ($\rvy_s$) required to generate a higher resolution image ($\rvx_s$) that is consistent with the average ($\rvx_{s+1}$) over the equivalent pixel space. At each resolution $s$, we use a CNF to reversibly map between $\rvy_s$ (or $\rvx_S$ when $s\mkern1mu{=}\mkern1mu S$) and a sample $\rvz_s$ from a known noise distribution. For generation, $\rvy_s$ only adds information missing in $\rvx_{s+1}$, but conditional on it.

This framework ensures a coarse image could generate several potential fine images, but these fine images share the same coarse image as their average. This fact is preserved across resolutions. Note that the 3 additional pixels in $\rvy_s$ per pixel in $\rvx_{s+1}$ are generated conditioned on the entire coarser image $\rvx_{s+1}$, thus maintaining consistency using full context.

In principle, any generative model could be used to map between the multi-resolution image and noise. Normalizing flows are good candidates for this as they are probabilistic generative models that perform exact likelihood estimates, and can be run in reverse to generate novel data from the model's distribution. This allows model comparison and measurement of generalization to unseen data. We choose to use the CNF variant of normalizing flows at each resolution. CNFs have recently been shown to be effective in modeling image distributions using a fraction of the number of parameters typically used in normalizing flows (and non flow-based approaches), and their underlying framework of Neural ODEs have been shown to be more robust than convolutional layers~\citep{yan2020robustness}.


\textbf{Training}: We train an MRCNF by maximizing the average log-likelihood of the images in the training dataset under the model.
The log probability of each image $\log p(\rvx)$ can be estimated recursively using the sequence of variables in \autoref{eq:multi-resolution}, and the corresponding simplification of the log-probability using \autoref{eq:multires_logp} as (here, $\rvx_1 = \rvx$):
\begin{align*}
\log p(\rvx) &= \Delta\log p_{\rvx_1 \rightarrow (\rvy_1, \rvx_2)} + \log p(\rvy_1, \rvx_2) ,\\
&= \Delta\log p_{\rvx_1 \rightarrow (\rvy_1, \rvx_2)} + \log p(\rvy_1 \mid \rvx_2) + \log p(\rvx_2) ,\\
&= \Delta\log p_{\rvx_1 \rightarrow (\rvy_1, \rvx_2)} + \log p(\rvy_1 \mid \rvx_2) \\
&\qquad + \Delta\log p_{\rvx_2 \rightarrow (\rvy_2, \rvx_3)} + \log p(\rvy_2 \mid \rvx_3) + \log p(\rvx_3) ,\\
\vdots\\
&= \sum_{s=1}^{S-1}\left(\Delta\log p_{\rvx_s \rightarrow (\rvy_s, \rvx_{s+1})} + \log p(\rvy_s \mid \rvx_{s+1})\right) + \log p(\rvx_S) . \numberthis
\label{eq:loglikelihood_recurse}
\end{align*}

where $\Delta\log p_{\rvx_s \rightarrow (\rvy_s, \rvx_{s+1})}$ is given by \autoref{eq:logabsdet0},
while $\log p(\rvy_s \mid \rvx_{s+1})$ and $\log p(\rvx_S)$ (at the coarsest resolution $S$) are given by \autoref{eq:cnf}:
\begin{align}
\label{eq:train_cnf}
&\rvz_s = g_s(\rvy_s \mid \rvx_{s+1}) ; \quad \log p(\rvy_s \mid \rvx_{s+1}) = \Delta\log p_{(\rvy_s \rightarrow \rvz_s) \mid \rvx_{s+1}} + \log p(\rvz_s) ,
\\
&\rvz_S = g_S(\rvx_S) ; \qquad \quad \log p(\rvx_S) = \Delta\log p_{\rvx_S \rightarrow \rvz_S} + \log p(\rvz_S) .
\end{align}

The coarsest resolution $S$ can be chosen such that the last CNF operates on the image distribution at a small enough resolution that is easy to model unconditionally. All other CNFs are conditioned on the immediate coarser image. The conditioning itself is achieved by concatenating the input image of the CNF with the coarser image. This model could be seen as a stack of CNFs connected in an autoregressive fashion.

Typically, likelihood-based generative models are compared using the metric of bits-per-dimension (BPD), i.e. the negative log likelihood per pixel in the image. Hence, we train our MRCNF to minimize the average BPD of the images in the training dataset, computed using \autoref{eq:bpd}:
\begin{align}
\text{BPD}(\rvx) = -\log p(\rvx)/\text{dims}(\rvx) .
\label{eq:bpd}
\end{align}

We use FFJORD~\citep{grathwohl2019ffjord} as the baseline model for our \gls{cnf}s. In addition, we use to two regularization terms introduced by RNODE~\citep{finlay2020rnode} to speed up the training of FFJORD models by stabilizing the learnt dynamics: the kinetic energy $\mathcal{K}(\theta)$ and the Jacobian norm $\mathcal{B}(\theta)$ of the flow $f(\rvv(t), t, \theta)$ described in \autoref{subsec:cnf}:
\begin{align}
\mathcal{K}(\theta) &= \int_{t_0}^{t_1} \| f(\rvv(t), t, \theta) \|_2^2\ \D t
\quad ; \\
\mathcal{B}(\theta) &= \int_{t_0}^{t_1} \| \epsilon^\top \nabla_z f(\rvv(t), t, \theta) \|_2^2\ \D t, \quad \epsilon \sim \mathcal{N}(0, I) .
\end{align}

\textbf{Parallel training:} Note that although the final log likelihood $\log p(\rvx)$ involves sequentially summing over values returned by all $S$ CNFs, the log likelihood term of each CNF is independent of the others. Conditioning is done using ground truth images. Hence, each CNF can be trained independently, in parallel.

\textbf{Generation}: Given an $S$-resolution model, we first sample $\rvz_s, s={1,\dots,S}$ from the latent noise distributions. The \gls{cnf} $g_s$ at resolution $s$ transforms the noise sample $\rvz_s$ to high-level information $\rvy_s$ conditioned on the immediate coarse image $\rvx_{s+1}$ (except $g_S$ which is unconditioned). $\rvy_s$ and $\rvx_{s+1}$ are then combined to form $\rvx_s$ using $M$ from \autoref{Mdefn}. This process is repeated progressively from coarser to finer resolutions, until the finest resolution image $\rvx_1$ is computed (see \autoref{fig:mrcnf_2d_gen}). It is to be noted that the generated image at one resolution is used to condition the CNF at the finer resolution.
\begin{align*}
\begin{cases}
\rvx_S = g_S^{-1}(\rvz_S) \qquad\qquad &s=S, \\
\rvy_s = g_s^{-1}(\rvz_s \mid \rvx_{s+1}) ; \quad
\rvx_s = M(\rvy_s, \rvx_{s+1}) \qquad\qquad
&s = S\text{-}1 \rightarrow 1 .
\end{cases}
\numberthis
\label{eq:gen_cnf}
\end{align*}



\subsection{Multi-resolution noise}
\label{subsec:mrnoise}

We further decompose the noise image as well into its respective coarser components. This means ultimately we use only one noise image at the finest level, and it is decomposed into multiple resolutions using \autoref{Mdefn}. $\rvx_{s+1}$ is mapped to noise of $1/4$ variance, $\rvy_s$ is mapped to noise of $c$-factored variance (see \autoref{fig:mrcnf_2d_gen}). Although this is optional, it preserves interpretation between the single- and multi-resolution models.

\subsection{8-bit to uniform}
The change-of-variables formula gives the change in log probability from the transformation of $\rvu$ to $\rvv$:
\begin{align*}
    \log p(\rvu) = \log p(\rvv) + \log \Abs{\det\frac{\D \rvv}{\D \rvu}} .
\end{align*}

Specifically, the change from an 8-bit image to one with pixel values in range [0, 1] is:
\begin{align*}
&\rvb_S^{(p)} = \frac{\rva_S^{(p)}}{256} ,\\
&\implies \log p(\rva_S) = \log p(\rvb_S) + \log \Abs{\det\frac{\D \rvb}{\D \rva}} ,\\
&\implies \log p(\rva_S) = \log p(\rvb_S) + \log \left(\frac{1}{256}\right)^{D_S}, \\
&\implies \log p(\rva_S) = \log p(\rvb_S) - D_S \log 256 .\\
\implies
&\text{bpd}(\rva_S)
= \frac{-\log p(\rva_S)}{D_S \log 2} ,\\
&= \frac{-(\log p(\rvb_S) - D_S \log 256)}{D_S \log 2} ,\\
&= \frac{-\log p(\rvb_S)}{D_S \log 2} + \frac{\log 256}{\log 2} ,\\
&= \text{bpd}(\rvx) + 8 ,
\end{align*}
where $\text{bpd}(\rvx)$ is given from~\autoref{eq:bpd}.

\subsection{Simple example of density estimation in MRCNF}

If Euler method is used as the ODE solver, density estimation \autoref{eq:neural_ode} reduces to:
\begin{align}
\rvv(t_1) = \rvv(t_0) + (t_1 - t_0)f_s(\rvv(t_0), t_0  \mid \rvc),
\end{align}
where $f_s$ is a neural network, $t_0$ is the ``time'' at which the state is image $\rvx$, and $t_1$ when it is noise $\rvz$.
We start at scale S with an image sample $\rvx_S$, and assume $t_0=0$ and $t_1=1$:
\begin{align}
\begin{cases}
&\rvz_S = \rvx_S + f_S(\rvx_S,\ t_0 \mid \rvx_{S-1}) ,\\
&\rvz_{S-1} = \rvx_{S-1} + f_{S-1}(\rvx_{S-1},\ t_0  \mid \rvx_{S-2}) ,\\
&\vdots\\
&\rvz_1 = \rvx_1 + f_1(\rvx_1,\ t_0 \mid \rvx_0) ,\\
&\rvz_0 = \rvx_0 + f_0(\rvx_0,\ t_0) .
\end{cases}
\label{eq:density_ex}
\end{align}
Then, density is estimated using \Cref{eq:loglikelihood_recurse}.

\subsection{Simple example of generation using MRCNF}

If Euler method is used as the ODE solver, generation \autoref{eq:neural_ode} reduces to:
\begin{align}
\rvv(t_0) = \rvv(t_1) + (t_0 - t_1)f_s(\rvv(t_1), t_1 \mid \rvc),
\end{align}
i.e. the state is integrated backwards from $t_1$ (i.e. $\rvz_s$) to $t_0$ (i.e. $\rvx_s$).
We start at scale 0 with a noise sample $\rvz_0$, and assume $t_0$ and $t_1$ are $0$ and $1$ respectively:
\begin{align}
\begin{cases}
&\rvx_0 = \rvz_0 - f_0(\rvz_0,\ t_1), \\
&\rvx_1 = \rvz_1 - f_1(\rvz_1,\ t_1 \mid \rvx_0) ,\\
&\vdots\\
&\rvx_{S-1} = \rvz_{S-1} - f_{S-1}(\rvz_{S-1},\ t_1 \mid \rvx_{S-2}),\\
&\rvx_S = \rvz_S - f_S(\rvz_{S},\ t_1 \mid \rvx_{S-1}).
\end{cases}
\label{eq:generation_ex}
\end{align}

\section{Related work}
\label{section:related}
Multi-resolution approaches already serve as a key component of state-of-the-art GAN~\citep{denton2015lapgan, karras2018progressive, karnewar2020msg} and VAE~\citep{razavi2019generating, vahdat2020nvae} based deep generative models.
The idea is to take advantage of the fact that much of the information in an image is contained in a coarsened version, which allows us to deal with simpler problems (coarser images) in a progressive fashion. This helps make models more efficient and effective.
Deconvolutional CNNs~\citep{long2015fully, radford2015unsupervised} use upsampling layers to generate images more effectively. Modern state-of-the-art generative models have also injected noise at different levels to improve sample quality \citep{brock2019biggan, karras2020analyzing,vahdat2020nvae}.
Several works~\citep{oord2016pixel, reed2017parallel, menick2019generating, razavi2019generating} have also shown how the inductive bias of the multi-resolution structure helps alleviate some of the problems of image quality in likelihood-based models.

Several prior works on normalizing flows~\citep{kingma2018glow, hoogeboom2019emerging, hoogeboom2019idf, song2019mintnet, ma2019macow, durkan2019spline, chen2020vflow, ho2019flow++, lee2020nanoflow, yu2020waveletflow} build on RealNVP~\citep{dinh2016density}. Although they achieve great results in terms of \gls{bpd} and image quality, they nonetheless report results from significantly higher number of parameters (some with 100x!), and several times GPU hours of training.

STEER~\citep{ghosh2020steer} introduced temporal regularization to CNFs by making the final time of integration stochastic. However, we found that this increased training time without significant BPD improvement.
\begin{align}
\text{STEER~\citep{ghosh2020steer}:}
\begin{cases}
\rvv(t_1) = \rvv(t_0) + \int_{t_0}^{T} f(\rvv(t), t)\ \D t ;
\\
T \sim \text{Uniform}(t_1-b, t_1+b);\ b < t_1 - t_0 .
\label{eq:steer}
\end{cases}
\end{align}

\textbf{``Multiple scales'' in prior normalizing flows}: Normalizing flows~\citep{dinh2016density, kingma2018glow, grathwohl2019ffjord} try to be ``multi-scale'' by transforming the input in a smart way (squeezing operation) such that the width of the features progressively reduces in the direction of image to noise, while maintaining the total dimensions. This happens while operating at a \textit{single resolution}. In contrast, our model stacks normalizing flows at \textit{multiple resolutions} in an autoregressive fashion by conditioning on the images at coarser resolutions.

Other classes of generative models that map from a complex distribution to a known noise distribution are Denoising diffusion probabilistic models (DDPM) \citep{sohl2015deep, ho2020ddpm, song2020sde} which use a predefined noising process, and score-based generative models \citep{song2019generative, song2020improved, jolicoeur2020adversarial, song2020sde} which estimate the gradient of the log density with respect to the input (i.e. the \textit{score}) of corrupted data with progressively lesser intensities of noise. In contrast, CNFs learn a reversible noising/denoising process using a Neural ODE.

\subsection{WaveletFlow}
\label{subsec:WaveletFlow}

WaveletFlow~\citep{yu2020waveletflow} is a recent innovation on the normalizing flow, wherein the image is decomposed into a lower-resolution average image, and 3 other informative components using the Discrete Wavelet Transformation. The 3 components at each resolution are mapped to noise using a normalizing flow conditioned on the average image at that resolution. WaveletFlow builds on the Glow~\citep{kingma2018glow} architecture. It uses an orthogonal transformation, which does not preserve range, and adds a constant term to the log likelihood at each resolution. Best results are obtained when WaveletFlow models with a high parameter count are trained for a long period of time. We fix these issues using our \gls{mrcnf}.

\textbf{Comparison to WaveletFlow}: We emphasize that there are important and crucial differences between our \gls{mrcnf} and WaveletFlow. We generalize the notion of a multi-resolution image representation (\autoref{subsec:y}), and show that Wavelets are one case of this general formulation. WaveletFlow builds on the Glow~\citep{kingma2018glow} architecture, while ours builds on \gls{cnf}s~\citep{grathwohl2019ffjord, finlay2020rnode}. We also make use of the notion of multi-resolution decomposition of the noise, which is optional, but is not taken into account by WaveletFlow. WaveletFlow uses an orthogonal transformation which does not preserve range ; our MRCNF uses \autoref{Mdefn} which is volume-preserving and range-preserving. Finally, WaveletFlow applies special sampling techniques to obtain better samples from its model. We have so far not used such techniques for generation, but we believe they can potentially help our models as well. By making these important changes, we fix many of the previously discussed issues with WaveletFlow. For a more detailed ablation study, please check \autoref{subsec:waveletflow_ablation}.

\section{Experimental details}
\label{mrcnf:results}

\subsection{Models}

We used the same neural network architecture as in RNODE~\citep{finlay2020rnode}. The CNF at each resolution consists of a stack of $bl$ blocks of a 4-layer deep convolutional network comprised of 3x3 kernels and softplus activation functions, with 64 hidden dimensions, and time t concatenated to the spatial input. In addition, except at the coarsest resolution, the immediate coarser image is also concatenated with the state. The integration time of each piece is [0, 1]. The number of blocks $bl$ and the corresponding total number of parameters are given in \autoref{tab:cifar_params}.

\begin{table}[htb]
\setlength{\tabcolsep}{2pt}
\centering
\caption{Number of parameters for different models with different total number of resolutions (res), and the number of channels (ch) and number of blocks (bl) per resolution.}
\begin{tabular}{ccc|c}
resolutions & ch & bl & Param\\
\hline
\multirow{3}{*}{1}
& 64 & 2 & \footnotesize{0.16M}\\[-0pt]
& 64 & 4 & \footnotesize{0.32M}\\[-0pt]
& 64 & 14 & \footnotesize{1.10M}\\
\hline
\multirow{3}{*}{2}
& 64 & 8 & \footnotesize{1.33M}\\[-0pt]
& 64 & 20 & \footnotesize{3.34M}\\[-0pt]
& 64 & 40 & \footnotesize{6.68M}\\[-0pt]
\hline
\multirow{3}{*}{3}
& 64 & 6 & \footnotesize{1.53M}\\[-0pt]
& 64 & 8 & \footnotesize{2.04M}\\[-0pt]
& 64 & 20 & \footnotesize{5.10M}\\[-0pt]
\end{tabular}
\label{tab:cifar_params}
\end{table}

\subsection{Gradient norm}

In order to avoid exploding gradients, We clipped the norm of the gradients~\citep{pascanu2013difficulty} by a maximum value of 100.0. In case of using adversarial loss, we first clip the gradients provided by the adversarial loss by 50.0, sum up the gradients provided by the log-likelihood loss, and then clip the summed gradients by 100.0.

\section{Experimental results}
\label{experiments}

\begin{table*}[!ht]
\begin{minipage}{\textwidth}
    \centering
    \caption{Unconditional image generation metrics (lower is better): parameters (P), bits-per-dimension, time (in hours). All our models were trained on only \textit{one} NVIDIA V100 GPU.
    }
    {\def\arraystretch{1.06}
    {\setlength{\tabcolsep}{.05em}
    \small
    \resizebox{\columnwidth}{!}{
    \begin{tabular}{|l|crc|rrr|crc|}
        \hline
        & \multicolumn{3}{c|}{\textbf{\textsc{CIFAR10}}} & \multicolumn{3}{c|}{\textbf{\textsc{ImageNet32}}} & \multicolumn{3}{c|}{\textbf{\textsc{ImageNet64}}} \\
        & \textsc{BPD} & \textsc{P} & \textsc{Time}
        & \textsc{BPD} & \textsc{P} & \textsc{Time}
        & \textsc{BPD} & \textsc{P} & \textsc{Time} \\
        \hline\hline
        \multicolumn{10}{|l|}{\textbf{Non Flow-based Prior Work}} \\
        \hline
        PixelRNN~\citep{oord2016pixel}
            & 3.00 & &
            & 3.86 & &
            & 3.63 & & \\[-2pt]
        Gated PixelCNN
        ~\citep{van2016conditional}
            & 3.03 &  &
            & 3.83 &  & 60
            & 3.57 &  & 60 \\[-2pt]
        Parallel Multiscale
        ~\citep{reed2017parallel}
            & & &
            & 3.95 & &
            & 3.70 & &  \\[-2pt]
        Image Transformer
        ~\citep{parmar2018image}
            & 2.90 &  &
            & 3.77 &  &
            &  &  &   \\[-2pt]
        PixelSNAIL
        ~\citep{chen2018pixelsnail}
            & 2.85 &  &
            & 3.80 &  &
            &  &  &   \\[-2pt]
        SPN
        ~\citep{menick2019generating}
            &  &  &
            & 3.85 & \footnotesize{150.0M} &
            & 3.53 & \footnotesize{150.0M} &   \\[-2pt]
        Sparse Transformer
        ~\citep{child2019generating}
            & 2.80 & \small{59.0M} &
            &  &  &
            & 3.44 & \footnotesize{152.0M} & \footnotesize{7days}  \\[-2pt]
        Axial Transformer
        ~\citep{ho2019axial}
            &  &  &
            & 3.76 &  &
            & 3.44 &  &   \\[-2pt]
        PixelFlow++
        ~\citep{nielsen2020closing}
            & 2.92 &  &
            &  &  &
            &  &  &  \\[-2pt]
        NVAE
        ~\citep{vahdat2020nvae}
            & 2.91 &  & 55
            & 3.92 &  & 70
            &  &  &   \\[-2pt]
        Dist-Aug
        Sparse Tx
        ~\citep{jun2020distaug}
            & 2.56 & \footnotesize{152.0M} &
            &  &  &
            & 3.42 & \footnotesize{152.0M} &  \\
        \hline
        \hline
        \multicolumn{10}{|l|}{\textbf{Flow-based Prior Work}} \\
        \hline
        IAF
        ~\citep{kingma2016improved}
            &  &  &
            & 3.11 & &
            &  &  &  \\
        RealNVP
        ~\citep{dinh2016density}
            & 3.49 & &
            & 4.28 & \small{46.0M} &
            & 3.98 & \small{96.0M} & \\[-2pt]
        Glow
        ~\citep{kingma2018glow}
            & 3.35 & \small{44.0M} &
            & 4.09 & \small{66.1M} &
            & 3.81 & \footnotesize{111.1M} & \\[-2pt]
        i-ResNets
        ~\citep{behrmann2019invertible}
            &  &  &
            &  &  &
            &  &  &  \\[-2pt]
        Emerging
        ~\citep{hoogeboom2019emerging}
            & 3.34 & \small{44.7M} &
            & 4.09 & \small{67.1M} &
            & 3.81 & \small{67.1M} & \\[-2pt]
        IDF
        ~\citep{hoogeboom2019idf}
            & 3.34 & &
            & 4.18 & &
            & 3.90 & & \\[-2pt]
        S-CONF
        ~\citep{karami2019invertible}
            & 3.34 & &
            & & &
            & & & \\[-2pt]
        MintNet
        ~\citep{song2019mintnet}
            & 3.32 & \small{17.9M} & \footnotesize{$\geq$5days}
            & 4.06 & \small{17.4M} &
            &  &  &  \\[-2pt]
        Residual Flow
        ~\citep{chen2019residual}
            & 3.28 & &
            & 4.01 & &
            & 3.76 & & \\[-2pt]
        MaCow
        ~\citep{ma2019macow}
            & 3.16 & \small{43.5M} &
            &  &  &
            & 3.69 & \footnotesize{122.5M} & \\[-2pt]
        Neural Spline Flows
        ~\citep{durkan2019spline}
            & 3.38 & \small{11.8M} &
            & & &
            & 3.82 & \small{15.6M} & \\[-2pt]
        Flow++
        ~\citep{ho2019flow++}
            & 3.08 & \small{31.4M} &
            & 3.86  & \footnotesize{169.0M} &
            & 3.69 & \small{73.5M} & \\[-2pt]
        ANF
        ~\citep{huang2020augmented}
            & 3.05 & &
            & 3.92 & &
            & 3.66 & & \\[-2pt]
        MEF
        ~\citep{xiao2020generative}
            & 3.32 & \small{37.7M} &
            & 4.05 & \small{37.7M} &
            & 3.73 & \small{46.6M} & \\[-2pt]
        VFlow
        ~\citep{chen2020vflow}
            & 2.98 & &
            & 3.83 & &
            &  &  &   \\[-2pt]
        Woodbury NF
        ~\citep{lu2020woodbury}
            & 3.47 & &
            & 4.20 & &
            & 3.87 & &  \\
        NanoFlow
        ~\citep{lee2020nanoflow}
            &  3.25 & &
            &  &  &
            &  &  &   \\
        ConvExp
        ~\citep{hoogeboom2020flow}
            &  3.218 & &
            &  &  &
            &  &  &  \\
        Wavelet Flow
        ~\citep{yu2020waveletflow}
            &  &  &
            & 4.08 & \small{64.0M} &
            & 3.78 & \small{96.0M} & 822  \\
        TayNODE
        ~\citep{kelly2020learning}
            & 1.039 & &
            & & &
            & & &  \\
        \hline
        \hline
        \multicolumn{10}{|l|}{\textbf{1-resolution Continuous Normalizing Flow}}\\
        \hline
        FFJORD
        ~\citep{grathwohl2019ffjord}
            & 3.40 & \small{0.9M} & \footnotesize{$\geq$5days}
            & $^{\ddagger}3.96$ & $^{\ddagger}$\small{2.0M} & $^{\ddagger}$\footnotesize{$>$5days}
            & x &  & x \\[-0pt]
        RNODE
        ~\citep{finlay2020rnode}
            & 3.38 & \small{1.4M} & 31.84
            & $^{\ddagger}2.36$ & \small{2.0M} & $^{\ddagger}30.1$
            & $^*3.83$ & \small{2.0M} & $^*256.4$ \\[-3pt]
            &  &  &
            & $^{\mathsection}3.49$ & $^\mathsection$\small{1.6M} & $^{\mathsection}40.39$
            & & &  \\
        FFJORD + STEER
        ~\citep{ghosh2020steer}
            & 3.40 & \small{1.4M} & 86.34
            & 3.84 & \small{2.0M} & \small{$>$5days}
            & & & \\
        RNODE + STEER
        ~\citep{ghosh2020steer}
            & 3.397 & \small{1.4M} & 22.24
            & 2.35 & \small{2.0M} & 24.90
            & & & \\[-3pt]
            & & &
            & $^\mathsection3.49$ & $^\mathsection$\small{1.6M} & $^\mathsection30.07$
            & & & \\
        \hline
        \hline
        \multicolumn{10}{|l|}{\textbf{\textsc{(Ours)} Multi-Resolution Continuous Normalizing Flow (MRCNF)}}\\
        \hline
        2-resolution MRCNF
            & 3.65 & \small{1.3M} & 19.79
            & 3.77 & \small{1.3M} & 18.18
            & 3.44& \small{2.0M} & 42.30 \\
        2-resolution MRCNF
            & 3.54 & \small{3.3M} & 36.47
            & 3.78 & \small{6.7M} & 17.98
            & x & \small{6.7M} & x \\
        3-resolution MRCNF
            & 3.79 & \small{1.5M} & 17.44
            & 3.97 & \small{1.5M} & 13.78
            & 3.55 & \small{2.0M} & 35.39 \\
        3-resolution MRCNF
            & 3.60 & \small{5.1M} & 38.27
            & 3.93 & \small{10.2M} & 41.20
            & x & \small{7.6M} & x\\
        \hline
    \end{tabular}
    }
    \label{tab:msflow}
    }
\footnotetext{
- Unreported values. \qquad
$^\dagger$As reported in~\citet{chen2019residual}.
    \quad
    $^\ddagger$As reported in~\citet{ghosh2020steer}.
    \\
    \quad
    $^\mathsection$Re-implemented by us.
    \quad
    `x': Fails to train.
    \quad
    $^*$RNODE~\citep{finlay2020rnode} used 4 GPUs to train.
    }
}
\end{minipage}
\end{table*}

We train \gls{msflow-image} models on  the CIFAR10~\citep{krizhevsky2009learning} dataset at finest resolution of 32x32, and the ImageNet~\citep{deng2009imagenet} dataset at 32x32, 64x64, 128x128. We build on top of the code provided in \citep{finlay2020rnode}\footnote{https://github.com/cfinlay/ffjord-rnode}. In all cases, we train using \textit{only one} NVIDIA RTX 20280 Ti GPU with 11GB.

In \autoref{tab:msflow}, we compare our results with prior work in terms of (lower is better in all cases) the \gls{bpd} of the images of the test datasets under the trained models, the number of parameters used by the model, and the number of GPU hours taken to train.
The most relevant models for comparison are the 1-resolution FFJORD~\citep{grathwohl2019ffjord} models, and their regularized version RNODE~\citep{finlay2020rnode}, since our model directly converts their architecture into multi-resolution. Other relevant comparisons are previous flow-based methods~\citep{dinh2016density, kingma2018glow, song2019mintnet, ho2019flow++, yu2020waveletflow}, however their core architecture (RealNVP~\citep{dinh2016density}) is quite different from FFJORD.

\textbf{BPD}: At lower resolution spaces, we achieve comparable \gls{bpd}s in less time with far fewer parameters than previous normalizing flows (and non flow-based approaches). However, the power of the multi-resolution formulation is more evident at higher resolutions: we achieve better BPD for ImageNet64 with significantly fewer parameters and less time using only one GPU.
A more complete table can be found in the appendix.


\textbf{Train time}: All our experiments used only one GPU, and took significantly less time to train than 1-resolution CNFs, and all prior works including flow-based and non-flow-based models.
For example on CIFAR-10, Glow~\citep{kingma2018glow} used 8 GPUs for 7 days,
MintNet~\citep{song2019mintnet} used 2 GPUs for $\approx$ 5 days,
1-resolution FFJORD~\citep{grathwohl2019ffjord} used 6 GPUs for $\approx$ 5 days. All our models used 1 GPU for $\leq$ 1 day.

To make a fair comparison with previous methods, we report the total time taken to train the \gls{cnf}s of all resolutions one after another on a single GPU. We also maintained the batch size of the finest resolution the same as that in the previous \gls{cnf} works, but used bigger batch sizes to train coarser resolutions.
However, since all the \gls{cnf}s can be trained in parallel, the actual training time in practice could be much lower.

\subsection{Super-resolution}

\begin{figure*}[!thb]
\begin{minipage}{\textwidth}
    \centering
    \includegraphics[width=.7\textwidth]{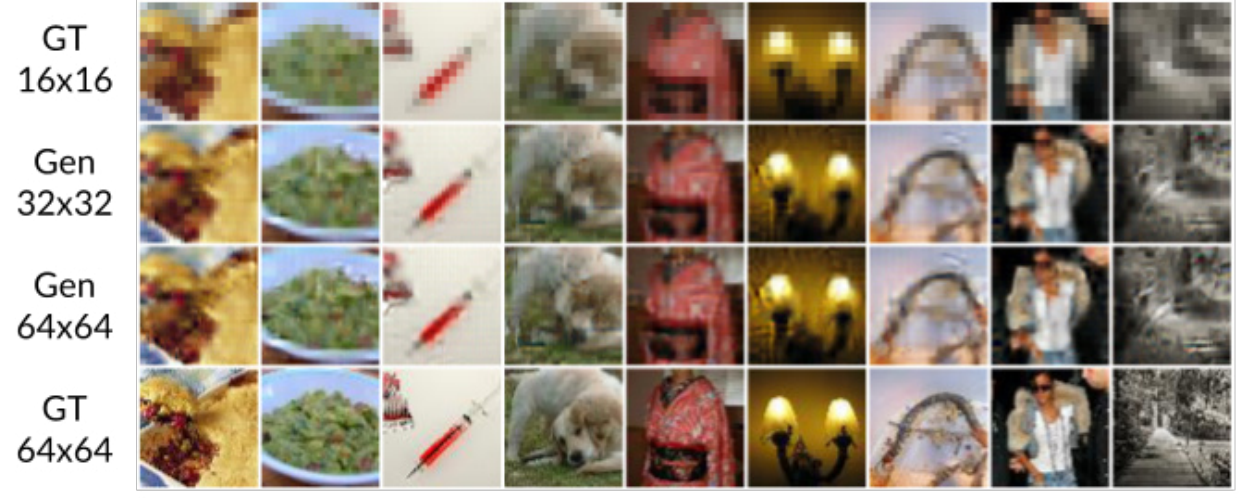}
    \caption{ImageNet: Example of super-resolving to 64x64 from ground truth 16x16. Row 1: ground truth 16x16, Row 2: generated 32x32, Row 3: generated 64x64 Row 4: ground truth 64x64. }
    \label{fig:super-res-images}
\end{minipage}
\end{figure*}

Our formulation also allows for super-resolution of images (\autoref{fig:super-res-images}) free of cost since our framework is autoregressive in resolution. At any stage, one can condition on a ground truth low-resolution image and generate the corresponding high-resolution image.

\subsection{Progressive training}

We trained an MRCNF model on ImageNet128 by training only the finest resolution (128x128) conditioned on the immediate coarser (64x64) images, and attached it to a 3-resolution model trained on ImageNet64. The resultant 4-resolution ImageNet128 model gives a \gls{bpd} of 3.31 (\autoref{tab:imagenet128}) with just 2.74M parameters in $\approx$60 GPU hours.

\begin{table}[!thb]
    \centering
    \caption{Metrics for unconditional ImageNet128 generation. \textsc{Param} is number of parameters, \textsc{Time} is in hours. `-' indicates unreported values.
    }
    \begin{tabular}{|l@{\hskip 4pt}|@{\hskip 4pt}c@{\hskip 8pt}r@{\hskip 8pt}c@{\hskip 4pt}|}
        \hline
        \textbf{\textsc{ImageNet128}}
        \hfill ($\downarrow$)
        & \textsc{BPD} & \textsc{Param} & \textsc{Time}\Tstrut
        \\\hline
        Parallel Multiscale~\citep{reed2017parallel} & 3.55 & - & - \Tstrut
        \\
        SPN~\citep{menick2019generating} & 3.08 & \small{250.00M} & -
        \\
        \hline
        \textbf{\textsc{(Ours)} 4-resolution MRCNF} & 3.31 & \small{2.74M} & 58.59\Tstrut
        \\\hline
    \end{tabular}
    \label{tab:imagenet128}
\end{table}


\subsection{Ablation study}
\label{subsec:waveletflow_ablation}

Our MRCNF method differs from WaveletFlow in three respects:

\begin{enumerate}
    \item we use CNFs, while WaveletFlow uses the discrete vairant of normalizing flows,
    \item we use \autoref{Mdefn} instead of \autoref{haar} as used by WaveletFlow,
    \item we use multi-resolution noise.
\end{enumerate}

We check the individual effects of these changes in an ablation study in \autoref{tab:ablation}, and conclude that:

\begin{enumerate}
\item Simply replacing the normalizing flows in WaveletFlow with CNFs does not produce the best results. It does improve the BPD and training time compared to WaveletFlow.
\item Using our unimodular transformation in \autoref{Mdefn} instead of the original Wavelet Transformation of \autoref{haar} not only improves the BPD, it also consistently decreases training time.
\item As expected, the use of multi-resolution noise does not have a critical impact on either BPD or training time. We use it anyway so as to retain interpretation with 1-resolution models.
\end{enumerate}

\begin{table*}[!htb]
    \centering
    {\def\arraystretch{1.1}
    \setlength{\tabcolsep}{.1em}
    \caption{Ablation study across using Wavelet in \autoref{haar}, and multi-resolution noise formulation in \autoref{subsec:mrnoise}. P is number of parameters, \textsc{Time} is in hours. Lower is better in all cases. `-' indicates unreported values. `x' : Fails to train3.}
    \begin{tabular}{|l|crr|crr|}
        \hline
        & \multicolumn{3}{c|}{\textbf{\textsc{CIFAR10}}} & \multicolumn{3}{c|}{\textbf{\textsc{ImageNet64}}} \\
        \hfill ($\downarrow$)
        & \textsc{BPD} & \textsc{P$\ \ $} & \textsc{Time}
        & \textsc{BPD} & \textsc{P$\ \ $} & \textsc{Time} \\
        \hline
        WaveletFlow \citep{yu2020waveletflow}
            & - & - & -
            & 3.78 & \footnotesize{98.0M} & 822.00 \\
        1-resolution CNF (RNODE) \citep{finlay2020rnode}
            & 3.38 & \footnotesize{1.4M} & 31.84
            & 3.83 & \footnotesize{2.0M} & 256.40 \\
        \hline\hline
        \multicolumn{7}{|l|}{\textbf{2-resolution}}\\
        \hline
        \cref{haar} WaveletFlow with CNF w/o multi-res noise
            & 3.68 & \footnotesize{1.3M} & 27.25
            & x & \footnotesize{2.0M} & x \\
        \cref{haar} WaveletFlow with CNF w/ multi-res noise
            & 3.69 & \footnotesize{1.3M} & 25.88
            & x & \footnotesize{2.0M} & x \\
        \cref{Mdefn} MRCNF w/o multi-res noise
            & 3.66 & \footnotesize{1.3M} & 19.79
            & 3.48 & \footnotesize{2.0M} & 42.33 \\
        \cref{Mdefn} MRCNF w/ multi-res noise \textbf{(Ours)}
            & 3.65 & \footnotesize{1.3M} & 19.69
            & 3.44 & \footnotesize{2.0M} & 42.30 \\
        \hline\hline
        \multicolumn{7}{|l|}{\textbf{3-resolution}}\\
        \hline
        \cref{haar} WaveletFlow with CNF w/o multi-res noise
            & 3.82 & \footnotesize{1.5M} & 22.99
            & 3.62 & \footnotesize{2.0M} & 43.37 \\
        \cref{haar} WaveletFlow with CNF w/ multi-res noise
            & 3.82 & \footnotesize{1.5M} & 25.28
            & 3.62 & \footnotesize{2.0M} & 44.21 \\
        \cref{Mdefn} MRCNF w/o multi-res noise
            & 3.79 & \footnotesize{1.5M} & 17.25
            & 3.57 & \footnotesize{2.0M} & 35.42 \\
        \cref{Mdefn} MRCNF w/ multi-res noise \textbf{(Ours)}
            & 3.79 & \footnotesize{1.5M} & 17.44
            & 3.55 & \footnotesize{2.0M} & 35.39 \\
        \hline
    \end{tabular}
    \label{tab:ablation}
    }
\end{table*}

Thus, our MRCNF model is not a trivial replacement of normalizing flows with CNFs in WaveletFlow.
We generalize the notion of multi-resolution image representation, in which the Discrete Wavelet Transform is one of many possibilities. We then derived a unimodular transformation that adds no change in likelihood.

\subsection{Adversarial loss}

Several works~\citep{makhzani2015adversarial, grover2017flow, Lee2018StochasticAV, beckham2019adversarial} have found it useful to add an adversarial loss to pre-existing losses to generate images that better resemble the true data distribution. Similar to \citep{grover2017flow}, we conducted experiments with an additional adversarial loss at each resolution. However in our experiments so far, we could achieve neither better \gls{bpd}s nor better \gls{fid}s~\citep{heusel2017gans}. As noted in \citep{theis2016note}, since likelihood-based models tend to cover all the modes by minimizing KL-divergence while GAN-based methods tend to mode collapse by minimizing JS-divergence, it is possible that the two approaches are incompatible, and so combining them is not trivial.

\subsection{FID v/s temperature}
Table \ref{tab:fid_vs_temp} lists the FID values of generated images from MRCNF models trained on CIFAR10, with different temperature settings on the Gaussian.
\begin{table}[!ht]
    \centering
    \begin{tabular}{|l@{\hskip 4pt}|@{\hskip 4pt}c@{\hskip 8pt}c@{\hskip 8pt}c@{\hskip 4pt}@{\hskip 4pt}c@{\hskip 8pt}c@{\hskip 8pt}c@{\hskip 4pt}|}
        \hline
        & \multicolumn{6}{c|}{\textbf{Temperature}} \\[2pt]
        & \textbf{1.0} & \textbf{0.9} & \textbf{0.8} & \textbf{0.7} & \textbf{0.6} & \textbf{0.5}
        \\[2pt]\hline
        \textbf{1-resolution CNF} & 138.82 & 147.62 & 175.93 & 284.75 & 405.34 & 466.16
        \\[2pt]
        \textbf{2-resolution MRCNF} & 89.55 & 106.21 & 171.53 & 261.64 & 370.38 & 435.17
        \\[2pt]
        \textbf{3-resolution MRCNF} & 88.51 & 104.39 & 152.82 & 232.53 & 301.89 & 329.12
        \\[2pt]
        \textbf{4-resolution MRCNF} & 92.19 & 104.35 & 135.58 & 186.71 & 250.39 & 313.39
        \\\hline
    \end{tabular}
    \caption{FID v/s temperature for \gls{msflow-image} models trained on CIFAR10.}
    \label{tab:fid_vs_temp}
\end{table}

\section{Examining Out-of-Distribution (OoD) behaviour}
\label{ood}

\begin{figure}[!thb]
\centering
\includegraphics[width=0.6\linewidth]{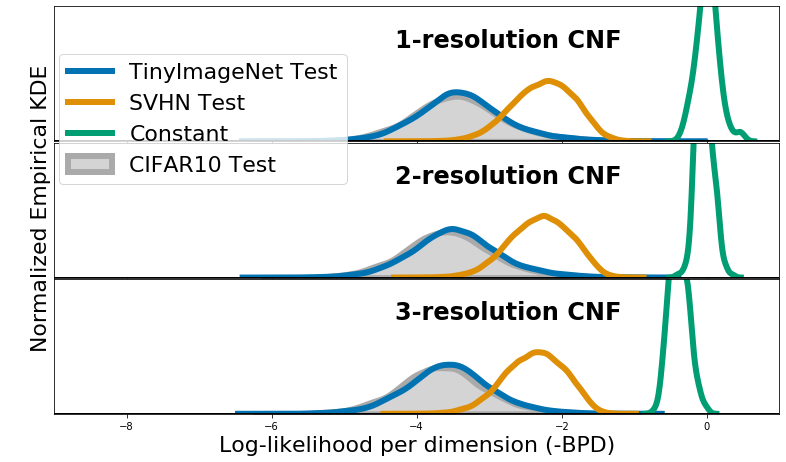}
\caption{Histogram of log likelihood per dimension i.e. $-$BPD (estimated using normalized empirical Kernel Density Estimation) of OoD datasets (TinyImageNet, SVHN, Constant) under (MR)CNF models trained on CIFAR10. As with other likelihood-based generative models such as Glow \& PixelCNN, OoD datasets have higher likelihood under (MR)CNFs.}
\label{fig:ood_bpd}
\end{figure}

The derivation of likelihood-based models suggests that the density of an image under the model is an effective measure of its likelihood of being in-distribution. However, recent works~\citep{theis2016note, nalisnick2018deep, serra2019input, nalisnick2019detecting} have pointed out that it is possible that images drawn from other distributions have higher model likelihood. Examples have been shown where normalizing flow models (such as Glow) trained on CIFAR10 images assign higher likelihood to SVHN~\citep{Netzer2011svhn} images. This could have serious implications on their practical applicability. Some also note that likelihood-based models do not generate images with good sample quality as they avoid assigning small probability to \gls{ood} data points, hence model likelihood (-\gls{bpd}) is not effective for detecting \gls{ood} data in such models.

We conduct the same experiments with (MR)CNFs, and find that similar conclusions can be drawn. \autoref{fig:ood_bpd} plots the histogram of log likelihood per dimension (-\gls{bpd}) of \gls{ood} images (SVHN, TinyImageNet) under \gls{msflow-image} models trained on CIFAR10. It can be observed that the likelihood of the \gls{ood} SVHN is higher than CIFAR10 for MRCNF, similar to the findings for Glow, PixelCNN, VAE in earlier works~\citep{nalisnick2018deep, choi2018waic, serra2019input, nalisnick2019detecting, kirichenko2020normalizing}.

One possible explanation put forward by \citet{nalisnick2019detecting} is that ``typical'' images are less ``likely'' than constant images, which is a consequence of the distribution of a Gaussian in high dimensions. Indeed, as our \autoref{fig:ood_bpd} shows, constant images have the highest likelihood under MRCNFs, while randomly generated (uniformly distributed) pixels have the least likelihood (not shown in figure due to space constraints).

\citep{choi2018waic, nalisnick2019detecting} suggest using ``typicality'' as a better measure of \gls{ood}.
However, \citep{serra2019input} observe that the complexity of an image plays a significant role in the training of likelihood-based generative models. They propose a new metric $S$ as an out-of-distribution detector:
\begin{align}
    S(\rvx) = \text{BPD}(\rvx) - L(\rvx).
\end{align}
where $L(\rvx)$ is the complexity of an image $\rvx$ measured as the length of the best compressed version of $\rvx$ (we use FLIF~\citep{sneyers2016flif} following \citet{serra2019input}) normalized by the number of dimensions.

We perform a similar analysis as \citet{serra2019input} to test how $S$ compares with -$\text{bpd}$ for \gls{ood} detection. For different \gls{msflow-image} models trained on CIFAR10, we compute the area under the receiver operating characteristic curve (auROC) using -bpd and $S$ as standard evaluation for the \gls{ood} detection task~\citep{hendrycks2019deep, serra2019input}. \autoref{tab:auROC} shows that $S$ does perform better than -$\text{bpd}$ in the case of (MR)CNFs, similar to the findings in \citet{serra2019input} for Glow and PixelCNN++. SVHN seems easier to detect as OoD for Glow than MRCNFs. However, OoD detection performance is about the same for TinyImageNet. We also observe that MRCNFs are better at OoD than CNFs.

Other \gls{ood} methods \citep{hendrycks2017baseline, liang2018enhancing, lee2018simple, sabeti2019data, host2019data, hendrycks2019deep} are not suitable, as identified in \citet{serra2019input}.

\begin{table}[!thb]
\setlength{\tabcolsep}{3pt}
\centering
\caption{auROC for \gls{ood} detection using -bpd and $S$\citep{serra2019input}, for models trained on CIFAR10.}
\begin{tabular}{lcccc}
\textbf{CIFAR10} & \multicolumn{2}{c}{SVHN} & \multicolumn{2}{c}{TIN}
\\
\cline{2-5}
(trained) & -BPD & $S$
& -BPD & $S$ \\
\hline
\multicolumn{1}{l|}{Glow} & 0.08 & 0.95 & 0.66 & 0.72
\\[0pt]
\multicolumn{1}{l|}{1-res CNF} & 0.07 & 0.16 & 0.48 & 0.60
\\[0pt]
\multicolumn{1}{l|}{2-res MRCNF} & 0.06 & 0.25 & 0.46 & 0.66
\\[0pt]
\multicolumn{1}{l|}{3-res MRCNF} & 0.05 & 0.25 & 0.46 & 0.66
\\
\hline
\end{tabular}
\label{tab:auROC}
\end{table}

\begin{figure*}[!htb]
    \centering
    \begin{subfigure}[b]{0.3\textwidth}
     \centering
     \includegraphics[width=\textwidth]{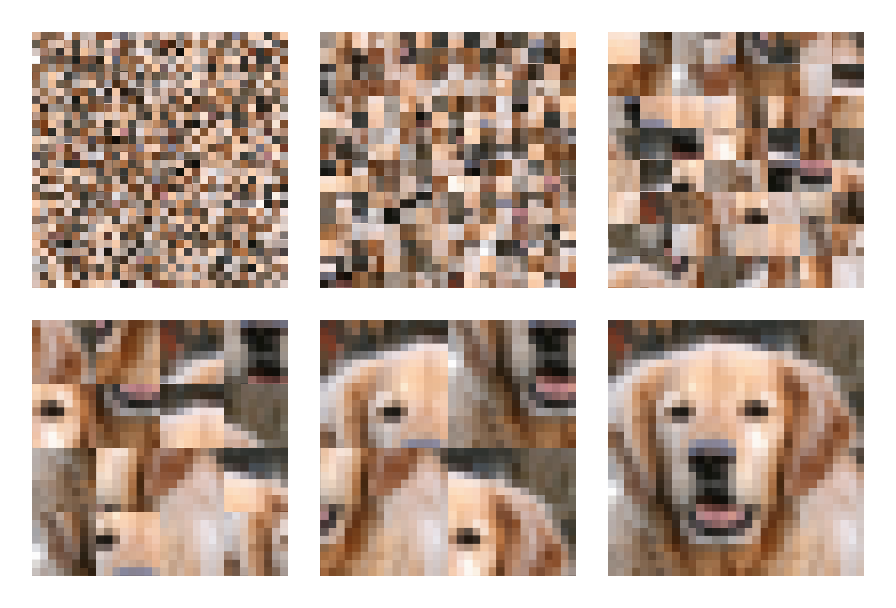}
     \caption{}
     \label{fig:1sc}
    \end{subfigure}
    \begin{subfigure}[b]{0.65\textwidth}
     \centering
     \includegraphics[width=\textwidth]{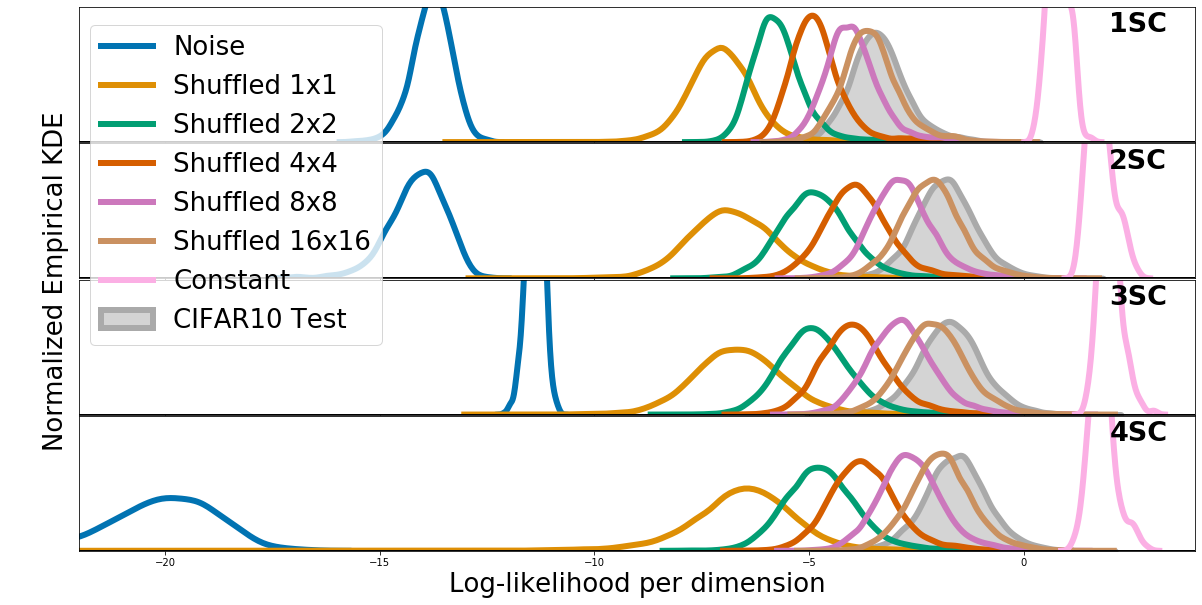}
     \caption{}
     \label{fig:2sc}
    \end{subfigure}
    \caption{(a) Example of shuffling different-sized patches of a 32x32 image: (left to right, top to bottom) 1x1, 2x2, 4x4, 8x8, 16x16, 32x32 (unshuffled) (b) Histogram of log likelihood per dimension (normalized empirical Kernel Density Estimate) for \gls{msflow-image} models at different resolutions, trained on CIFAR10.}
    \label{fig:ood_shuffle}
\end{figure*}

\subsection{Shuffled in-distribution images}

Prior work \citep{kirichenko2020normalizing} concludes that normalizing flows do not represent images based on their semantic contents, but rather directly encode their visual appearance. We verify this for continuous normalizing flows by estimating the density of in-distribution test images, but with patches of pixels randomly shuffled. \autoref{fig:ood_shuffle} (a) shows an example of images of shuffled patches of varying size, \autoref{fig:ood_shuffle} (b) shows the graph of the their log-likelihoods.

That shuffling pixel patches would render the image semantically meaningless is reflected in the FID between CIFAR10-Train and these sets of shuffled images --- 1x1: 340.42, 2x2: 299.99, 4x4: 235.22, 8x8: 101.36, 16x16: 33.06, 32x32 (i.e. CIFAR10-Test): 3.15. However, we see that images with large pixel patches shuffled are quite close in likelihood to the unshuffled images (\autoref{fig:ood_shuffle} (b)), suggesting that since their visual content has not changed much they are almost as likely as unshuffled images under MRCNFs.

\section{Qualitative samples}

\begin{figure*}[!htb]
    \centering
    \begin{subfigure}[b]{0.45\textwidth}
     \centering
     \includegraphics[width=\textwidth]{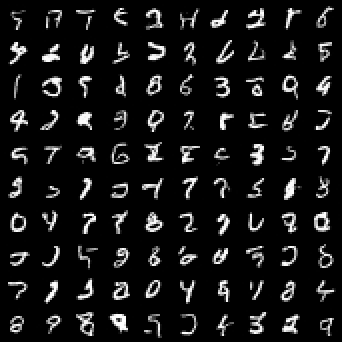}
     \caption{Generated samples at 16x16}
     \label{fig:mnist16}
    \end{subfigure}
    \hfill
    \begin{subfigure}[b]{0.5\textwidth}
     \centering
     \includegraphics[width=\textwidth]{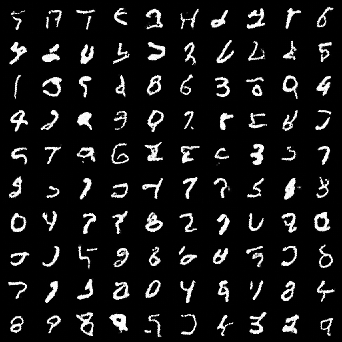}
     \caption{Corresponding generated samples at 32x32}
     \label{fig:mnist32}
    \end{subfigure}
    \caption{Generated samples from MNIST at different resolutions.}
    \label{fig:mrcnf-mnist}
\end{figure*}

\begin{figure*}[!htb]
    \centering
    \begin{subfigure}[b]{0.25\textwidth}
     \centering
     \includegraphics[width=\textwidth]{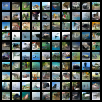}
     \caption{8x8}
    \end{subfigure}
    \hfill
    \begin{subfigure}[b]{0.3\textwidth}
     \centering
     \includegraphics[width=\textwidth]{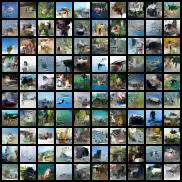}
     \caption{16x16}
    \end{subfigure}
    \hfill
    \begin{subfigure}[b]{0.4\textwidth}
     \centering
     \includegraphics[width=\textwidth]{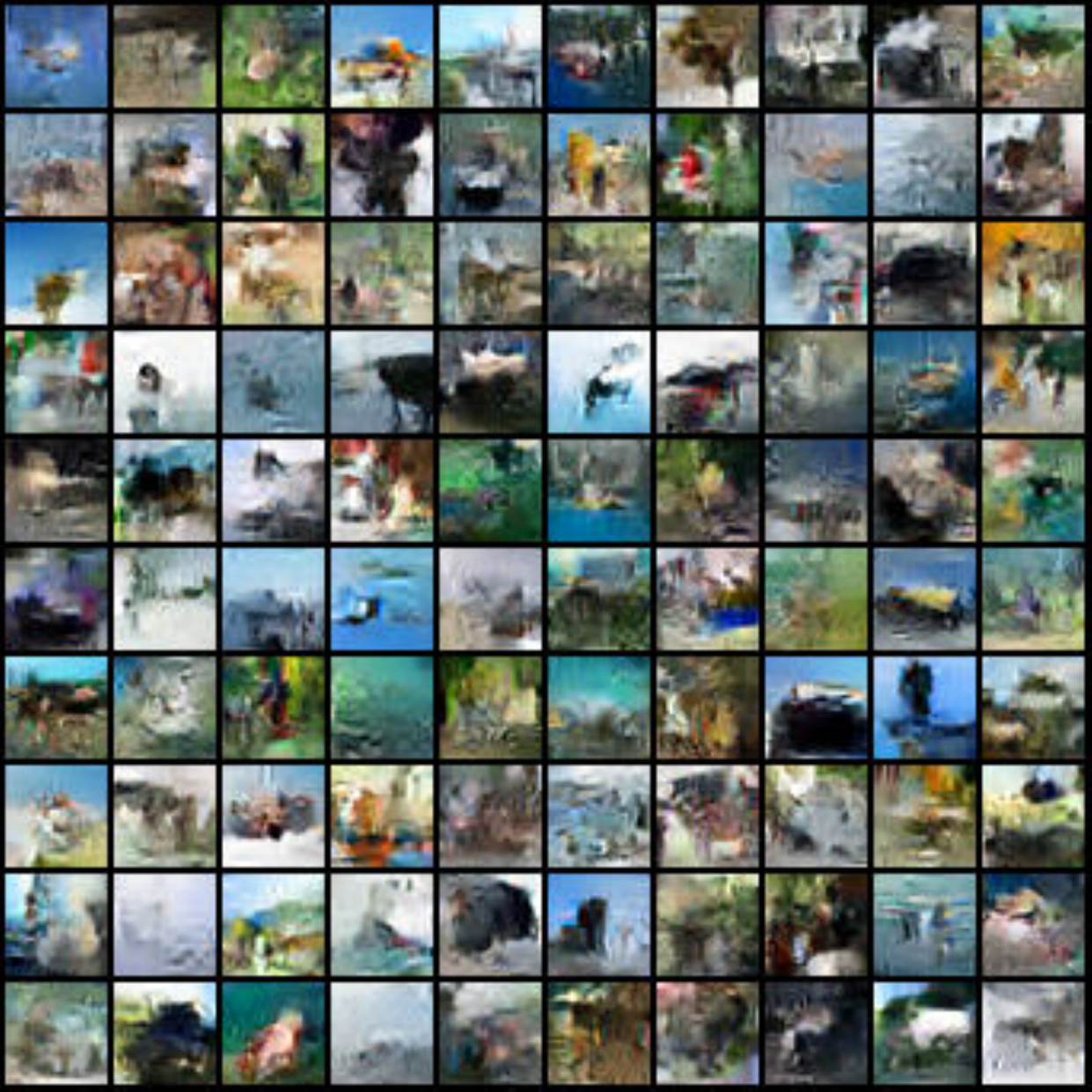}
     \caption{32x32}
    \end{subfigure}
    \caption{Generated samples from CIFAR10 at different resolutions.}
    \label{fig:mrcnf-cifar10}
\end{figure*}

We show qualitative examples from the MNIST~\citep{deng2012mnist} and CIFAR10~\citep{cifar10} in \cref{fig:mrcnf-mnist,fig:mrcnf-cifar10}.

\section{Conclusion}
\label{mrcnf-conclusion}
We have presented a Multi-Resolution approach to Continuous Normalizing Flows (MRCNF). MRCNF models achieve comparable or better performance in significantly less training time, training on a single GPU, with a fraction of the number of parameters of other competitive models. Although the likelihood values for 32x32 resolution datasets such as CIFAR10 and ImageNet32 do not improve over the baseline, ImageNet64 and above see a marked improvement.
The performance is better for higher resolutions, as seen in the case of ImageNet128. We also conducted an ablation study to note the effects of each change we introduced in the formulation.

In addition, we show that (Multi-Resolution) Continuous Normalizing Flows have similar out-of-distribution properties as other Normalizing Flows.

%% file: ProtoResSMPL.tex
\renewcommand{\vec}[1]{\mathbf{#1}}
\renewcommand{\Re}{\mathbb{R}}

\anglais
\counterwithin{figure}{chapter}
\counterwithin{table}{chapter}

\chapter{Neural Inverse Kinematics with 3D Human Pose Prior~\citep{voleti2022smpl}}
\label{chap:ProtoResSMPL}

\setcounter{section}{-1}
\section{Prologue to article}
\label{chap:pro_ProtoResSMPL}

\subsection{Article details}

\textbf{SMPL-IK: Learned Morphology-Aware Inverse Kinematics for AI Driven Artistic Workflows}. Vikram Voleti, Boris Oreshkin, Florent Bocquel\'et, Felix Harvey, Louis-Simon M\'enard, Christopher Pal. \textit{SIGGRAPH Asia 2022 (Technical Communications)}

\textit{Personal contribution:} The project began with discussions between the authors during Vikram's internship at Unity Technologies partly funded by MITACS, Canada. This project was in collaboration with Boris Oreshkin, and the DeepPose team at Unity Labs, Montreal. Christopher Pal provided advice and guidance throughout the project. Vikram performed a literature survey on the state-of-the-art pose estimation models from image/video, identifying and comparing 20+ datasets of 2D and 3D human pose in various formats, and 20+ methods of 3D human pose estimation. Vikram also learned the technical details of 3D character editing in Unity, 3D graphics pipelines, etc. Boris Oreshkin proposed the main idea of leveraging a 3D human pose model as a prior to improve existing pose estimation methods. Vikram analyzed the literature, identified a strong 3D human pose prior (SMPL), and convinced the team to use SMPL prior for 3D animation tasks moving forward. Boris Oreshkin, Florent Bocquel\'et, and Vikram collectively worked on integrating SMPL into the team's state-of-the-art human pose estimation model ProtoRes. Vikram identified a huge human pose dataset in the SMPL format called AMASS. Boris and Vikram worked on training SMPL-integrated ProtoRes on AMASS, and testing on other standard datasets (Human3.6M, etc.). Vikram and Boris also worked on stacking an image-to-pose model (ROMP) before this to extract 3D pose from an image, and calculated metrics on standard datasets. Vikram worked with Louis-Simon M\'enard on integrating this into the Unity Labs software. Boris and Florent proposed shape inversion to expand the pipeline to work on non-human characters. Project progress was managed and tracked by the full team using JIRA, as well as through weekly team meetings and regular chats on Slack. Vikram made a demo of the full pipeline from image to 3D pose editing of human and non-human 3D characters. The team compiled the work into a research paper published in the Technical Communications track at SIGGRAPH Asia 2022. Vikram made a video for the conference (SIGGRAPH Asia) explaining the paper, methodology, and results.

\subsection{Context}

Inverse Kinematics (IK) systems are often rigid with respect to their input character, thus requiring user intervention to be adapted to new skeletons. Many computer vision pose estimation algorithms naturally operate in the Skinned Multi-Person Linear (SMPL) space, and this extension would open new content authoring opportunities. However, to date SMPL models have not been integrated with advanced machine learning IK tools, and this represents a clear research gap.

\subsection{Contributions}

In this paper we aim at creating a flexible, learned IK solver applicable to a wide variety of human morphologies. We extend a state-of-the-art machine learning IK solver to operate on the well known Skinned Multi-Person Linear model (SMPL). We call our model SMPL-IK, and show that when integrated into real-time 3D software, this extended system opens up opportunities for defining novel AI-assisted animation workflows. For example, when chained with existing pose estimation algorithms, SMPL-IK accelerates posing by allowing users to bootstrap 3D scenes from 2D images while allowing for further editing. Additionally, we propose a novel SMPL Shape Inversion mechanism (SMPL-SI) to map arbitrary humanoid characters to the SMPL space, allowing artists to leverage SMPL-IK on custom characters. In addition to qualitative demos showing proposed tools, we present quantitative SMPL-IK baselines on the H36M and AMASS datasets. Our code is publicly available \url{https://github.com/boreshkinai/smpl-ik}, and a video explaining the paper at SIGGRAPH Asia 2022 is available at \url{https://www.youtube.com/watch?v=FixF406owB4}.

\subsection{Research impact}

This work integrated the use of SMPL in the context of pose estimation using ProtoRes. Although there are follow-up works that also show the validity of our approach~\citep{ma2022pretrained}, the adoption of our work hasn’t been as widespread yet as we had hoped in the research community. Our work at Unity continues to be used in Unity's products, specifically in the Unity Labs software, allowing animators to edit 3D characters with flexibility.

\newpage

\begin{figure}[!htb]
  \centering
  \includegraphics[width=\textwidth]{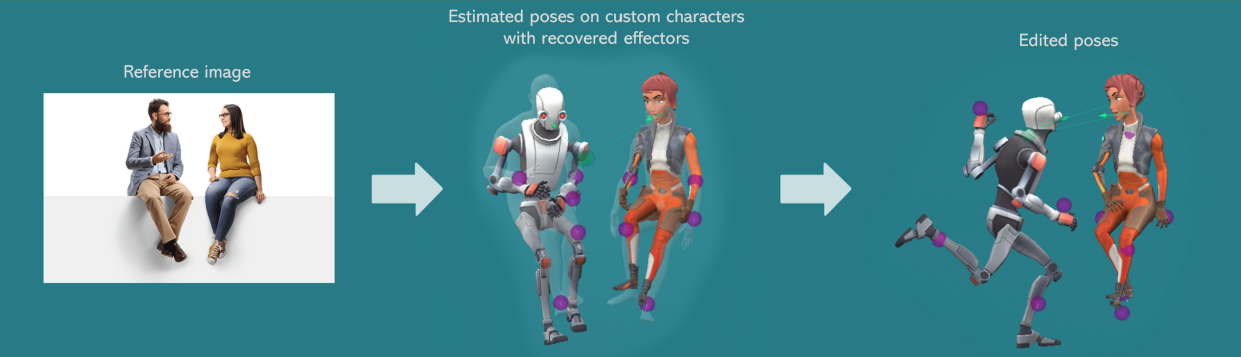}
  \caption{Our approach unlocks novel artistic workflows such as the one depicted above. An animator uses an image to initialize an editable 3D scene. Thus a multi-person 3D scene acquired from an RGB picture is populated with custom user-defined characters whose 3D poses are further edited with the state-of-the-art machine learning inverse kinematics tool integrated in a real-time 3D development software. 
  }
  \label{fig:unity_teaser}
\end{figure}

\section{Introduction}

\gls{ik} is the problem of estimating 3D positions and rotations of body joints given some end-effector locations \citep{kawato1993supervised, aristidou2018inverse}. IK is an ill-posed nonlinear problem with multiple solutions. For example, given the 3D location of the right hand, what is a realistic human pose? It has been shown recently that machine learning IK model can be integrated with 3D content authoring user interface to produce a very effective pose authoring tool~\citep{oreshkin2021protores,bocquelet2022ai}. Using this tool, an animator provides a terse pose definition via a limited set of positional and angular constraints. The computer tool fills in the rest of the pose, minimizing pose authoring overhead.

The Skinned Multi-Person Linear (SMPL) model is a principled and popular way of jointly modelling human mesh, skeleton and pose~\citep{loper2015smpl}. It would seem natural to extend this model with inverse kinematics capabilities: making both human shape/mesh and pose editable using independent parameters. Additionally, many computer vision pose estimation algorithms naturally operate in the SMPL space making them natively compatible with a hypothetical SMPL IK model. This extension would open new content authoring opportunities. However, to date SMPL models have not been integrated with advanced machine learning IK tools, and this represents a clear research gap.

In our work we close this gap, exploring and solving two inverse problems in the context of the SMPL human mesh representation: SMPL-IK, an Inverse Kinematics model, and SMPL-SI, a Shape Inversion model. We show how these new components can be used to create new artistic workflows driven by AI algorithms. For example, we demonstrate the tool integrating SMPL-IK and SMPL-SI with an off-the-shelf image-to-pose model, initializing a multi-person 3D scene editable via flexible and easy-to-use user controls.

We demonstrate the tool integrating SMPL-IK with an off-the-shelf image-to-pose model, and use that as a starting point for an editable SMPL character whose gender, shape and pose are editable via flexible and easy-to-use user controls. Furthermore, we show that by including the proposed SMPL-SI model in the workflow we add the additional flexibility of handling custom user supplied characters in the same universal SMPL space by finding an SMPL approximation of the user supplied character via SMPL-SI.

In summary, SMPL-IK accelerates posing by allowing users to bootstrap 3D scenes from 2D images, while allowing for further realistic editing. Additionally, we propose a novel SMPL shape inversion mechanism to map arbitrary humanoid characters to the SMPL space, allowing artists to leverage SMPL-IK on custom 3D characters.

Our main contributions are as follows:
\begin{itemize}
    \item made SMPL-IK by integrating SMPL into a state-of-the-art 3D human pose estimation model called ProtoRes~\citep{oreshkin2021protores}.
    \item achieved 3D human pose estimation in SMPL format with only partial input pose required, by training SMPL-IK on AMASS dataset~\citep{amass}.
    \item calculated 3D human pose metrics on standard 3D human pose datasets (AMASS and Human3.6M) in \Cref{tab:protores}.
    \item expanded this model's capability to non-human bodies by proposing Shape Inversion.
    \item stacked an image-to-pose model (ROMP~\citep{sun2021romp}) before this, thus making an 3D human pose estimator from images.
    \item unified everything into a pipeline that makes a realistic 3D pose editor requiring only partial pose input, initialized by a 3D human pose from an image.
\end{itemize}

\section{Background}
\label{protores:bg}

\subsection{Skinned Multi-Person Linear model (SMPL)}

Skinned Multi-Person Linear model (SMPL) is a realistic 3D human body model
based on skinning and blend shapes \citep{loper2015smpl}. 
It is parameterized by two kinds of parameters: shape/beta parameters that control the body shape, and pose parameters that control pose-dependent deformations. SMPL realistically represents a wide range of human body morphologies, and pose-dependent deformations of the body. There have been some extensions to the SMPL model such as SMPL+H~\citep{MANO:SIGGRAPHASIA:2017}, SMPL-X~\citep{pavlakos2019expressive}, STAR~\citep{STAR:2020}, etc. SMPL remains a widely accepted model to represent realistic human body pose and is prevalently used for 3D pose estimation of humans in images and video~\citep{bogo2016keep,Luo_2020_ACCV,li2021hybrik,rajasegaran2021tracking,sun2021monocular}. 

\subsection{3D human pose datasets}

In \Cref{table:datasets}, we identify various datasets that contain ground truth information on human motion. We also add comments on the type of data contained in the datasets i.e. whether they are images or videos or only 3D content such as Motion Capture (MoCap), whether the dataset contains indoor or outdoor scenes, whether they are of a single person or multiple people. In addition, we specify some details each dataset such as the type of people, how the data was processed, number of images, number of subjects, number of hours of recording, etc.

\afterpage{
\footnotesize{
\begin{longtable}{| p{.2\textwidth} | p{.25\textwidth} | p{.5\textwidth} |}
\caption{Table of 3D human pose datasets}
\label{table:datasets}
\\ \hline
\textbf{Dataset} & \textbf{Type of data} & \textbf{Comments} \\ \hline\hline

\multicolumn{3}{|c|}{\textbf{images + 3D pose}}\\\hline

\href{http://mocap.cs.cmu.edu/faqs.php}{CMU MoCap data} & images + 3D pose & -  $>$11k motions, 40hrs of motion data, 300+ subjects \\
& (NOT synchronized) & - MoCap NOT synced with images!
\\
& & - GT using cameras, etc.\\[8pt]\hline\hline

\multicolumn{3}{|c|}{\textbf{video + 3D pose}}\\\hline

\href{http://humaneva.is.tue.mpg.de/}{HumanEva} & video + 3D pose & - 40k frames+3D pose @60Hz with 7 Vicon cameras \\
\citep{sigal2010humaneva} & - Single person, indoor & - 30k frames of pure MoCap \\
& & - GT with MoCap software \\[8pt]\hline

\href{https://www2.eecs.berkeley.edu/Research/Projects/CS/vision/shape/h3d/}{H3D (UCB)} & video + 3D pose & - 2k frames (1500 train, 500 test) \\
\citep{PoseletsICCV09} & - Single person, outdoor & - 19 keypoints (joints, eyes, nose, etc.)\\
& (small dataset) & - GT from manual annotation \\[8pt]\hline

\href{http://vision.imar.ro/human3.6m/papers.php}{Human 3.6M} & video + 3D pose & - 3.6M frames+poses indoor with 4 Vicon cameras \\
\citep{ionescu2014h36m} & - Single person, indoor & - 11 professional actors (5F, 6M) imitate real poses \\
& & - Hybrid dataset: virtual character in real video \\
& & - GT using 4 cameras, software, etc. \\[8pt]\hline

\href{https://vcai.mpi-inf.mpg.de/3dhp-dataset/}{MPI-INF-3DHP} & video + 3D pose & - $>$1.3M frames, 14 cameras (500k frames, 5 cameras at chest height) \\
\citep{mehta20173dhp} & - Single person, indoor & - 8 actors (4M, 4F), 8 activity sets each of $\approx$1min \\
& & \quad - walking, sitting, exercise poses, dynamic actions \\
& & \quad - more pose classes than Human3.6m \\
& & \quad - 2 sets of clothing: casual everyday, plain-colored \\
& & - GT from multi-view marker-less MoCap \\[8pt]\hline

\href{https://cvssp.org/data/totalcapture/}{TotalCapture} & video + 3D pose & - 1.9M frames using 8 Vicon cameras \\
\citep{Trumble2017BMVC} & - Single person, indoor & - 5 people (4M, 1F) perform 5 actions repeated 3 times \\
& & - GT 3D MoCap w/ 8 Vicon cameras + IMUs \\[8pt]\hline

\href{https://vcai.mpi-inf.mpg.de/projects/SingleShotMultiPerson/}{MuCo-3DHP} & video + 3D pose & - Composited from MPI-INF-3DHP \\
\citep{singleshotmultiperson2018} & - Multi-person, indoor & - GT 3D pose using multi-view marker-less MoCap \\[8pt]\hline\hline

\multicolumn{3}{|c|}{\textbf{video + 3D pose in SMPL format}}\\\hline

\href{https://files.is.tuebingen.mpg.de/classner/up/}{UP-3D} & video + 3D pose (SMPL) & - 7k frames \\
\citep{Lassner2017UP} & - Single person, outdoor & \quad - 5.5k from LSP, LSP-extended, MPII-HumanPose\\
& & \quad - 1.5k from FashionPose \\
& & - GT 3D pose by fitting SMPL on 2D pose \\[8pt]\hline

\href{https://vcai.mpi-inf.mpg.de/projects/SingleShotMultiPerson/}{MuPoTS-3D} & video + 3D pose (SMPL) & - $>$8k frames, 20 sequences, 8 subjects \\
\citep{singleshotmultiperson2018} (eval only) & - Multi-person, indoor \& outdoor & - GT 3D pose using multi-view marker-less MoCap \\[8pt]\hline

\href{https://europe.naverlabs.com/research/computer-vision/mannequin-benchmark/}{Mannequin} & video + 3D pose (SMPL) &  - 24k frames, 567 scenes, 742 subjects ($\leq$5 per frame) \\
\citep{singleshotmultiperson2018} & - Multi-person, outdoor & - Videos from mannequin challenge \\
& - Videos: people as mannequins & - GT using optimization of SfM and tracking to SMPL  \\[8pt]\hline

\href{https://virtualhumans.mpi-inf.mpg.de/3DPW/}{3DPW} & video + 3D pose (SMPL) &  - $>$51k frames, 60 sequences, 1700secs @ 30Hz \\
\citep{von2018recovering} & - Multi-person, outdoor & - 7 actors in 18 clothing styles \\
& & - GT ``MoCap'' using IMUs $\to$ 3D pose by fitting SMPL  \\[8pt]\hline

\href{https://github.com/ChenFengYe/SportsCap}{SMART} & Sport video + 3D pose (SMPL) &  - 45k frames of 30 athletes, 9 activities \\
\citep{chen2021sportscap} & - Multi-person, indoor & - GT marker-based MoCap using 12 Vicon cameras $\to$ SMPL \\[8pt]\hline

\href{https://www.biomotionlab.ca/movi/}{MoVi} & video + 3D pose (SMPL) &  - 700k frames with 90 subjects (60 female, 30 male, 5 left-handed) \\
\citep{ghorbani2021movi} & - Multi-person, indoor & \quad - 20 pre-defined actions and 1 self-chosen movement \\
& & \quad - 5 data capture rounds, only `S1' and `S2' for video+3D \\
& & - GT using 4 cameras + IMU $\to$ 3D pose using MoSh++ \\[8pt]\hline\hline

\multicolumn{3}{|c|}{\textbf{Only 3D pose}}\\\hline

\href{https://amass.is.tue.mpg.de/}{AMASS} & Only 3D pose (SMPL) & - 42 hours of MoCap, 346 subjects, 11451 motion \\
\citep{amass} & & - Combines CMU, MPI-HDM05, MPIPose Limits, KIT, BioMotion Lab, TCD, ACCAD \\
& & - SMPL 3D shape (16), DMPL soft tissue coeffs (8), and full SMPL pose (90) \\[8pt]\hline

\href{https://amass.is.tue.mpg.de/}{Synchronized Scans and Markers (SSM)} & Only 3D pose (MoCap+Body) & Part of AMASS training:  dense 3D meshes in motion, with marker-based mocap \\[8pt]\hline

\end{longtable}
}
}

\subsection{Inverse Kinematics (IK)}

Inverse Kinematics (IK) is the estimation of 3D positions and rotations of body joints given some end-effector locations. It is a prominent problem in robotics and animation, and is traditionally solved by analytical or iterative optimization methods comprehensively reviewed by~\citet{aristidou2018inverse}. Solving IK using machine learning techniques has consistently attracted attention~\citep{bocsi2011learning, d2001learning, de2008learning}, with more work focusing on neural networks based methods~\citep{el2018comparative,bensadoun2022neural,mourot2022survey}.

In the animation space, the current neural IK state-of-the-art is ProtoRes~\citep{oreshkin2021protores}. It takes a variable set of effector positions, rotations or look-at targets as inputs, and performs IK to reconstruct all joint locations and rotations. Its effectiveness in editing complex 3D character poses has been demonstrated in a real-time live demo~\citep{bocquelet2022ai}. One limitation of ProtoRes is that, being trained on a fixed skeleton, it does not explicitly include any learnt body shape prior.

In this work, we relax this limitation by integrating SMPL into ProtoRes, and training on a large dataset of SMPL human pose data called AMASS~\citep{amass}.

\subsection{Retargeting}

Retargeting is the task of transferring the pose of a source character to a target character with a different morphology (bone lengths) and possibly a different topology (number of joints, connectivity, etc.)~\citep{gleicher1998retargetting}. Retargeting is a ubiquitous task in animation, and procedural tools exist for retargeting between skeletons of different morphologies and topologies~\citep{unity2022retargeting}. 


\subsection{SMPL and IK}

There is very little prior work that attempts to use an IK-enabled SMPL model for 3D character animation. \citet{bebko2021bmlSUP} pose SMPL characters in the Unity platform, but do not perform any IK. \citet{minimalIk} performs IK on SMPL parameters using standard optimization, but only in the full pose context, which has very limited applicability for artistic pose editing.  VPoser~\citep{pavlakos2019expressive} trains a Variational Auto-Encoder (VAE) to work as a prior on 3D human pose obtained from SMPL. This VAE is used as an iterative IK solver for a pose defined via keypoints. However, the VPoser architecture only works with relatively dense \emph{positional} inputs (no ability to handle sparse heterogeneous effector scenarios has been demonstrated). It also requires on-line L-BFGS optimization, making it too rigid and computationally expensive for pose authoring. There is a clear gap between SMPL and IK: existing IK models suitable for artistic pose editing do not support SMPL, and existing SMPL-based models have insufficient IK cababilities.


\newpage
\section{SMPL Inverse Kinematics (SMPL-IK)}
\label{sec:smpl-ik}

We propose SMPL-IK, a learned morphology-aware inverse kinematics module that accounts for SMPL shape and gender information to compute the full pose including the root joint position and 3D rotations of all SMPL joints based on a partially-defined pose specified by SMPL $\beta$-parameters (body shape), gender flag, and a few input effectors (positions, rotations, or look-at targets). SMPL-IK supports effector combinations of arbitrary number and type. SMPL-IK extends the learned inverse kinematics model ProtoRes~\citep{oreshkin2021protores}. ProtoRes only deals with a fixed morphology scenario in which an ML-based IK model is trained on a fixed skeleton. We remove this limitation by conditioning the ProtoRes computation on the SMPL $\beta$-parameters and gender
(see \Cref{app:smpl_ik_neural_network} for technical details). This results in an IK model that can operate on the wide range of morphologies incorporated in the expansive dataset used to create the SMPL model itself.

There are multiple advantages of this extension, including the following. First, rich public datasets can be used to train a learned IK model, in our case we train on the large AMASS dataset~\citep{amass}. Second, an animator can now edit both the pose and the body shape of a flexible SMPL-based puppet using a state-of-the-art learned IK tool, which we demonstrate in \Cref{app:editing_both_shape_and_pose}. Third, training IK in SMPL space unlocks a seamless interface with off-the-shelf AI algorithms operating in a standardized SMPL space, such as computer vision pose estimation backbones.

\section{SMPL Shape Inversion (SMPL-SI)} \label{sec:smpl_is}

SMPL-SI maps arbitrary humanoid skeletons onto their SMPL approximations by learning a mapping from skeleton features to the corresponding SMPL $\beta$-parameters (solving the inverse shape problem). Therefore, it can be used to map arbitrary user-supplied skeletons in the SMPL representation, and hence integrate SMPL-IK with custom user skeletons. Recall that the SMPL model implements the following forward equation: 
\begin{align} \label{eqn:smpl_forward_equation}
    \vec{p} = \smpl(\beta, \theta),
\end{align}
mapping shape parameters $\beta \in \Re^{10}$ and pose angles $\theta \in \Re^{22\times 3}$ into SMPL joint positions $\vec{p} \in \Re^{24\times 3}$. Datasets such as H36M contain multiple tuples $(\vec{p}_i, \beta_i, \theta_i)$. In principle, the pairs of H36M's skeleton features $\vec{f}_i$ extracted from  $(\vec{p}_i, \theta_i)$ and corresponding labels $\beta_i$, could be used for training a shape inversion model:
\begin{align}
    \hat\beta = \smplis(\vec{f}).
\end{align}
However, the H36M training set contains only 6 subjects, meaning the entire dataset has only 6 distinct $\beta_i$ vectors, thus insufficient for learning any meaningful $\smplis$.

Accordingly, we propose to train the $\smplis$ model as follows. We randomly sample 20k tuples $(\vec{p}_i, \beta_i)$ with $\widetilde\beta_i = [\epsilon_i, s_i]$, where $\epsilon_i \in \Re^{10}$ is a sample from uniform distribution $\mathcal{U}(-5, 5)$ and $s_i \in \Re$ is the scale sampled from $\mathcal{U}(0.2, 2)$. Scale $s_i$ is used to account for the variation in the overall size of the user-supplied characters relative to the standard SMPL model. We then use the $\smpl$ forward equation~\eqref{eqn:smpl_forward_equation} to compute joint positions $\widetilde{\vec{p}}_i$ corresponding to $\widetilde\beta_i$ and $\theta_i$ set to the T-pose. We then compute skeleton features $\widetilde{\vec{f}}_i$ for each $\widetilde{\vec{p}}_i$ as distances between the following pairs of joints: (right hip, right knee), (right knee, right ankle), (head, right ankle), (head, right wrist), (right shoulder, right elbow), (right elbow, right wrist). Finally, we implement the kernel density estimator using the 20k samples and the user skeleton features $\vec{f}$ to estimate the shape parameters $\widetilde\beta$ of the $\smpl$ model:
\begin{align} \label{smplsi_main_equation}
    \hat\beta = \sum_i \frac{\widetilde\beta_i w_i}{\sum_j w_j}, \quad w_i = \kernel((\vec{f} - \widetilde{\vec{f}}_i) / h).
\end{align}
The implementation uses a Gaussian kernel $\kernel$ with width $h=0.02$. The reason for this is that since there can be multiple equally plausible $\beta$'s for each skeleton (making SMPL-SI an ill-defined problem, like many other inverse problems), a point solution of the inverse problem may be degenerate. To address this, the general solution is formulated in probabilistic Bayesian terms based on the joint generative distribution of skeleton shape and features $p(\widetilde\beta, \vec{f})$. The Bayesian $\beta$-estimator is then derived from the corresponding posterior distribution of $\beta$ parameters given features:
\begin{align}
    \hat\beta = \int \widetilde\beta p(\widetilde\beta | \vec{f}) d \widetilde\beta,
\end{align}
Note that $\hat\beta$ mixes a few likely values of $\widetilde\beta$ corresponding to posterior distribution modes. Decomposing $p(\widetilde\beta, \vec{f}) = p(\vec{f} | \widetilde\beta) p(\widetilde\beta)$ we get:
\begin{align} \label{eqn:beta_estimate_decomposition}
    \hat\beta = \int \frac{\widetilde\beta p(\vec{f} | \widetilde\beta) p(\widetilde\beta) d \widetilde\beta}{\int p(\vec{f} | \widetilde\beta) p(\widetilde\beta) d\widetilde\beta}. 
\end{align}
Since the joint distribution $p(\widetilde\beta, \vec{f})$ is unknown, we approximate it using a combination of kernel density estimation and Monte-Carlo sampling. Assuming conservative uniform prior for $p(\widetilde\beta)$, we sample $\beta$ as described above and we use a kernel density estimator $p(\vec{f} | \widetilde\beta) \approx \frac{1}{hN} \sum_i \kernel(\frac{\vec{f} - \widetilde{\vec{f}}_i}{h})$. Using this in~\eqref{eqn:beta_estimate_decomposition} together with Monte-Carlo sampling from $p(\widetilde\beta)$, results in~\eqref{smplsi_main_equation}.

\section{Proposed workflow} \label{sec:proposed_ai_driven_flow}

\begin{figure}[!htb] 
  \centering
  \includegraphics[width=\linewidth]{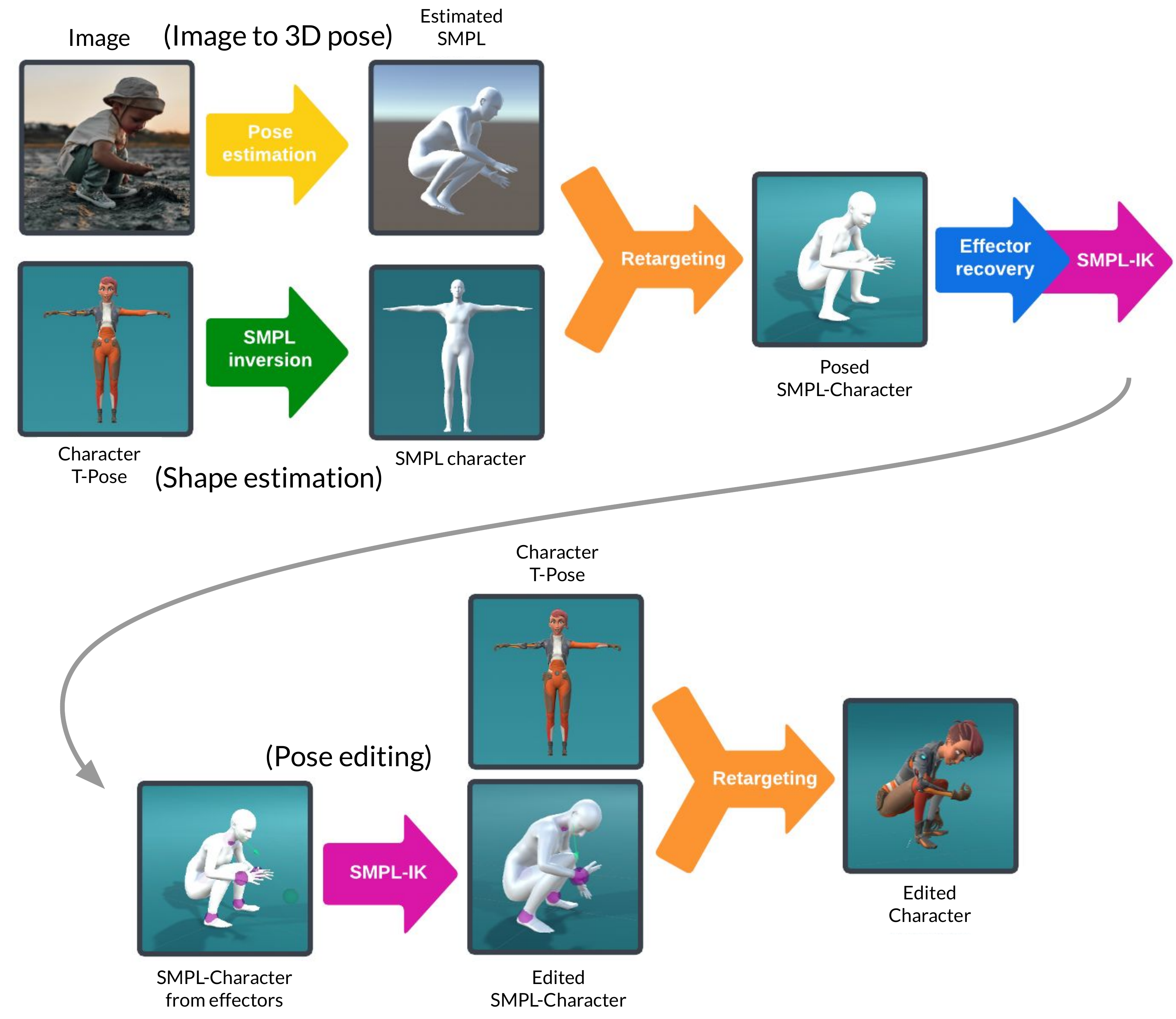}
  \caption{Pipeline for pose estimation and editing from a 2D image to a custom humanoid 3D character in the same pose as the human in the 2D image. Using any advanced animation tool, our approach enables an animator to apply poses extracted from an image to a custom user-defined 3D character and edit them further using SMPL-IK, our state-of-the-art machine learning inverse kinematics tool, integrated in real-time 3D development software (such as Unity as shown here). Given an RGB image of a human in a certain pose, a pose estimation algorithm is used to estimate the human's pose in terms of SMPL parameters. Our novel SMPL-SI then maps the skeleton of the user-defined 3D character onto its SMPL approximation, and the estimated human pose is retargeted onto the SMPL shape approximation obtained from SMPL-SI. This gives an SMPL mesh in the human pose with the morphology of the custom 3D character. From here, the animator can perform pose editing in the SMPL space of pose parameters, using our proposed SMPL-IK, while keeping the shape parameters fixed. To provide the user with effectors that control the pose of the 3D character (shown in purple in the demo videos), our proposed Effector Recovery method extracts only a few effectors to create an editable initial pose, while best preserving the estimated pose.
  }
  \label{fig:pipeline}
\end{figure}

\Cref{fig:unity_teaser} presents a high-level summary of the proposed artistic workflow for 3D scene authoring from an image, while \Cref{fig:pipeline} provides the detailed overview of how it is implemented for a user-defined humanoid character. \Cref{app:editing_both_shape_and_pose}, \Cref{app:pose_authoring} and \Cref{app:labeling_tool_demo} depict simpler workflows for authoring SMPL poses, image labeling in the SMPL space and authoring poses on custom characters from scratch. These were implemented in the 3D real-time Unity engine for validation. These workflows leverage SMPL-IK and SMPL-SI building blocks as well as some others described in the rest of this section.

\subsection{Overall pipeline}

Our overall pipeline from an image with a human to a custom humanoid 3D character in the same pose as the human is as is shown in \Cref{fig:pipeline}. The individual components are expanded upon in the below subsections.

\textbf{Image to 3D pose:} First, given an RGB image of a human in a certain pose, a state-of-the-art pose estimation algorithm is used to extract an estimate of the human's pose in terms of simple parameters. In practice, we use the ROMP model~\citep{sun2021romp} for pose estimation, but any other pose estimation model that outputs SMPL parameters could be used. As mentioned before, SMPL parameters consist of shape/beta parameters that control the body shape, and pose parameters that control pose-dependent deformations.

\textbf{Custom character shape estimation using SMPL-SI:} Then, we use our novel framework of SMPL Shape Inversion (SMPL-SI) to map the skeleton of a user-defined humanoid 3D character onto its SMPL approximation. SMPL-SI learns a mapping from skeleton features to the corresponding simple shape/beta parameters, thus solving the inverse shape problem. These features are the distance between key joints in the canonical pose, like distance between the elbow and the wrist of the character. We achieve this by implementing the kernel density estimator for the shape parameters of the SMPL model, approximating the user-supplied skeleton.

\textbf{Retarget pose to custom character:} We then retarget the initial pose estimation result onto the SMPL shape approximation of the custom 3D character obtained via SMPL-SI. This gives us an SMPL mesh in the pose provided in the image, but with the morphology of the custom 3D character.

\textbf{Pose editing:} From here, we perform pose editing using our proposed SMPL-IK in the SMPL space of pose parameters, keeping the shape parameters fixed. The user is provided with effectors that control the pose of the 3D character (shown in purple in the demo videos). When a user edits the pose of one or more effectors, a new full pose is estimated by SMPL-IK.

SMPL-IK computes the full pose, including the root joint position and 3D rotations of all SMPL joints, based on a partially defined pose specified by SMPL beta parameters, that is the body shape, gender flag, and a few input effectors, such as the positions, rotations, or look-at targets. 
SMPL-IK supports effector combinations of arbitrary number and type.

\textbf{Effector recovery:} We also propose effector recovery to extract only a few effectors to create an editable initial pose, while best preserving the estimated pose. More effectors means better reconstruction at the cost of less freedom to the posing model. 

Hence, given the edited SMPL character with the new full pose, this pose is then retargeted back to the custom 3D character. Because the morphology of the SMPL character is similar to that of the custom character, this retargeting best preserves the edited pose.

Each of these stages is expanded upon below.

\subsection{Image to 3D pose}

We propose to process a monocular RGB image to initialize an editable 3D scene as shown in \Cref{fig:unity_teaser}. Several methods exist for pose estimation from RGB inputs, most recent of which include ROMP~\citep{sun2021monocular} and HybrIK~\citep{li2021hybrik}. In our approach, we use a pre-trained ROMP model that predicts shape, 3D joint rotations and 3D root joint location for each human instance in the image. The outputs of pose estimation can be directly used to edit the estimated 3D SMPL mesh using SMPL-IK, leading to advanced 3D labelling tools that can be used to refine pose estimation and augmented reality datasets, as described in \Cref{app:labeling_tool_demo}. Alternatively, pose estimation results can be retargeted to user-supplied 3D characters, in which case the 3D scene with retargeted characters is further edited through the combination of SMPL-IK and SMPL-SI as explained below.

\subsection{Custom character shape estimation using SMPL-SI}

Pose estimation algorithms output pose in the standardized SMPL space, whereas users may wish to repurpose the pose towards their own custom character. We use SMPL-SI to find the best approximation of the custom character by estimating its corresponding SMPL $\beta$ parameters from the custom skeleton features (e.g. certain bone lengths). The SMPL character created using SMPL-SI provides a good approximation of the user character hence providing for its smooth integration with SMPL-IK, operating in the standard SMPL space.

\subsection{Retarget pose to custom character}

In \Cref{fig:pipeline}, procedural retargeting first retargets the initial pose estimation result onto the SMPL approximation of the user-defined character obtained via SMPL-SI. Second, it retargets the pose edited by the animator with SMPL-IK back on the user character. On both occasions, SMPL-SI makes the job of procedural retargeting easier. First, it aligns the topology of user character with the SMPL space. Second, the SMPL character derived via SMPL-SI is a close approximation of the user character, simplifying the transfer of the pose edited with SMPL-IK back onto the user character.


\subsection{Pose editing}

Pose editing relies on the Unity UX integration of SMPL-IK similar to one of ProtoRes and augmented with the SMPL shape editing controls as well as pose estimation, SMPL-SI, retargeting and effector recovery integrations. Editing happens directly in the user character space following the WYSIWYG paradigm. A full pipeline demo is presented in~\Cref{app:full_pipeline_demo}.

\subsection{Effector recovery}

Pose estimation outputs a full pose (24 3D joint angles and 3D root joint location) of each human in the scene. The pose editing process constrained by this information would be very tedious. SMPL-IK makes pose authoring efficient using very sparse constraints (e.g. using 5-6 effectors). Therefore, we propose to extract only a few effectors to create an editable initial pose. We call this \emph{Effector Recovery}, which proceeds starting from an empty set of effectors, given the full pose provided by the computer vision backbone, in an iterative greedy fashion. Out of the remaining effectors, we add one at a time, run a new candidate effector configuration through SMPL-IK, and obtain the pose reconstructed from this configuration. We then choose a new effector configuration by retaining the candidate effector set that minimizes the L2 joint reconstruction error in the character space. We repeat this process until either the maximum number of allowed effectors is reached, or the reconstruction error falls below a fixed threshold. We find this greedy algorithm very effective in producing a minimalistic set of effectors most useful in retaining the initial pose, which is shown in supplementary video discussed in \Cref{app:effector_recovery}.

\section{SMPL-IK details} \label{app:smpl_ik_details}

\subsection{SMPL-IK neural network diagram} \label{app:smpl_ik_neural_network}

\begin{figure}[!htb]
  \centering
  \includegraphics[width=\columnwidth]{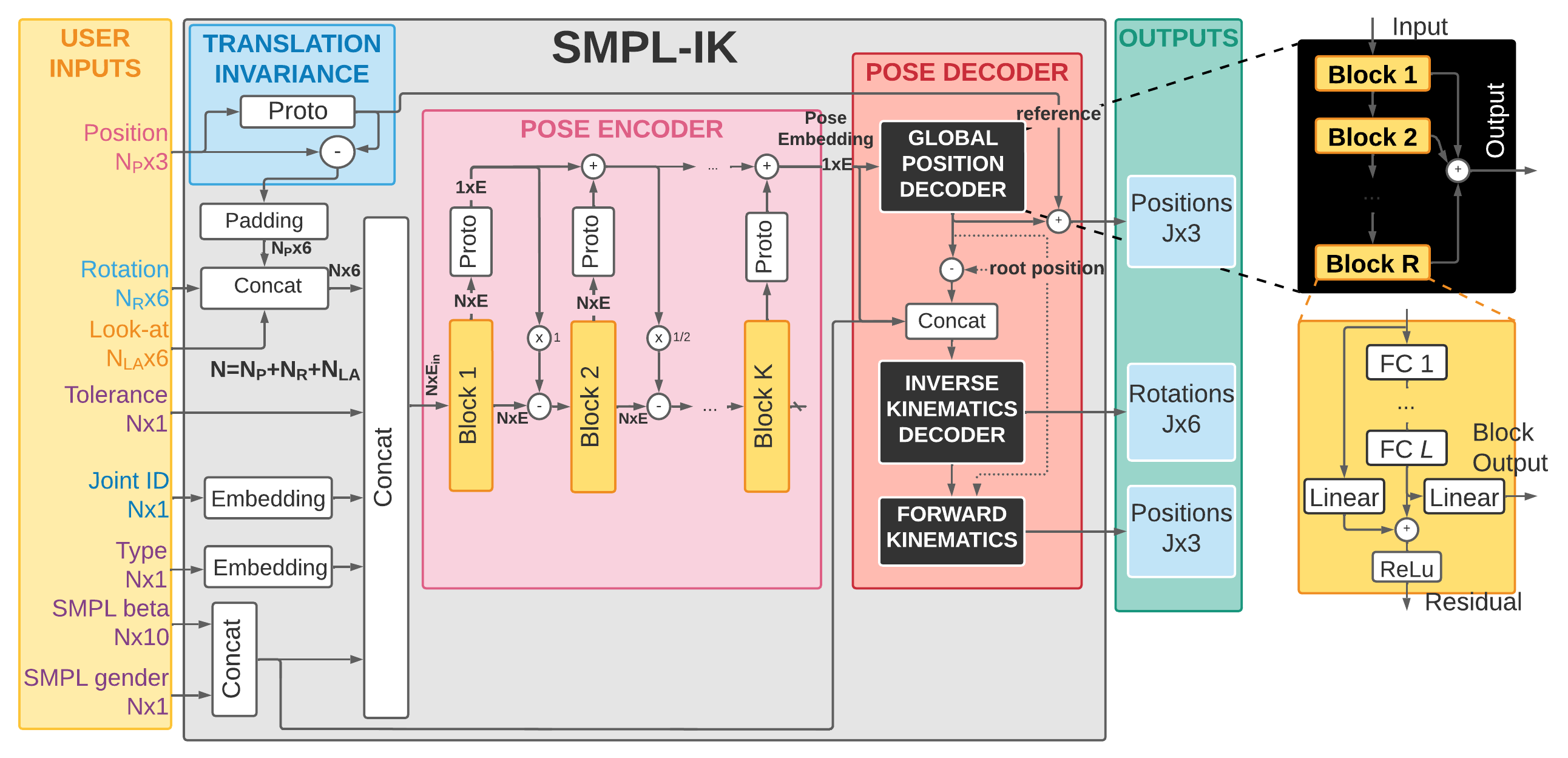}
  \caption{SMPL-IK neural network diagram. Note input conditioning on $\beta$ and gender inputs.}
  \label{fig:smpl_ik_neural_network}
\end{figure}

\Cref{fig:smpl_ik_neural_network} shows the overall architecture of the model. The inputs to the model are a variable number of 3D positions, rotations, look-at (direction of the head), the tolerance in the estimation, the ID of the joints being given as input, their type (position or rotation or look-at), and the SMPL parameters of body shape and gender. These inputs are fed to a variant of the ProtoRes model~\citep{oreshkin2021protores}. The power of the ProtoRes model is that it is capable of handling a variable number of inputs using a specific architecture in the Pose Encoder module. Then, a Pose Decoder module transforms the features encoded by the Pose Encoder into meaningful outputs for the full pose : all joint positions and rotations. The Pose Decoder consists of a global position decoder, an inverse kinematics decoder that outputs the joint rotations, and a forward kinematics decoder that outputs the joint positions.

It is in the Pose Decoder that we incorporate SMPL. The inverse and forward kinematics are handled by the relevant equations in SMPL, conditioned on the body shape and gender provided by the user.

\subsection{SMPL-IK training details} \label{app:splik_training_details}

The training process overall follows that of ProtoRes~\citep{oreshkin2021protores}. We found that naively training on AMASS works well, while for H36M, simply training on its 6 subjects results in the lack of generalization in the $\beta$ subspace of inputs. To overcome this, we used the following $\beta$ augmentation strategy. For each sample drawn from the H36M dataset, we added white Gaussian noise with unit variance, and recalculated joint positions based on the augmented $\beta$ value and the pose $\theta$ from the dataset. The model was trained on the augmented H36M dataset. We found that overall, the model quality was better when it was trained on the AMASS dataset, although the quality of the H36M model was also acceptable. 

\subsection{SMPL-IK evaluation details}
\label{app:splik_eval_details}

The datasets used to measure quantitative generalization results are H36M and AMASS. We used H36M train and test splits derived in ROMP~\citep{sun2021monocular}, which in turn follow H36M Protocol 2 (subjects S1, S5, S6, S7, S8 for training and S9, S11 for test, plus 1:10 subsampling of the training set). For AMASS, we take the train/validation/test splits from~\citet{amass} (valdation datasets: HumanEva, MPI\_HDM05, SFU, MPI\_mosh; test datasets: Transitions\_mocap, SSM\_synced; training datasets: everything else).

The evaluation metrics we chose to quantify SMPL-IK are commonly used in the context of H36M and AMASS datasets: MPJPE, PA-MPJPE, and the geodesic rotation error, which was shown to be important in quantifying the quality of realistic poses in~\citet{oreshkin2021protores}. The metrics are defined as follows. 

Mean Per Joint Position Error (MPJPE) is computed by flattening all poses and joints into the leading dimension, resulting in the ground truth $\vec{p} \in \Re^{N \times 3}$ and its prediction $\widehat{\vec{p}}$:
\begin{align} 
\mpjpe(\vec{p}, \widehat{\vec{p}}) = \frac{1}{N} \sum_{i=1}^N \| \vec{p}_i - \widehat{\vec{p}}_i \|_2.
\end{align}

PA-MPJPE, Procrustes aligned MPJPE, is MPJPE calculated after each estimated 3D pose in the batch is aligned to its respective ground truth by the Procrustes method, which is simply a similarity transformation.

GE, geodesic error, between a rotation matrix $\vec{R}$ and its prediction $\widehat{\vec{R}}$,~\citet{salehi2018real}:
\begin{align} \label{eqn:geodesic_loss}
\geodesic(\vec{R}, \widehat{\vec{R}}) = \arccos\left[(\tr(\widehat{\vec{R}}^T \vec{R}) - 1) / 2  \right].
\end{align}

All metrics in \Cref{table:protorez_vs_baselines} are computed on test sets of AMASS and H36M using models trained on respective training sets using the randomized effector benchmark framework described in detail in~\citet{oreshkin2021protores}. Notably, we evaluate the model's performance by assessing pose reconstruction quality from sparse variable inputs. We randomly sample 2 to 64 effectors to be used as inputs and average reconstruction errors across multiple iterations.

\section{Empirical results}
\label{protores:res}

\begin{table}[!ht] 
    \centering
    \caption{SMPL-IK benchmark following the randomized effector scheme~\citep{oreshkin2021protores} on AMASS and H36M datasets, based on MPJPE (Mean Per Joint Position Error), PA-MPJPE (Procrustes-Aligned MPJPE), and GE (Geodesic Error) metrics.}
    \label{table:protorez_vs_baselines}
    \begin{tabular}{ccc|ccc} 
        \toprule
        \multicolumn{3}{c|}{AMASS} & \multicolumn{3}{c}{H36M} \\
        \cmidrule(lr){1-3} \cmidrule(lr){4-6} 
        MPJPE & PA-MPJPE  & GE & MPJPE & PA-MPJPE  & GE \\
        \hline
        59.3 & 52.5 & 0.1602  & 65.8 & 57.9 & 0.224  \\ 
        \bottomrule
    \end{tabular}
    \label{tab:protores}
\end{table}

In \autoref{table:protorez_vs_baselines}, we report pose reconstruction errors of our SMPL-IK approach for two datasets: AMASS~\citep{amass} and Human3.6M~\citep{ionescu2014h36m}
(see \Cref{app:splik_eval_details} for more details). Since the evaluation metrics are reconstruction errors from sparse inputs, it is not possible to compare with prior methods that don't use this ramndomized effector scheme~\citep{oreshkin2021protores}.

\section{Limitations}
\label{protores:lim}

SMPL-IK and SMPL-SI are most effective when dealing with realistic human shapes and poses, because they are trained on the SMPL model and realistic 3D pose data from the AMASS dataset. Obviously, they perform worse when dealing with unrealistic and disproportionate human body types, such as those of certain cartoon characters. 
SMPL-SI relies on a set of joints to compute user character features. These joints are present in most characters, but without them its operation is not viable. 





\section{Demos}
\label{protores:demos}

All demo videos are available here: \url{https://drive.google.com/drive/u/1/folders/1bHwoZjAX9njFCGszzLpUtOFGXxs0sWKW}.

\subsection{Editing both shape and pose demo} \label{app:editing_both_shape_and_pose}
 \begin{figure}[!ht]
  \centering
  \includegraphics[width=\linewidth]{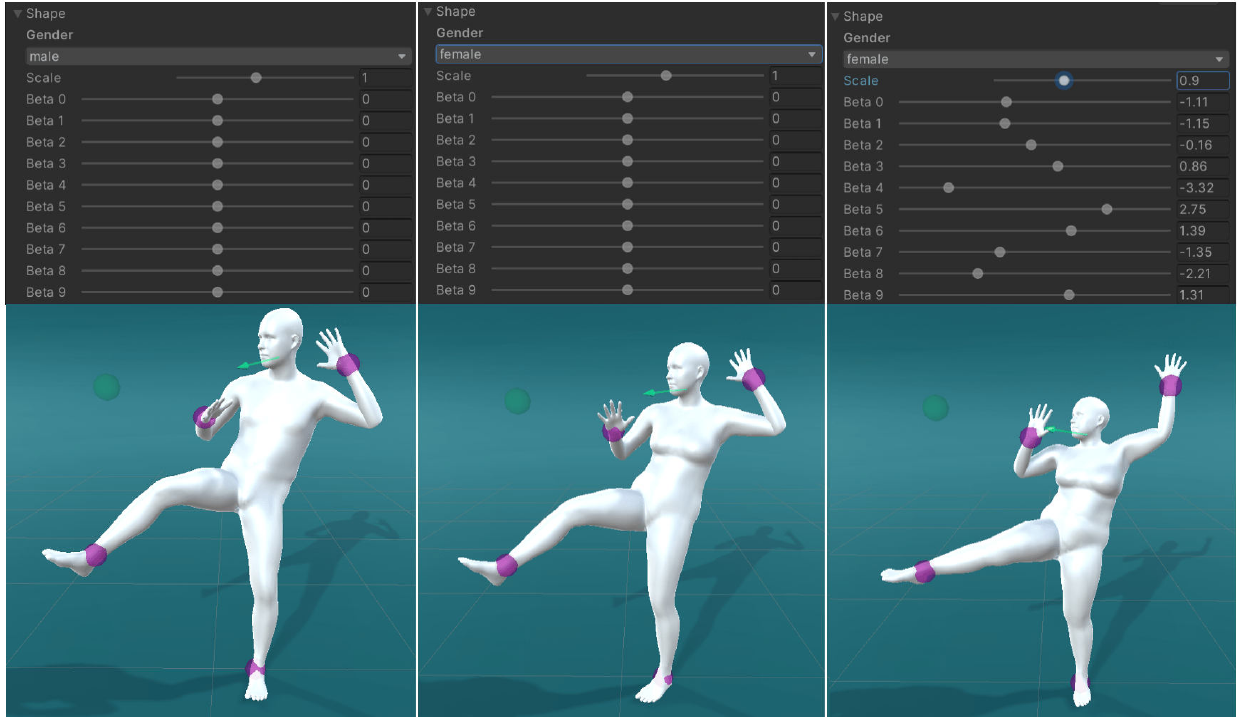}
  \caption{Morphology-aware learned IK. Left: posing the average male SMPL character. Center: result of modifying only the SMPL gender parameter. Right: result of additionally modifying the SMPL $\beta$ shape parameters.}
  \label{fig:shape_and_demo_posing}
\end{figure}
This is shown in the video \url{Demo\_Pose\_and\_Shape\_Editing.mp4}.
Compared to ProtoRes, SMPL-IK adds the additional flexibility of editing body shape together with pose. \Cref{fig:shape_and_demo_posing} demonstrates the user interface of shape editing, including the gender setting and the controls for the 10 SMPL $\beta$ parameters. In addition, the demo video shows how pose and shape of the SMPL character can be edited simultaneously. In this video, we demonstrate the benefit of SMPL-IK in pose authoring. First, we show how different effectors can be successfully manipulated using SMPL-IK leading to different realistic poses of the same body. Then, we show how changing body type, described by gender and scale, leads to different realistic versions of the same pose for different bodies. Finally, we show fine-grained modification of the body type by manipulating the SMPL shape parameters of the body. At every step, the corresponding pose is estimated using our SMPL-IK approach.

\subsection{Labeling tool demo} \label{app:labeling_tool_demo}
\begin{figure}[ht]
  \centering
  \includegraphics[width=\columnwidth]{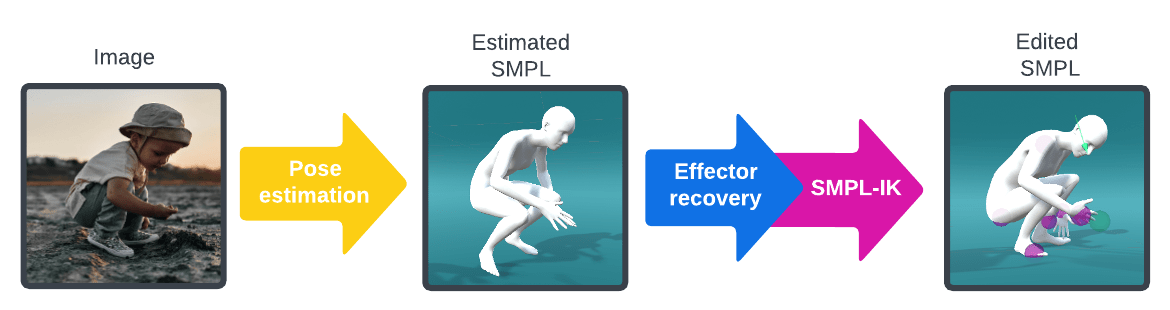}
  \caption{Pipeline for 2D image labeling with accurate 3D pose.}
  \label{fig:pipeline_labeling}
\end{figure}

This is shown in the video \url{Demo\_Labeling\_Tool.mp4}.
One of the drawbacks of the current pose estimation datasets based on real data is that only 3D or 2D positions of joints are actually labeled. However, it was shown that rotations are very important for representing a believable naturally looking pose~\citep{oreshkin2021protores}. SMPL-IK can be used as a labeling tool to add the missing 3D-rotation information to existing datasets, elevating them to the next level with minimal human effort. Given an image of a human, our SMPL-IK approach (combined with an off-the-shelf image-to-pose estimator) provides an editable 3D SMPL model in a pose close to the one in the image  (see \Cref{fig:pipeline_labeling}). The labeling tool based on SMPL-IK and its integration with Unity can be used to correct the joint rotations and specify the correct lookat (head/eyes direction) that is most often estimated incorrectly by the current state-of-the-art pose estimation algorithms due to the absence of this information in the current pose estimation datasets.

 \subsection{Pose authoring on a custom character} \label{app:pose_authoring}
\begin{figure}[ht]
  \centering
  \includegraphics[width=\linewidth]{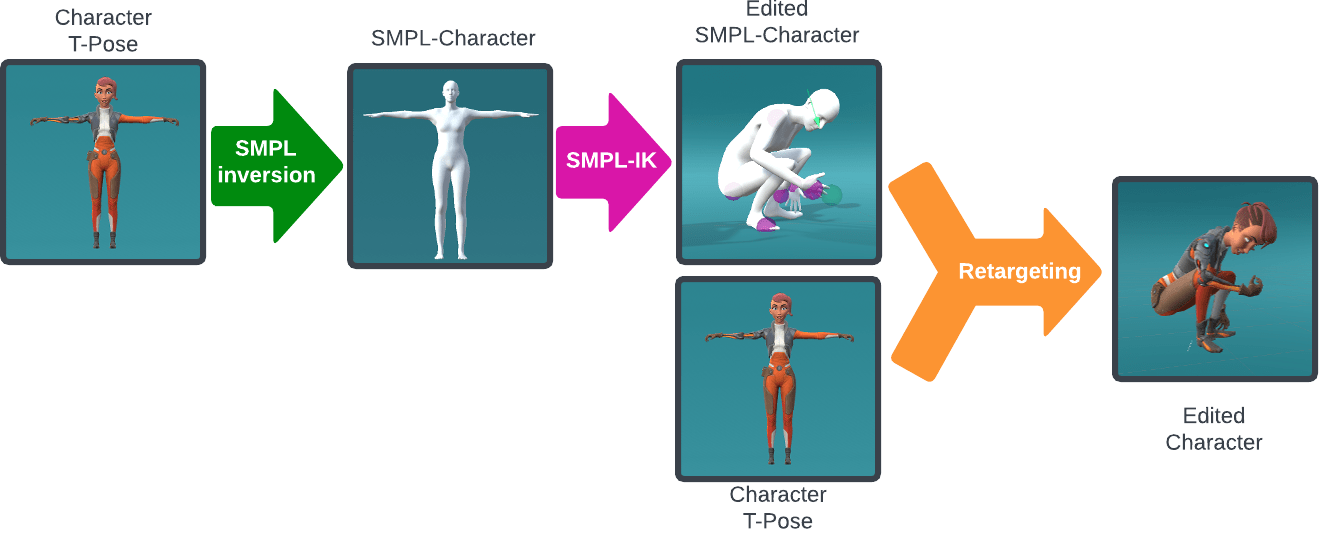}
  \caption{Pose authoring with a custom humanoid character via SMPL-IK and SMPL-SI.}
  \label{fig:pipeline_custom_posing}
\end{figure}

\Cref{fig:pipeline_custom_posing} depicts the simplified workflow that is used for authoring a pose for the custom user defined character using a combination of SMPL-IK and SMPL-SI. 
Supplementary videos \url{Demo_Authoring_Pose_Child.mp4},  \url{Demo_Authoring_Pose_Child.mp4}, \url{Demo_Authoring_Pose_Female.mp4}, \url{Demo_Authoring_Pose_Male.mp4}, \url{Demo_Authoring_Pose_Strong.mp4} show how SMPL-SI can be used to manipulate 4 custom characters (child, female, male and strong male) with different proportions and morphologies.

\subsection{Effector recovery} \label{app:effector_recovery}

The video \url{Demo_Effector_Recovery.mp4} demonstrates the effector recovery mechanism in action.
It shows the effect of changing the maximum number of effectors hyperparameter as well as the effect of changing number of recovered effectors on the initial pose extracted from image. It is clear that a relatively small number of effectors are sufficient to recover a good initialization for the editable pose.

\subsection{Full pipeline demos} \label{app:full_pipeline_demo}

\subsubsection{\url{Demo\_Crouch\_FineTuning.mp4}}:

In this video, we show how to edit a pose in Unity using our approach. First, given a user-provided 3D character, SMPL-SI is used to estimate the SMPL body shape parameters that best fit the character. This SMPL-Character is shown in the video transparently along with the 3D character, and is also shown in the second image in \Cref{app:pose_authoring}. Then, given an image of a human in a pose, such as the crouched baby in \Cref{app:labeling_tool_demo}, an off-the-shelf image-to-pose estimator is used to obtain its SMPL pose parameters. Then, the SMPL-Character is retargeted onto the estimated pose. Next, for further editing of the character from the new pose, Effector Recovery is performed to recover the best effectors that describe that pose for that character. The effectors are shown in purple. These effectors can now be used to edit the pose as the user wishes. Optionally, more effectors could be activated for further fine-tuned editing, including both positional and rotational effectors.

\subsubsection{\url{Demo\_Sitting\_Editing.mp4}}:

In this video, we demonstrate the case shown in \Cref{fig:unity_teaser}, with two humanoid 3D characters. As image of two people sitting is loaded, the poses of the two people are estimated using an off-the-shelf image-to-pose model, and the two 3D characters are retargeted to these estimated poses. Further, the pose of the 3D characters are then edited by manipulating the effectors. The video shows the various effectors and the effect of manipulating them. Every manipulation uses our SMPL-IK approach to estimate the realistic pose of that character.

\section{Conclusion}

In this work, we introduce SMPL-IK, an integration of SMPL into ProtoRes, a state-of-the-art inverse kinematics-based 3D human pose estimation model. Using this integrated model, we achieved full 3D human pose estimation from only partial input pose by training SMPL-IK on AMASS dataset. This included a thorough investigation into publicly available 3D human pose datasets. We then expanded this model’s capability to non-human bodies by proposing Shape Inversion, a retargeting technique. We also extended its capability to 3D human pose estimation from images, by stacking an image-to-pose model (ROMP) before SMPL-IK. Our model also allows pose editing by running SMPL-IK on newer poses formed by change in few pose effectors. We also proposed Effector Recovery, an iterative method to extract the minimal set of pose effectors that critically define a full pose. Finally, we unified everything into a pipeline that makes a realistic 3D pose editor requiring only partial pose input, initialized by a 3D human pose from an image.

%% file: NIDDPM.tex
\anglais
\counterwithin{figure}{chapter}
\counterwithin{table}{chapter}

\chapter{Non-Isotropic Denoising Diffusion Models~\citep{voletiv2022ddpm}}
\label{chap:NIDDPM}

\setcounter{section}{-1}
\section{Prologue to article}
\label{chap:pro_NIDDPM}

\subsection{Article details}

\textbf{Score-based Denoising Diffusion with Non-Isotropic Gaussian Noise Models}. Vikram Voleti, Christopher Pal, Adam Oberman. \textit{Advances in Neural Information Processing Systems (NeurIPS) 2022 Workshop}.

\textit{Personal contribution}:
The project began with discussions between the authors at Mila. Vikram worked on deriving a non-isotropic formulation of the noise process in denoising diffusion models. Adam Oberman and Christopher Pal provided advice and guidance throughout the project, including explaining the mathematical details of Stochastic Differential Equations, providing text books and source material to understand Gaussian processes better. Vikram derived the relevant equations for the isotropic Gaussian formulations (DDPM and SMLD covered in \Cref{chap:Background}), then derived the equations for the non-isotropic Gaussian variant. Adam Oberman provided the literature on Gaussian Free Fields (GFFs). Vikram Voleti wrote the code for the non-isotropic variant of DDPM, GFF, and conducted experiments on image generation.

\subsection{Context}

Generative models based on denoising diffusion techniques have led to an unprecedented increase in the quality and diversity of imagery that is now possible to create with neural generative models. However, most contemporary state-of-the-art methods are derived from a standard isotropic Gaussian formulation. A non-isotropic Gaussian variant had not been explored yet.

\subsection{Contributions}

In this work, we examine the situation where non-isotropic Gaussian distributions are used. We present the key mathematical derivations for creating denoising diffusion models using an underlying non-isotropic Gaussian noise model. We also provide initial experiments with the CIFAR10 dataset to help verify empirically that this more general modelling approach can also yield high-quality samples.

\subsection{Research impact}

This work derived the use of non-isotropic Gaussian process in the context of denoising diffusion models. Other works then derived the use of non-Gaussian noise processes such as Gamma noise~\citep{nachmani2021ddgm}, Poisson noise~\citep{xu2022Poisson}, Heat dissipation process~\citep{Rissanen2022heat}. The results on image generation using this non-isotropic formulation are by themselves not much better than those using the isotropic formulation. However, other work has utilized the real power of this formulation, and expanded the modality to a continuous domain i.e. a function space. Thus, as the forward process perturbs input functions gradually using a Gaussian process, instead of the data itself, it is now possible to model infinite-dimensional data using denoising diffusion models. This is shown by \citet{hagemann2023multilevel,bond2023infty}, and concurrently by \citet{lim2023scorebased} --- a project Vikram contributed to in collaboration with NVIDIA.

\newpage

\begin{figure}[!thb]
\vspace{-.5cm}
\centering
\begin{subfigure}{.18\textwidth}
  \centering
  \includegraphics[width=\linewidth]{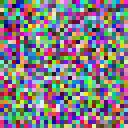}
  \caption{Isotropic}
  \label{fig:sub1}
\end{subfigure}
\begin{subfigure}{.18\textwidth}
  \centering
  \includegraphics[width=\linewidth]{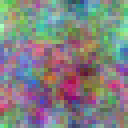}
  \caption{Non-isotropic}
  \label{fig:sub2}
\end{subfigure}
\vspace{-.2cm}
\caption{Gaussian noise samples.
}
\label{fig:gff_sample}
\end{figure}

\section{Introduction}

Score-based denoising diffusion models~\cite{song2019generative, ho2020ddpm, song2020sde} have seen great success as generative models for images~\cite{dhariwal2021diffusion, song2020improved}, as well as other modes such as video~\cite{ho2022VDM, yang2022ResidualVideoDiffusion, voleti2022MCVD}, audio~\cite{kong2021diffwave, chen2021wavegrad}, etc. The underlying framework relies on a noising "forward" process that adds noise to real images (or other data), and a denoising "reverse" process that iteratively removes noise. In most cases, the noise distribution used is the isotropic Gaussian i.e. noise samples are independently and identically distributed (IID) as the standard normal at each pixel.

In this work, we lay the theoretical foundations and derive the key mathematics for a non-isotropic Gaussian formulation for denoising diffusion models. It is our hope that these insights may open the door to new classes of models.
One type of non-isotropic Gaussian noise arises in a family of models known as  Gaussian Free Fields (GFFs)~\cite{sheffield2007gff, berestycki2015gff, bramson2016gff, werner2020gff} (a.k.a. Gaussian Random Fields). GFF noise can be obtained by either convolving isotropic Gaussian noise with a filter, or applying frequency masking of noise.
In either case this procedure allows one to model or generate smoother and correlated types of Gaussian noise.
In \Cref{fig:gff_sample,fig:GFFs}, we compare examples of isotropic Gaussian noise with GFF noise obtained using a frequency space window function consisting of $w(f)=\frac{1}{f}$.

Our contributions here consist of the following: (1) deriving the key mathematics for score-based denoising diffusion models using non-isotropic multivariate Gaussian distributions, (2) examining the special case of a GFF and the corresponding non-Isotropic Gaussian noise model, and (3) showing that diffusion models trained (eg. on the CIFAR-10 dataset~\cite{krizhevsky2009learning}) using a GFF noise process are also capable of yielding high-quality samples comparable to models based on isotropic Gaussian noise.

\Cref{sec:niddpm} and \Cref{sec:nismld} contain detailed derivations of our Non-Isotropic DDPM (NI-DDPM) and Non-Isotropic SMLD (NI-SMLD) denoising diffusion models.
\Cref{ni_comp} provides a direct comparison between DDPM~\citep{ho2020ddpm} and our NI-DDPM. \Cref{GFF} derives Gaussian Free Fields (GFFs) in connection with the previous sections. \Cref{NIDDPM_exps} provides results of image generation experiments using DDPM and our NI-DDPM.

\section{Non-isotropic DDPM (NI-DDPM)}
\label{sec:niddpm}

\subsection{Forward (data to noise) for NI-DDPM}
For a fixed sequence of positive scales $0 < \beta_1 < \cdots < \beta_L < 1$, $\balpha_t = \prod_{s=1}^t (1 - \beta_s)$, the \textbf{transition ``forward''} process is:
\begin{align}
&p_{t}^{\GFFDDPM} (\rvx_t \mid \rvx_{t-1}) = \gN(\rvx_t \mid \sqrt{1 - \beta_t} \rvx_{t-1}, \beta_t \mSigma)
\implies &\rvx_t = \sqrt{1 - \beta_t}\rvx_{t-1} + \sqrt{\beta_t} \sqrt{\mSigma} \rvz_{t-1},
\label{eq:NIDDPM_forward}
\end{align}
where $\sqrt{\mSigma}$ is the matrix square root of $\mSigma$ (e.g. as given by Cholesky decomposition).

\textit{Note} : $\sqrt{\mSigma}$ is \textit{not} the element-wise square root of $\mSigma$.

The \textbf{cumulative ``forward''} process can be derived as:
\begin{align}
&p_{t}^{\GFFDDPM} (\rvx_t \mid \rvx_0) = \gN(\rvx_t \mid \sqrt{\balpha_t} \rvx_0, (1 - \balpha_t) \mSigma). \\
\implies &\rvx_t = \sqrt{\balpha_t}\rvx_0 + \sqrt{1 - \balpha_t} \sqrt{\mSigma} \rvepsilon \implies \rvepsilon = \sqrt{\mSigma^{-1}}\frac{\rvx_t - \sqrt{\balpha_t}\rvx_0}{\sqrt{1 - \balpha_t}}.
\label{eq:GFF_DDPM_noise}
\end{align}

\subsection{Score for NI-DDPM}
\label{sec:niddpm_score}
\begin{align}
&\nabla_{\rvx_t} \log p_{t}^{\GFFDDPM} (\rvx_t \mid \rvx_0) = -\mSigma^{-1}\frac{\rvx_t - \sqrt{\balpha_t} \rvx_0}{1 - \balpha_t} = -\frac{1}{\sqrt{1 - \balpha_t}}\sqrt{\mSigma^{-1}}\rvepsilon.
\label{eq:score-noise-NIDDPM}
\end{align}

Derivation of the score value:
\begin{align*}
&p_{t}^{\GFFDDPM} (\rvx_t \mid \rvx_0) = \gN(\rvx_t \mid \sqrt{\balpha_t} \rvx_0, (1 - \balpha_t) \mSigma), \\
&= \frac{1}{(2\pi)^{D/2}((1 - \balpha_t)|\mSigma|)^{1/2}}\exp\left( -\frac{1}{2(1 - \balpha_t)}(\rvx_t - \sqrt{\balpha_t}\rvx_0)^\trns\mSigma^{-1}(\rvx_t - \sqrt{\balpha_t}\rvx_0) \right). \\
\implies &\log p_{t}^{\GFFDDPM} (\rvx_t \mid \rvx_0) = -\log ((2\pi)^{D/2}((1 - \balpha_t)|\mSigma|)^{1/2})
\\
&\qquad\qquad\qquad\qquad\qquad\quad - \frac{1}{2(1 - \balpha_t)}(\rvx_t - \sqrt{\balpha_t}\rvx_0)^\trns\mSigma^{-1}(\rvx_t - \sqrt{\balpha_t}\rvx_0), \\
\implies &\nabla_{\rvx_t} \log p_{t}^{\GFFDDPM} (\rvx_t \mid \rvx_0) = -\frac{1}{2(1 - \balpha_t)}2\mSigma^{-1}(\rvx_t - \sqrt{\balpha_t}\rvx_0) = -\frac{1}{\sqrt{1 - \balpha_t}}\sqrt{\mSigma^{-1}}\rvepsilon.
\end{align*}

\subsection{Score-matching objective for NI-DDPM}

The objective for score estimation in NI-DDPM at noise level $\balpha_t$ is:
\begin{align*}
\loss^{\GFFDDPM}(\rvtheta; \balpha_t) &\triangleq\ \frac{1}{2} \EE_{p_{t}^{\GFFDDPM}(\rvx_t \mid \rvx_0) p(\rvx_0)} \left[ \Norm{\rvs_\rvtheta (\rvx_t, \balpha_t) + \mSigma^{-1} \frac{\rvx_t - \sqrt{\balpha_t} \rvx_0}{1 - \balpha_t} }_2^2 \right],
\numberthis \\
&\triangleq\ \frac{1}{2} \EE_{p_{t}^{\GFFDDPM}(\rvx_t \mid \rvx_0) p(\rvx_0)} \left[ \Norm{\rvs_\rvtheta (\rvx_t, \balpha_t) +  \frac{1}{\sqrt{1 - \balpha_t}}\sqrt{\mSigma^{-1}}\rvepsilon }_2^2 \right].
\end{align*}

\subsection{Variance of score for NI-DDPM}
\begin{align*}
&\EE\left[ \Norm{ \nabla_{\rvx_t} \log q_{\balpha_t}^{\GFFDDPM} (\rvx_t \mid \rvx_0) }_2^2 \right],
= \EE\left[ \Norm{ -\mSigma^{-1}\frac{\rvx_t - \sqrt{\balpha_t} \rvx_0}{1 - \balpha_t} }_2^2 \right], \\
&= \EE\left[ \Norm{ \mSigma^{-1}\frac{\sqrt{1 - \balpha_t}\sqrt{\mSigma}\rvepsilon}{1 - \balpha_t} }_2^2 \right]
= \frac{1}{1 - \balpha_t}\mSigma^{-1}\EE\left[ \Norm{\rvepsilon}_2^2 \right] = \frac{1}{1 - \balpha_t} \mSigma^{-1}.
\numberthis
\end{align*}

\subsection{Overall objective function for NI-DDPM}
\label{subsec:niddpm_objective}

The overall objective function weights the score-matching objective by the inverse of the variance of the score at each time step:
\begin{align}
& \gL^{\GFFDDPM}(\rvtheta; \{\balpha_t\}_{t=1}^{L}) \triangleq \EE_t\ \lambda(\balpha_t) \ \loss^{\GFFDDPM}(\rvtheta; \balpha_t).
\end{align}
We consider three possibilities for the loss weight $\lambda(\balpha_t)$:

(a) \boldmath
$ \lambda_a(\balpha_t) = (1 - \balpha_t)\mSigma$.
\unboldmath
\begin{align*}
\gL^{\GFFDDPM}_a(\rvtheta; \{\balpha_t\}_{t=1}^{L})
&\triangleq \EE_{t, p_{t}(\rvx_t \mid \rvx_0) p(\rvx_0)} \left[ \Norm{ \sqrt{1 - \balpha_t}\sqrt{\mSigma} \rvs_\rvtheta (\rvx_t, \balpha_t) + \sqrt{\mSigma^{-1}}\frac{(\rvx_t - \sqrt{\balpha_t} \rvx_0)}{\sqrt{1 - \balpha_t}} }_2^2 \right], \\
&= \EE_{t, \rvepsilon, \rvx_0} \left[ \Norm{ \sqrt{1 - \balpha_t}\sqrt{\mSigma} \rvs_\rvtheta (\rvx_t, \balpha_t) + \rvepsilon }_2^2 \right].
\numberthis
\end{align*}

(b) \boldmath
$ \lambda_b(\balpha_t) = (1 - \balpha_t)$.
\unboldmath
\begin{align*}
\gL^{\GFFDDPM}_b(\rvtheta; \{\balpha_t\}_{t=1}^{L})
&\triangleq \EE_{t, p_{\balpha_t}(\rvx_t \mid \rvx_0) p(\rvx_0)} \left[ \Norm{ \sqrt{1 - \balpha_t} \rvs_\rvtheta (\rvx_t, \balpha_t) + \mSigma^{-1}\frac{(\rvx_t - \sqrt{\balpha_t} \rvx_0)}{\sqrt{1 - \balpha_t}} }_2^2 \right], \\
&= \EE_{t, \rvepsilon, \rvx_0} \left[ \Norm{ \sqrt{1 - \balpha_t} \rvs_\rvtheta (\rvx_t, \balpha_t) + \sqrt{\mSigma^{-1}} \rvepsilon }_2^2 \right].
\numberthis
\end{align*}

(c) \boldmath
$ \lambda_c(\balpha_t) = (1 - \balpha_t)\mSigma^2$.
\unboldmath
\begin{align*}
\gL^{\GFFDDPM}_c(\rvtheta; \{\balpha_t\}_{t=1}^{L})
&\triangleq \EE_{t, p_{t}(\rvx_t \mid \rvx_0) p(\rvx_0)} \left[ \Norm{ \sqrt{1 - \balpha_t} \mSigma \rvs_\rvtheta (\rvx_t, \balpha_t) + \frac{(\rvx_t - \sqrt{\balpha_t} \rvx_0)}{\sqrt{1 - \balpha_t}} }_2^2 \right], \\
&= \EE_{t, \rvepsilon, \rvx_0} \left[ \Norm{ \sqrt{1 - \balpha_t} \mSigma \rvs_\rvtheta (\rvx_t, \balpha_t) + \sqrt{\mSigma} \rvepsilon }_2^2 \right].
\numberthis
\end{align*}

\subsection{Noise-matching objective for NI-DDPM}

A score model that matches the actual score-noise relationship in \cref{eq:score-noise-NIDDPM} is:
\begin{align}
\rvs_\rvtheta(\rvx_t, \balpha_t) = -\frac{1}{\sqrt{1 - \balpha_t}}\sqrt{\mSigma^{-1}}\rvepsilon_\rvtheta(\rvx_t, \balpha_t).
\end{align}

In this case, the overall objective function changes to the noise-matching objective:
\begin{align*}
&\gL^{\GFFDDPM}_{a}(\rvtheta; \{\balpha_t\}_{t=1}^{L}) \triangleq \EE_{t, \rvepsilon, \rvx_0} \bigg[ \Norm{\rvepsilon - \rvepsilon_\rvtheta (\rvx_t, \balpha_t)}_2^2 \bigg]
\numberthis. \\
&\gL^{\GFFDDPM}_{b}(\rvtheta; \{\balpha_t\}_{t=1}^{L}) \triangleq \EE_{t, \rvepsilon, \rvx_0} \bigg[ \Norm{\sqrt{\mSigma^{-1}}\rvepsilon  - \sqrt{\mSigma^{-1}}\rvepsilon_\rvtheta (\rvx_t, \balpha_t)}_2^2 \bigg].
\numberthis \\
&\gL^{\GFFDDPM}_{c}(\rvtheta; \{\balpha_t\}_{t=1}^{L}) \triangleq \EE_{t, \rvepsilon, \rvx_0} \bigg[ \Norm{\sqrt{\mSigma}\rvepsilon - \sqrt{\mSigma}\rvepsilon_\rvtheta (\rvx_t, \balpha_t) }_2^2 \bigg].
\numberthis
\end{align*}

\subsection{Reverse (noise to data) for NI-DDPM}

The goal is to estimate the \textbf{reverse} transition probability $q_t(\rvx_{t-1} \mid \rvx_t)$. This is intractable, but it is possible to estimate it conditioned on $\rvx_0$ i.e. $q_t(\rvx_{t-1} \mid \rvx_t, \rvx_0)$.

We know from the forward process that:
\begin{align*}
&p_{t}^{\GFFDDPM} (\rvx_t \mid \rvx_0) = \gN(\rvx_t \mid \sqrt{\balpha_t} \rvx_0, (1 - \balpha_t) \mSigma). \\
\implies &\hat\rvx_0 = \frac{1}{\sqrt{\balpha_t}}(\rvx_t - \sqrt{1 - \balpha_t}\sqrt{\mSigma}\rvepsilon_{\rvtheta^*}(\rvx_t, \balpha_t)).
\numberthis
\label{eq:NIDDPM_reverse}
\end{align*}

From Bayes' theorem, we compute the parameters of $q_t(\rvx_{t-1} \mid \rvx_t, \rvx_0)$ i.e. the reverse process additionally conditioning on $\rvx_0$,  with the help of \cite{bishop2006pattern} 2.116.
For a variable $\rvu$ distributed as a normal with mean $\vmu$ and covariance matrix $\mLambda^{-1}$, and a dependent variable $\rvv$ conditionally distributed as a normal with mean $\rmA\rvu + \rvb$ and covariance matrix $\rmL^{-1}$, the marginal distribution of $\rvv$, and the other conditional distribution $p(\rvu \mid \rvv)$ are given as:
\begin{subequations}
\begin{empheq}[box=\widefbox]{align*}
p(\rvu) &= \gN(\rvu \mid \vmu, \mLambda^{-1}),
\\
p(\rvv \mid \rvu) &= \gN(\rvv \mid \rmA\rvu + \rvb, \rmL^{-1})
\\
\implies
p(\rvv) &= \gN(\rvv \mid \rmA\vmu + \rvb, \rmL^{-1} + \rmA \mLambda^{-1} \rmA^\trns),
\\
\implies p(\rvu \mid \rvv) &= \gN(\rvu \mid \rmC(\rmA^\trns \rmL (\rvv - \rvb) + \mLambda \vmu), \rmC), \rmC = (\mLambda + \rmA^\trns \rmL \rmA)^{-1}.
\end{empheq}
\end{subequations}
For NI-DDPM, $p(\rvu) = p_{t-1}^{\GFFDDPM}(\rvx_{t-1} \mid \rvx_0), p(\rvv) = p_t^{\GFFDDPM}(\rvx_{t})$:
\begin{align*}
p_{t-1}^{\GFFDDPM}(\rvx_{t-1} \mid \rvx_{0}) &= \gN(\rvx_{t-1} \mid \sqrt{\balpha_{t-1}}\rvx_0, (1 - \balpha_{t-1})\mSigma) .
\\
p_{t}^{\GFFDDPM}(\rvx_t \mid \rvx_{t-1}, \rvx_0) &= \gN(\rvx_t \mid \sqrt{1 - \beta_t}\rvx_{t-1}, \beta_t\mSigma).
\\
\implies \text{Given } p(\rvu) &= p_{t-1}^{\GFFDDPM}(\rvx_{t-1} \mid \rvx_0), p(\rvv \mid \rvu) = p_t^{\GFFDDPM}(\rvx_t \mid \rvx_0), \\
&\text{ we need } p(\rvu \mid \rvv) = q_{t-1}^{\GFFDDPM}(\rvx_{t-1} \mid \rvx_t, \rvx_0).
\\
\implies \vmu &= \sqrt{\balpha_{t-1}}\rvx_0,  \mLambda^{-1} = (1 - \balpha_{t-1})\mSigma, \rmA = \sqrt{1 - \beta_t}, \rvb = \vzero, \rmL^{-1} = \beta_t\mSigma.
\\
\implies \rmC &= \left(\frac{1}{1 - \balpha_{t-1}}\mSigma^{-1} + (1 - \beta_t)\frac{1}{\beta_t}\mSigma^{-1}\right)^{-1} = \left(\frac{\cancel{\beta_t} + 1 - \cancel{\beta_{t}} - \alpha_t}{(1 - \balpha_{t-1})\beta_t}\mSigma^{-1}\right)^{-1}, \\
&= \frac{1 - \balpha_{t-1}}{1 - \balpha_t}\beta_t\mSigma = \tilde\beta_t\mSigma.
\\
\implies \rmC(\rmA^\trns \rmL (\rvy - \rvb) &+ \mLambda \vmu) = \frac{1 - \balpha_{t-1}}{1 - \balpha_t}\beta_t\mSigma \left(\sqrt{1 - \beta_t}\frac{1}{\beta_t}\mSigma^{-1}\rvx_{t} + \frac{1}{1 - \balpha_{t-1}}\mSigma^{-1}\sqrt{\balpha_{t-1}}\rvx_0\right), \\
&= \frac{\sqrt{\balpha_{t-1}}\beta_t}{1 - \balpha_t}\rvx_0 + \frac{\sqrt{1 - \beta_t}(1 - \balpha_{t-1})}{1 - \balpha_t}\rvx_t.
\end{align*}

Thus, the parameters of the distribution of the reverse process are:
\begin{align*}
\therefore &q_t^{\GFFDDPM}(\rvx_{t-1} \mid \rvx_t, \rvx_0) = \gN(\rvx_{t-1} \mid \tilde\vmu_{t-1}(\rvx_t, \rvx_0), \tilde\beta_{t-1}\mSigma) \text{ , where} \\
&\tilde\vmu_{t-1}(\rvx_t, \rvx_0) = \frac{\sqrt{\balpha_{t-1}}\beta_t}{1 - \balpha_t}\rvx_0 + \frac{\sqrt{1 - \beta_t}(1 - \balpha_{t-1})}{1 - \balpha_t}\rvx_t ;\quad \tilde\beta_{t-1} = \frac{1 - \balpha_{t-1}}{1 - \balpha_t}\beta_t.
\numberthis \\
\implies &\rvx_{t-1}
= \tilde\vmu_{t-1}(\rvx_t, \hat\rvx_0) + \sqrt{\tilde\beta_{t-1}} \sqrt{\mSigma} \rvz,
\numberthis
\label{eq:NIDDPM_step2}
\\ &\text{where } \\
&\tilde\vmu_{t-1}(\rvx_t, \hat\rvx_0) = \frac{\sqrt{\balpha_{t-1}}\beta_t}{1 - \balpha_t}\left(\frac{1}{\sqrt{\balpha_t}}\big(\rvx_t - \sqrt{1 - \balpha_t}\sqrt{\mSigma}\rvepsilon^*\big)\right) + \frac{\sqrt{1 - \beta_t}(1 - \balpha_{t-1})}{1 - \balpha_t}\rvx_t, \\
&\quad = \frac{\sqrt{\balpha_{t-1}}}{\sqrt{\balpha_t}} \frac{\beta_t}{1 - \balpha_t}\rvx_t - \frac{\sqrt{\balpha_{t-1}}}{\sqrt{\balpha_t}} \frac{\beta_t}{\sqrt{1 - \balpha_t}} \sqrt{\mSigma}\rvepsilon^* + \frac{\sqrt{1 - \beta_t}}{1 - \balpha_t}\bigg(1 - \balpha_{t-1}\bigg)\rvx_t, \\
&\quad = \frac{1}{\sqrt{1 - \beta_t}} \frac{\beta_t}{1 - \balpha_t}\rvx_t + \frac{\sqrt{1 - \beta_t}}{1 - \balpha_t}\left(1 - \frac{\balpha_t}{1 - \beta_t}\right)\rvx_t - \frac{1}{\sqrt{1 - \beta_t}} \frac{\beta_t}{\sqrt{1 - \balpha_t}} \sqrt{\mSigma}\rvepsilon^*, \\
&\quad = \frac{1}{\sqrt{1 - \beta_t}}\left( \frac{\beta_t}{1 - \balpha_t}\rvx_t + \frac{1 - \beta_t}{1 - \balpha_t}\left(1 - \frac{\balpha_t}{1 - \beta_t}\right)\rvx_t - \frac{\beta_t}{\sqrt{1 - \balpha_t}} \sqrt{\mSigma}\rvepsilon^* \right), \\
&\quad = \frac{1}{\sqrt{1 - \beta_t}}\left( \frac{\cancel{\beta_t} + \bcancel{1 \cancel{- \beta_t} - \balpha_t}}{\bcancel{1 - \balpha_t}}\rvx_t - \frac{\beta_t}{\sqrt{1 - \balpha_t}} \sqrt{\mSigma}\rvepsilon^* \right), \\
&\implies \tilde\vmu_{t-1}(\rvx_t, \hat\rvx_0) = \frac{1}{\sqrt{1 - \beta_t}}\left( \rvx_t - \frac{\beta_t}{\sqrt{1 - \balpha_t}} \sqrt{\mSigma}\rvepsilon_{\rvtheta^*}(\rvx_t) \right).
\numberthis \\
\numberthis 
\\
\implies \rvx_{t-1}
&= \frac{1}{\sqrt{1 - \beta_t}}\left(\rvx_t - \frac{\beta_t}{\sqrt{1 - \balpha_t}}\sqrt{\mSigma}\ \rvepsilon_{\rvtheta^*}(\rvx_t)\right) + \sqrt{\tilde\beta_{t-1}} \sqrt{\mSigma}\ \rvz,
\numberthis
\\
&= \frac{1}{\sqrt{1 - \beta_t}}\big(\rvx_t + \beta_t \mSigma\ \rvs_{\rvtheta^*}(\rvx_t, \balpha_t)\big) + \sqrt{\tilde\beta_{t-1}} \sqrt{\mSigma}\ \rvz.
\numberthis
\end{align*}

\subsection{Sampling in NI-DDPM}
\label{subsec:sampling_niddpm}

We perform sampling at each time step in 2 parts:
\begin{align*}
&\text{\textit{Step 1 (from \cref{eq:NIDDPM_reverse}):  }} \hat\rvx_0 = \frac{1}{\sqrt{\balpha_t}}(\rvx_t - \sqrt{1 - \balpha_t} \sqrt{\mSigma} \rvepsilon_{\rvtheta^*}(\rvx_t, \balpha_t)).
\numberthis
\\
&\text{\textit{Step 2 (from \cref{eq:NIDDPM_step2}):  }} \rvx_{t-1} = \frac{\sqrt{\balpha_{t-1}}\beta_t}{1 - \balpha_t}\hat\rvx_0 + \frac{\sqrt{1 - \beta_t}(1 - \balpha_{t-1})}{1 - \balpha_t}\rvx_t + \sqrt{\tilde\beta_{t-1}} \sqrt{\mSigma} \rvz.
\numberthis
\end{align*}

\subsection{Sampling in NI-DDPM using DDIM}

DDIM replaces \textit{Step 2} with the NI-DDPM forward process \cref{eq:GFF_DDPM_noise}:
\begin{align*}
\text{\textit{Step 1: }} \hat\rvx_0 &= \frac{1}{\sqrt{\balpha_t}}(\rvx_t - \sqrt{1 - \balpha_t}\sqrt{\mSigma}\rvepsilon_{\rvtheta^*}(\rvx_t)).
\numberthis\\
\text{\textit{Step 2: }} \rvx_{t-1} &= \sqrt{\balpha_{t-1}} \hat\rvx_0 + \sqrt{1 - \balpha_{t-1}} \sqrt{\mSigma} \rvepsilon_{\rvtheta^*}(\rvx_t).
\numberthis
\label{eq:NIDDPM_DDIM_step2}
\end{align*}

This is derived from the following distributions from \cite{song2020ddim}:
\begin{align*}
&p_{L}^{\GFFDDPM}(\rvx_L \mid \rvx_0) = \gN(\rvx_L \mid \sqrt{\balpha_L}\rvx_0, (1 - \balpha_L)\mSigma).
\numberthis
\\
&q_t^{\GFFDDIM}(\rvx_{t-1} \mid \rvx_t, \rvx_0) = \gN\left( \rvx_{t-1} \mid \sqrt{\balpha_{t-1}}\rvx_0 + \sqrt{1 - \balpha_{t-1}}\frac{\rvx_t - \sqrt{\balpha_t}\rvx_0}{\sqrt{1 - \balpha_t}}, \vzero \right).
\numberthis
\\
\implies &p_{t}^{\GFFDDIM}(\rvx_{t} \mid \rvx_0) = \gN\left( \rvx_{t} \mid \sqrt{\balpha_{t}}\rvx_0, (1 - \balpha_{t})\mSigma \right).
\numberthis
\end{align*}

\subsection{Expected Denoised Sample (EDS) for NI-DDPM}

From \cite{saremi2019neb}, we know that the expected denoised sample $\rvx_0^*(\rvx_t, \balpha_t) \triangleq \EE_{\rvx_0 \sim q_{t}(\rvx_0 \mid \rvx_t)}[\rvx_0]$ and the optimal score $\rvs_{\rvtheta^*}(\rvx_t, \balpha_t)$ are related as (as mentioned earlier in \cref{eq:EDS_defn}):
\begin{align}
&\rvs_{\rvtheta^*}(\rvx_t, \balpha_t) = \EE\left[ \Norm{ \nabla_{\rvx_t} \log p_{t} (\rvx_t \mid \rvx_0) }_2^2 \right] (\rvx_0^* (\rvx_t, \balpha_t) - \rvx_t).
\end{align}
For NI-DDPM with non-isotropic Gaussian noise of covariance $(1 - \balpha_t)\mSigma$, 
\begin{align}
&\rvs_{\rvtheta^*}(\rvx_t, \balpha_t) = \frac{1}{1 - \balpha_t}\mSigma^{-1}(\rvx_0^*(\rvx_t, \balpha_t) - \rvx_t).
\\
\implies &\rvx_0^{*}(\rvx_t, \balpha_t) = \rvx_t + (1 - \balpha_t) \mSigma\ \rvs_{\rvtheta^*}(\rvx_t, \balpha_t) = \rvx_t - \sqrt{1 - \balpha_t} \sqrt{\mSigma}\ \rvepsilon_{\rvtheta^*}(\rvx_t).
\end{align}

\subsection{SDE formulation : Non-Isotropic Variance Preserving (NIVP) SDE}
\label{subsec:niddpm_sde}

For NI-DDPM i.e. Non-Isotropic Variance Preserving (NIVP) SDE, the forward equation and transition probability are derived (below) as:

\begin{align}
\D \rvx &= -\frac{1}{2}\beta(t)\rvx\ \D t + \sqrt{\beta(t)}\sqrt{\mSigma}\ \D \rvw. \\
p_{0t}^\NIVP(\rvx(t) \mid \rvx(0)) &= \gN \left(\rvx(t) \mid \rvx(0)\ e^{-\frac{1}{2}\int_0^t \beta(s) \D s}, \mSigma(\rmI - \rmI e^{-\int_0^t \beta(s) \D s}) \right).
\end{align}

\subsubsection{Derivations}:

\textbf{Forward process}: We know from \cref{eq:NIDDPM_forward} that:
\begin{align*}
\rvx_t &= \sqrt{1 - \beta_t}\rvx_{t-1} + \sqrt{\beta_t} \sqrt{\mSigma} \rvepsilon_{t-1} . \\
\implies \rvx(t + \Delta t) &= \sqrt{1 - \beta(t + \Delta t) \Delta t}\ \rvx(t) + \sqrt{\beta(t + \Delta t) \Delta t}\ \sqrt{\mSigma}\rvepsilon(t), \\
&\approx \left(1 - \frac{1}{2}\beta(t + \Delta t)\Delta t\right) \rvx(t) + \sqrt{\beta(t + \Delta t) \Delta t}\ \sqrt{\mSigma}\rvepsilon(t), \\
&\approx \rvx(t) - \frac{1}{2}\beta(t)\Delta t\ \rvx(t) + \sqrt{\beta(t) \Delta t}\ \sqrt{\mSigma}\rvepsilon(t). \\
\implies \D \rvx &= -\frac{1}{2}\beta(t)\rvx\ \D t + \sqrt{\beta(t)}\sqrt{\mSigma}\ \D \rvw.
\numberthis
\end{align*}


\textbf{Mean} (from eq. 5.50 in Sarkka \& Solin (2019)):
\begin{align*}
&\D \rvx = \rvf\ \D t + \rmG\ \D \rvw \implies \frac{\D \vmu}{\D t} = \EE_\rvx[\rvf] .\\
&\therefore \frac{\D \vmu_{\GFFDDPM}(t)}{\D t} = \EE_\rvx[-\frac{1}{2}\beta(t)\rvx] = -\frac{1}{2}\beta(t)\EE_\rvx(\rvx) = -\frac{1}{2}\beta(t)\vmu_{\GFFDDPM}(t) , \\
&\implies \frac{\D \vmu_{\GFFDDPM}(t)}{\vmu_{\GFFDDPM}(t)} = -\frac{1}{2}\beta(t) \D t \implies \log \vmu_{\GFFDDPM}(t) \vert_{0}^{t} = -\frac{1}{2}\int_0^t \beta(s) \D s , \\
&\implies \log \vmu_{\GFFDDPM}(t) - \log \vmu(0) = -\frac{1}{2}\int_0^t \beta(s) \D s 
\implies \log \frac{\vmu_{\GFFDDPM}(t)}{\vmu(0)} = -\frac{1}{2}\int_0^t \beta(s) \D s , \\
&\implies \vmu_{\GFFDDPM}(t) = \vmu(0)\ e^{-\frac{1}{2}\int_0^t \beta(s) \D s}. \\
\end{align*}

\textbf{Covariance} (from eq. 5.51 in Sarkka \& Solin (2019)):
\begin{align*}
&\D \rvx = \rvf\ \D t + \rmG\ \D \rvw \implies \frac{\D \mSigma_{\text{cov}}}{\D t} = \EE_\rvx[\rvf (\rvx - \vmu)^\trns] + \EE_\rvx[(\rvx - \vmu) \rvf^\trns] + \EE_\rvx[\rmG\rmG^\trns]. \\
&\therefore \frac{\D \mSigma_{\GFFDDPM}(t)}{\D t} = \EE_\rvx[-\frac{1}{2}\beta(t)\rvx\rvx^\trns] + \EE_\rvx[\rvx(-\frac{1}{2}\beta(t)\rvx)^\trns] + \EE_\rvx[\sqrt{\beta(t)}\sqrt{\mSigma}\sqrt{\beta(t)}\sqrt{\mSigma}] , \\
&\qquad\qquad\qquad = -\beta(t)\mSigma_{\GFFDDPM}(t) + \beta(t)\mSigma = \beta(t)(\mSigma - \mSigma_{\GFFDDPM}(t)), \\
&\implies \frac{\D \mSigma_{\GFFDDPM}(t)}{\mSigma - \mSigma_{\GFFDDPM}(t)} = \beta(t) \D t \implies -\log(\mSigma - \mSigma_{\GFFDDPM}(t)) \vert_{0}^{t} = \int_0^t \beta(s) \D s, \\
&\implies -\log(\mSigma - \mSigma_{\GFFDDPM}(t)) + \log(\mSigma - \mSigma_{\rvx}(0)) = \int_0^t \beta(s) \D s, \\
&\implies \frac{\mSigma - \mSigma_{\GFFDDPM}(t)}{\mSigma - \mSigma_{\rvx}(0)} = e^{-\int_0^t \beta(s) \D s} \implies \mSigma_{\GFFDDPM}(t) = \mSigma - e^{-\int_0^t \beta(s) \D s} (\mSigma - \mSigma_{\rvx}(0)), \\
&\implies \mSigma_{\GFFDDPM}(t) = \mSigma + e^{-\int_0^t \beta(s) \D s} (\mSigma_{\rvx}(0) - \mSigma).
\end{align*}

For each data point $\rvx(0)$, $\vmu(0) = \rvx(0)$, $\mSigma_{\rvx}(0) = \vzero$:
\begin{align*}
&\implies \vmu_{\GFFDDPM}(t) = \rvx(0)\ e^{-\frac{1}{2}\int_0^t \beta(s) \D s}, \\
&\qquad \quad \mSigma_{\GFFDDPM}(t) = \mSigma + e^{-\int_0^t \beta(s) \D s} (\vzero - \mSigma) = \mSigma(\rmI - \rmI e^{-\int_0^t \beta(s) \D s}). \\
&\therefore \GFFDDPM \text{ i.e. } p_{0t}^{\NIVP}(\rvx(t) \mid \rvx(0)) = \gN \left(\rvx(t) \mid \rvx(0)\ e^{-\frac{1}{2}\int_0^t \beta(s) \D s}, \mSigma(\rmI - \rmI e^{-\int_0^t \beta(s) \D s}) \right).
\end{align*}

\newpage
\section{Non-isotropic SMLD (NI-SMLD)}
\label{sec:nismld}

\subsection{Forward (data to noise) for NI-SMLD}

In NI-SMLD, for a fixed sequence of positive scales $0 < \sigma_1 < \cdots \sigma_L < 1$, and a noise sample $\rvepsilon_t \sim \gN(\vzero, \rmI)$ from a standard normal distribution, and a clean data point $\rvx_0$, the \textbf{cumulative} ``\textbf{forward}'' process is:
\begin{align}
&q_{\sigma_i}^{\GFFSMLD} (\rvx_i \mid \rvx_0) = \gN(\rvx_i \mid \rvx_0, \sigma_i^2 \mSigma). \\
\implies &\rvx_i = \rvx_0 + \sigma_i \sqrt{\mSigma} \rvepsilon \implies \rvepsilon = \sqrt{\mSigma^{-1}}\frac{\rvx_i - \rvx_0}{\sigma_i}.
\end{align}
The \textbf{transition} ``\textbf{forward}'' process can be derived as:
\begin{align}
&q_{\sigma_i}^{\GFFSMLD} (\rvx_{i+1} \mid \rvx_i) = \gN(\rvx_{i+1} \mid \rvx_i, (\sigma_{i+1}^2 - \sigma_i^2)\mSigma). \\
\implies &\rvx_i = \rvx_{i-1} + \sqrt{\sigma_i^2 - \sigma_{i-1}^2}\sqrt{\mSigma}\rvepsilon_{i-1}.
\label{eq:NISMLD_forward}
\end{align}

\subsection{Score for NI-SMLD}

\begin{align}
&q_{\sigma_i}^{\GFFSMLD} (\rvx_i \mid \rvx_0) = \gN(\rvx_i \mid \rvx_0, \sigma_i^2 \mSigma). \\
\implies &\nabla_{\rvx_i} \log q_{\sigma_i}^{\GFFSMLD} (\rvx_i \mid \rvx_0) = -\mSigma^{-1}\frac{\rvx_i - \rvx_0}{\sigma_i^2} = -\sqrt{\mSigma^{-1}}\frac{\rvepsilon}{\sigma_i}.
\end{align}

\subsection{Score-matching objective function for NI-SMLD}

The objective function for SMLD at noise level $\sigma$ is:
\begin{align}
\loss^{\GFFSMLD}(\rvtheta; \sigma_i) &\triangleq\ \frac{1}{2} \EE_{q_{\sigma_i}^{\GFFSMLD}(\rvx_i \mid \rvx_0) p(\rvx_0)} \bigg[ \Norm{\rvs_\rvtheta (\rvx_i, \sigma_i) + \mSigma^{-1}\frac{\rvx_i - \rvx_0}{\sigma_i^2} }_2^2 \bigg] ,\\
&=\ \frac{1}{2} \EE_{q_{\sigma_i}^{\GFFSMLD}(\rvx_i \mid \rvx_0) p(\rvx_0)} \bigg[ \Norm{\rvs_\rvtheta (\rvx_i, \sigma_i) + \frac{1}{\sigma_i}\sqrt{\mSigma^{-1}}\rvepsilon }_2^2 \bigg].
\end{align}

\subsection{Variance of score for NI-SMLD}
\begin{align*}
&\EE\left[ \Norm{ \nabla_{\rvx_i} \log q_{\sigma_i}^{\GFFSMLD} (\rvx_i \mid \rvx_0) }_2^2 \right] = \EE\left[ \Norm{ -\mSigma^{-1}\frac{\rvx_i - \rvx_0}{\sigma_i^2} }_2^2 \right], \\
&= \EE\left[ \Norm{ \mSigma^{-1} \frac{\sigma_i\sqrt{\mSigma}\rvepsilon}{\sigma_i^2} }_2^2 \right] = \frac{1}{\sigma_i^2}\mSigma^{-1}\EE\left[ \Norm{\rvepsilon}_2^2 \right] = \frac{1}{\sigma_i^2} \mSigma^{-1}.
\numberthis
\end{align*}

\subsection{Overall objective function for NI-SMLD}

The overall objective function weights the score-matching objective $\loss^{\GFFSMLD}(\rvtheta; \sigma_i)$ by the inverse of the variance of the score at each time step $\implies \lambda(\sigma_i) = \sigma_i^2\mSigma$:
\begin{align*}
\gL^{\GFFSMLD}(\rvtheta; \{\balpha_t\}_{t=1}^{L}) &\triangleq \EE_t\ \lambda(\balpha_t) \ \loss^{\GFFSMLD}(\rvtheta; \balpha_t), \\
&= \frac{1}{2L} \sum_{i=1}^{L} \EE_{q_{\sigma_i}^{\GFFSMLD}(\rvx_i \mid \rvx_0) p(\rvx_0)} \bigg[ \Norm{\sigma_i \sqrt{\mSigma} \rvs_\rvtheta (\rvx_i, \sigma_i) + \sqrt{\mSigma^{-1}} \frac{(\rvx_i - \rvx_0)}{\sigma_i} }_2^2 \bigg], \\
&= \frac{1}{2L} \sum_{i=1}^{L} \EE_{q_{\sigma_i}^{\GFFSMLD}(\rvx_i \mid \rvx_0) p(\rvx_0)} \bigg[ \Norm{\sigma_i \sqrt{\mSigma} \rvs_\rvtheta (\rvx_i, \sigma_i) + \rvepsilon }_2^2 \bigg].
\numberthis
\end{align*}

\subsection{Unconditional NI-SMLD score estimation}

An unconditional score model is:
\begin{align}
\rvs_\rvtheta(\rvx_i, \sigma_i) = -\sqrt{\mSigma^{-1}} \frac{1}{\sigma_i} \rvepsilon_\rvtheta(\rvx_i).
\end{align}

In this case, the overall objective function changes to:
\begin{align*}
\gL^{\GFFSMLD}(\rvtheta; \{\sigma_i\}_{i=1}^{L}) 
&\triangleq \frac{1}{2L} \sum_{i=1}^{L} \EE_{q_{\sigma_i}^{\GFFSMLD}(\rvx_i \mid \rvx_0) p(\rvx_0)} \bigg[ \Norm{ \rvepsilon - \rvepsilon_\rvtheta (\rvx_i) }_2^2 \bigg], \\
&= \frac{1}{2L} \sum_{i=1}^{L} \EE_{q_{\sigma_i}^{\GFFSMLD}(\rvx_i \mid \rvx_0) p(\rvx_0)} \bigg[ \Norm{ \rvepsilon - \rvepsilon_\rvtheta (\rvx_0 + \sigma_i\sqrt{\mSigma}\rvepsilon) }_2^2 \bigg].
\numberthis
\end{align*}

\subsection{Sampling i.e. Reverse (noise to data) for NI-SMLD}

Similar to SMLD, NI-SMLD does not explicitly define a reverse process. Instead, similar to \citet{song2019generative, song2020improved,jolicoeur2021gotta}, we derive Annealed Langevin Sampling below to transform from noise to data. 

\underline{Forward} : $\rvx_i = \rvx_{i-1} + \sqrt{\sigma_i^2 - \sigma_{i-1}^2}\sqrt{\mSigma}\rvepsilon_{i-1}$.

\underline{Reverse}:
Annealed Langevin Sampling for NI-SMLD:
\begin{align*}
&\rvx_L^0 \sim \gN(\vzero, \sigma_{\max} \sqrt{\mSigma}).\\
&\begin{rcases}
&\rvx_i^0 = \rvx_{i+1}^M,
\\
&\rvx_i^{m+1} \leftarrow \rvx_i^m + \alpha_i \rvs_{\rvtheta^*}(\rvx_i^m, \sigma_i) + \sqrt{2\alpha_i} \sqrt{\mSigma} \rvepsilon_i^{m+1} , m=1,\cdots,M.
\end{rcases}
i = L, \cdots, 1
\numberthis
\\
&\alpha_i = \eps \sigma_i^2/\sigma_L^2.
\end{align*}

Consistent Annealed Sampling~\citep{jolicoeur2020adversarial} for NI-SMLD:
\begin{align*}
&\rvx_{i-1} \leftarrow \rvx_i + \alpha_i \rvs_{\rvtheta^*}(\rvx_i, \sigma_i) + \beta \sigma_{i-1} \sqrt{\mSigma} \rvepsilon_{i-1}, i = L, \cdots, 1.
\numberthis
\\
&\alpha_i = \eps \sigma_t^2/\sigma_{\min}^2;\ \beta = \sqrt{1 - \gamma^2(1 - \eps/\sigma_{\min}^2)^2};\ \gamma = \sigma_t/\sigma_{t-1} ; \sigma_t > \sigma_{t-1}.
\end{align*}

\subsection{Expected Denoised Sample (EDS) for NI-SMLD}

From \cite{saremi2019neb}, assuming non-isotropic Gaussian noise of covariance $\sigma^2\mSigma$, we know that the EDS $\rvx_0^*(\rvx_t, \sigma) \triangleq \EE_{\rvx_0 \sim q_\sigma(\rvx_0 \mid \rvx_t)}[\rvx]$ and optimal score $\rvs_{\rvtheta^*}(\rvx_I, \sigma)$ are related as:
\begin{align*}
&\rvs_{\rvtheta^*}(\rvx_i, \sigma) = \frac{1}{\sigma^2}\mSigma^{-1}(\rvx_0^*(\rvx_i, \sigma) - \rvx_t). \\
\implies &\rvx_0^*(\rvx_i, \sigma) = \rvx_i + \sigma^2 \mSigma\ \rvs_{\rvtheta^*}(\rvx_i, \sigma) = \rvx_i - \sigma \sqrt{\mSigma}\ \rvepsilon_{\rvtheta^*}(\rvx_i).
\numberthis
\end{align*}

\subsection{SDE formulation : Non-Isotropic Variance Exploding (NIVE) SDE}

For NI-SMLD i.e. Non-Isotropic Variance Exploding (NIVE) SDE, the forward equation and transition probability are derived (below) as:
\begin{align}
    \D \rvx &= \sqrt{\frac{\D [\sigma^2(t)]}{\D t}}\ \sqrt{\mSigma}\ \D \rvw \label{eq:NIVESDE_d}. \\
    p_{0t}^{\NIVE}(\rvx(t) \mid \rvx(0)) &= \gN \left(\rvx(t) \mid \rvx(0), \sigma^2(t) \mSigma \right) \label{eq:NIVESDE_p}.
\end{align}

\subsubsection{Derivations}:

\textbf{Forward process}: We know from \cref{eq:NISMLD_forward} that:
\begin{align*}
\rvx_i &= \rvx_{i-1} + \sqrt{\sigma_i^2 - \sigma_{i-1}^2} \sqrt{\mSigma} \rvepsilon_{i-1}. \\
\implies \rvx(t + \Delta t) &= \rvx(t) + \sqrt{(\sigma^2(t + \Delta t) - \sigma^2(t)) \Delta t}\ \sqrt{\mSigma}\rvepsilon(t), \\
&\approx \rvx(t) + \sqrt{\frac{\D [\sigma^2(t)]}{\D t} \Delta t} \sqrt{\mSigma} \rvw(t). \\
\implies \D \rvx &= \sqrt{\frac{\D [\sigma^2(t)]}{\D t}}\ \sqrt{\mSigma} \ \D \rvw.
\numberthis
\end{align*}
Thus, \cref{eq:NIVESDE_d} is derived.

\vspace{1.5em}

\textbf{Mean} and \textbf{Covariance} (from eq. 5.50 and eq.5.51 in \cite{sarkka2019applied}) for a random variable $\rvx$ that changes according to a stochastic process with drift and diffusion coefficients $\rvf$ and $\rmG$, change as:
\begin{align}
\D \rvx = \rvf\ \D t + \rmG\ \D \rvw 
\implies 
&\frac{\D \vmu}{\D t} = \EE_\rvx[\rvf], \\
&\frac{\D \mSigma_{\text{cov}}}{\D t} = \EE_\rvx[\rvf (\rvx - \vmu)^\trns] + \EE_\rvx[(\rvx - \vmu) \rvf^\trns] + \EE_\rvx[\rmG\rmG^\trns].
\numberthis
\end{align}

For NI-SMLD, $\rvf = \vzero, \vmu = \vzero, \rmG = \sqrt{\frac{\D [\sigma^2(t)]}{\D t}} \sqrt{\mSigma}$.
\begin{align}
&\frac{\D \vmu_{\GFFSMLD}(t)}{\D t} = \EE_\rvx[\vzero] = \vzero, \\
\implies &\vmu_{\GFFSMLD}(t) = \vmu(0) = \rvx(0) . \\
&\frac{\D \mSigma_{\GFFSMLD}(t)}{\D t} = \EE_\rvx\left[\vzero + \vzero + \sqrt{\frac{\D [\sigma^2(t)]}{\D t}}\sqrt{\mSigma} \sqrt{\frac{\D [\sigma^2(t)]}{\D t}} \sqrt{\mSigma} \right] = \frac{\D [\sigma^2(t)]}{\D t}\mSigma, \\
\implies &\mSigma_{\GFFSMLD}(t) = \sigma^2(t) \mSigma.
\end{align}

$\therefore \text{NI-SMLD } \text{ i.e. } p_{0t}^{\NIVE}(\rvx(t) \mid \rvx(0)) = \gN \left(\rvx(t) \mid \rvx(0), \sigma^2(t) \mSigma \right)$.
Thus, \cref{eq:NIVESDE_p} is derived.

\subsection{Initial noise scale for NI-SMLD}

Building on Proposition 1 in \citet{song2020improved}, let $\hat{p}_{\sigma_{1}}(\rvx) \triangleq \frac{1}{N} \sum_{i=1}^N p^{(i)}(\rvx)$, where $p^{(i)}(\rvx) \triangleq \gN(\rvx \mid \rvx^{(i)}, \sigma_1^2 \mSigma)$. With $r^{(i)}(\rvx) \triangleq \frac{p^{(i)}(\rvx)}{\sum_{k=1}^N  p^{(k)}(\rvx) }$, the score function is $\nabla_\rvx \log \hat{p}_{\sigma_{1}}(\rvx) = \sum_{i=1}^N r^{(i)}(\rvx) \nabla_\rvx \log p^{(i)}(\rvx)$.

We know that:
\begin{align*}
\gN(\rvx \mid \rvx^{(i)}, \sigma_1^2 \mSigma) = \frac{1}{(2\pi)^{D/2} \sigma_1^D \abs{\mSigma}^{1/2}} \exp \left( -\frac{1}{2\sigma_1^2} (\rvx - \rvx^{(i)})^\trns \mSigma^{-1} (\rvx - \rvx^{(i)}) \right).
\end{align*}

\begin{align*}
    &\EE_{p^{(i)}(\rvx)}[r^{(j)}(\rvx)] = \int \frac{p^{(i)}(\rvx)p^{(j)}(\rvx)}{\sum_{k=1}^N p^{(k)}(\rvx)} \ud \rvx \leq \int \frac{p^{(i)}(\rvx)p^{(j)}(\rvx)}{p^{(i)}(\rvx) + p^{(j)}(\rvx)} \ud \rvx, \\
    &= \frac{1}{2} \int \frac{2}{\frac{1}{p^{(i)}(\rvx)} + \frac{1}{p^{(j)}(\rvx)}} \ud \rvx \leq \frac{1}{2} \int \sqrt{p^{(i)}(\rvx) p^{(j)}(\rvx)} \ud \rvx, \\
    &= \frac{1}{2} \frac{1}{(2\pi)^{D/2}\sigma_1^D \abs{\mSigma}^{1/2}} \int \exp \bigg(-\frac{1}{4\sigma_1^2}\bigg( (\rvx - \rvx^{(i)})^\trns \mSigma^{-1} (\rvx - \rvx^{(i)}) + (\rvx - \rvx^{(j)})^\trns \mSigma^{-1} (\rvx - \rvx^{(j)}) \bigg) \ud \rvx, \\
    &= \frac{1}{2} \frac{1}{(2\pi)^{D/2}\sigma_1^D \abs{\mSigma}^{1/2}} \int \exp \bigg(-\frac{1}{4\sigma_1^2}\bigg( (\rvx - \rvx^{(i)})^\trns \mSigma^{-1} (\rvx - \rvx^{(i)}) + (\rvx - \rvx^{(j)})^\trns \mSigma^{-1} (\rvx - \rvx^{(i)}), \\
    &\qquad \qquad \qquad \qquad \qquad \qquad \qquad + (\rvx - \rvx^{(j)})^\trns \mSigma^{-1} (\rvx^{(i)} - \rvx^{(j)}) \bigg) \ud \rvx, \\
    &= \frac{1}{2} \frac{1}{(2\pi)^{D/2}\sigma_1^D \abs{\mSigma}^{1/2}} \int \exp \bigg(-\frac{1}{4\sigma_1^2}\bigg( (\rvx - \rvx^{(i)})^\trns \mSigma^{-1} (\rvx - \rvx^{(i)}) + (\rvx - \rvx^{(i)})^\trns \mSigma^{-1} (\rvx - \rvx^{(i)}), \\
    &\qquad \qquad \qquad \qquad \qquad \qquad \qquad + (\rvx^{(i)} - \rvx^{(j)})^\trns \mSigma^{-1} (\rvx - \rvx^{(i)}) + (\rvx - \rvx^{(i)})^\trns \mSigma^{-1} (\rvx^{(i)} - \rvx^{(j)}), \\
    &\qquad \qquad \qquad \qquad \qquad \qquad \qquad + (\rvx^{(i)} - \rvx^{(j)})^\trns \mSigma^{-1} (\rvx^{(i)} - \rvx^{(j)}) \bigg) \ud \rvx, \\
    &= \frac{1}{2} \frac{1}{(2\pi)^{D/2}\sigma_1^D \abs{\mSigma}^{1/2}} \int \exp \bigg(-\frac{1}{2\sigma_1^2}\bigg( (\rvx - \rvx^{(i)})^\trns \mSigma^{-1} (\rvx - \rvx^{(i)}) + 2(\rvx - \rvx^{(i)})^\trns \mSigma^{-1} \frac{(\rvx^{(i)} - \rvx^{(j)})}{2}, \\
    &\qquad \qquad \qquad \qquad \qquad \qquad \qquad + \frac{(\rvx^{(i)} - \rvx^{(j)})}{2}^\trns \mSigma^{-1} \frac{(\rvx^{(i)} - \rvx^{(j)})}{2} - \frac{(\rvx^{(i)} - \rvx^{(j)})}{2}^\trns \mSigma^{-1} \frac{(\rvx^{(i)} - \rvx^{(j)})}{2}, \\
    &\qquad \qquad \qquad \qquad \qquad \qquad \qquad + \frac{1}{2}(\rvx^{(i)} - \rvx^{(j)})^\trns \mSigma^{-1} (\rvx^{(i)} - \rvx^{(j)}) \bigg) \ud \rvx, \\
    &= \frac{1}{2} \exp \bigg( -\frac{1}{8\sigma_1^2} (\rvx^{(i)} - \rvx^{(j)})^\trns \mSigma^{-1} (\rvx^{(i)} - \rvx^{(j)}) \bigg).
\end{align*}
It is desirable to choose $\sigma_1$ that is proportional to the numerator term so that $\EE_{p^{(i)}(\rvx)}[r^{(j)}(\rvx)]$ has a reasonably large value. To make this happen, from \citet{song2020improved}:
\begin{align*}
    &\implies \frac{1}{\sigma_1^2} (\rvx^{(i)} - \rvx^{(j)})^\trns \mSigma^{-1} (\rvx^{(i)} - \rvx^{(j)}) \approx 1, \\
    &\implies (\sqrt{\mSigma^{-1}} (\rvx^{(i)} - \rvx^{(j)}))^\trns (\sqrt{\mSigma^{-1}} (\rvx^{(i)} - \rvx^{(j)})) \approx \sigma_1^2, \\
    &\implies \Norm{ \sqrt{\mSigma^{-1}} (\rvx^{(i)} - \rvx^{(j)}) }_2 \approx \sigma_1, \\
    &\implies \Norm{ \sigma_N\ \Real \big( \rmW_N^{-1} \rmK \rmW_N (\rvx^{(i)} - \rvx^{(j)}) \big) }_2 \approx \sigma_1, \\
    &\implies \Norm{\sigma_N \rmW_N^{-1} \rmK \rmW_N \rvx^{(i)} - \sigma_N \rmW_N^{-1} \rmK \rmW_N \rvx^{(j)} }_2 \approx \sigma_{1}.
\end{align*}

For CIFAR10, this $\sigma_{1} \approx 20$ for NI-SMLD (whereas for SMLD $\sigma_1 \approx 50$).

\subsection{Other noise scales}

Building on Proposition 2 in \citet{song2020improved}, let image $\rvx \in \mathbb{R}^{D} \sim \gN (\vzero, \sigma^2 \rmI)$, and $r = ||\rvx||_2$. For simplification of analysis, because image dimensions $D$ are typically quite large, we can assume:
\begin{align*}
p_{\sigma_i}(r) &= \gN(r \mid m_i, s_i^2), \text{where } m_i \triangleq \sqrt{D}\sigma_i ; s_i^2 \triangleq \sigma_i^2 / 2.
\end{align*}
Using the three-sigma rule of thumb~\citep{grafarend2006linear}, $p_{\sigma_i}(r)$ has high density in:
\begin{align*}
\gI_i &\triangleq (m_i - 3s_i, m_i + 3s_i).
\end{align*}
Given the discrete nature of $\sigma_i$s, we need the radial components of $p_{\sigma_i}(\rvx)$ and $p_{\sigma_{i-1}}(\rvx)$ to have large overlap. This naturally leads us to fix $p_{\sigma_i}(r \in \gI_{i-1})$ to be a moderately high constant, \citet{song2020improved} chose $0.5$. Given $\Phi(.)$ is the Cumulative Density Function (CDF) of standard Gaussian, and $\gamma \triangleq \sigma_{i-1}/\sigma_{i}$ considering a geometric progression of $\sigma_i$s:
\begin{align*}
p_{\sigma_i}(r \in \gI_{i-1})
&= \Phi\left( \frac{(m_{i-1} + 3s_{i-1}) - m_i}{s_i} \right) - \Phi\left( \frac{(m_{i-1} - 3s_{i-1}) - m_i}{s_i} \right), \\
&= \Phi\left(\frac{\sqrt{2}}{\sigma_i} (\sqrt{D}\sigma_{i-1} + \frac{3\sigma_{i-1}}{\sqrt{2}} - \sqrt{D}\sigma_i) \right) - \Phi\left(\frac{\sqrt{2}}{\sigma_i} (\sqrt{D}\sigma_{i-1} - \frac{3\sigma_{i-1}}{\sqrt{2}} - \sqrt{D}\sigma_i) \right), \\
&= \Phi\left(\frac{1}{\sigma_i} (\sqrt{2D}(\sigma_{i-1} - \sigma_i) + 3\sigma_{i-1}) \right) - \Phi\left(\frac{1}{\sigma_i} (\sqrt{2D}(\sigma_{i-1} - \sigma_i) - 3\sigma_{i-1}) \right), \\
&= \Phi\left( \sqrt{2D}(\gamma - 1) + 3\gamma \right) - \Phi \left( \sqrt{2D}(\gamma - 1) - 3\gamma \right) \approx 0.5.
\end{align*}

From the previous discussion, given $\sigma_1=20$, and setting $\sigma_L=0.01$, $L=207$, $\gamma = 1.0376$ (whereas for SMLD $\sigma_1=50, L=232$).

\subsection{Configuring annealed Langevin dynamics}

Building on Proposition 3 in \citet{song2020improved}, $\gamma = \frac{\sigma_{i-1}}{\sigma_i}$. For $\alpha = \epsilon\cdot  \frac{\sigma_i^2}{\sigma_L^2}$, we have $\rvx_T \sim \gN(\vzero, \Var[\rvx_T])$, where
\begin{align}
    \frac{\Var[\rvx_T]}{\sigma_i^2} &= \gamma^2 \rmP^T \mSigma \rmP^T + \frac{2\epsilon}{\sigma_L^2} \sum_{t=0}^{T-1} ( \rmP^t \mSigma \rmP^t),
\end{align}
where $\rmP = \rmI - \frac{\alpha}{\sigma_i^2}\mSigma^{-1} = \rmI - \frac{\epsilon}{\sigma_L^2}\mSigma^{-1}$ (proof below).

Hence, we choose $\epsilon$ s.t. $\frac{Var[\rvx_T]}{\sigma_i^2} \approx 1$,
$\implies \epsilon = 3.1\mathrm{e}{-7}$ for $T=5$, $\epsilon = 2.0\mathrm{e}{-6}$ for $T=1$ (whereas for SMLD $\epsilon = 6.2\mathrm{e}{-6}$ for $T=5$)

\textbf{Proof:}

First, the conditions we know are:
\begin{gather*}
\rvx_0 \sim p_{\sigma_{i-1}}(\rvx) = \gN(\rvx \mid \vzero, \sigma_{i-1}^2 \mSigma) = \frac{1}{(2\pi)^{D/2}\sigma_{i-1}^D\abs{\mSigma}^{1/2}} \exp \left( -\frac{1}{2\sigma_{i-1}^2} \rvx^\trns \mSigma^{-1} \rvx \right),\\
\nabla_\rvx \log p_{\sigma_i}(\rvx_t) = \nabla_\rvx  \left(-\log(\text{const.}) - \frac{1}{2\sigma_i^2} \rvx_t^\trns \mSigma^{-1} \rvx_t \right) = -\frac{1}{\sigma_i^2} \mSigma^{-1} \rvx_t, \\
\rvx_{t+1} \gets \rvx_{t} + \alpha \nabla_\rvx \log p_{\sigma_i}(\rvx_t) + \sqrt{2\alpha} \rvg_t = \rvx_t - \alpha \frac{1}{\sigma_i^2}\mSigma^{-1}\rvx_t + \sqrt{2\alpha} \rvg_t,
\end{gather*}
where $\rvg_t \sim \gN(\vzero, \mSigma)$, $\alpha = \epsilon \frac{\sigma_i^2}{\sigma_L^2}$. Therefore, the variance of $\rvx_t$ satisfies
\begin{align*}
    \Var[\rvx_t] = \begin{cases}
        \sigma_{i-1}^2 \mSigma \quad & \text{if $t = 0$}\\
        \Var[\big(\rmI - \frac{\alpha}{\sigma_i^2}\mSigma^{-1}\big) \rvx_{t-1}] + 2\alpha \mSigma \quad & \text{otherwise}.
    \end{cases}
\end{align*}

\begin{align*}
\Var[\rmA \rvx] = \rmA \Var[\rvx] \rmA^\trns
&\implies \Var[\rvx_t]
= \big(\rmI - \frac{\alpha}{\sigma_i^2}\mSigma^{-1}\big) \Var[\rvx_{t-1}] \big(\rmI - \frac{\alpha}{\sigma_i^2}\mSigma^{-1}\big) + 2\alpha \mSigma.
\end{align*}

Let $\rmP = \rmI - \frac{\alpha}{\sigma_i^2}\mSigma^{-1} = \rmI - \frac{\epsilon}{\sigma_L^2}\mSigma^{-1}$.
\begin{align*}
\implies \Var[\rvx_t]
&= \rmP \Var[\rvx_{t-1}] \rmP + 2\alpha \mSigma = \rmP (\rmP \Var[\rvx_{t-2}] \rmP + 2\alpha \mSigma) \rmP + 2\alpha \mSigma\\
&= \rmP \rmP \Var[\rvx_{t-2}] \rmP \rmP + 2\alpha ( \rmP \mSigma \rmP + \mSigma) = \rmP^{(2)} \Var[\rvx_{t-2}] \rmP^{(2)} + 2\alpha ( \rmP \mSigma \rmP + \mSigma), \\
\implies \Var[\rvx_T]
&= \rmP^{(T)} \Var[\rvx_0] \rmP^{(T)} + 2\alpha \sum_{t=0}^{T-1} ( \rmP^{(t)} \mSigma \rmP^{(t)}) = \sigma_{i-1}^2 \rmP^{(T)} \mSigma \rmP^{(T)} + 2\epsilon \frac{\sigma_i^2}{\sigma_L^2} \sum_{t=0}^{T-1} ( \rmP^{(t)} \mSigma \rmP^{(t)}), \\
\implies \frac{\Var[\rvx_T]}{\sigma_i^2} &= \gamma^2 \rmP^{(T)} \mSigma \rmP^{(T)} + \frac{2\epsilon}{\sigma_L^2} \sum_{t=0}^{T-1} ( \rmP^{(t)} \mSigma \rmP^{(t)}).
\end{align*}

\subsection{Optimal conditional score function}

\citet{jolicoeur2020adversarial} discovered in Appendix E that for SMLD, the conditional score estimate in expectation does not match the theoretic value when derived from the unconditional score model in the case of a single data point $\rvx_0$:
\begin{align}
\text {unconditional } \rvs_\rvtheta(\rvx_i) &= \frac{1}{L}\sum_{i=1}^L \left( \frac{\EE_{\rvx_0 \sim q_{\sigma_i}(\rvx_0 \mid \rvx_i)}[\rvx_0] - \rvx_i}{\sigma_i} \right) = \frac{\rvx_0 - \rvx_i}{\sigma_H}, \\
\implies \text {conditional } \rvs_\rvtheta(\rvx_i, \sigma_i) &= \frac{1}{\sigma_i}\rvs_\rvtheta(\rvx_i) = \frac{\rvx_0 - \rvx_i}{\sigma_i \sigma_H} \neq \frac{\rvx_0 - \rvx_i}{\sigma_i^2},
\end{align}
where $\frac{1}{\sigma_H} = \frac{1}{L}\sum_{i=1}^L \frac{1}{\sigma_i}$, i.e. $\sigma_H$ is the harmonic mean of the $\sigma_i$s used to train.

In the case of NI-SMLD, using calculus of variations,
\begin{align*}
\frac{\partial \gL}{\partial \rvs} &= \int \int q_{\sigma}(\rvx_i, \rvx, \sigma) \left( \rvs(\rvx_i) + \sqrt{\mSigma^{-1}}\frac{\rvx_i - \rvx}{\sigma} \right) \D\rvx \D \sigma = 0, \\
&\iff \rvs(\rvx_i)q(\rvx_i) = \sqrt{\mSigma^{-1}} \int \int q_{\sigma}(\rvx_i, \rvx)p(\sigma) \left( \frac{\rvx_i - \rvx}{\sigma} \right) \D\rvx \D \sigma, \\
&\iff \rvs(\rvx_i)q(\rvx_i) = \sqrt{\mSigma^{-1}} \EE_{\sigma \sim p(\sigma)} \left[ \int q_{\sigma}(\rvx_i \mid \rvx) q(\rvx_i) \left( \frac{\rvx_i - \rvx}{\sigma} \right) \D\rvx \right], \\
&\iff \rvs(\rvx_i) = \sqrt{\mSigma^{-1}} \EE_{\sigma \sim p(\sigma)} \left[ \int q_{\sigma}(\rvx_i \mid \rvx) \left( \frac{\rvx_i - \rvx}{\sigma} \right) \D\rvx \right], \\
&\iff \text{unconditional } \rvs(\rvx_i) = \sqrt{\mSigma^{-1}} \EE_{\sigma \sim p(\sigma)} \left[ \frac{\EE_{\rvx \sim q_\sigma(\rvx \mid \rvx_i)}[\rvx] - \rvx}{\sigma} \right].
\end{align*}
For SMLD, in the case of a single data point $\rvx_0$:
\begin{align*}
\text{unconditional } \rvs(\rvx_i) &= \sqrt{\mSigma^{-1}} \frac{\rvx_0 - \rvx_i}{\sigma_H}, \\
\implies \text{conditional } \rvs_\rvtheta(\rvx_i, \sigma_i) &= \frac{1}{\sigma_i}\sqrt{\mSigma^{-1}}\rvs_\rvtheta(\rvx_i) = \mSigma^{-1} \frac{\rvx_0 - \rvx_i}{\sigma_i \sigma_H} \neq \mSigma^{-1} \frac{\rvx_0 - \rvx_i}{\sigma_i^2}.
\numberthis
\end{align*}

Hence, even for NI-SMLD, the discrepancy between the theoretical value of the score, and the estimated conditional score derived from the unconditional score persists. We correct for this discrepancy while sampling:
\begin{align}
\text{conditional } \rvs_\rvtheta(\rvx_i, \sigma_i) = \frac{\sigma_H}{\sigma_i}\left[\frac{1}{\sigma_i}\sqrt{\mSigma^{-1}} \rvs_\rvtheta(\rvx_i)\right] = \frac{\sigma_H}{\sigma_i^2} \sqrt{\mSigma^{-1}} \rvs_\rvtheta(\rvx_i)
.
\end{align}

\section{Isotropic DDPM v/s Non-isotropic DDPM}
\label{ni_comp}

Below is an essential summary of isotropic and non-isotropic Gaussian denoising diffusion models for a more direct comparison. 

\begin{figure}[!htb]
\centering
\includegraphics[width=\linewidth]{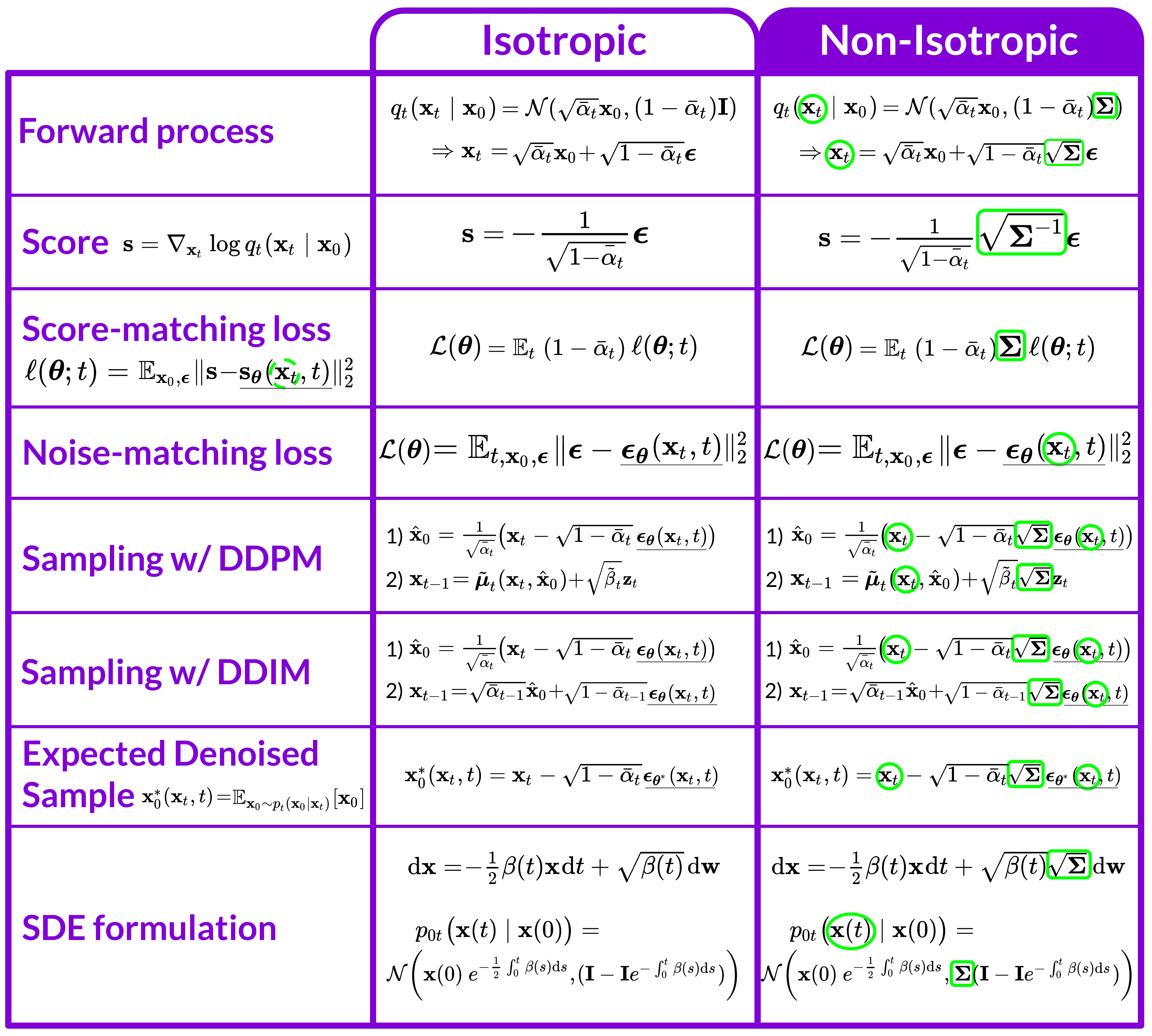}
\caption{Isotropic v/s Non-Isotropic Denoising Diffusion Probabilistic Models}
\label{figs:IvsNIDDPM}
\end{figure}

\subsection{Isotropic Gaussian denoising diffusion models (DDPM)}

We perform our analysis below within the Denoising Diffusion Probabilistic Models (DDPM)~\citep{ho2020ddpm} framework, but our analysis is valid for all other types of score-based denoising diffusion models.

In DDPM, for a fixed sequence of positive scales $0 < \beta_1 < \cdots < \beta_L < 1$, $\balpha_t = \prod_{s=1}^t (1 - \beta_s)$, and a noise sample $\rvepsilon \sim \gN(\vzero, \rmI)$, the cumulative ``\textbf{forward}'' noising process is:
\begin{align}
&q_t (\rvx_t \mid \rvx_0) = \gN(\sqrt{\balpha_t} \rvx_0, (1 - \balpha_t) \rmI) \implies \rvx_t = \sqrt{\balpha_t}\rvx_0 + \sqrt{1 - \balpha_t} \rvepsilon.
\label{eq:DDPM_noise1}
\end{align}
The ``\textbf{reverse}'' process involves iteratively \textbf{sampling} $\rvx_{t-1}$ from $\rvx_t$ conditioned on $\rvx_0$ i.e. $p_{t-1}(\rvx_{t-1} \mid \rvx_t, \rvx_0)$, obtained from $q_t(\rvx_t \mid \rvx_0)$ using Bayes' rule. For this, first $\rvepsilon$ is estimated using a neural network $\rvepsilon_\rvtheta(\rvx_t, t)$. Then, using $\hat\rvx_0 = \big(\rvx_t - \sqrt{1 - \balpha_t}\rvepsilon_{\rvtheta}(\rvx_t, t)\big)/\sqrt{\balpha_t}$ from \cref{eq:DDPM_noise1}, $\rvx_{t-1}$ is sampled:
\begin{gather*}
p_{t-1}(\rvx_{t-1} \mid \rvx_t, \hat{\rvx}_0) = \gN(\ \tilde\vmu_{t}(\rvx_t, \hat{\rvx}_0), \tilde\beta_{t}\rmI \ ) \implies \rvx_{t-1} = \tilde\vmu_{t}(\rvx_t, \hat{\rvx}_0) + \sqrt{\tilde\beta_{t}}\rvz_t \quad \text{, where} \numberthis \\
\tilde\vmu_{t}(\rvx_t, \hat{\rvx}_0) = \frac{\sqrt{\balpha_{t-1}}\beta_t}{1 - \balpha_t}\hat{\rvx}_0 + \frac{\sqrt{1 - \beta_t}(1 - \balpha_{t-1})}{1 - \balpha_t}\rvx_t \ ;\ \tilde\beta_t = \frac{1 - \balpha_{t-1}}{1 - \balpha_t}\beta_t \ ;\ \rvz_t \sim \gN(\vzero, \rmI)\ .
\numberthis
\label{eq:reverse}
\end{gather*}
The \textbf{objective} to train $\rvepsilon_\rvtheta(\rvx_t, t)$ is simply an expected reconstruction loss with the true $\rvepsilon$:
\begin{align}
\gL_\rvepsilon(\rvtheta) = \EE_{t \sim \gU({1,\cdots,L}), \rvx_0 \sim p(\rvx_0), \rvepsilon \sim \gN(\vzero, \rmI)} \left[ \Norm{ \rvepsilon - \rvepsilon_\rvtheta\big(\sqrt{\balpha_t}\rvx_0 + \sqrt{1 - \balpha_t} \rvepsilon, t \big)}_2^2 \right].
\label{eq:eps_loss}
\end{align}
From the perspective of score matching, the \textbf{score} of the DDPM forward process is:
\begin{align}
\text{Score }\ \rvs = \ &\nabla_{\rvx_t} \log q_{t} (\rvx_t \mid \rvx_0) = -\frac{1}{(1 - \balpha_t)}(\rvx_t - \sqrt{\balpha_t} \rvx_0) = -\frac{1}{\sqrt{1 - \balpha_t}}\rvepsilon .
\label{eq:score}
\end{align}
Thus, the overall \textbf{score-matching objective} for a score estimation network $\rvs_\rvtheta(\rvx_t, t)$ is the weighted sum of the loss $\loss_\rvs(\rvtheta; t)$ for each $t$, the weight being the inverse of the score \textbf{variance} at $t$ i.e. $(1 - \balpha_t)$:
\begin{align}
&\gL_\rvs(\rvtheta) = \EE_t\ (1 - \balpha_t)\ \loss_\rvs(\rvtheta; t) = \EE_{t, \rvx_0, \rvepsilon} \bigg[ \Norm{\sqrt{1 - \balpha_t}\rvs_\rvtheta (\rvx_t, t) + \rvepsilon }_2^2 \bigg] .
\end{align}
When the score network output is redefined as per the \textbf{score-noise relationship} in \cref{eq:score}:
\begin{align}
\rvs_\rvtheta(\rvx_t, t) = -\frac{1}{\sqrt{1 - \balpha_t}}\rvepsilon_\evtheta(\rvx_t, t)\ \implies\ \gL_\rvs(\rvtheta) = \EE_{t, \rvx_0, \rvepsilon} \bigg[ \Norm{-\rvepsilon_\rvtheta (\rvx_t, t) + \rvepsilon }_2^2 \bigg] \quad = \gL_\rvepsilon(\rvtheta)
\end{align}
Thus, $\gL_\rvs = \gL_\rvepsilon$ i.e. the score-matching and noise reconstruction objectives are equivalent.
\begin{align*}
&\rvs_{\rvtheta^*}(\rvx_t, t) = \EE\left[ \Norm{ \nabla_{\rvx_t} \log q_{t} (\rvx_t \mid \rvx_0) }_2^2 \right] \big(\rvx_0^* (\rvx_t, t) - \rvx_t\big) = \frac{1}{1 - \balpha_t}\big(\rvx_0^* (\rvx_t, t) - \rvx_t\big), \numberthis \\
\implies
&\rvx_0^*(\rvx_t, t) = \rvx_t + (1 - \balpha_t)\ \rvs_{\rvtheta^*}(\rvx_t, t) = \rvx_t - \sqrt{1 - \balpha_t} \ \rvepsilon_{\rvtheta^*}(\rvx_t, t).
\numberthis
\end{align*}
The EDS is often used to further improve the quality of the final image at $t=0$.


\subsection{Non-isotropic Gaussian denoising diffusion models (NI-DDPM)}

We formulate the Non-Isotropic DDPM (\textbf{NI-DDPM}) using a non-isotropic Gaussian noise distribution with a positive semi-definite covariance matrix $\mSigma$ in the place of $\rmI$.

The \textbf{forward} noising process is:
\begin{align}
&q_t (\rvx_t \mid \rvx_0) = \gN(\sqrt{\balpha_t} \rvx_0, (1 - \balpha_t) \mSigma)
\implies \rvx_t = \sqrt{\balpha_t}\rvx_0 + \sqrt{1 - \balpha_t} \sqrt{\mSigma} \rvepsilon.
\label{eq:niddpm_noise}
\end{align}
Thus, the \textbf{score} of NI-DDPM is (see \Cref{sec:niddpm_score} for derivation):
\begin{align}
&\text{Score }\ \rvs = \nabla_{\rvx_t} \log q_t (\rvx_t \mid \rvx_0) = -\mSigma^{-1}\frac{\rvx_t - \sqrt{\balpha_t} \rvx_0}{1 - \balpha_t} = -\frac{1}{\sqrt{1 - \balpha_t}}\sqrt{\mSigma^{-1}}\rvepsilon.
\label{eq:GFF_score}
\end{align}
The \textbf{score-matching objective} for a score estimation network $\rvs_\rvtheta(\rvx_t, t)$ at each noise level $t$ is now:
\begin{align}
&\loss(\rvtheta; t) = \EE_{\rvx_0 \sim p(\rvx_0), \rvepsilon \sim \gN(\vzero, \rmI)} \bigg[ \Norm{\rvs_\rvtheta (\sqrt{\balpha_t}\rvx_0 + \sqrt{1 - \balpha_t} \rvepsilon, t) + \frac{1}{\sqrt{1 - \balpha_t}}\sqrt{\mSigma^{-1}}\rvepsilon }_2^2 \bigg].
\label{eq:GFF_score_loss_t}
\end{align}
The \textbf{variance} of this score is:
\begin{align*}
&\EE\left[ \Norm{ \nabla_{\rvx_t} \log q_{t}(\rvx_t \mid \rvx_0) }_2^2 \right]
= \EE\left[ \Norm{ -\frac{1}{\sqrt{1 - \balpha_t}}\sqrt{\mSigma^{-1}}\rvepsilon }_2^2 \right]
= \frac{1}{1 - \balpha_t}\mSigma^{-1}\EE\left[ \Norm{\rvepsilon}_2^2 \right].
\label{eq:GFF_variance}
\numberthis
\end{align*}
The \textbf{overall objective} is a weighted sum, the weight being the inverse of the score variance $(1 - \balpha_t)\mSigma$:
\begin{align}
\gL(\rvtheta) &= \EE_{t \sim \gU({1,\cdots,L})}\ (1 - \balpha_t)\mSigma \ \loss(\rvtheta; t) = \EE_{t, \rvx_0, \rvepsilon} \bigg[ \Norm{\sqrt{1 - \balpha_t}\sqrt{\mSigma}\rvs_\rvtheta (\rvx_t, t) + \rvepsilon }_2^2 \bigg].
\end{align}
Following the \textbf{score-noise relationship} in \cref{eq:GFF_score}:
\begin{align}
&\rvs_\rvtheta(\rvx_t, t) = -\frac{1}{\sqrt{1 - \balpha_t}}\sqrt{\mSigma^{-1}}\rvepsilon_\rvtheta(\rvx_t, t).
\label{eq:niddpm_stheta}
\end{align}
The \textbf{objective function} now becomes (expanding $\rvs_\rvtheta$ as per \cref{eq:niddpm_stheta}):
\begin{align}
&\gL(\rvtheta) = \EE_{t \sim \gU({1,\cdots,L}), \rvx_0 \sim p(\rvx_0), \rvepsilon \sim \gN(\vzero, \rmI)} \bigg[ \Norm{-\rvepsilon_\rvtheta (\sqrt{\balpha_t}\rvx_0 + \sqrt{1 - \balpha_t} \sqrt{\mSigma} \rvepsilon, t) + \rvepsilon }_2^2 \bigg].
\end{align}
This objective function for NI-DDPM seems like $\gL_\epsilon$ of DDPM, but DDPM's $\rvepsilon_\rvtheta$ network cannot be re-used here since their forward processes are different. DDPM produces $\rvx_t$ from $\rvx_0$ using \cref{eq:DDPM_noise1}, while NI-DDPM uses \cref{eq:niddpm_noise}. See \Cref{subsec:niddpm_objective} for alternate formulations of the score network.

\textbf{Sampling} involves computing $p_{t-1}(\rvx_{t-1} \mid \rvx_t, \hat{\rvx}_0)$ (see \Cref{subsec:sampling_niddpm} for derivation):
\begin{align*}
q_{t} (\rvx_t \mid \rvx_0) = \gN(\sqrt{\balpha_t} \rvx_0, (1 - \balpha_t) \mSigma)
&\implies \hat\rvx_0 = \frac{1}{\sqrt{\balpha_t}}\big(\rvx_t - \sqrt{1 - \balpha_t}\sqrt{\mSigma}\rvepsilon_{\rvtheta}(\rvx_t, t)\big).
\label{eq:niddpm_x0hat}
\numberthis
\\
p_{t-1}(\rvx_{t-1} \mid \rvx_t, \hat{\rvx}_0) = \gN( \tilde\vmu_t(\rvx_t, \hat{\rvx}_0), \tilde\beta_t\mSigma)
&\implies \rvx_{t-1}
= \tilde\vmu_t(\rvx_t, \hat{\rvx}_0) + \sqrt{\tilde\beta_t} \sqrt{\mSigma} \rvz_t.
\numberthis
\end{align*}
where $\tilde\vmu_t$, $\tilde\beta_t$ and $\rvz_t$ are the same as \cref{eq:reverse}.

Alternatively, \cite{song2020sde} mentions using $\beta_t$ instead of $\tilde\beta_t$:
\begin{align*}
p_{t-1}^{\beta_t}(\rvx_{t-1} \mid \rvx_t, \hat{\rvx}_0) = \gN( \tilde\vmu_t(\rvx_t, \hat{\rvx}_0), \beta_t\mSigma)
&\implies \rvx_{t-1} = \tilde\vmu_t(\rvx_t, \hat{\rvx}_0) + \sqrt{\beta_t} \sqrt{\mSigma} \rvz_t.
\numberthis
\end{align*}
Alternatively, sampling using \textbf{DDIM}~\cite{song2020ddim} invokes the following distribution for $\rvx_{t-1}$:
\begin{align*}
&p_{t-1}^{\text{DDIM}}(\rvx_{t-1} \mid \rvx_t, \hat \rvx_0) = \gN\left( \sqrt{\balpha_{t-1}}\hat{\rvx}_0 + \sqrt{1 - \balpha_{t-1}}\frac{\rvx_t - \sqrt{\balpha_t}\hat{\rvx}_0}{\sqrt{1 - \balpha_t}}, \ \ \vzero \right).
\numberthis
\\
\implies
&\rvx_{t-1} = \sqrt{\balpha_{t-1}} \hat\rvx_0 + \sqrt{1 - \balpha_{t-1}} \sqrt{\mSigma} \rvepsilon_{\rvtheta}(\rvx_t, t).
\numberthis
\end{align*}
The \textbf{Expected Denoised Sample} $\rvx_0^*(\rvx_t, t)$ and the optimal score $\rvs_{\rvtheta^*}$ are now related as:
\begin{align*}
&\rvs_{\rvtheta^*}(\rvx_t, t) = \EE\left[ \Norm{ \nabla_{\rvx_t} \log q_{t} (\rvx_t \mid \rvx_0) }_2^2 \right] \big(\rvx_0^* (\rvx_t, t) - \rvx_t\big) = \frac{1}{1 - \balpha_t} \mSigma^{-1} \big(\rvx_0^* (\rvx_t, t) - \rvx_t\big). \numberthis \\
&\implies
\rvx_0^*(\rvx_t, t) = \rvx_t + (1 - \balpha_t)\ \mSigma \rvs_{\rvtheta^*}(\rvx_t, t) = \rvx_t - \sqrt{1 - \balpha_t} \ \sqrt{\mSigma} \rvepsilon_{\rvtheta^*}(\rvx_t, t).
\numberthis
\end{align*}

\textbf{SDE formulation}: Score-based diffusion models have also been analyzed as stochastic differential equations (SDEs)~\cite{song2020sde}. The SDE version of NI-DDPM, which we call Non-Isotropic Variance Preserving (NIVP) SDE, is (see \Cref{subsec:niddpm_sde} for derivation):
\begin{align*}
&\D \rvx = -\frac{1}{2}\beta(t)\rvx\ \D t + \sqrt{\beta(t)}\sqrt{\mSigma}\ \D \rvw.
\numberthis
\\
\implies &p_{0t}\big(\rvx(t) \mid \rvx(0)\big) = \gN \left( \rvx(0)\ e^{-\frac{1}{2}\int_0^t \beta(s) \D s},\ \mSigma(\rmI - \rmI e^{-\int_0^t \beta(s) \D s}) \right).
\numberthis
\end{align*}

Finally, \Cref{sec:ddpm} and \Cref{sec:niddpm} contain more detailed derivations of the above equations for DDPM~\cite{ho2020ddpm} and our NI-DDPM. See \Cref{sec:smld} and \Cref{sec:nismld} for the equivalent derivations for Score Matching Langevin Dynamics (SMLD)~\cite{song2019generative,song2020improved}, and our Non-Isotropic SMLD (NI-SMLD).

\section{Gaussian Free Field (GFF) images}
\label{GFF}

An instantiation of non-isotropic Gaussian noise is Gaussian Free Field~\citep{sheffield2007gff} (GFF). In 2D, this manifests as a GFF image. A GFF image $\rvg$ can be obtained from a normal noise image $\rvz$ as follows~\citep{sheffield2007gff}:

\begin{figure}[!tbh]
\centering
\includegraphics[width=.95\linewidth]{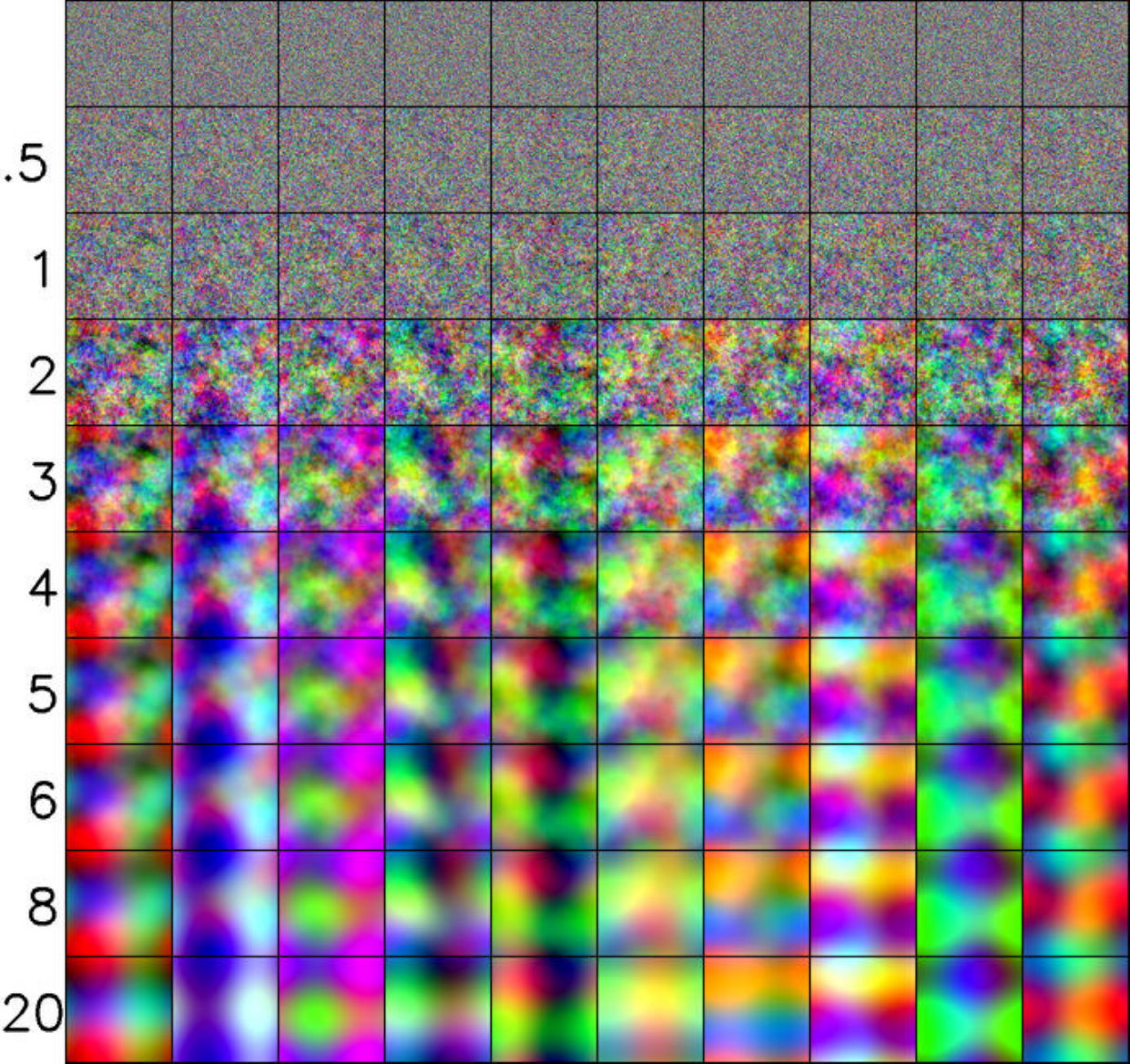}
\caption{(left to right) 10 GFF images (each varying downwards) as a function of the power $\gamma$ of the index (mentioned on the left).}
\label{fig:GFFs}
\end{figure}

\begin{enumerate}
\item
First, sample an $n \times n$ noise image $\rvz$ from the 
standard complex normal distribution with covariance matrix $\Gamma = \rmI_N$ where $N=n^2$ is the total number of pixels, and pseudo-covariance matrix $C = \vzero$: $\rvz \sim \gC \gN(\vzero, \rmI_N, \vzero)$.

The standard \textbf{complex} normal distribution is one where the real part $\rvx$ and imaginary part $\rvy$ are each distributed as the standard normal distribution with variance $\frac{1}{2}\rmI_N$. Let $\mSigma_{\rva \rvb}$ be the covariance matrix between $\rva$ and $\rvb$. We know that $\mSigma_{\rvx \rvx} = \mSigma_{\rvy \rvy} = \frac{1}{2}\rmI_N$, and $\mSigma_{\rvx \rvy} = \mSigma_{\rvy \rvx} = \vzero_N$ Then:
\begin{align}
\Gamma &= \EE_\rvz[\rvz \rvz^\conj] = \mSigma_{\rvx \rvx} + \mSigma_{\rvy \rvy} + i(\mSigma_{\rvy \rvx} - \mSigma_{\rvx \rvy}) = \rmI_N.
\label{eq:cov_z}\\
C &= \EE_\rvz[\rvz \rvz^\trns] = \mSigma_{\rvx \rvx} - \mSigma_{\rvy \rvy} + i(\mSigma_{\rvy \rvx} + \mSigma_{\rvx \rvy}) = \vzero_N.
\label{eq:psudocov_z}
\end{align}
\item
Apply the Discrete Fourier Transform using the $N \times N$ weights matrix $\rmW_N$: $\rmW_N \rvz$.
\item
Consider a diagonal $N \times N$ matrix of the reciprocal of an index value $k_{ij}$ per pixel $(i,j)$ in Fourier space : $\rmK^{-1} = [1/|k_{ij}|]_{(i,j)}$, and multiply this with the above: $\rmK^{-1} \rmW_N \rvz$.
\item
Take its Inverse Discrete Fourier Transform ($\rmW_N^{-1}$) to make the raw GFF image: $\rmW_N^{-1} \rmK^{-1} \rmW_N \rvz$.
However, this results in a GFF image with a small non-unit variance.
\item
Normalize the above GFF image with the standard deviation $\sigma_N$ at its resolution $N$, so that it has unit variance (see \Cref{subsec:gff_prob} for derivation of $\sigma_N$):
\begin{align*}
\rvg_\text{complex} &= \frac{1}{\sigma_N}\rmW_N^{-1} \rmK^{-1} \rmW_N \rvz
\numberthis
\quad \iff \quad \rvz = \sigma_N \rmW_N^{-1} \rmK \rmW_N \rvg_\text{complex}.
\end{align*}
\item
Extract only the real part of $\rvg_\text{complex}$, and normalize (see \Cref{subsec:GFF_real} for derivation):
\begin{align*}
\rvg &= \frac{1}{\sqrt{2N}\sigma_N} \Real \left(\rmW_N^{-1} \rmK^{-1} \rmW_N \rvz \right).
\numberthis
\label{eq:g}
\iff \rvz = \sqrt{2N}\sigma_N \Real \left(\rmW_N^{-1} \rmK \rmW_N \rvg \right).
\end{align*}
\end{enumerate}

See \Cref{fig:gff_sample,fig:GFFs} for examples of GFF images. Effectively, this prioritizes lower frequencies over higher, making noise smoother and correlated.

The probability distribution of GFF images $\rvg$ can be seen as a non-isotropic multivariate Gaussian with mean $\vzero$, and a non-diagonal covariance matrix $\mSigma$ (see \Cref{subsec:gff_prob,subsec:GFF_real} for derivation):
\begin{align}
p(\rvg) &= \gN(\vzero, \mSigma). \\
\mSigma &= \sqrt{\mSigma} \sqrt{\mSigma}^\trns. \\
\sqrt{\mSigma} &= \frac{1}{\sqrt{2N}\sigma_N}\Real\left(\rmW_N^{-1} \rmK^{-1} \rmW_N\right). \\
\implies \rvg &= \sqrt{\mSigma}\rvz.
\numberthis
\label{eq:covG2}
\end{align}

\subsection{Probability distribution of GFF}
\label{subsec:gff_prob}

Let the probability distribution of GFF images be $\mathcal{G}$. This can be seen as a non-isotropic multivariate Gaussian with a non-diagonal covariance matrix $\mSigma$:
\begin{align}
\rvg \sim \gG
= \gN(\vmu, \mSigma)
= \gN(\vzero_N, \mSigma).
\end{align}

We know from the properties of Discrete Fourier Transform (following the normalization convention of the Pytorch / Numpy implementation) that, given the Discrete Fourier Transform matrix $\rmW_N$:
\begin{align}
\rmW_N = \rmW_N^\trns; \rmW_N^{-1} = {\rmW_N^{-1}}^\trns; 
\rmW_N^{-1} = \frac{1}{N}\rmW_N^* = \frac{1}{N}\rmW_N^\conj.
\label{eq:fourier}
\end{align}

\label{subsec:GFF_real}

Here, $\rvg$ is real and $\rvz$ is complex, and $\rvg_\complex$ is complex (real+imaginary).

$\vmu$ is given by:
\begin{align*}
\vmu
&= \EE_{\rvg} [ \rvg ]
= \EE_{\rvg} [\ \frac{1}{2\sigma_N}(\rvg_\complex + \rvg_\complex^*) \ ]
= \frac{1}{2\sigma_N} \left( \EE_\rvg [\rvg_\complex] + \EE_\rvg [\rvg_\complex^*] \right), \\
&= \frac{1}{2\sigma_N} \left( \EE_\rvz [\rmW_N^{-1} \rmK^{-1} \rvz] + \EE_\rvz [(\rmW_N^{-1} \rmK^{-1} \rvz)^*] \right), \\
&= \frac{1}{2\sigma_N} \left( \rmW_N^{-1} \rmK^{-1} \EE_\rvz [\rvz] + \rmW_N^{-1 *} \rmK^{-1} \EE_\rvz [\rvz^*] \right), \\
\implies \vmu &= \vzero_N. \qquad \qquad [\because \EE_\rvz [\rvz] = \EE_\rvz [\rvz^*] = \vzero_N]
\numberthis
\label{eq:mu_G}
\end{align*}

$\mSigma$ is given by:
\begin{align*}
\mSigma
&= \EE_{\rvg} [\ \rvg \ \rvg^\trns \ ], \\
&= \EE_\rvg [\ \frac{1}{2\sigma_N}(\rvg_\complex + \rvg^*_\complex) \ \frac{1}{2\sigma_N}(\rvg_\complex + \rvg_\complex^*)^\trns \ ], \\
&= \frac{1}{4\sigma_N^2} \EE_\rvg [\ (\rvg_\complex + \rvg_\complex^*) \ (\rvg_\complex^\trns + \rvg_\complex^\conj) \ ], \\
&= \frac{1}{4\sigma_N^2} \EE_\rvg [\ \rvg_\complex \rvg_\complex^\trns + \rvg_\complex \rvg_\complex^\conj + \rvg_\complex^* \rvg_\complex^\trns + \rvg_\complex^* \rvg_\complex^\conj \ ], \\
&= \frac{1}{4\sigma_N^2}\left( \EE_\rvg [\rvg_\complex \rvg_\complex^\trns] + \EE_\rvg [\rvg_\complex \rvg_\complex^\conj] + \EE_\rvg [\rvg_\complex^* \rvg_\complex^\trns] + \EE_\rvg [\rvg_\complex^* \rvg_\complex^\conj] \right).
\\
\EE_\rvg [\rvg_\complex \rvg_\complex^\trns]
&= \EE_\rvz [\rmW_N^{-1} \rmK^{-1} \rvz \ (\rmW_N^{-1} \rmK^{-1} \rvz)^\trns], \\
&= \EE_\rvz [\rmW_N^{-1} \rmK^{-1} \rvz \ \rvz^\trns \rmK^{-1} \rmW_N^{-\trns} ], \quad [\because \rmK^{-1} \text{ is diagonal }] \\
&= \rmW_N^{-1} \rmK^{-1} \EE_\rvz [\rvz \rvz^\trns] \rmK^{-1} \rmW_N^{-T}, \\
&= \vzero_N. \qquad \qquad \qquad \qquad \qquad \qquad \quad [\because \EE_\rvz [\rvz \rvz^\trns] = \vzero_N \text{ (\cref{eq:psudocov_z})}]
\\
\EE_\rvg [\rvg_\complex \rvg_\complex^\conj]
&= \EE_\rvz [\rmW_N^{-1} \rmK^{-1} \rvz \ (\rmW_N^{-1} \rmK^{-1} \rvz)^\conj], \\
&= \EE_\rvz [\rmW_N^{-1} \rmK^{-1} \rvz \ \rvz^\conj \rmK^{-1} \rmW_N^{-\conj} ], \quad [\because \mK^{-1} \text{ is real diagonal }] \\
&= \rmW_N^{-1} \rmK^{-1} \EE_\rvz [\rvz \rvz^\conj] \rmK^{-1} \frac{1}{N}\rmW_N,  \quad [\because \rmW_N^{-1} = \frac{1}{N}\rmW_N^\conj \text{ (\cref{eq:fourier})} ] \\
&= \frac{1}{N} \rmW_N^{-1} \rmK^{-1} \rmK^{-1} \rmW_N. \qquad \qquad [\because \EE_\rvz [\rvz \rvz^\conj] = \rmI_N \text{ (\cref{eq:cov_z})}]
\\
\EE_\rvg [\rvg_\complex^* \rvg_\complex^\trns]
&= \EE_\rvz [(\rmW_N^{-1} \rmK^{-1} \rvz)^* \ (\rmW_N^{-1} \rmK^{-1} \rvz)^\trns], \\
&= \EE_\rvz [\rmW_N^{-1 *} \rmK^{-1} \rvz^* \ \rvz^\trns \rmK^{-1} \rmW_N^{-1} ], \\
&= \frac{1}{N}\rmW_N \rmK^{-1} \EE_\rvz [\rvz^* \rvz^\trns] \rmK^{-1} \rmW_N^{-1}, \quad [\because \rmW_N^{-1} = \frac{1}{N}\rmW_N^* \text{ (\cref{eq:fourier})} ] \\
&= \frac{1}{N} \rmW_N \rmK^{-1} \rmK^{-1} \rmW_N^{-1}. \quad [\because \EE_\rvz [\rvz^* \rvz^\trns] = \EE_\rvz [\rvz \rvz^\conj]^* = \rmI_N \text{ (\cref{eq:cov_z})}]
\\
\EE_\rvg [\rvg_\complex^* \rvg_\complex^\conj]
&= \EE_\rvz [(\rmW_N^{-1} \rmK^{-1} \rvz)^* \ (\rmW_N^{-1} \rmK^{-1} \rvz)^\conj], \\
&= \EE_\rvz [\rmW_N^{-1 *} \rmK^{-1} \rvz^* \ \rvz^\conj \rmK^{-1} \rmW_N^{-\conj} ], \\
&= \rmW_N^{-1 *} \rmK^{-1} \EE_\rvz [\rvz^* \rvz^\conj] \rmK^{-1} \rmW_N^{-\conj}, \\
&= \vzero_N. \qquad \qquad \qquad \qquad [\because \EE_\rvz [\rvz^* \rvz^\conj] = \EE_\rvz [\rvz \rvz^\trns]^* = \vzero_N \text{ (\cref{eq:cov_z})}]
\\
\implies \mSigma &= \frac{1}{4\sigma_N^2}\left( \vzero_N + \frac{1}{N} \rmW_N^{-1} \rmK^{-1} \rmK^{-1} \rmW_N + \frac{1}{N} \rmW_N \rmK^{-1} \rmK^{-1} \rmW_N^{-1} + \vzero_N \right), \\
&= \frac{1}{4N\sigma_N^2}\left(\rmW_N^{-1} \rmK^{-1} \rmK^{-1} \rmW_N + \rmW_N \rmK^{-1} \rmK^{-1} \rmW_N^{-1} \right), \\
&= \frac{1}{4N\sigma_N^2}\left(\rmW_N^{-1} \rmK^{-1} \rmK^{-1} \rmW_N + (N \rmW_N^{-1 *}) \rmK^{-1} \rmK^{-1} (\frac{1}{N}\rmW_N^*) \right), \\
&= \frac{1}{2N\sigma_N^2}\left(\frac{1}{2}\left(\rmW_N^{-1} \rmK^{-1} \rmK^{-1} \rmW_N + (\rmW_N^{-1} \rmK^{-1} \rmK^{-1} \rmW_N)^*\right) \right), \\
\implies \mSigma &= \frac{1}{2N\sigma_N^2}\ \Real \left( \rmW_N^{-1} \rmK^{-1} \rmK^{-1} \rmW_N \right).
\numberthis
\label{eq:covG}
\\
\sqrt{\mSigma} &= \frac{1}{\sqrt{2N}\sigma_N}\ \Real \left( \rmW_N^{-1} \rmK^{-1} \rmW_N \right),
\numberthis
\label{eq:covG_sqrt}
\\
\mSigma^{-1} &= 2N\sigma_N^2\ \Real \left( \rmW_N^{-1} \rmK \rmK \rmW_N \right),
\numberthis
\label{eq:covGinv}
\\
\sqrt{\mSigma^{-1}} &= \sqrt{2N}\sigma_N\ \Real \left( \rmW_N^{-1} \rmK \rmW_N \right).
\numberthis
\label{eq:covGinv_sqrt}
\end{align*}

\subsection{Log probability of transformation}
\label{subsec:gff_logprob}
\begin{align*}
\rvg = \sqrt{\mSigma}\rvz
\implies \log p(\rvg) &= \log p(\rvz) - \log \Abs{\det \frac{\D \rvg}{\D \rvz}} = \log p(\rvz) - \log \Abs{\det \sqrt{\mSigma}} \\
&= \log p(\rvz) - \log \Abs{\det \frac{1}{\sqrt{2N}\sigma_N}\Real\left(\rmW_N^{-1} \rmK^{-1} \rmW_N\right) } \\
&= \log p(\rvz) - \frac{1}{\sqrt{2N}\sigma_N} \log \Abs{\det\rmK^{-1} }.
\end{align*}

This is useful for building (normalizing) flows using non-isotropic Gaussian noise.

\subsection{Varying $\rmK$}

The index matrix $\rmK$ involves computation of an index value $k_{ij}$ per pixel $(i,j)$. However, this index value could be raised to any power $\gamma$ i.e. $|k_{ij}|^\gamma$. The effect of varying $\gamma$ can be seen in \Cref{fig:GFFs} : greater the $\gamma$, the more correlated are neighbouring pixels.



\section{Experimental results}
\label{NIDDPM_exps}

\begin{table}[!htb]
\caption{Image generation metrics FID, Precision (P), and Recall (R) for CIFAR10 using DDPM and NI-DDPM, with different generation steps.\vspace{-1.5em}}
\centering
\def\arraystretch{1.4}
\begin{tabular}{lrrcc}
\multicolumn{1}{l}{}                & \multicolumn{1}{l}{}              & \multicolumn{1}{l}{}                    & \multicolumn{1}{l}{}                 \\ \hline
\multicolumn{1}{|l|}{CIFAR10}                    & \multicolumn{1}{l|}{\textbf{steps}} & \multicolumn{1}{l|}{\textbf{FID} $\downarrow$} & \multicolumn{1}{l|}{\textbf{P} $\uparrow$} & \multicolumn{1}{l|}{\textbf{R} $\uparrow$} \\ \hline
\multicolumn{1}{|l|}{\multirow{5}{*}{DDPM}}         & \multicolumn{1}{r|}{1000}           & \multicolumn{1}{r|}{6.05}         & \multicolumn{1}{r|}{0.66}               & \multicolumn{1}{r|}{0.54}            \\ \cline{2-5} 
\multicolumn{1}{|l|}{}                                        & \multicolumn{1}{r|}{100}            & \multicolumn{1}{r|}{12.25}        & \multicolumn{1}{r|}{0.62}               & \multicolumn{1}{r|}{0.48}            \\ \cline{2-5} 
\multicolumn{1}{|l|}{}                                        & \multicolumn{1}{r|}{50}             & \multicolumn{1}{r|}{16.61}        & \multicolumn{1}{r|}{0.60}               & \multicolumn{1}{r|}{0.43}            \\ \cline{2-5} 
\multicolumn{1}{|l|}{}                                        & \multicolumn{1}{r|}{20}             & \multicolumn{1}{r|}{26.35}        & \multicolumn{1}{r|}{0.56}               & \multicolumn{1}{r|}{0.24}            \\ \cline{2-5} 
\multicolumn{1}{|l|}{}                                        & \multicolumn{1}{r|}{10}             & \multicolumn{1}{r|}{44.95}        & \multicolumn{1}{r|}{0.49}               & \multicolumn{1}{r|}{0.24}            \\ \hline
\multicolumn{1}{|l|}{\multirow{5}{*}{NI-DDPM}}              & \multicolumn{1}{r|}{1000}           & \multicolumn{1}{r|}{6.95}         & \multicolumn{1}{r|}{0.62}               & \multicolumn{1}{r|}{0.53}            \\ \cline{2-5} 
\multicolumn{1}{|l|}{}                                        & \multicolumn{1}{r|}{100}            & \multicolumn{1}{r|}{12.68}        & \multicolumn{1}{r|}{0.60}               & \multicolumn{1}{r|}{0.49}            \\ \cline{2-5} 
\multicolumn{1}{|l|}{}                                        & \multicolumn{1}{r|}{50}             & \multicolumn{1}{r|}{16.91}        & \multicolumn{1}{r|}{0.57}               & \multicolumn{1}{r|}{0.45}            \\ \cline{2-5} 
\multicolumn{1}{|l|}{}                                        & \multicolumn{1}{r|}{20}             & \multicolumn{1}{r|}{30.41}        & \multicolumn{1}{r|}{0.52}               & \multicolumn{1}{r|}{0.35}            \\ \cline{2-5} 
\multicolumn{1}{|l|}{}                                        & \multicolumn{1}{r|}{10}             & \multicolumn{1}{r|}{60.32}        & \multicolumn{1}{r|}{0.43}               & \multicolumn{1}{r|}{0.23}            \\ \hline
\end{tabular}
\label{tab:1}
\end{table}

We train two models on CIFAR10, one using DDPM and the other using NI-DDPM with the exact same hyperparameters (batch size, learning rate, etc.) for 300,000 iterations. We then sample 50,000 images from each, and calculate the image generation metrics of Fr\'echet Inception Distance (FID)~\cite{heusel2017gans}, Precision (P), and Recall (R).
Although the models were trained on 1000 steps between data and noise, we report these metrics while sampling images using 1000, and smaller steps: 100, 50, 20, 10.

As can be seen from \Cref{tab:1}, our non-isotropic variant performs comparable to the isotropic baseline. The difference between them increases with decreasing number of steps between noise and data. This provides a reasonable proof-of-concept that non-isotropic Gaussian noise works just as well as isotropic noise when used in denoising diffusion models for image generation. These conclusions could be generalized to denoising diffusion models of any modality, since the theoretical framework of Non-Isotropic Denoising Diffusion Models remains intact irrespective of the modality of data.

\section{Conclusion}
We have presented the key mathematics behind non-isotropic Gaussian DDPMs, as well as a complete example using a GFF. We then noted quantitative comparison of using GFF noise vs. regular noise on the CIFAR-10 dataset. In the appendix, we also include further derivations for non-isotropic SMLD models. GFFs are just one example of a well known class of models that are a subset of non-isotropic Gaussian distributions. 
In the same way that other work has examined non-Gaussian distributions such as the Gamma distribution~\citep{nachmani2021ddgm}, Poisson distribution~\citep{xu2022Poisson}, and Heat dissipation processes~\citep{Rissanen2022heat}, we hope that our work here may lay the foundation for other new denoising diffusion formulations. 

%% file: MCVD.tex
\anglais
\counterwithin{figure}{chapter}
\counterwithin{table}{chapter}

\chapter{MCVD: Masked Conditional Video Diffusion~\citep{voleti2022MCVD}}
\label{chap:MCVD}

\anglais
\counterwithin{figure}{chapter}
\counterwithin{table}{chapter}

\setcounter{section}{-1}
\section{Prologue to article}
\label{chap:pro_MCVD}

\subsection{Article details}

\textbf{MCVD: Masked Conditional Video Diffusion for Prediction, Generation, and Interpolation}. Vikram Voleti*, Alexia Jolicoeur-Martineau*, Christopher Pal (*denotes equal contribution).
\textit{Advances in Neural Information Processing Systems (NeurIPS) 2022}.

\textit{Personal contribution}:
The project began with discussions between the authors at Mila. The idea was to apply denoising diffusion models to model the modality of video. Since denoising diffusion models were shown to work very well on images, it was to be seen whether they could be trained to generate videos. Vikram Voleti and Alexia Jolicoeur-Martineau initially discussed some ideas with other members of the research community, however they did not work as expected. Then Vikram Voleti worked on an initial idea, wrote the code, and conducted preliminary experiments using Moving MNIST that proved that it was indeed possible for denoising diffusion models to model video successfully. Vikram Voleti and Alexia Jolicoeur-Martineau then joined forces to scale this up to bigger and more complex datasets. Christopher Pal provided advice and guidance throughout the project, provided the idea of masking the past and future frames so that a single model can solve all video tasks simultaneously, and wrote parts of the paper. Vikram and Alexia both contributed to the code, experimental design, experiments, improvements to the model architecture, metrics for evaluation, sampling techniques, writing of the final publication, rebuttal, release of code and model checkpoints.

\subsection{Context}

Video prediction is a challenging task. The quality of video frames from current state-of-the-art generative models tends to be poor and generalization beyond the training data is difficult. Furthermore, existing prediction frameworks are typically not capable of simultaneously handling other video-related tasks such as unconditional generation or interpolation. The recently proposed denoising diffusion models, although very successful at generating images, had not been applied to the modality of video yet.

\subsection{Contributions}

In this work, we devise a general-purpose framework called Masked Conditional Video Diffusion (MCVD) for all of these video synthesis tasks using a probabilistic conditional score-based denoising diffusion model, conditioned on past and/or future frames. We train the model in a manner where we randomly and independently mask all the past frames or all the future frames. This novel but straightforward setup allows us to train a single model that is capable of executing a broad range of video tasks, specifically: future/past prediction -- when only future/past frames are masked; unconditional generation -- when both past and future frames are masked; and interpolation -- when neither past nor future frames are masked. Our experiments show that this approach can generate high-quality frames for diverse types of videos. Our MCVD models are built from simple non-recurrent 2D-convolutional architectures, conditioning on blocks of frames and generating blocks of frames. We generate videos of arbitrary lengths autoregressively in a block-wise manner. Our approach yields state-of-the-art results across standard video prediction and interpolation benchmarks, with computation times for training models measured in 1-12 days using $\le$ 4 GPUs. 

Project page: \url{https://mask-cond-video-diffusion.github.io}

Code: \url{https://mask-cond-video-diffusion.github.io/}

\subsection{Research impact}

Denoising diffusion models have quickly become the default generative models for most modalities: images, video, audio, 3D, etc. Since our work was published, several works (including a few concurrent works) have used denoising diffusion models to generate videos~\citep{ho2022VDM,yang2022ResidualVideoDiffusion,harvey2022flexible,hoppe2022diffusion,yu2023video,he2022latent,nikankin2022sinfusion,luo2023decomposed,yin2023nuwa}. Others have further conditioned them on text prompts to make text-to-video models, including Make-A-Video by Meta~\citep{singer2022make}, ImagenVideo by Google~\citep{ho2022imagevideo}, Phenaki by Google~\citep{villegas2022phenaki}, MagicVideo by ByteDance~\citep{zhou2022magicvideo}, Gen2 by RunwayML~\citep{esser2023structure}, etc. This has been taken to the next level in Make-A-Video3D by Meta~\citep{singer2023textto4d}, which Vikram contributed to during an internship at Meta, to generate 4D scenes from text prompts i.e. text-to-4D based on text-to-video models.

\begin{figure}[!thb]
    \centering
    \includegraphics[width=\textwidth,interpolate=false]{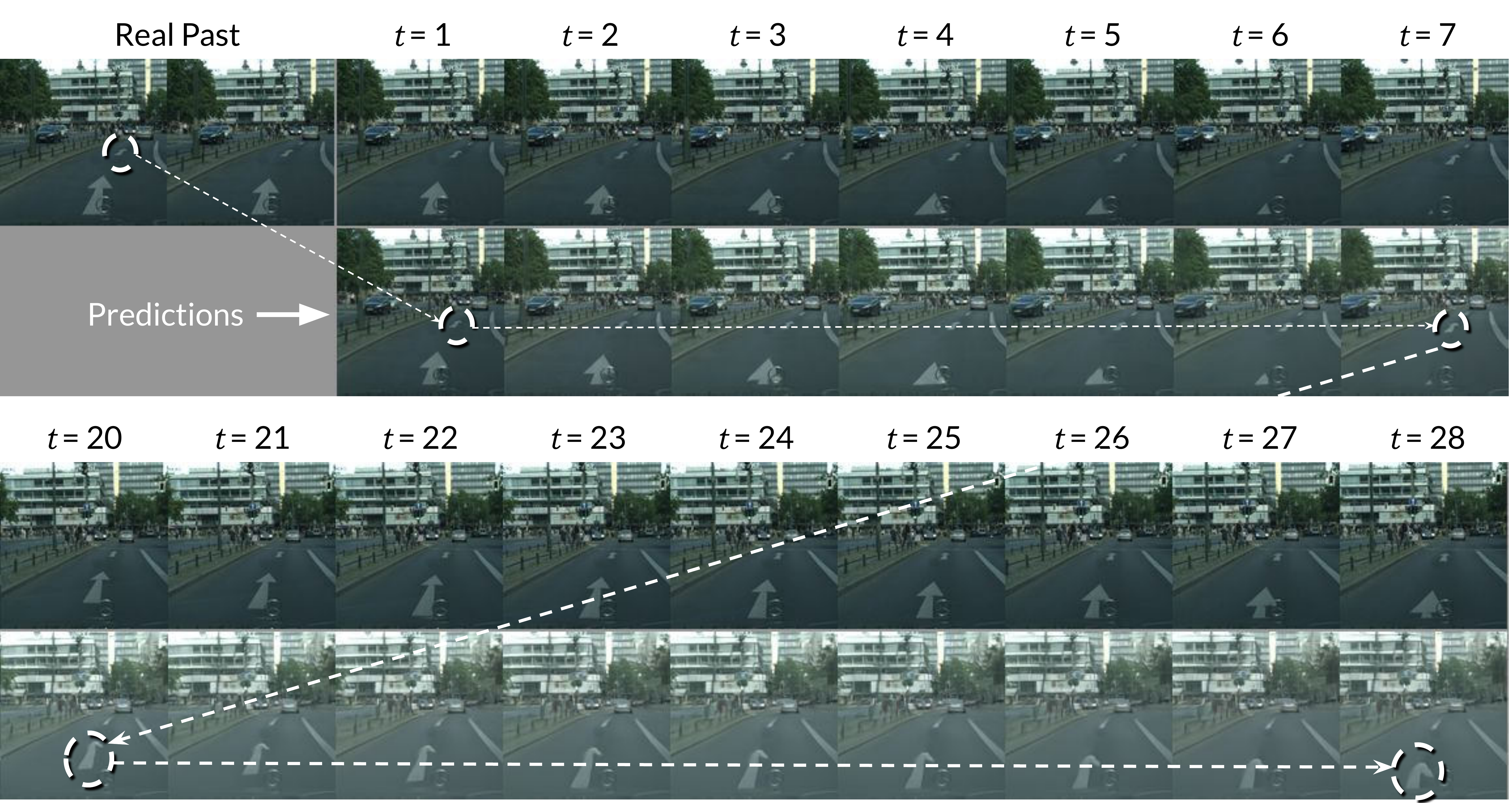}
    \caption{Our approach generates high quality frames many steps into the future: Given two conditioning frames from the Cityscapes \citep{cordts2016cityscapes} validation set (top left), we show 7 predicted future frames in row 2 below, then skip to frames 20-28, autoregressively predicted in row 4. Ground truth frames are shown in rows 1 and 3. Notice the initial large arrow advancing and passing under the car. In frame 20 (the far left of the 3rd and 4th row), the initially small and barely visible second arrow in the background of the conditioning frames has advanced into the foreground. Result generated by our \textbf{MCVD} method (concat variant). Note that some Cityscapes videos contain brightness changes, which may explain the brightness change in this sample.}
    \label{fig:teaser}
\end{figure}

\section{Introduction}

Predicting what one may visually perceive in the future is closely linked to the dynamics of objects and people. As such, this kind of prediction relates to many crucial human decision-making tasks ranging from making dinner to driving a car. If video models could generate full-fledged videos in pixel-level detail with plausible futures, agents could use them to make better decisions, especially safety-critical ones. Consider, for example, the task of driving a car in a tight situation at high speed. Having an accurate model of the future could mean the difference between damaging a car or something worse. We can obtain some intuitions about this scenario by examining the predictions of our model in Figure \ref{fig:teaser}, where we condition on two frames and predict 28 frames into the future for a car driving around a corner. We can see that this is enough time for two different painted arrows to pass under the car. If one zooms in, one can inspect the relative positions of the arrow and the Mercedes hood ornament in the real versus predicted frames. Pixel-level models of trajectories, pedestrians, potholes, and debris on the road could one day improve the safety of vehicles.

Although beneficial to decision making, video generation is an incredibly challenging problem; not only must high-quality frames be generated, but the changes over time must be plausible and ideally drawn from an accurate and potentially complex distribution over probable futures. Looking far in time is exceptionally hard given the exponential increase in possible futures. Generating video from scratch or unconditionally further compounds the problem because even the structure of the first frame must be synthesized. 
Also related to video generation are the simpler tasks of a) video prediction, predicting the future given the past, and b) interpolation, predicting the in-between given past and future. Yet, both problems remain challenging. Specialized tools exist to solve the various video tasks, but they rarely solve more than one task at a time.



Given the monumental task of general video generation, current approaches are still very limited despite the fact that many state of the art methods have hundreds of millions of parameters \citep{wu2021greedy,weissenborn2019scaling,villegas2019high,babaeizadeh2021fitvid}. While industrial research is capable of looking at even larger models, current methods frequently underfit the data, leading to blurry videos, especially in the longer-term future and recent work has examined ways in improve parameter efficiency \citep{babaeizadeh2021fitvid}. Our objective here is to devise a video generation approach that generates high-quality, time-consistent videos within our computation budget of $\leq$ 4 GPU) and computation times for training models $\leq$ two weeks. Fortunately, diffusion models for image synthesis have demonstrated wide success, which strongly motivated our use of this approach. Our qualitative results in \autoref{fig:teaser} also indicate that our particular approach does quite well at synthesizing frames in the longer-term future (i.e., frame 29 in the bottom right corner).

One family of diffusion models might be characterized as Denoising Diffusion Probabilistic Models (DDPMs) \citep{sohl2015deep, ho2020ddpm, dhariwal2021diffusion}, while another as Score-based Generative Models (SGMs) \citep{song2019generative, li2019learning, song2020improved, jolicoeur2020adversarial}. However, these approaches have effectively merged into a field we shall refer to as score-based diffusion models, which work by defining a stochastic process from data to noise and then reversing that process to go from noise to data. Their main benefits are that they generate very 1) high-quality and 2) diverse data samples. One of their drawbacks is that solving the reverse process is relatively slow, but there are ways to improve speed \citep{song2020ddim,jolicoeur2021gotta,salimans2022progressive,liu2022pseudo,xiao2021tackling}. Given their massive success and attractive properties, we focus here on developing our framework using score-based diffusion models for video prediction, generation, and interpolation.

Our work makes the following contributions:
\begin{enumerate}
    \item A conditional video diffusion approach for video prediction and interpolation that yields state-of-the-art (SOTA) results.
    \item A conditioning procedure based on masking past and/or future frames in a blockwise manner giving a single model the ability to solve multiple video tasks: future/past prediction, unconditional generation, and interpolation. 
    \item A sliding window \emph{blockwise autoregressive} conditioning procedure to allow fast and coherent long-term generation (\autoref{fig:autoregression}).
    \item  A convolutional U-net neural architecture integrating recent developments with a conditional normalization technique we call SPAce-TIme-Adaptive Normalization (SPATIN) (\autoref{fig:our_net}).
\end{enumerate}

By conditioning on blocks of frames in the past and optionally blocks of frames even further in the future, we are able to better ensure that temporal dynamics are transferred across blocks of samples, i.e. our networks can learn \emph{implicit} models of spatio-temporal dynamics to inform frame generation. Unlike many other approaches, we do not have explicit model components for spatio-temporal derivatives or optical flow or recurrent blocks.  





\section{Masked conditional diffusion for video}
\label{MCVD:method}

Let $\x_0 \in \mathbb{R}^d$ be a sample from the data distribution $p_{\text{data}}$. A sample $\x_0$ can corrupted from $t=0$ to $t=T$ through the Forward Diffusion Process (FDP) with the following transition kernel:
\begin{equation}
  q_t(\bx_t | \bx_{t-1}) = \mathcal{N}(\bx_t;\sqrt{1-\beta_t}\bx_{t-1},\beta_t \bI).
\end{equation}

Furthermore, $\x_t$ can be sampled directly from $\x_0$ using the following accumulated kernel:
\begin{align}
  q_t(\bx_t|\bx_0) = \mathcal{N}(\bx_t; \sqrt{\balpha_t}\bx_0, (1-\bar\alpha_t)\bI) \label{eq:q_marginal_arbitrary_t} \implies \rvx_t = \sqrt{\balpha_t}\rvx_0 + \sqrt{1 - \balpha_t} \rvepsilon,
\end{align}
where $\bar\alpha_t = \prod_{s=1}^t (1 - \beta_s)$, and $\rvepsilon \sim \gN(\vzero, \rmI)$.

Generating new samples can be done by reversing the FDP  and solving the Reverse Diffusion Process (RDP) starting from Gaussian noise $\rvx_T$. It can be shown (\citet{song2020sde, ho2020ddpm}) that the RDP can be computed using the following transition kernel:
\begin{align*}
    &p_t(\bx_{t-1}|\bx_t,\bx_0) =  \mathcal{N}(\bx_{t-1}; \tilde\bmu_t(\bx_t, \bx_0), \tilde\beta_t \bI), \\
    \text{where}\quad \tilde\bmu_t(\bx_t, \bx_0) &= \frac{\sqrt{\bar\alpha_{t-1}}\beta_t }{1-\bar\alpha_t}\bx_0 + \frac{\sqrt{\alpha_t}(1- \bar\alpha_{t-1})}{1-\bar\alpha_t} \bx_t \quad \text{and} \quad
    \tilde\beta_t = \frac{1-\bar\alpha_{t-1}}{1-\bar\alpha_t}\beta_t .
    \numberthis
    \label{eq:RDP}
\end{align*}

Since $\x_0$ given $\x_t$ is unknown, it can be estimated using eq. (\ref{eq:q_marginal_arbitrary_t}): $\hat\rvx_0 = \big(\rvx_t - \sqrt{1 - \balpha_t}\rvepsilon\big)/\sqrt{\balpha_t}$, where $\rvepsilon_\theta(\rvx_t|t)$ estimates $\rvepsilon$ using a time-conditional neural network parameterized by $\theta$. This allows us to reverse the process from noise to data. The loss function of the neural network is:
\begin{align}
 L(\theta) = \Eb{t, \bx_0 \sim p_{\text{data}}, \bepsilon \sim \mathcal{N}(\vzero, \bI)}{ \left\| \bepsilon - \bepsilon_\theta(\sqrt{\bar\alpha_t} \bx_0 + \sqrt{1-\bar\alpha_t}\bepsilon \mid t) \right\|_2^2} . \label{eq:training_objective_simple}
\end{align}

Note that estimating $\epsilon$ is equivalent to estimating a scaled version of the score function (i.e., the gradient of the log density) of the noisy data:
\begin{align}
&\nabla_{\rvx_t} \log q_t (\rvx_t \mid \rvx_0) = -\frac{1}{1 - \balpha_t}(\rvx_t - \sqrt{\balpha_t} \rvx_0) = -\frac{1}{\sqrt{1 - \balpha_t}}\rvepsilon .
\end{align}

Thus, data generation through denoising depends on the score-function, and can be seen as noise-conditional score-based generation.

Score-based diffusion models can be straightforwardly adapted to video by considering the joint distribution of multiple continuous frames. While this is sufficient for unconditional video generation, other tasks such as video interpolation and prediction remain unsolved.

A conditional video prediction model can be approximately derived from the unconditional model using imputation~\citep{song2020sde}; indeed, the contemporary work of \cite{ho2022VDM} attempts to use this technique; however, their approach is based on an approximate conditional model.

\subsection{Video prediction}

We first propose to directly model the conditional distribution of video frames in the immediate future given past frames. Assume we have $p$ past frames $\rvp = \left\{ \rvp^i \right\}_{i=1}^{p}$ and $k$ current frames in the immediate future $\x_0 = \left\{ \x_0^i \right\}_{i=1}^{k}$. We condition the above diffusion models on the past frames to predict the current frames:
\begin{align}\label{eqn:vidpred}
 L_{\text{vidpred}}(\theta) = \Eb{t, [\rvp, \x_0] \sim p_{\text{data}}, \bepsilon \sim \mathcal{N}(\vzero, \bI)} {\Norm{ \bepsilon - \bepsilon_\theta(\sqrt{\balpha_t} \x_0 + \sqrt{1-\bar\alpha_t}\bepsilon \mid \rvp,  t)}^2} .
\end{align}

\begin{figure}[!htb]
    \centering
    \includegraphics[width=.75\textwidth,interpolate=false]{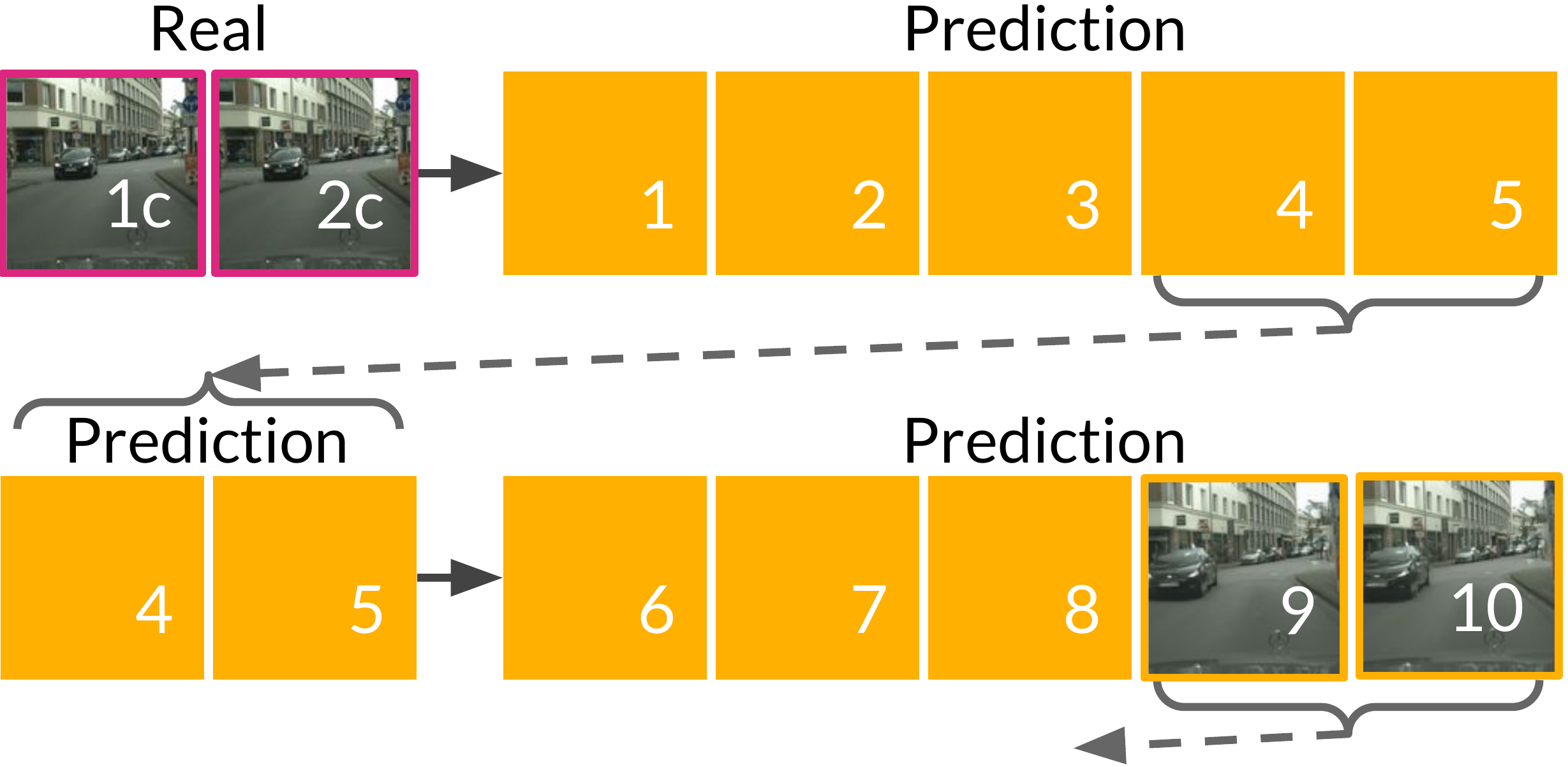}
    \caption{Blockwise autoregressive prediction with our model. }
    \label{fig:autoregression}
\end{figure}

\begin{figure}[!htb]
    \centering
    \includegraphics[width=\textwidth]{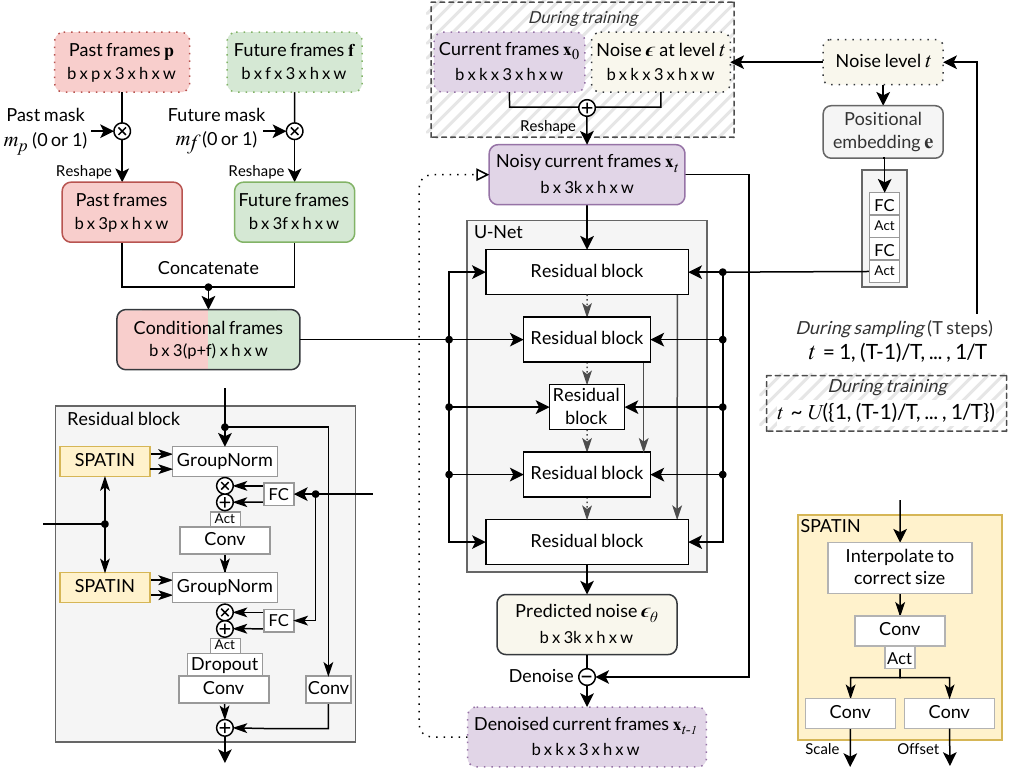}
    \caption{Noisy current frames are given to a U-Net whose residual blocks receive conditional information from past/future frames and noise-level. The output is the predicted noise in the current frames, which is used to denoise the current frames.}
    \label{fig:our_net}
\end{figure}

Given a model trained as above, video prediction for subsequent time steps can be achieved by blockwise autoregressively predicting current video frames conditioned on previously predicted frames (see \autoref{fig:autoregression}). We use variants of the network shown in \autoref{fig:our_net} to model $\bepsilon_\theta$ in \autoref{eqn:vidpred} here, and for \autoref{eqn:vidgen} and \autoref{eqn:vidgeneral} below.

\begin{figure}[!htb]
    \centering
    \includegraphics[width=\textwidth,interpolate=false]{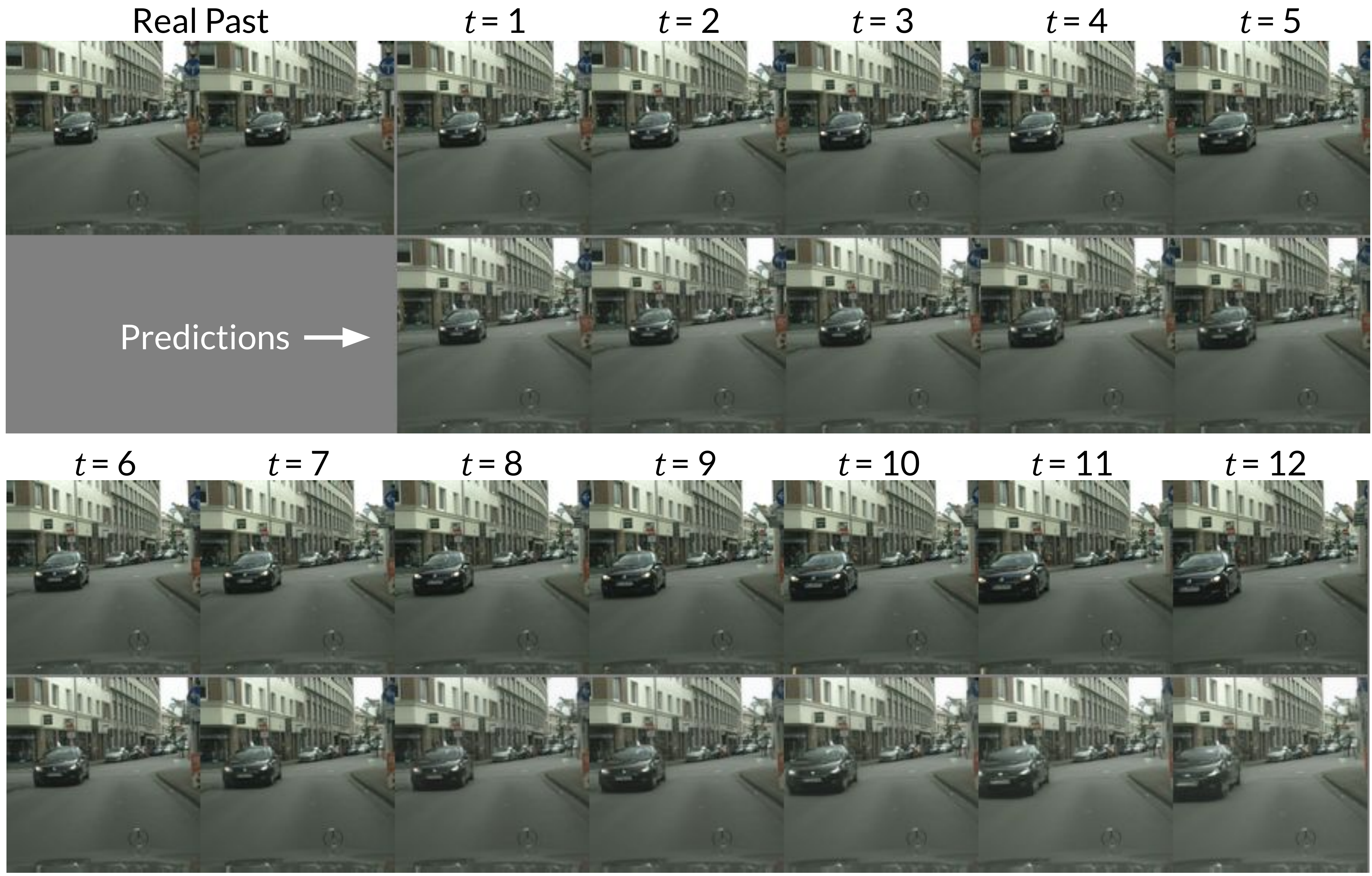}
    \caption{This figure shows our block autoregressive strategy where the top row and third row are ground truth, and the second and fourth rows show the blockwise autoregressively generated frames using our approach.}
    \label{fig:autoregression2}
\end{figure}

\subsection{Video prediction + generation}

Our approach above allows video prediction, but not unconditional video generation. As a second approach, we extend the same framework to video generation by masking (zeroing-out) the past frames with probability $p_{\text{mask}}=\nicefrac{1}{2}$ using binary mask $m_p$. The network thus learns to predict the noise added without any past frames for context. Doing so means that we can perform conditional as well as unconditional frame generation, i.e., video prediction and generation with the same network. This leads to the following loss ($\mathcal{B}$ is the Bernouilli distribution):
\begin{align}\label{eqn:vidgen}
 L_{\text{vidgen}}(\theta) = \Eb{t, [\rvp, \x_0] \sim p_{\text{data}}, \bepsilon \sim \mathcal{N}(\vzero, \bI), m_p \sim \mathcal{B}(p_{\text{mask}})}{ \left\| \bepsilon - \bepsilon_\theta(\sqrt{\bar\alpha_t} \x_0 + \sqrt{1-\bar\alpha_t}\bepsilon \mid m_p\rvp,  t) \right\|^2} .
\end{align}
We hypothesize that this dropout-like \citep{srivastava14a} approach will also serve as a form of regularization, improving the model's ability to perform predictions conditioned on the past. We see positive evidence of this effect in our experiments -- see the MCVD past-mask model variants in \Cref{tab:bair_pred,tab:SMMNIST_ablation} versus without past-masking. Note that random masking is used only during training.
\subsection{Video prediction + generation + interpolation}

We now have a design for video prediction and generation, but it still cannot perform video interpolation nor past prediction from the future. As a third and final approach, we show how to build a general model for solving all four video tasks. Assume we have $p$ past frames, $k$ current frames, and $f$ future frames $\rvf = \left\{ \rvf^i \right\}_{i=1}^{f}$. We randomly mask the $p$ past frames with probability $p_{mask}=\nicefrac{1}{2}$, and similarly randomly mask the $f$ future frames with the same probability (but sampled separately). Thus, future or past prediction is when only future or past frames are masked. Unconditional generation is when both past and future frames are masked. Video interpolation is when neither past nor future frames are masked. The loss function for this general video machinery is:
\begin{align}\label{eqn:vidgeneral}
 L(\theta) = \Eb{t, [\rvp, \rvx_0, \rvf] \sim p_{\text{data}}, \bepsilon \sim \mathcal{N}(\vzero, \bI), ( m_p, m_f ) \sim \mathcal{B}(p_{\text{mask}})}{ \left\| \bepsilon - \bepsilon_\theta(\sqrt{\bar\alpha_t} \x_0 + \sqrt{1-\bar\alpha_t}\bepsilon \mid m_p\rvp, m_f\rvf,  t) \right\|^2} .
\end{align}

The three video tasks and our solutions are visualized in \Cref{fig:tasks}.

\subsection{Network architecture}
\label{MCVD:arch}

For our denoising network we use a U-net architecture \citep{ronneberger2015u, honari2016recombinator, salimans2017pixelcnn++} combining the improvements from \citet{song2020sde} and \citet{dhariwal2021diffusion}. This architecture uses a mix of 2D convolutions \citep{fukushima1982neocognitron}, multi-head self-attention \citep{cheng2016long}, and adaptive group-norm \citep{wu2018group}. We use positional encodings of the noise level ($t \in [0,1]$) and process it using a transformer style positional embedding:
%
\vspace{-.2cm}
\begin{equation}
\textbf{e}(t)=\left[ \ldots,\cos \left(t c^{\frac{-2d}{D}} \right),  \sin \left(t c^{\frac{-2d}{D}} \right)
, \ldots  \right]^{\mathrm{T}},
\end{equation}
where $d=1,\ldots,D/2$ , $D$ is the number of dimensions of the embedding, and $c=10000$. This embedding vector is passed through a fully connected layer, followed by an activation function and another fully connected layer. Each residual block has an fully connected layer that adapts the embedding to the correct dimensionality.

To provide $\x_t$, $\rvp$, and $\rvf$ to the network, we separately concatenate the past/future frames and the noisy current frames in the channel dimension. 
The concatenated noisy current frames are directly passed as input to the network. Meanwhile, the concatenated conditional frames are passed through an embedding that influences the conditional normalization akin to SPatially-Adaptive (DE)normalization (SPADE) \citep{park2019semantic} to account for the effect of time/motion, we call this approach SPAce-TIme-Adaptive Normalization (SPATIN).
%

%
%
%
%
%
%
%
We also try passing the direct concatenation of the conditional and noisy current frames as the input. In our experiments below, we show some results with SPATIN and some with concatenation (concat).
%
%
%
%
For simple video prediction with \autoref{eqn:vidpred}, we experimented with 3D convolutions and 3D attention. However, this requires an exorbitant amount of memory, and we found no benefit in using 3D layers over 2D layers at the same memory (i.e. the biggest model that fits in 4 GPUs). We also tried and found no benefit from gamma noise \citep{nachmani2021ddgm}, L1 loss, and F-PNDM sampling \citep{liu2022pseudo}.


\section{Related work}
\label{MCVD:related}


Score-based diffusion models have been used for image editing~\citep{meng2022sdedit, saharia2021palette, nichol2021glide} and our approach to video generation might be viewed as an analogy to classical image inpainting, but in the temporal dimension. 
The GLIDE or Guided Language to Image Diffusion for Generation and Editing approach of ~\cite{nichol2021glide} uses CLIP-guided diffusion for image editing, while Denoising Diffusion Restoration Models (DDRM)~\cite{kawar2022Ddrm} additionally condition on a corrupted image to restore the clean image. Adversarial variants of score-based diffusion models have been used to enhance quality \citep{jolicoeur2020adversarial} or speed \citep{xiao2021tackling}.

Contemporary work to our own such as that of ~\cite{ho2022VDM} and \cite{yang2022ResidualVideoDiffusion} also examine video generation using score-based diffusion models. However, the Video Diffusion Models (VDMs) work of~\citet{ho2022VDM} approximates conditional distributions using a gradient method for conditional sampling from their unconditional model formulation. In contrast, our approach directly works with a conditional diffusion model, which we obtain through masked conditional training, thereby giving us the exact conditional distribution as well as the ability to generate unconditionally. Their experiments focus on: a) unconditional video generation, and b) text-conditioned video generation, whereas our work focuses primarily on predicting future video frames from the past, using our masked conditional generation framework.  The Residual Video Diffusion (RVD) of ~\cite{yang2022ResidualVideoDiffusion} is only for video prediction, and it uses a residual formulation to generate frames autoregressively one at a time. Meanwhile, ours directly models the conditional frames to generate multiple frames in a block-wise autoregressive manner.

\begin{figure}
    \includegraphics[width=\textwidth]{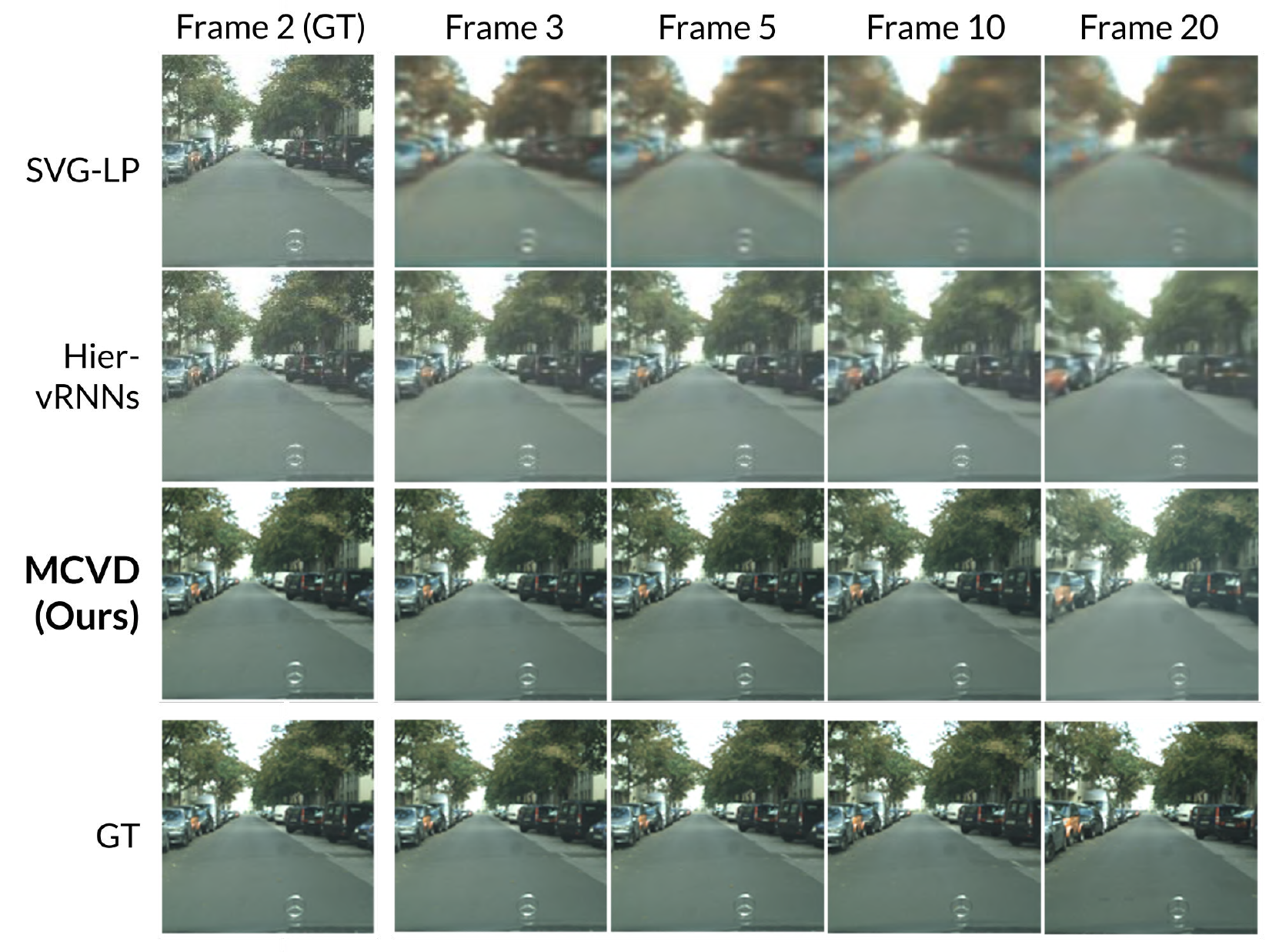}
    \caption{Comparing future prediction methods on Cityscapes:  SVG-LP (Top Row), Hier-vRNNs (Second Row), Our Method (Third Row), Ground Truth (Bottom Row). Frame 2, a ground truth conditioning frame is shown in first column, followed by frames: 3, 5, 10 and 20 generated by each method vs the ground truth at the bottom. 
    }
    \label{fig:RNN_compare}
\end{figure}

Recurrent neural network (RNN) techniques were early candidates for modern deep neural architectures for video prediction and generation. Early work combined RNNs with a stochastic latent variable (SV2P) \cite{babaeizadeh2018stochastic} and was optimized by variational inference. 
The stochastic video generation (SVG) approach of \cite{denton2018SVGLP} learned both prior and a per time step latent variable model, which influences the dynamics of an LSTM at each step. The model is also trained in a manner similar to a variational autoencoder, i.e., it was another form of variational RNN (vRNN). To address the fact that vRNNs tend to lead to blurry results, \cite{Castrejn2019ImprovedCV} (Hier-vRNN) increased the expressiveness of the latent distributions using a hierarchy of latent variables. We compare qualitative result of SVG and Hier-vRNN with the concat variant of our MCVD method in \Cref{fig:RNN_compare}. Other vRNN-based models include SAVP~\cite{Lee2018StochasticAV}, SRVP~\cite{franceschi2020stochastic}, SLAMP~\cite{akan2021slamp}. 

The well known Transformer paradigm \citep{vaswani2017attention} from natural language processing has also been explored for video. The Video-GPT work of \cite{yan2021videogpt} applied an autoregressive GPT style \citep{brown2020language} transformer to the codes produced from a VQ-VAE \citep{van2017neural}. 
The Video Transformer work of \cite{weissenborn2019scaling} models video using 3-D spatio-temporal volumes without linearizing positions in the volume. They examine local self-attention over small non-overlapping sub-volumes or 3D blocks. This is done partly to accelerate computations on TPU hardware.
%
%
Their work also observed that the peak signal-to-noise ratio (PSNR) metric and the mean-structural similarity (SSIM) metrics \citep{wang2004image} were developed for images, and have serious flaws when applied to videos. PSNR prefers blurry videos and SSIM does not correlate well to perceptual quality. Like them, we focus on the recently proposed Frechet Video Distance (FVD) \citep{unterthiner2018towards}, computed over entire videos and which is sensitive to visual quality, temporal coherence, and diversity of samples. \citet{rakhimov2020latent} (LVT) used transformers to predict the dynamics of video in latent space. \citet{le2021ccvs} (CCVS) also predict in latent space, that of an adversarially trained autoencoder, and also add a learnable optical flow module.

Generative Adversarial Network (GAN) based approaches to video generation have also been studied extensively. \citet{vondrick2016generating} proposed an early GAN architecture for video, using a spatio-temporal CNN.
\citet{villegas2017decomposing} proposed a strategy for separating motion and content into different pathways of a convolutional LSTM based encoder-decoder RNN.
\citet{saito2017temporal} (TGAN) predicted a sequence of latents using a temporal generator, and then the sequence of frames from those latents using an image generator. TGANv2~\cite{saito2020train} improved its memory efficiency.
MoCoGAN~\cite{tulyakov2018mocogan} explored style and content separation, but within a CNN framework. \citet{yushchenko2019markov} used the MoCoGAN framework by re-formulating the video prediction problem as a Markov Decision Process (MDP).
FutureGAN~\cite{aigner2018futuregan} used spatio-temporal 3D convolutions in an encoder decoder architecture, and elements of the progressive GAN~\cite{karras2018progressive} approach to improve image quality. TS-GAN~\cite{munoz2021temporal} facilitated information flow between consecutive frames. TriVD-GAN~\cite{luc2020transformation} proposes a novel recurrent unit in the generator to handle more complex dynamics, while DIGAN~\cite{yu2022generating} uses implicit neural representations in the generator.

Video interpolation was the subject of a flurry of interest in the deep learning community a number of years ago \citep{niklaus2017video, jiang2018super, xue2019video, bao2019depth}. However, these architectures tend to be fairly specialized to the interpolation task, involving optical flow or motion field modelling and computations. Frame interpolation is useful for video compression; therefore, many other lines of work have examined interpolation from a compression perspective. However, these architectures tend to be extremely specialized to the video compression task \citep{yang2020learning}.  

The Cutout approach of \cite{devries2017improved} has examined the idea of cutting out small continuous regions of an input image, such as small squares.
Dropout \citep{srivastava14a} at the FeatureMap level was proposed and explored under the name of SpatialDropout in \cite{tompson2015efficient}.
Input Dropout \citep{de2020input} has been examined in the context of dropping different channels of multi-modal input imagery, such as the dropping of the RGB channels or depth map channels during training, then using the model without one of the modalities during testing, e.g. in their work they drop the depth channel.

Regarding our block-autoregressive approach, previous video prediction models were typically either 1) non-recurrent: predicting all $n$ frames simultaneously with no way of adding more frames (most GAN-based methods), or 2) recurrent in nature, predicting 1 frame at a time in an autoregressive fashion. The benefit of the non-recurrent type is that you can generate videos faster than 1 frame at a time while allowing for generating as many frames as needed. The disadvantage is that it is slower than generating all frames at once, and takes up more memory and compute at each iteration. Our model finds a sweet spot in between in that it is block-autoregressive: generating $k < n$  frames at a time recurrently to finally obtain $n$ frames.

%
%
%

%
\section{Experiments}
\label{MCVD:exp}
\subsection{Tasks}

We show state-of-the-art results on three video tasks:
\begin{enumerate}
    \item \textbf{Video prediction} i.e. prediction of future frames conditioned on past frames,
    \item \textbf{Video generation} i.e. unconditional generation of video frames, and
    \item \textbf{Video interpolation} i.e. prediction of intermediate frames conditioned on past and future frames
\end{enumerate}

\begin{figure}[!tbh]
    \begin{minipage}[b]{0.46\linewidth}
    \centering
    \includegraphics[width=\textwidth]{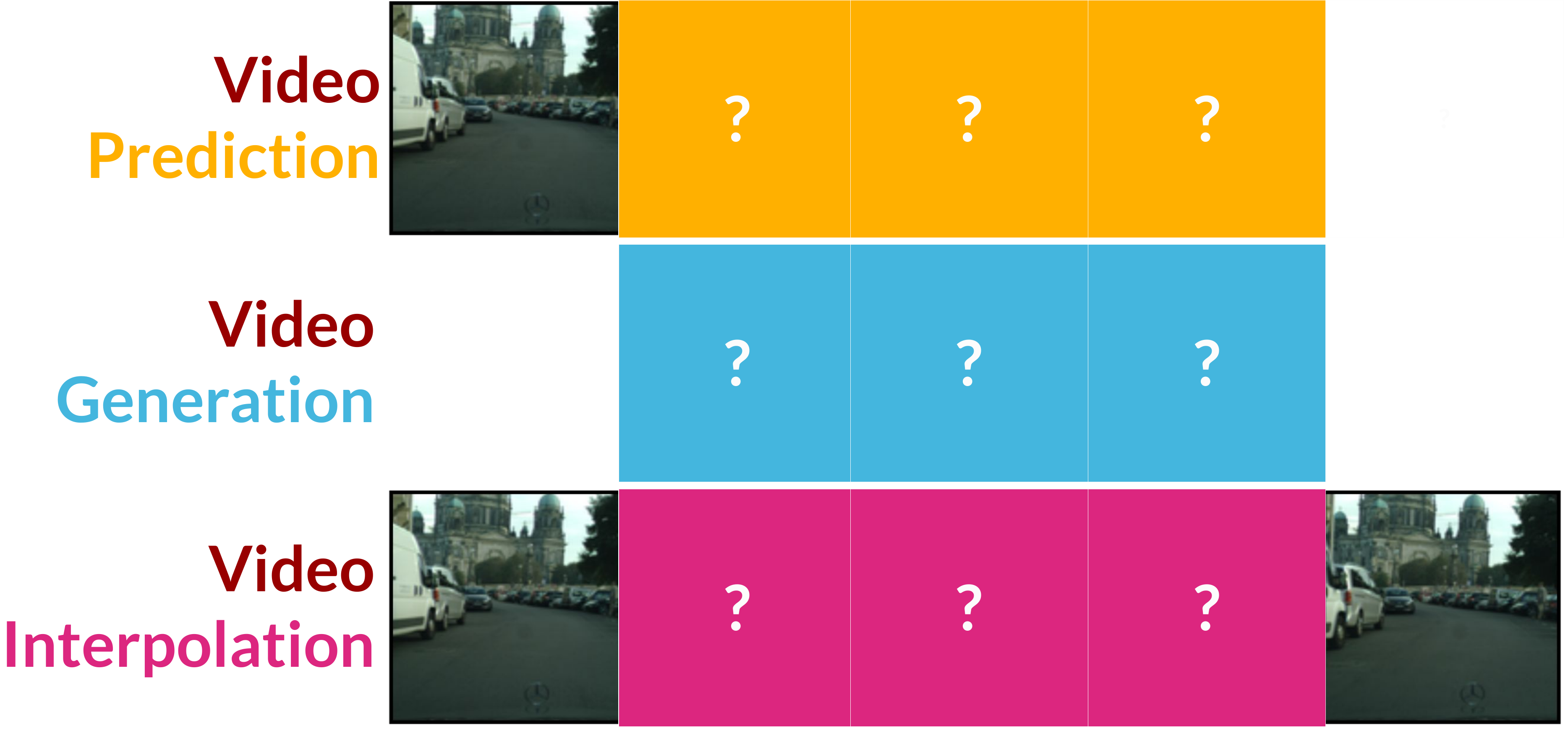}
    \subcaption{Tasks}
    \end{minipage}
    \hfill
    \begin{minipage}[b]{0.52\linewidth}
    \centering
    \includegraphics[width=\textwidth]{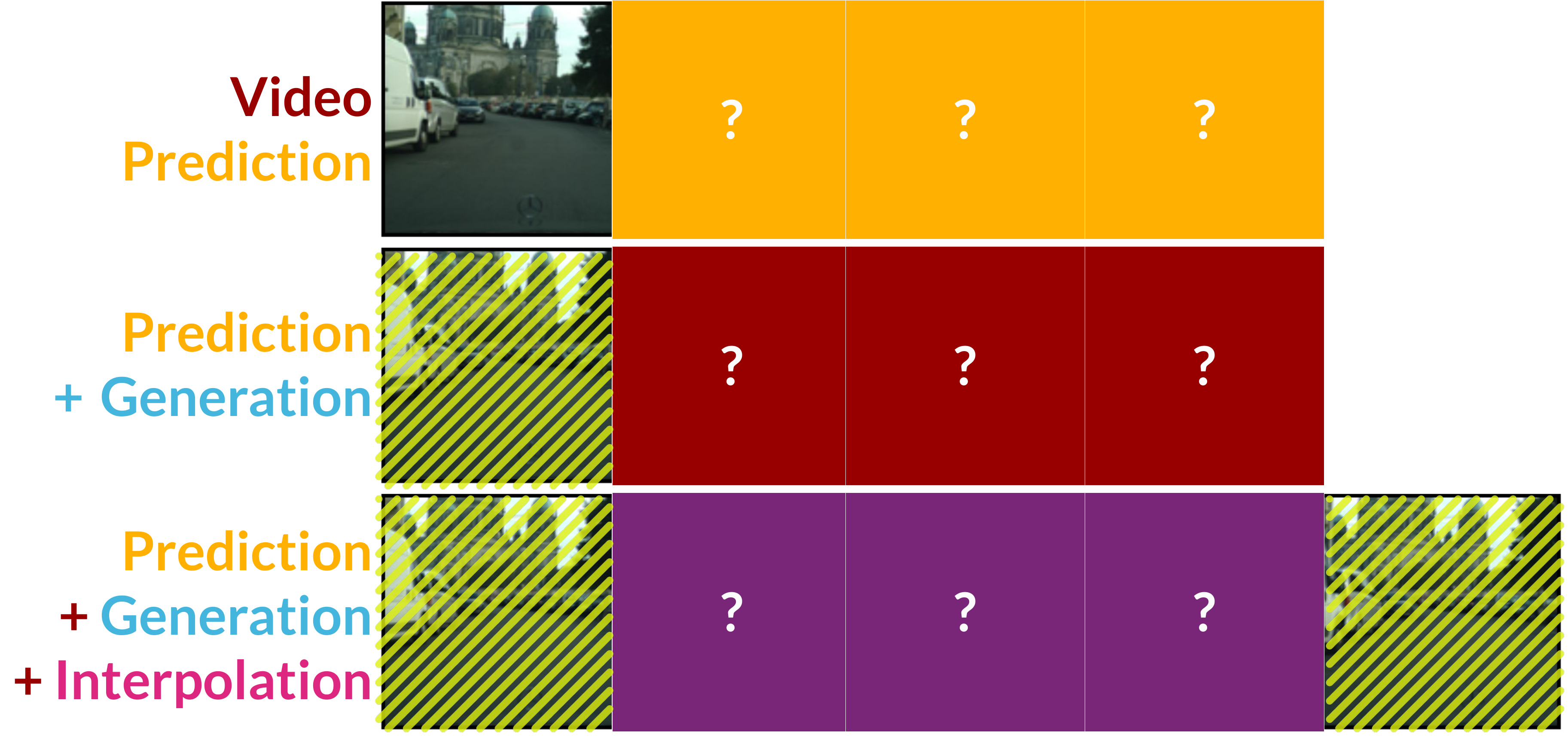}
    \subcaption{Masked solutions (green is mask)}
    \end{minipage}
    \caption{Visualization of video (a) tasks and (b) our proposed masked solutions.}
    \label{fig:tasks}
\end{figure}

These tasks and our proposed solutions are visualized in \Cref{fig:tasks}.

\subsection{Datasets}

Our choice of datasets is in order of progressive difficulty: 1) SMMNIST: black-and-white digits; 2) KTH: grayscale single-humans; 3) BAIR: color, multiple objects, simple scene; 4) Cityscapes: color, natural complex natural driving scene; 5) UCF101: color, 101 categories of natural scenes. We process these datasets similarly to prior works. For Cityscapes, each video is center-cropped, then resized to $128\times128$. For UCF101, each video clip is center-cropped at 240×240 and resized to 64×64, taking care to maintain the train-test splits. We generate 128x128 images for Cityscapes and 64x64 images for the other datasets. 
 
\begin{enumerate}
    \item \textbf{Video prediction} : We show the results of our video prediction experiments on test data that was never seen during training in Tables \ref{tab:SMMNIST}~-~\ref{tab:bair_pred} 
    for Cityscapes~\citep{cordts2016cityscapes}, Stochastic Moving MNIST (SMMNIST)~\citep{denton2018SVGLP,srivastava2015unsupervised}, KTH~\citep{schuldt2004recognizing} and  BAIR~\citep{ebert2017self} respectively.
    \item \textbf{Video generation} : We present unconditional generation results for BAIR in \autoref{tab:bair_gen} and UCF-101~\citep{soomro2012ucf101} in \autoref{tab:ucf_gen}.
    \item \textbf{Video interpolation} : We show interpolation results for SMMNIST, KTH, and BAIR in \autoref{tab:interp}.
\end{enumerate}

\subsection{Training details}

Unless otherwise specified, we set the mask probability to 0.5 when masking was used. For sampling, we report results using the sampling methods DDPM~\citep{ho2020ddpm} or DDIM~\citep{song2020ddim} with only 100 sampling steps, though our models were trained with 1000, to make sampling faster. We observe that the metrics are generally better using DDPM than DDIM (except for UCF-101). Using 1000 sampling steps could yield better results.

Note that all our models are trained to predict only 4-5 current frames at a time, unlike other models that predict $\ge$10. We use these models to then autoregressively predict longer sequences for prediction or generation. This was done in order to fit the models in our GPU memory budget. Despite this disadvantage, we find that our MCVD models perform better than many previous SOTA methods. 

We provide some additional information regarding model size, memory requirements, batch size and computation times in \autoref{tab:memory}. This is followed by additional results and visualizations for SMMNIST, KTH, BAIR, UCF-101 and Cityscapes.

\begin{table}[!ht]
\small
\caption{Compute used. ``steps'' indicates the checkpoint with the best approximate FVD, ``GPU hours'' is the total training time up to ``steps''.}
\label{tab:memory}
\def\arraystretch{1.1}
\begin{tabular}{l|rrrrrrr}
\toprule
Dataset, & params & CPU mem & batch & GPU & GPU mem & steps & GPU \\ 
model & & (GB) & size & & (GB) & & hours \\ \hline
SMMNIST concat & 27.9M & 3.6 & 64 & Tesla V100 & 14.5 & 700000  & 78.9 \\
SMMNIST spatin & 53.9M & 3.3 & 64 & RTX 8000 & 23.4  & 140000  & 39.7 \\
KTH concat & 62.8M & 3.2 & 64 & Tesla V100 & 21.5  & 400000  & 65.7 \\
KTH spatin & 367.6M & 8.9 & 64 & A100 & 145.9  & 340000  & 45.8 \\
BAIR concat & 251.2M & 5.1 & 64 & Tesla V100 & 76.5  & 450000  & 78.2 \\
BAIR spatin & 328.6M & 9.2 & 64 & A100 & 86.1  & 390000  & 50.0 \\
Cityscapes concat & 262.1M & 6.2 & 64 & Tesla V100 & 78.2  & 900000  & 192.83 \\
Cityscapes spatin & 579.1M & 8.9 & 64 & A100 & 101.2  & 650000  & 96.0 \\
UCF concat & 565.0M & 8.9 & 64 & Tesla V100 &  100.1 & 900000  & 183.95 \\
UCF spatin & 739.4M & 8.9 & 64 & A100 & 115.2  & 550000  & 79.5 \\
\bottomrule
\end{tabular}
\end{table}

\subsection{Metrics}

As mentioned earlier, we primarily use the FVD metric for comparison across models as FVD measures both fidelity and diversity of the generated samples. Previous works compare Frechet Inception Distance (FID)~\citep{heusel2017gans} and Inception Score (IS)~\citep{salimans2016improved}, adapted to videos by replacing the Inception network with a 3D-convolutional network that takes video input. FVD is computed similarly to FID, but using an I3D network trained on the huge video dataset Kinetics-400. We also report PSNR and SSIM.

We tried to add the older FID and IS metrics (as opposed to the newer FVD metric which we used above) for UCF-101 as proposed in \citet{saito2017temporal}, but we
had difficulties integrating the chainer~\citep{tokui2019chainer} based implementation of these metrics into our PyTorch \citep{paszke2019pytorch} code base.

\section{Results}

Our MCVD concat past-future-mask and past-mask results are of particular interest as they yield SOTA results across many benchmark configurations. 

\subsection{Video prediction}

\begin{threeparttable}[b]
\centering
\caption{Video prediction on Cityscapes ($128\times128$), 2 past frames, predicting 28.
}
\label{tab:cityscape}
\setlength{\tabcolsep}{2pt}
{\def\arraystretch{0.95}
\begin{tabular}{l|r|lll}
\toprule
\textbf{Cityscapes} ($128\times128$) [2 $\rightarrow$ 28; trained on $k$] & $k$ & FVD$\downarrow$      & LPIPS$\downarrow$ & SSIM$\uparrow$   \\ \hline
SVG-LP~\citep{denton2018SVGLP}     & 10 & 1300.26              & 0.549 $\pm$ 0.06  & 0.574 $\pm$ 0.08 \Tstrut \\
vRNN 1L ~\citep{Castrejn2019ImprovedCV} & 10 & \ \ 682.08           & 0.304 $\pm$ 0.10  & 0.609 $\pm$ 0.11 \\
Hier-vRNN ~\citep{Castrejn2019ImprovedCV} & 10 & \ \ 567.51           & 0.264 $\pm$ 0.07  & 0.628 $\pm$ 0.10 \\
GHVAE~\citep{wu2021greedy}              & 10 & \ \ 418.00 & 0.193 $\pm$ 0.014 & \textbf{0.740} $\pm$ 0.04 \\
\textbf{MCVD} spatin past-mask (Ours)                  &  \textbf{5} &              \ \ 184.81      &     0.121 $\pm$ 0.05              & 0.720 $\pm$ 0.11 \\ 
\textbf{MCVD} concat past-mask (Ours)                  &  \textbf{5} &              \ \ \textbf{141.31}      &     \textbf{0.112} $\pm$ 0.05              & 0.690 $\pm$ 0.12 \\ 
\bottomrule
\end{tabular}
}
\end{threeparttable}

\begin{table}[b!]
\caption{Video prediction results on SMMNIST ($64\times64$) for 10 predicted frames conditioned on 5 past frames. We predicted 10 trajectories per real video, and report the average FVD and maximum SSIM, averaged across 256 test videos. }
\vspace{-0.8em}
\label{tab:SMMNIST}
{\def\arraystretch{1.00}
\begin{tabular}{l|r|ll}
\toprule
\textbf{SMMNIST} [5 $\rightarrow$ 10; trained on $k$] & $k$ & \textbf{FVD$\downarrow$}      & \textbf{SSIM$\uparrow$}       \\ \hline
SVG \citep{denton2018SVGLP}
& 10 & 90.81                & 0.688     \Tstrut           \\
vRNN 1L \citep{Castrejn2019ImprovedCV}
& 10 & 63.81                & 0.763                \\
Hier-vRNN  \citep{Castrejn2019ImprovedCV}
& 10 & 57.17                & 0.760                \\
\textbf{MCVD} concat (Ours) & \textbf{5} & 25.63 & \textbf{0.786}    \\
\textbf{MCVD} spatin (Ours) & \textbf{5} & \textbf{23.86} & 0.780    \\
\bottomrule
\end{tabular}
}
\end{table}

\begin{table}[b!]
\caption{Video prediction results on KTH ($64\times64$), predicting 30/40 frames from 10 past frames using models trained to predict $k$ frames at a time. All models test on 256 videos.}
\label{tab:KTH_appendix}
\vspace{-0.8em}
{\def\arraystretch{1.00}
\begin{tabular}{l|rc|lll}
\toprule
\textbf{KTH} [10 $\rightarrow$ $pred$; trained on $k$] & $k$ & $pred$ & FVD$\downarrow$    & PSNR$\uparrow$  & SSIM$\uparrow$  \\ \hline
SV2P~\citep{babaeizadeh2018stochastic} & 10 & 30 & 636 $\pm$ 1  & 28.2  & 0.838              \Tstrut \\
SVG-LP~\citep{denton2018SVGLP}      & 10 & 30 & 377 $\pm$ 6 & 28.1 & 0.844 \\
SAVP~\citep{Lee2018StochasticAV}         & 10 & 30 & 374 $\pm$ 3 & 26.5  & 0.756 \\
\textbf{MCVD} spatin (Ours) & \textbf{5} & 30 & 323 $\pm$ 3 & 27.5   &  0.835    \\
\textbf{MCVD} concat past-future-mask (Ours) & \textbf{5} & 30 & 294.9 & 24.3   &  0.746    \\
SLAMP~\citep{akan2021slamp}              & 10 & 30 & 228 $\pm$ 5 & 29.4 & 0.865 \\
SRVP~\citep{franceschi2020stochastic}    & 10 & 30 & 222 $\pm$ 3 & 29.7 & 0.870 \\
\hline
Struct-vRNN~\citep{minderer2019unsupervised}            & 10 & 40   & 395.0  & 24.29 & 0.766 \Tstrut \\
\textbf{MCVD} concat past-future-mask (Ours) & \textbf{5} & 40 & 368.4 & 23.48   &  0.720    \\
\textbf{MCVD} spatin (Ours)                                    & \textbf{5} & 40   & 331.6 $\pm$ 5  &     26.40  &  0.744    \\
\textbf{MCVD} concat (Ours)                                    & \textbf{5} & 40   & 276.6 $\pm$ 3 &     26.20  &  0.793    \\
SV2P time-invariant~\citep{babaeizadeh2018stochastic} & 10 & 40   & 253.5 & 25.70 & 0.772 \\
SV2P time-variant ~\citep{babaeizadeh2018stochastic}  & 10 & 40   & 209.5 & 25.87 & 0.782 \\
SAVP~\citep{Lee2018StochasticAV}                        & 10 & 40   & 183.7 & 23.79 & 0.699 \\
SVG-LP~\citep{denton2018SVGLP}                     & 10 & 40   & 157.9 & 23.91 & 0.800 \\
SAVP-VAE~\citep{Lee2018StochasticAV}                    & 10 & 40   & 145.7 & 26.00 & 0.806 \\
Grid-keypoints~\citep{gao2021accurate}                  & 10 & 40   & 144.2 & 27.11 & 0.837 \\
\bottomrule
\end{tabular}
}
\end{table}

\begin{threeparttable}[!ht]
	\caption{Video prediction results on  BAIR ($64\times64$) conditioning on $p$ past frames and predicting $pred$ frames in the future, using models trained to predict $k$ frames at at time.
	}
	\label{tab:bair_pred_appendix}
	\centering
 {\def\arraystretch{0.95}
	\begin{tabular}{l|crc|rcc}
	\toprule
		BAIR [past (p) $\rightarrow$ pred (pr) ; trained on k] & p & k & pr & FVD$\downarrow$ & PSNR$\uparrow$ & SSIM$\uparrow$ \\
		\cmidrule(){1-7}
        LVT \citep{rakhimov2020latent} & 1 & 15 & 15 & 125.8 & -- & -- \\
        DVD-GAN-FP \citep{clark2019adversarial} & 1 & 15 & 15 & 109.8 & -- & -- \\
		\textbf{MCVD} spatin (Ours) & 1 & \textbf{5} & 15 & 103.8 & 18.8 & 0.826  \\
        TrIVD-GAN-FP \citep{luc2020transformation} & 1 & 15 & 15 & 103.3 & -- & -- \\
		VideoGPT \citep{yan2021videogpt} & 1 & 15 & 15 & 103.3 & -- & -- \\
		CCVS \citep{le2021ccvs} & 1 & 15 & 15 & 99.0 & -- & -- \\
	    \textbf{MCVD} concat (Ours) & 1 & \textbf{5} & 15  & 98.8 & 18.8 & 0.829 \\
		\textbf{MCVD} spatin past-mask (Ours) & 1 & \textbf{5} & 15 & 96.5 & 18.8 & 0.828 \\
		\textbf{MCVD} concat past-mask (Ours) & 1 & \textbf{5} & 15 & 95.6 & 18.8 & 0.832  \\
        Video Transformer \citep{weissenborn2019scaling} & 1 & 15 & 15 & 94-96\tnote{a} & -- & -- \\
		FitVid \citep{babaeizadeh2021fitvid} & 1 & 15 & 15 & 93.6 & -- & -- \\
		\textbf{MCVD} concat past-future-mask (Ours) & 1 & \textbf{5} & 15 & \textbf{89.5} & 16.9 & 0.780  \\
		\cmidrule(){1-7}
        SAVP \citep{Lee2018StochasticAV} & 2 & 14 & 14 & 116.4 & -- & -- \\
		\textbf{MCVD} spatin (Ours) & 2 & \textbf{5} & 14 & 94.1 & 19.1 & 0.836  \\
		\textbf{MCVD} spatin past-mask (Ours) & 2 & \textbf{5} & 14 & 90.5 & 19.2 & 0.837  \\
		\textbf{MCVD} concat (Ours)  & 2 & \textbf{5} & 14 & 90.5 & 19.1 & 0.834  \\
		\textbf{MCVD} concat past-future-mask (Ours) & 2 & \textbf{5} & 14 & 89.6 & 17.1 & 0.787  \\
		\textbf{MCVD} concat past-mask (Ours) & 2 & \textbf{5} & 14 & \textbf{87.9} & 19.1 & 0.838  \\
		\cmidrule(){1-7}
		SVG-LP~\citep{akan2021slamp}    & 2    & 10   & 28   & 256.6          &      --            & 0.816    \\
        SVG~\citep{akan2021slamp}.   & 2 & 12 & 28 & 255.0 & 18.95 & 0.8058 \\
        SLAMP \citep{akan2021slamp} & 2 & 10 & 28 & 245.0 & 19.7 & 0.818  \\
        SRVP \citep{franceschi2020stochastic} & 2 & 12 & 28 & 162.0 & 19.6 & 0.820 \\
		WAM \citep{jin2020exploring} & 2 & 14 & 28 & 159.6 & 21.0 & 0.844 \\
        SAVP~\citep{Lee2018StochasticAV} & 2 & 12 & 28 & 152.0 & 18.44 & 0.7887 \\
        vRNN 1L~\cite{Castrejn2019ImprovedCV}                                & 2    & 10   & 28   & 149.2          &        --          & 0.829   \\
        SAVP~\citep{Lee2018StochasticAV} & 2    & 10   & 28   & 143.4          &        --          & 0.795  \\
        Hier-vRNN ~\citep{Castrejn2019ImprovedCV}                             & 2    & 10   & 28   & 143.4           &         --         & 0.822   \\
		
		\textbf{MCVD} spatin (Ours) & 2 & \textbf{5} & 28 & 132.1 & 17.5 & 0.779  \\
		\textbf{MCVD} spatin past-mask (Ours) & 2 & \textbf{5} & 28 & 127.9 & 17.7 & 0.789 \\
		\textbf{MCVD} concat (Ours)  & 2 & \textbf{5} & 28 & 120.6 & 17.6 & 0.785  \\
		\textbf{MCVD} concat past-mask (Ours) & 2 & \textbf{5} & 28 & 119.0 & 17.7 & 0.797 \\
		\textbf{MCVD} concat past-future-mask (Ours) & 2 & \textbf{5} & 28 & \textbf{118.4} & 16.2 & 0.745 \\
		\bottomrule
	\end{tabular}
    \label{tab:bair_pred}
    }
    \begin{tablenotes}
    \item[a] 94 on only the first frames, 96 on all subsquences of test frames
    \end{tablenotes}
\end{threeparttable}

\subsection{Video generation}

\begin{table}[!ht]
	\caption{Unconditional generation of BAIR video frames.}
	\label{tab:bair_gen}
	\centering
 {\def\arraystretch{1.00}
	\begin{tabular}{l|c|c}
	\toprule
		\textbf{BAIR} ($64\times64$) [0 $\rightarrow$ $pred$; trained on 5] & $pred$ & FVD$\downarrow$ \\
		\cmidrule(){1-3}
		\textbf{MCVD} spatin past-mask (Ours) & 16 & 267.8 \\
		\textbf{MCVD} concat past-mask (Ours) & 16 & \textbf{228.5} \\
		\cmidrule(){1-3}
		\textbf{MCVD} spatin past-mask (Ours) & 30 & 399.8 \\
		\textbf{MCVD} concat past-mask (Ours) & 30 & \textbf{348.2} \\
		\bottomrule
	\end{tabular}
    }
\end{table}

\begin{table}[!ht]
	\centering
	\caption{Unconditional generation of UCF-101 video frames.}
    {\def\arraystretch{1.00}
	\begin{tabular}{l|r|c}
	\toprule
		\textbf{UCF-101} ($64\times64$)  [0 $\rightarrow$ 16; trained on $k$] & $k$ & FVD$\downarrow$ \\
		\cmidrule(){1-3}
		MoCoGAN-MDP \citep{yushchenko2019markov} & 16 & 1277.0 \\
		\textbf{MCVD} concat past-mask (Ours) & \textbf{4} & 1228.3  \\
		TGANv2 \citep{saito2020train} & 16 & 1209.0  \\
		\textbf{MCVD} spatin past-mask (Ours) & \textbf{4} & 1143.0 \\
		DIGAN \citep{yu2022generating} & 16 & \textbf{655.0} \\
		\bottomrule
	\end{tabular}
	\label{tab:ucf_gen}
    }
\end{table}

\subsection{Video interpolation}

\begin{table}[!htb]
\caption{Video Interpolation results (64 $\times$ 64). Given $p$ past + $f$ future frames $\rightarrow$ interpolate $k$ frames. Reporting average of the best metrics out of $n$ trajectories per test sample. $\downarrow\!(p\!+\!f)$ and $\uparrow\!k$ is harder. We used MCVD spatin past-mask for SMMNIST and KTH, and MCVD concat past-future-mask for BAIR. We also include results on SMMNIST for a "pure" model trained without any masking.}
\label{tab:interp}
\setlength{\tabcolsep}{2pt}
\centering
\resizebox{\textwidth}{!}{
\def\arraystretch{1.1}
\begin{tabular}{l||ccc|cc||ccc|cc||ccc|cc}
\toprule
 & \multicolumn{5}{c||}{\textbf{SMMNIST} ($64\times64$)} & \multicolumn{5}{c||}{\textbf{KTH} ($64\times64$)} & \multicolumn{5}{c}{\textbf{BAIR} ($64\times64$)} \\
 & $p\!\!+\!\!f$ & $k$ & $n$ & PSNR$\uparrow$      & SSIM$\uparrow$
 & $p\!\!+\!\!f$ & $k$ & $n$ & PSNR$\uparrow$      & SSIM$\uparrow$
 & $p\!\!+\!\!f$ & $k$ & $n$ & PSNR$\uparrow$      & SSIM$\uparrow$ \\ \hline
SVG-LP & 18 & 7 & 100 & 13.543 & 0.741 & 18 & 7 & 100 & 28.131 & 0.883 & 18 & 7 & 100 & 18.648 & 0.846  \Tstrutsmall  \\
FSTN & 18 & 7 & 100 & 14.730 & 0.765 & 18 & 7 & 100 & 29.431 & 0.899 & 18 & 7 & 100 & 19.908 & 0.850 \\
SepConv & 18 & 7 & 100 & 14.759 & 0.775 & 18 & 7 & 100 & 29.210 & 0.904 & 18 & 7 & 100 & 21.615 & 0.877    \\
SuperSloMo & 18 & 7 & 100 & 13.387 & 0.749 & 18 & 7 & 100 & 28.756 & 0.893 & -- & -- & -- & -- & -- \\
SDVI full & 18 & 7 & 100 & 16.025 & 0.842 & 18 & 7 & 100 & 29.190 & 0.901 & 18 & 7 & 100 & 21.432 & 0.880    \\
SDVI & 16 & 7 & 100 & 14.857 & 0.782 & 16 & 7 & 100 & 26.907 & 0.831 & 16 & 7 & 100 & 19.694 & 0.852    \\\hline
\multirow{3}{*}{\textbf{MCVD} (Ours)} & \textbf{10} & \textbf{10} & 100 & 20.944 & 0.854 & \textbf{15} & \textbf{10} & 100 & 34.669 & 0.943 & \textbf{4} & \textbf{5} & 100 & 25.162 & 0.932  \Tstrut  \\
& \textbf{10} & \textbf{5} & \textbf{10} & 27.693 & 0.941 & \textbf{15} & \textbf{10} & \textbf{10} & 34.068 & 0.942 & \textbf{4} & \textbf{5} & \textbf{10} & 23.408 & 0.914   \\
& \multicolumn{3}{c|}{pure} & 18.385 & 0.802 & \textbf{10} & \textbf{5} & \textbf{10} & 35.611 & 0.963 &  &  &  &  & \\
\bottomrule
\end{tabular}
}
\end{table}

We compare our MCVD method with previous methods of video interpolation on the standard datasets of SMMNIST, KTH, and BAIR: SVG-LP~\cite{denton2018SVGLP}, FSTN~\cite{lu2017flexible}, SepConv~\cite{niklaus2017video}, SuperSloMo~\cite{jiang2018super}, and SDVI~\cite{xu2020stochastic}.

\subsection{Ablation studies}
\label{MCVD:ablation}

We were able to draw the following conclusions from ablation studies, expanded upon below:
\begin{itemize}
    \item A model trained on multiple tasks performs better than one trained on individual tasks. This shows that solving tasks like interpolation helps in solving more complex tasks like prediction and generation.
    \item In general, the concat variant performs better than the spatin variant.
\end{itemize}

In \autoref{tab:bair_pred} we compare models that use concatenated raw pixels as input to U-Net blocks (concat) to SPATIN variants. We also compare no-masking to past-masking variants, i.e. models which are only trained predict the future vs. models which are regularized by being trained for prediction and unconditional generation. It can be seen that our model works across different choices of past frames and generates better quality for shorter videos. This is expected from models of this kind. Moreover, it can be seen that the model trained on the two tasks of Prediction and Generation (i.e., the models with past-mask) performs better than the model trained only on Prediction!

\begin{figure}[!htb]
\includegraphics[width=\linewidth]{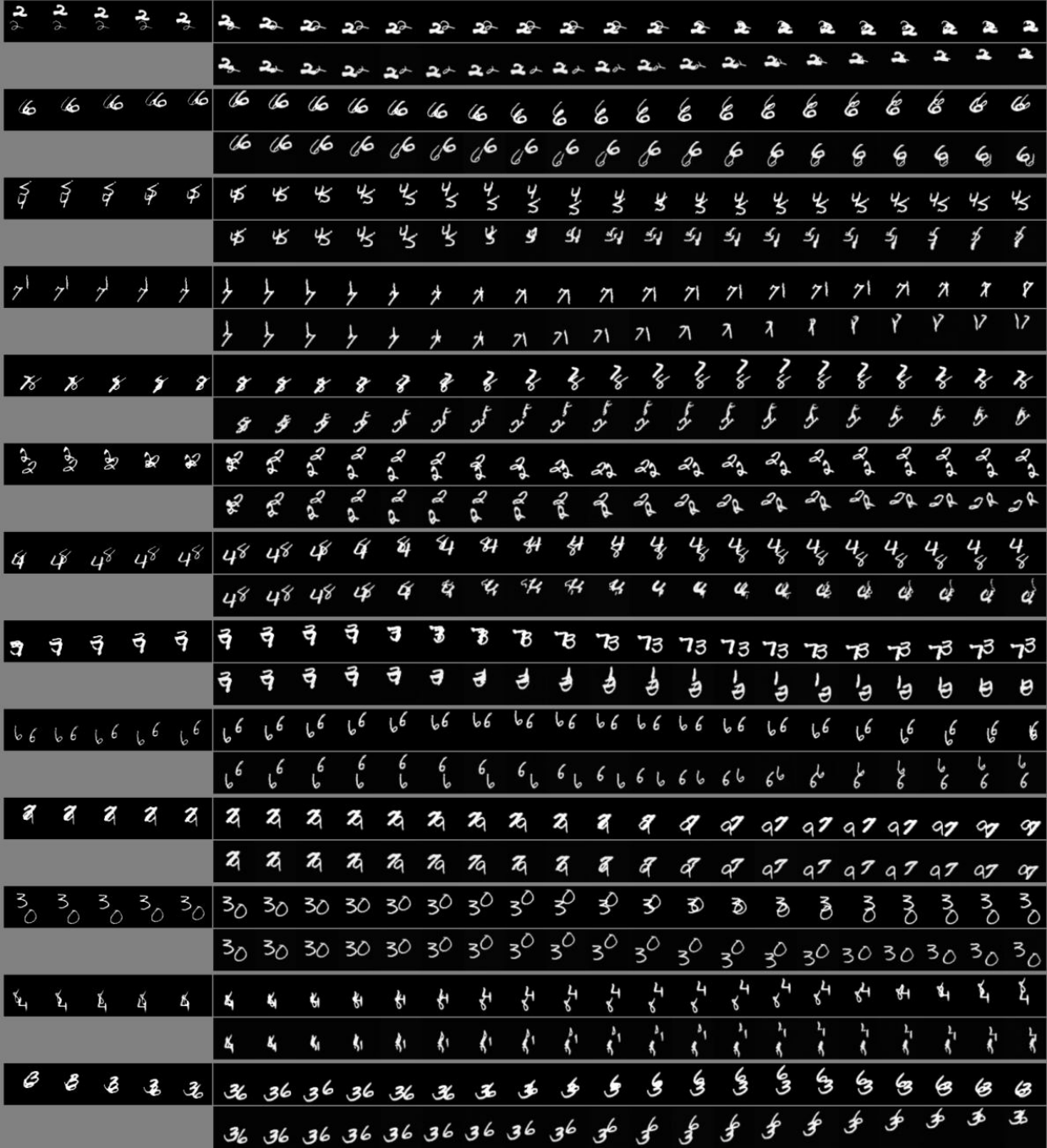}
\caption{SMMNIST 5 $\to$ 10, trained on 5 (prediction). For each sample, top row is real ground truth, bottom is predicted by our MCVD model.}
\label{fig:SMMNIST_2c5_images_1}
\end{figure}

In \autoref{tab:SMMNIST_ablation}, we provide results for an ablation study using SMMNIST on the different design choices: concat vs concat past-future-mask vs spatin vs spatin future-mask vs spatin past-future-mask. In \autoref{fig:SMMNIST_2c5_images_1} we provide some visual results for SMMNIST.

\begin{table}[!ht]
\caption{Results on the SMMNIST evaluation, conditioned on 5 past frames, predicting 10 frames using models trained to predict 5 frames at a time.}
\label{tab:SMMNIST_ablation}
\begin{tabular}{l|lllll}
\toprule
\textbf{SMMNIST} [5 $\rightarrow$ 10; trained on 5] & FVD$\downarrow$  & PSNR$\uparrow$  & SSIM$\uparrow$ & LPIPS$\downarrow$ & MSE$\downarrow$ \\ \hline
\textbf{MCVD} spatin future-mask        & 44.14 $\pm$ 1.73  &     16.31  &  0.758 & 0.141 & 0.027    \\
\textbf{MCVD} spatin past-future-mask & 36.12 $\pm$ 0.63 & 16.15 & 0.748 & 0.146 & 0.027  \\
\textbf{MCVD} concat                  & 25.63 $\pm$ 0.69  &     17.22  &  0.786 & 0.117 & 0.024  \\
\textbf{MCVD} spatin                  & 23.86 $\pm$ 0.67  &     17.07  &  0.785 & 0.129 & 0.025  \\
\textbf{MCVD} concat past-future-mask & \textbf{20.77} $\pm$ 0.77  &     16.33  & 0.753  & 0.139  &  0.028  \\
\bottomrule
\end{tabular}
\end{table}

It can be seen that concat is, in general, better than spatin. It can also be seen that the past-future-mask variant, which is a general model capable of all three tasks, performs better at the individual tasks than the models trained only on the individual task. This was demonstrated in \autoref{tab:bair_pred} as well. This shows that the model gains helpful insights while generalizing to all three tasks, which it does not while training only on the individual task.

We conducted preliminary experiments with a larger number of frames. Since the models with a larger number of frames were bigger, we could only run them for a shorter time with a smaller batch size than the smaller models. In general, we found that larger models did not substantially improve the results. We attribute this to the fact that using more frames means that the model should be given more capacity, but we could not increase it due to our computational budget constraints. We emphasize that our method works very well with fewer computational resources.

Examining these results we remark that we have SOTA performance for prediction on SMMNIST, BAIR and the challenging Cityscapes evaluation. Our Cityscapes model yields an FVD of 145.5, whereas the best previous result of which we are aware is 418. The quality of our Cityscapes results are illustrated visually in \autoref{fig:teaser} and \autoref{fig:autoregression}. While our completely unconditional generation results are strong, we note that when past masking is used to regularize future predicting models, we see clear performance gains in \autoref{tab:bair_pred}. Finally, in \autoref{tab:interp} we see that our interpolation results are SOTA by a wide margin, across experiments on SMMNIST, KTH and BAIR -- even compared to architectures much more specialized for interpolation.

It can be seen that our proposed method generates better quality videos, even though it was trained on a shorter number of frames than other methods. It can also be seen that training on multiple tasks using random masking improves the quality of generated frames than training on the individual tasks.

\section{Conclusion}

We have shown how to obtain SOTA video prediction and interpolation results with randomly masked conditional video diffusion models using a relatively simple architecture. We found that past-masking was able to improve performance across all model variants and configurations tested. 
We believe our approach may pave the way forward toward high quality larger-scale video generation.

\subsection{Limitations} Videos generated by these models are still small compared to real movies, and they can still become blurry or inconsistent when the number of generated frames is very large. Our unconditional generation results on the highly diverse UCF-101 dataset are still far from perfect. More work is clearly needed to scale these models to larger datasets with more diversity and with longer duration video. As has been the case in many other settings, simply using larger models with many more parameters is a strategy that is likely to improve the quality and flexibility of these models -- we were limited to 4 GPUs for our work here. There is also a need for faster sampling methods capable of maintaining quality over time. %
%
%

Given our strong interpolation results, conditional diffusion models which generate skipped frames could make it possible to generate much longer, but consistent video through a strategy of first generating sparse distant frames in a block, followed by an interpolative diffusion step for the missing frames. 
%

\subsection{Broader Impacts} 
High-quality video generation is potentially a powerful technology that could be used by malicious actors for applications such as creating fake video content. Our formulation focuses on capturing the distributions of real video sequences. High-quality video prediction could one day find use in applications such as autonomous vehicles, where the cost of errors could be high. Diffusion methods have shown great promise for covering the modes of real probability distributions. In this context, diffusion-based techniques for generative modelling may be a promising avenue for future research where the ability to capture modes properly is safety critical. Another potential point of impact is the amount of computational resources being spent for these applications involving the high fidelity and voluminous modality of video data. We emphasize the use of limited resources in achieving better or comparable results. Our submission provides evidence for more efficient computation involving fewer GPU hours spent in training time.


\section{Qualitative results}
\label{MCVD:videos}


Below are prediction/generation examples from Stochastic Moving MNIST, KTH, BAIR, UCF-101, and Cityscapes. For more examples including videos, please visit \url{https://mask-cond-video-diffusion.github.io}

\newpage

\subsection{Stochastic Moving MNIST}

\clearpage

\subsection{KTH}

\begin{figure}[!htb]
\includegraphics[width=.95\linewidth]{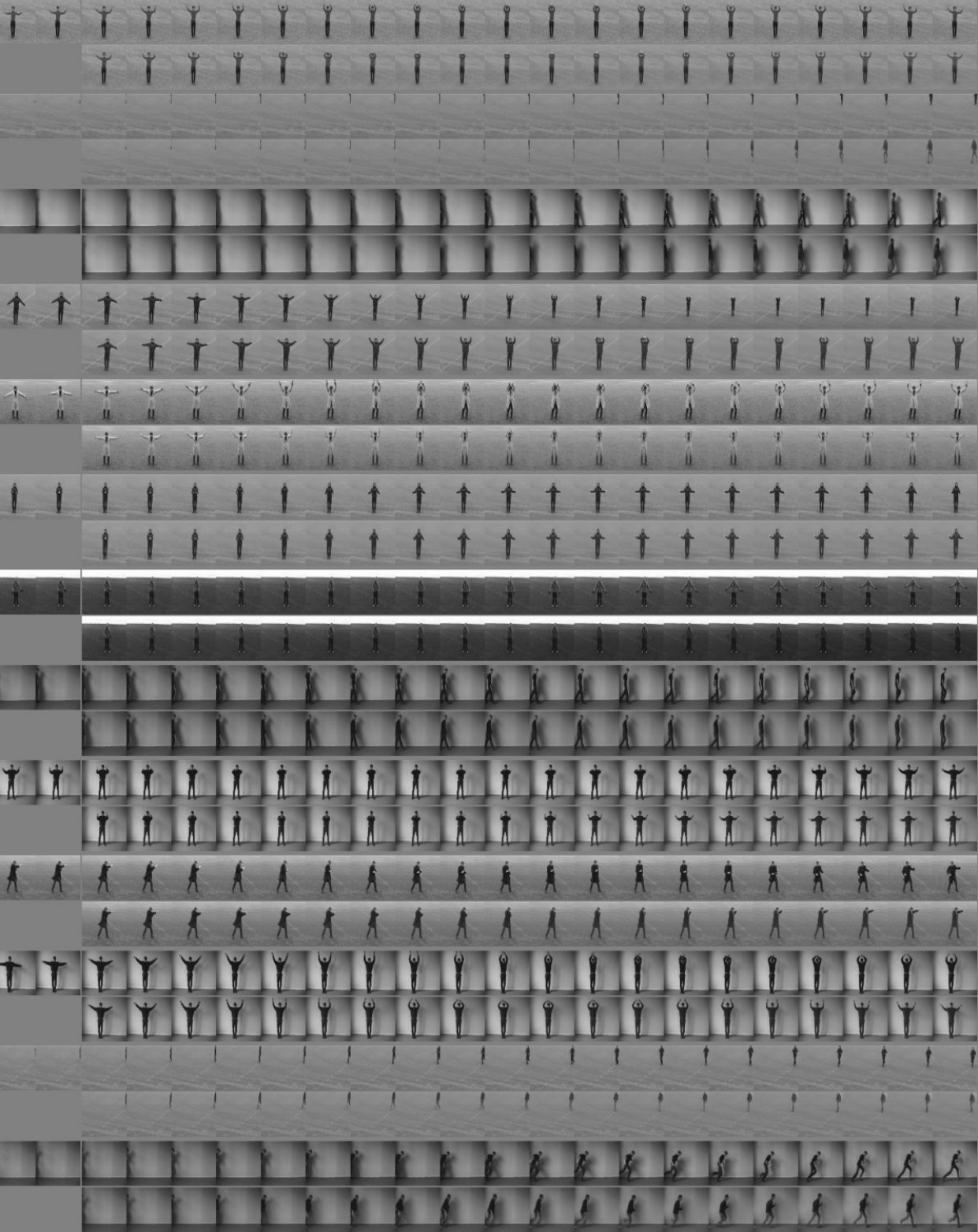}
\caption{KTH 5 $\to$ 20, trained on 5 (prediction). For each sample, top row is real ground truth, bottom is predicted by our MCVD model. (We show only 2 conditional frames here)}
\label{fig:KTH_5c10_images}
\end{figure}

\clearpage

\subsection{BAIR}

\begin{figure}[!htb]
\includegraphics[width=\linewidth]{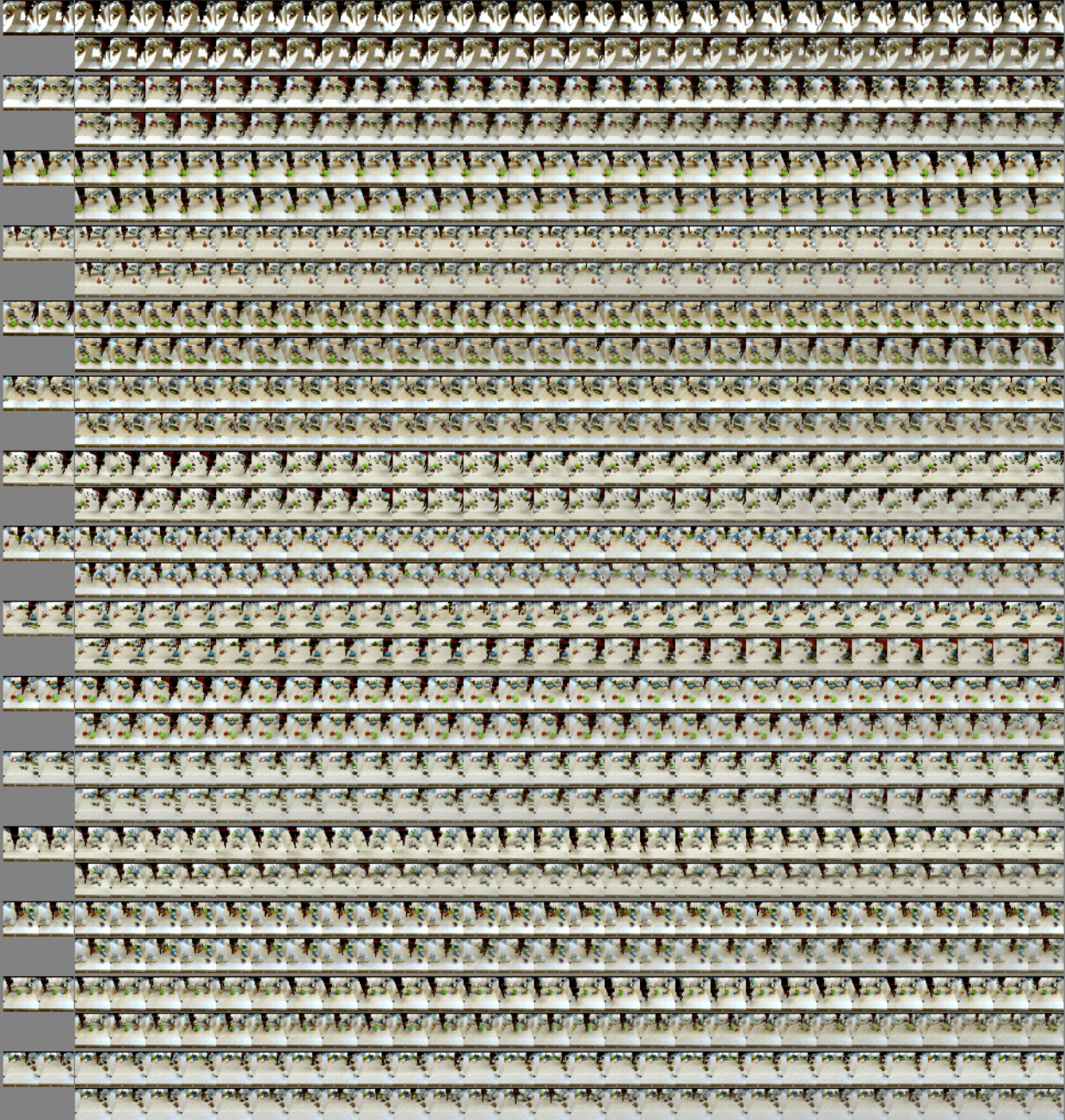}
\caption{BAIR 2 $\to$ 28, trained on 5 (prediction). For each sample, top row is real ground truth, bottom is predicted by our MCVD model.}
\label{fig:BAIR_2c2_images}
\end{figure}

\clearpage

\subsection{UCF-101}

\begin{figure}[!htb]
\includegraphics[width=\linewidth]{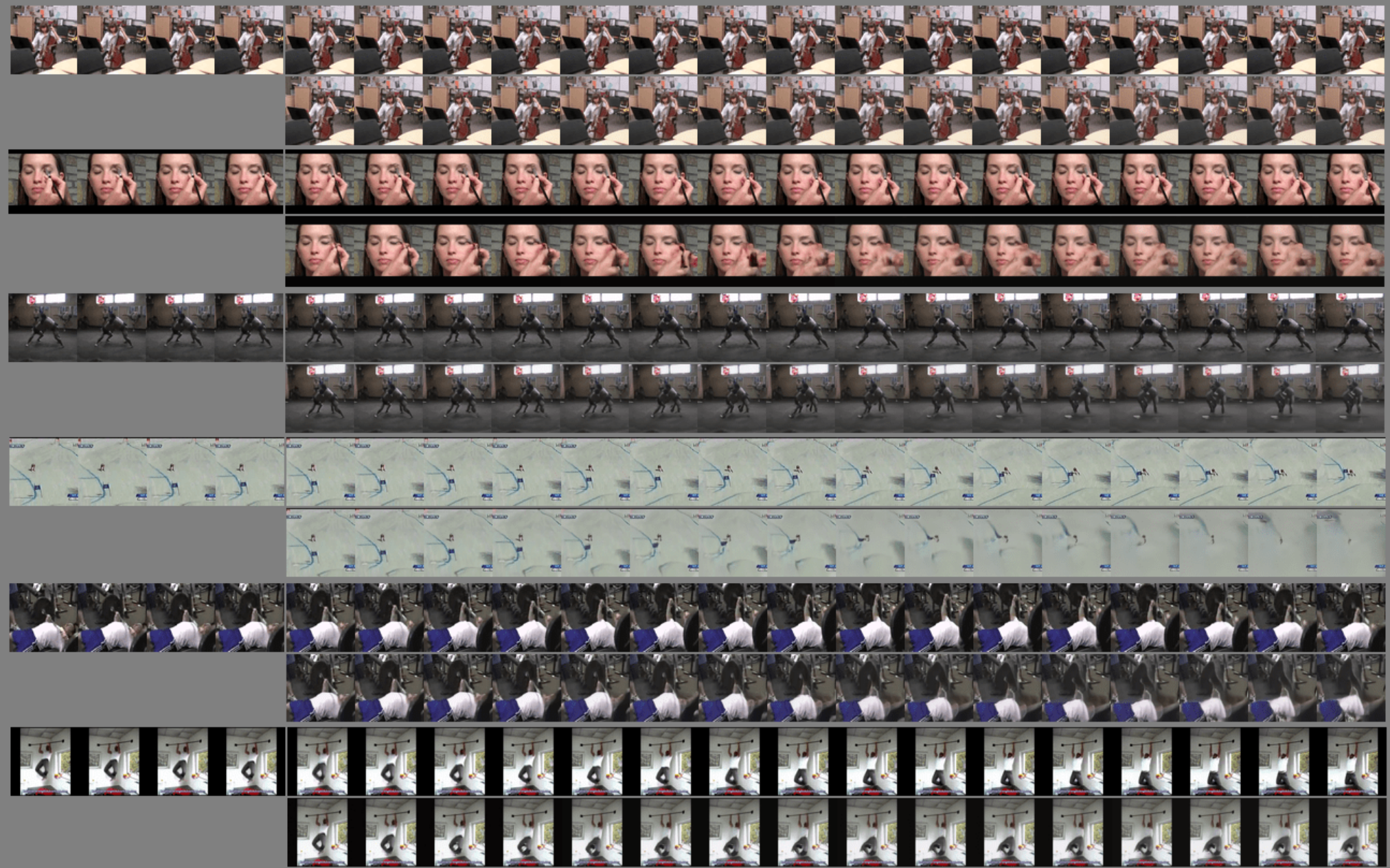}
\caption{UCF-101 4 $\to$ 16, trained on 4 (prediction). For each sample, top row is real ground truth, bottom is predicted by our MCVD model.}
\label{fig:UCF101_4c4_images_pred}
\end{figure}

\begin{figure}[!htb]
\includegraphics[width=.98\linewidth]{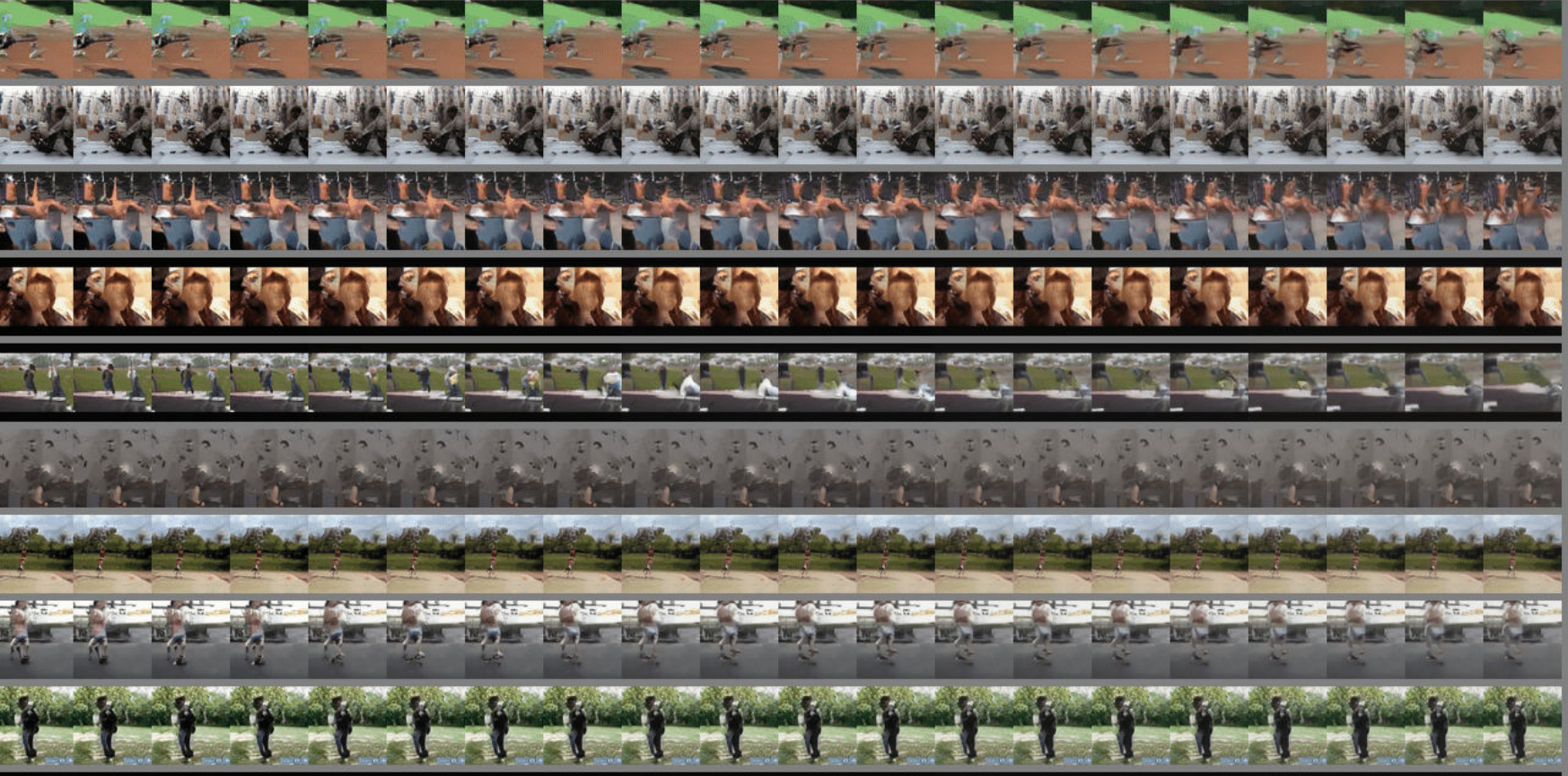}
\caption{UCF-101  0 $\to$ 4 (generation)}
\label{fig:UCF101_4c4_images_gen}
\end{figure}

\clearpage

\subsection{Cityscapes}

Here we provide some examples of future frame prediction for Cityscapes sequences conditioning on two frames and predicting the next 7 frames.

\begin{figure}[!ht]
    \centering
    \includegraphics[width=\textwidth]{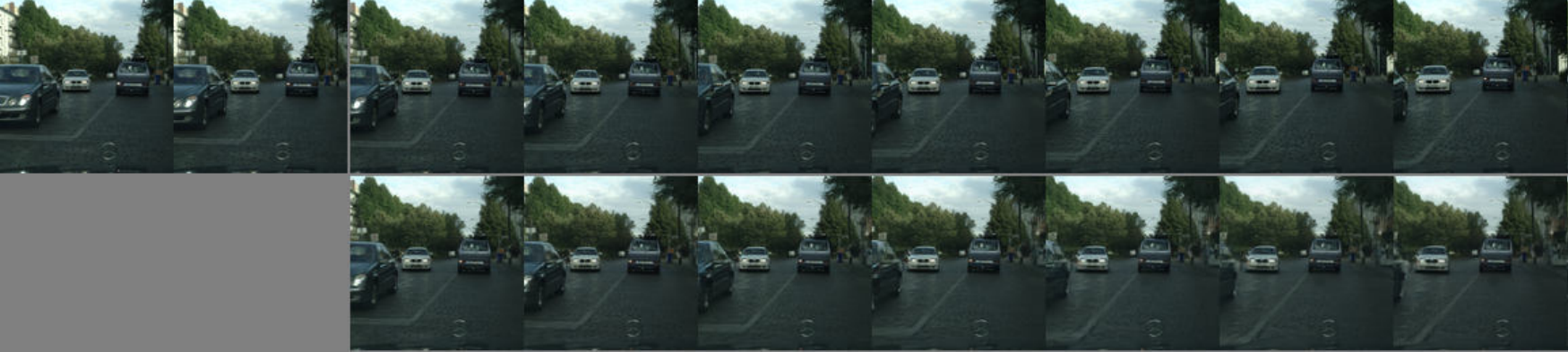}
    \includegraphics[width=\textwidth]{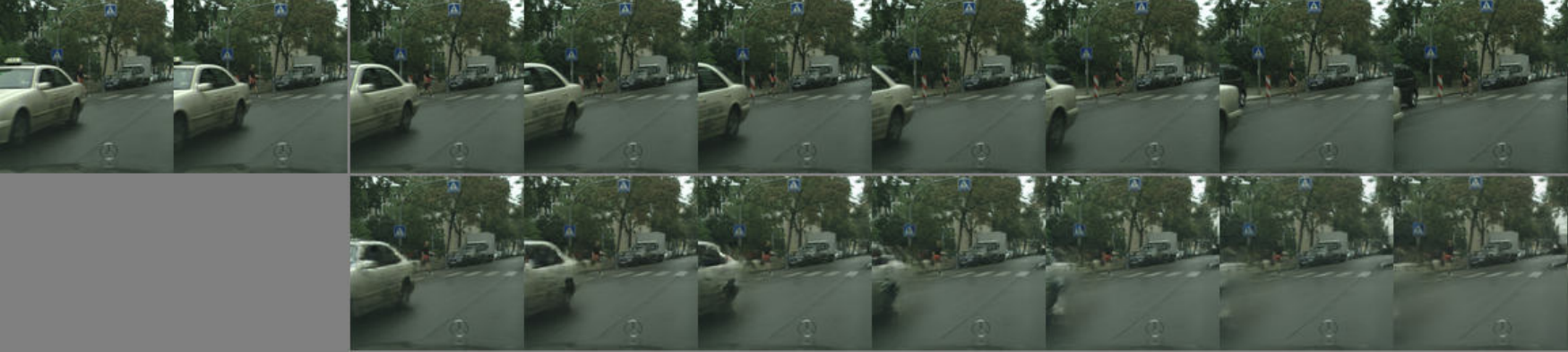}
    \includegraphics[width=\textwidth]{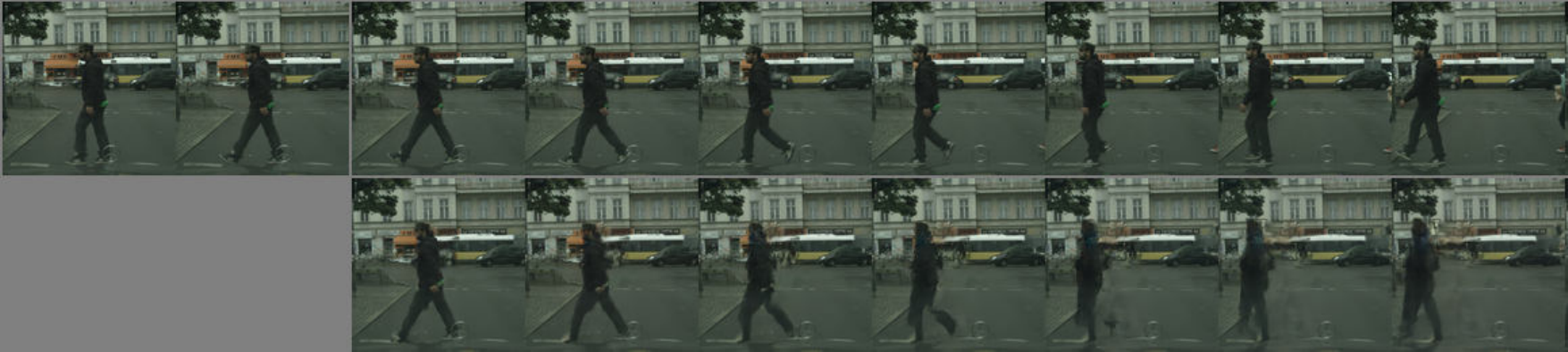}
    \includegraphics[width=\textwidth]{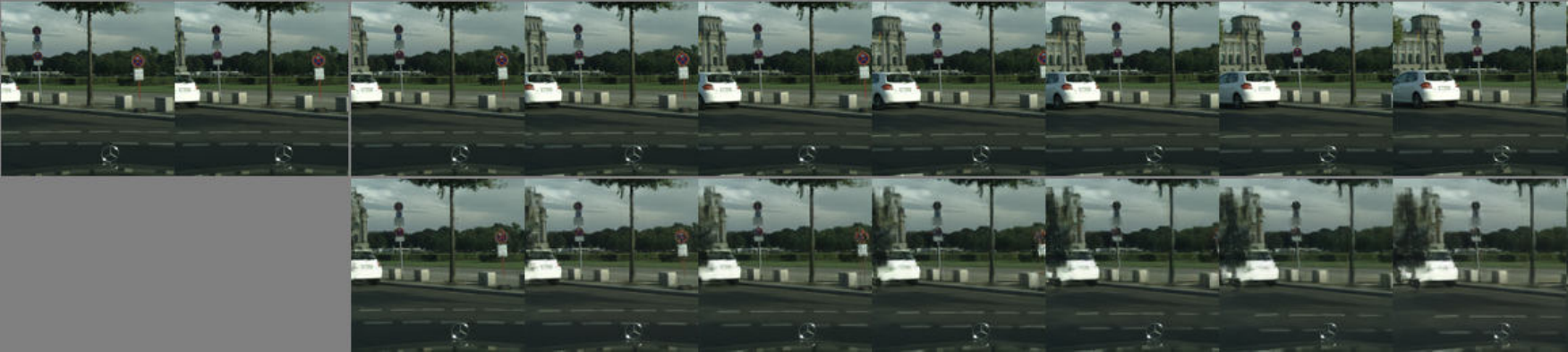}
    \caption{Cityscapes: 2 $\to$ 7, trained on 5 (prediction); Conditioning on the two frames in the top left corner of each block of two rows of images, we generate the next 7 frames. The top row is the true frames, bottom row contains the generated frames. We use the MCVD concat model variant.}
    \label{fig:more_city}
\end{figure}

%% file: Conclusion.tex
\anglais

\counterwithin{figure}{chapter}
\counterwithin{table}{chapter}

\chapter{Conclusion}

The articles presented in this thesis reside at the intersection of generative modeling and computer vision, focusing on inventing novel ideas and combinations of techniques to advance the state of the art in image generation, improved 3D animation, and enhanced video prediction, generation, and interpolation. Our endeavors of the diverse projects within this thesis have embraced a multifaceted notion of improvement, encompassing aspects such as quality enhancement as well as computational efficiency. Leveraging the abundant intellectual (and other) resources at Mila, University of Montreal, we delved into various successful projects, contributing to the advancement of generative modeling in computer vision and paving the way for future exploration and innovation in this exciting field:
\begin{enumerate}
\item \textbf{Neural ODEs for video prediction}: novel use of Neural ODEs to model the dynamics of a video~\citep{voleti2019simple}.
\item \textbf{Multi-Resolution Continuous Normalizing Flows (MRCNF)}: a normalizing flow approach that uses fewer parameters and significantly less time to train than existing works~\citep{voleti2019simple}.
\item \textbf{Neural inverse kinematics with 3D human pose prior}: integrating the most useful inverse kinematics approach for animators (ProtoRes) with the best 3D human pose prior (SMPL) trained on the biggest 3D human pose dataset (AMASS)~\citep{voleti2022smpl}.
\item \textbf{Non-isotropic denoising diffusion models}: a novel formulation of denoising diffusion models that expands their capabilities beyond the original formulation~\citep{voletiv2022ddpm}.
\item \textbf{Denoising diffusion models for video prediction, generation, interpolation}: novel use of denoising diffusion models to solve all three video tasks simultaneously with the same model~\citep{voleti2022MCVD}.
\end{enumerate}

Towards the end of the last project on video prediction using denoising diffusion models, diffusion-based text-to-image models gained significant traction, followed by text-to-3D object, text-to-video, and text-to-4D. The success of denoising diffusion models has sparked a surge of interest in this area, with many researchers exploring novel ideas and techniques to improve upon existing methods and further advance the field of generative modeling. As the field continues to grow, there are several areas of focus that researchers are likely to explore in the future.

One area of interest is the development of more efficient and scalable generative models that can handle larger datasets and produce more realistic results. Text has emerged as a powerful input modality in driving the generation of images, 3D models, and videos, showcasing significant progress in recent years. Text-to-image models include DALL-E(2)~\citep{ramesh2021zero,dalle2}, Imagen~\citep{saharia2022photorealistic}, and the open source Stable Diffusion~\citep{rombach2022high}. Text-to-3D models such as DreamFusion~\cite{poole2022dreamfusion}, Magic3D~\citep{lin2023magic3d}, Shap-E~\citep{jun2023shape} leverage text-to-image models to synthesize 3D objects from text. Recent text-to-video models include Make-A-Video by Meta~\citep{singer2022make}, ImagenVideo by Google~\citep{ho2022imagevideo}, Gen2 by RunwayML~\citep{esser2023structure}, etc. Make-A-Video3D~\citep{singer2023textto4d}, which Vikram had contributed to, enables the generation of 4D scenes from text prompts, utilizing the foundations of text-to-video models.

Advanced techniques of controlling and editing the synthesized results of generative models have grown immensely in the recent past as well. Diffusion-based controllable methods include Textual Inversion~\citep{gal2022textual}, Dreambooth~\citep{ruiz2023dreambooth}, Imagic~\cite{kawar2023imagic}, ControlNet~\citep{zhang2023adding}, InstructPix2Pix~\citep{brooks2023instructpix2pix}, etc. Among recent GAN-based methods, DragGAN~\citep{pan2023draggan} exhibits promising potential in offering unprecedented levels of controllability in image editing using generative models. A recently developed work called DragDiffusion~\citep{shi2023dragdiffusion} showcases this controllability using denoising diffusion models.

Another area of interest in the research community is the integration of generative models with other machine learning techniques, such as reinforcement learning, to enable the creation of more intelligent and interactive synthetic environments. Additionally, there is likely to be increased focus on the ethical implications of generative models, including issues related to bias, privacy, and intellectual property.

In conclusion, the future of generative modeling holds tremendous promise, offering a multitude of exciting possibilities for further research and innovation in the domains of image generation, 3D modeling, and video synthesis. The advancements achieved thus far have propelled the field forward, pushing the boundaries of what is conceivable and laying the foundation for groundbreaking applications. As the field continues to evolve, we can expect breakthroughs that redefine our understanding and capabilities in generating images, 3D models, and videos. The potential applications are vast, ranging from entertainment and creative industries to healthcare, robotics, and beyond.

I hope this thesis has contributed to the understanding and advancement of generative modeling, exploring its applications in computer vision, discussing various techniques, and highlighting the significance of conditional variants. As the field continues to mature, it is essential to embrace new challenges, address limitations, and explore novel directions. The journey towards harnessing the full potential of generative modeling has just begun, and the future holds tremendous opportunities for further research, innovation, and transformative applications in the world of computer vision.